\documentclass[10pt]{svjour3}
\usepackage{amssymb,amsmath}
\usepackage{amssymb,amsmath}
\usepackage{soul}
\usepackage{graphics}
\usepackage{psfig}
\usepackage{epsfig}
\usepackage{graphicx}
\usepackage{algorithmic}
\usepackage[ruled]{algorithm2e}
\usepackage{morefloats}
\usepackage{color,soul}
\usepackage{url}
\usepackage{mathtools, cuted}
\usepackage[most]{tcolorbox}
\usepackage{hyperref}
\hypersetup{
    colorlinks=true,
    linkcolor=blue,
    filecolor=magenta,      
    urlcolor=cyan,
    pdftitle={Sharelatex Example},
    bookmarks=true,
}

\begin{document}

\title{State-of-The-Art Fuzzy Active Contour Models for Image Segmentation}

\author{Ajoy~Mondal and Kuntal Ghosh}
\institute{CVIT,
International Institute of Information Technology, \\
Hyderabad, India \\
email: ajoy.mondal83@gmail.com \\
MIU,
Indian Statistical Institute, \\
Kolkata, India \\
email: kuntal@isical.ac.in }
\maketitle

\begin{abstract}

Image segmentation is the initial step for every image analysis task. A large variety of segmentation algorithm has been proposed in the literature during several decades with some mixed success. Among them, the fuzzy energy based active contour models get attention to the researchers during last decade which results in development of various methods. A good segmentation algorithm should perform well in a large number of images containing noise, blur, low contrast, region in-homogeneity, etc. However, the performances of the most of the existing fuzzy energy based active contour models have been evaluated typically on the limited number of images. In this article, our aim is to review the existing fuzzy active contour models from the theoretical point of view and also evaluate them experimentally on a large set of images under the various conditions. The analysis under a large variety of images provides objective insight into the strengths and weaknesses of various fuzzy active contour models. Finally, we discuss several issues and future research direction on this particular topic.


\end{abstract}

\section{Introduction}\label{introduction}

Image segmentation is a fundamental task in image analysis, computer vision, medical image processing, etc~\cite{gonzalez,GARCIALAMONT2018}. Segmentation is a process of partitioning an image into various regions which are homogeneous with respect to their features (e.g. intensity, color, texture, etc)~\cite{gonzalez,zaitoun2015survey}. Various image segmentation algorithms have been developed during several decades~\cite{fu1981survey,sahoo1988survey,pal1993review,khan2014survey,zaitoun2015survey}. Among them, clustering and active contour models ({\sc acm}s) are most commonly used for image segmentation. Fuzzy logic has been used to solve various problems: decision making~\cite{amin2019dealer,fahmi2019trapezoidal}, pattern recognition~\cite{melin2018type,MITCHELL2005409}, image segmentation~\cite{naz2010image,zhang2017improved}, etc. Fuzzy clustering has been successfully considered from the early stage of the image segmentation task up to now~\cite{dunn1973fuzzy,bezdek1981pattern}. It can retain more information from the original image than crisp clustering by introducing the degree of belongingness of each image pixel to the clusters~\cite{bezdek1981pattern}. Fuzzy clustering using global image information is not robust for images which are corrupted by various types of noise~ \cite{dunn1973fuzzy,bezdek1981pattern}. A large number of modified fuzzy clustering techniques has been proposed by incorporating local information which is derived from the image to improve segmentation accuracy~\cite{cai2007fast,krinidis2010robust,gong2013fuzzy,liu2015incorporating,zhang2017improved}. Due to incorporation of local information, these methods are robust to the noisy environment to some extend. 

Active contour model ({\sc acm}) developed by Kass \emph{et al.}~\cite{kass_active_contour_1988}
is successfully applied for image segmentation task~\cite{active_contour_nguyen_2012}. 
The main idea of this technique is deformation of initial curve. Finally, it evolves towards the object boundary under some constraints. It produces closed parametric curve which represents object boundary~\cite{cremers_2007_review}. Results are highly depend on the initial contour position and model is not robust for noisy and/or blurred images. To overcome these problems, various modified {\sc acm}s are invented in the literature~\cite{caselles1993geometric,malladi_geometric_active_contour_1995,caselles_1997_geodesic,yezzi_1997_geometric,cohen_1997_global,li2010distance,gunn_1997,zhang2013local}. {\sc acm}s consider image gradient to attract the contour towards object boundary. They fail to extract the contour of noisy, blurred and discontinuous edged images.   

Different from image gradient based {\sc acm} models, Chen and Vese~\cite{chan_2001_active} proposed an active contour (referred as, Chen-Vese) model which depends on the region information of the image. In this model, two regions inside and outside of the contour are assumed to be homogeneous. They formulated it as an energy minimization problem and included global image statistics in this energy function. The energy function is then transferred into level set formulation. Chen-Vese model is robust to the initial contour position and noisy, blurred and discontinuous edged images. However, Chen-Vese model fails to segment images having intensity in-homogeneity. To solve this problem, various modification on Chen-Vese model have been done by incorporating local image statistics~\cite{li_2008_minimization,lankton_2008_localizing,zhang2010active,zhang2010activelocal,wang2017active}.  

Krinidis and Chatzis~\cite{fuzzy_krinidis_2009} formulated region based active contour model as a minimization of fuzzy energy function different from Chen-Vese formulation. This model is referred to as fuzzy energy based active contour model which can handle objects whose boundaries are not necessarily defined by gradient, objects with very smooth or even with discontinuous boundaries. The fuzzy logic has been intensively used in clustering (e.g. image segmentation) but not in active contour models. Generally, fuzzy methods provide more accurate and robust clustering, thus the authors combine fuzzy logic with active contour method to introduce fuzzy energy based active contour model to segment images. The fuzziness of this energy function has the ability to reject local minima and it is able to produce better segmentation results. Recent days, the researchers developed various new fuzzy energy based models~\cite{shyu_2012_global,tran2014image,tran2014zernike,mondal2016robust} by modifying   
the base model proposed by Krinidis and Chatzis~\cite{fuzzy_krinidis_2009}
which are able to segment images in various complex environments. Recently, the fuzzy energy based active contour models get attention from the researchers for image segmentation tasks. Image segmentation using fuzzy energy based active contour models have indeed progressed to impressive, even amazing individual results. But as long as most of image segmentation papers (using fuzzy energy based active contour models) still use a limited number of images to test the performance of their approach, it is very difficult to conclude anything on the robustness of the  methods in a variety of circumstances. We feel the time is ripe for an experimental survey for various conditions.

The aim of this survey is to access the state of the art fuzzy energy based active models for image segmentation with an emphasis on the accuracy and the robustness of the models on varying environments. We aim to group the models on the basis of global or local or both image information considered to formulate energy function. We also experimentally evaluate the performance of the existing models on the large number of images having various complexities with respect to various measures: Jacard error and F-measure. We have gathered 100 images of various categories under varying environments. Finally, we provide several issues associated these methods and future research direction. We hope that this survey will help the reader to understand insight into the strengths and weaknesses of various fuzzy energy based active contour models from both the theoretical and practical point of views.

Rest of the paper is organised as follows. We discuss global fuzzy energy based active models where energy term is defined with the global image feature in Section~\ref{global_fuzzy_energy_based_active_contour}. Section~\ref{local_fuzzy_active_contour_model} reviews on various local fuzzy energy based active contour models while Section~\ref{global_local_fuzzy_active_contour_model} analyzes global and local fuzzy energy based active models from theoretical point of view. Experimental results and performance analysis of the existing fuzzy active models on various categories of images under several complex conditions are presented in Section~\ref{experiment}. We discuss various issues and future research direction in Section~\ref{discussion} and finally conclusive remark is given in Section~\ref{conclusion}.        

\section{Global Fuzzy Energy based Active Contour Models} \label{global_fuzzy_energy_based_active_contour}

Krinidis and Chatzis~\cite{fuzzy_krinidis_2009} proposed a fuzzy energy based active contour model for image segmentation. In this section, we discuss their model and its variations.

Let $I(X): \Omega  \to \Re ^d$ be a given vector valued image, where $\Omega  \subset \Re^2$ and $d \ge 1$ are the image domain and the dimension of the vector $I(X)$, respectively. In case of the gray level images, $d=1$ while $d=3$ for the color images. Let $C$ be a closed contour in the image domain $\Omega$ which separates $\Omega$ into two regions: $\Omega _1  = inside\left( C \right)$ and $\Omega _2  = outside\left( C \right)$. When a given image $I(X)$ is approximated by two regions over the image domain $\Omega$, the energy function is defined as

\begin{eqnarray}
\begin{array}{l}
 F\left( {C,c_1,c_2 ,u} \right) = \mu .L\left( C \right) + \lambda _1 \int\limits_\Omega  {\left[ {u\left( X \right)} \right]} ^m \left\| {I(X) - c_1 } \right\|^2 dX \\
  + \lambda _2 \int\limits_\Omega  {\left[ {1 - u\left( X \right)} \right]} ^m \left\| {I(X) - c_2 } \right\|^2 dX   \label{energy_function},
 \end{array}
\end{eqnarray}
where $c_1 \in \Re^d$ and $c_2 \in \Re^d$ (depending on $C$) are the average prototypes of the regions inside and outside of the contour $C$, respectively. The membership function $u\left( X \right) \in \left[ {0,1} \right]$ is the degree of membership of a pixel $X$ for belonging to the inside of $C$. $m>1$ is a weight exponent which controls the degree of `fuzziness' of each membership value. $L(C)$ is the length of the curve ($C$). The authors considered global image statistics for energy formulation and it is referred to as global fuzzy energy based active contour model.

The energy function defined in Eq.~(\ref{energy_function}) will be minimized when the contour $C$ is exactly lie on the object boundary and average prototypes $c_1$ and $c_2$ optimally approximate the given image $I(X)$ into two regions i.e. inside and outside of $C$, respectively. The objective is to find such a contour $C$ which will minimize the energy function defined in Eq. (\ref{energy_function}) over all pixels $X \in \Omega$. Let us assume, $C_{O}$ be the optimal extracted boundary of the object, then the energy function given in Eq.~(\ref{energy_function}) will be minimized if $C = C_O$, i.e.
\begin{eqnarray}
\begin{array}{l}
C_O  = \mathop {\arg \min }\limits_C F\left( {C,c_1,c_2 ,u} \right)\label{minimization}.
\end{array}
\end{eqnarray}
Therefore, the curve $C_O$ is the solution to the segmentation problem.

\paragraph{\textbf{Pseudo Level Set Formulation}}

Krinidis and Chatzis~\cite{fuzzy_krinidis_2009} defined a pseudo level set formulation, similar to the level set method~\cite{osher1988fronts}, based on the membership values $u$, where the curve $C \subset \Omega$ is implicitly represented by the pseudo zero level set of Lipschitz similar function $u: \Omega  \to \Re$, such that

\begin{eqnarray}
\begin{array}{l}
u(X) = \left\{
     \begin{array}{lr}
     C = \left\{ {X \in \Omega:u\left( X \right) = 0.5} \right\}, \\
     inside\left( C \right) = \left\{ {X \in \Omega:u\left( X \right) > 0.5} \right\}, \\
     outside\left( C \right) = \left\{ {X \in \Omega:u\left( X \right) < 0.5} \right\}. \\
     \end{array}
   \right. \label{level_set}
\end{array}
\end{eqnarray}

Here, it is noted that the regularization or penalty term of the proposed energy function $F(C,c_1,c_2,u)$ given in Eq.~(\ref{energy_function}) is $L(C) = \int\limits_\Omega  {\left| {\nabla H\left( {u\left( X \right) - 0.5} \right)} \right|} dX$, where $H\left( s \right)$ is a Heaviside function~\cite{osher1988fronts}.

Therefore, the energy function in Eq.~(\ref{energy_function}) can be rewritten as

\begin{eqnarray}
\begin{array}{l}
 F\left( {C,c_1,c_2 ,u} \right) = \mu \int\limits_\Omega  {\left| {\nabla H\left( {u\left( X \right) - 0.5} \right)} \right|} dX + \lambda _1 \int\limits_\Omega  {\left[ {u\left( X \right)} \right]} ^m \left\| {I(X) - c_1 } \right\|^2 dX \\
  + \lambda _2 \int\limits_\Omega  {\left[ {1 - u\left( X \right)} \right]} ^m \left\| {I(X) - c_2 } \right\|^2 dX   \label{energy1_function},
 \end{array}
\end{eqnarray}

When the energy function defined in Eq.~(\ref{energy1_function}) is minimized, the values of $u$ for the pixels located inside the contour $C$ are different from its values for the pixels located outside the contour. However, the values of $u$ for the pixels located inside the contour $C$ are similar. This is same for the pixels located outside the contour $C$.

Keeping $u$ fixed, the authors minimized Eq.~(\ref{energy1_function}) with respect to $c_{1}$ and $c_{2}$, these can be expressed as function of $u$ as

\begin{eqnarray}
\begin{array}{l}
 c_1 = \frac{\int\limits_\Omega  {\left[ {u\left( X \right)} \right]}^m I(X) dX }{\int\limits_\Omega  {\left[ {u\left( X \right)} \right]}^m dX} \label{c1_function}
 \end{array}
\end{eqnarray}

and

\begin{eqnarray}
\begin{array}{l}
 c_1 = \frac{\int\limits_\Omega  {\left[ {1 - u\left( X \right)} \right]} ^m I(X) dX}{\int\limits_\Omega  {\left[ {1 - u\left( X \right)} \right]} ^m dX} \label{c2_function}.
 \end{array}
\end{eqnarray}

Keeping the values of $c_{1}$ and $c_{2}$ fixed, the membership of each pixel, $u(X)$, is then computed using:

\begin{eqnarray}
\begin{array}{l}
 c_1 = \frac{1}{1+\left(\frac{\lambda _1 \left\| {I(X) - c_1 }\right\|^2}  {\lambda _2 \left\| {I(X) - c_2 }\right\|^2} \right)^{\frac{1}{m-1}}} \label{member_function}.
 \end{array}
\end{eqnarray}

For simplicity, without loss of generality, the above minimization (Eq.~(\ref{member_function})) has been done without considering the length term (i.e. $\mu=0$).

Due to success of this model on segmentation of blurred, noisy and discontinuous edged images, the researcher's considered this model or developed various models by modifying fuzzy energy based active contour model~\cite{fuzzy_krinidis_2009} in the literature to segment variety of images~\cite{tran2014image,wu2015novel,pereira2011fuzzy,gong2015efficient,badshah2018segmentation}. Model proposed by Krinidis and Chatzis~\cite{fuzzy_krinidis_2009} are considered to analyze liver CT images in~\cite{sajithregion} and segment brain tissue in~\cite{chen2008fuzzy}.

Fuzzy energy based active contour model produces good segmentation results for blurred, noisy and discontinuous edged images. However, it fails for images containing gradual tonality variations, region in-homogeneity, background clutter, object occlusion, etc. To overcome drawbacks of this model, several new methods are invented in the literature~\cite{tran2014image,wu2015novel,pereira2011fuzzy,gong2015efficient,badshah2018segmentation}. Pereira {\em et al.}~\cite{pereira2011fuzzy} observed that fuzzy active contour model~\cite{fuzzy_krinidis_2009} fails to properly segment images containing gradual tonality variations. To properly segment multi-tonality images, they modified fuzzy active contour model~\cite{fuzzy_krinidis_2009} by incorporating type-$2$ fuzzy set which increases the fuzziness of the energy function. Instead of using $u$, which is type-$1$ fuzzy membership, they applied the type-$2$ values $\tilde{u}$ in Eqs. (\ref{c1_function}) and (\ref{c2_function}), represented by $\tilde{c_1}$ and $\tilde{c_2}$, respectively. The authors defined energy function as
\begin{eqnarray}
\begin{array}{l}
F\left( {C,\tilde{c_1}, \tilde{c_2}, \tilde{u}} \right) = \mu .L\left( C \right) + \lambda _1 \int\limits_\Omega  {\left[ {\tilde{u}\left( X \right)} \right]} ^m \left\| {I(X) - \tilde{c_1} } \right\|^2 dX \\
+ \lambda _2 \int\limits_\Omega  {\left[ {1 - \tilde{u\left( X \right)}} \right]} ^m \left\| {I(X) - \tilde{c_2} } \right\|^2 dX   \label{energy2_function},
\end{array}
\end{eqnarray}

where $\tilde{u}$ is type-$2$ membership function. The authors used four different functions (Triangular, Exponential, Logarithmic and Circular) for calculating type-$2$ membership values. They are defined as 

Triangular: \begin{eqnarray}
\begin{array}{l}
\tilde{u}(X) = u(X)- \frac{1-u(X)}{2} \label{triagular_function},
\end{array}
\end{eqnarray}

Exponential: \begin{eqnarray}
\begin{array}{l}
\tilde{u}(X) = e^{[u(X)]^{2}}-1 \label{exponential_function},
\end{array}
\end{eqnarray}

Circular: \begin{eqnarray}
\begin{array}{l}
\tilde{u}(X) = \left\{
     \begin{array}{lr}
     \sqrt{u(X)-[u(X)]^{2}}, \text{ if } u(X)<0.5\\
     1-\sqrt{u(X)-[u(X)]^{2}}, \text{ else }  \\
     \end{array}
   \right. \label{circular_function},
\end{array}
\end{eqnarray}

Logarithmic: \begin{eqnarray}
\begin{array}{l}
\tilde{u}(X) = log_{2}\left(u(X)-1\right) \label{logarithmic_function},
\end{array}
\end{eqnarray}

Minimization of energy function defined in Eq.~(\ref{energy1_function}) in fuzzy active contour model~\cite{fuzzy_krinidis_2009} is time consuming. To minimize the computational cost of this energy function, Gong {\em et al.} proposed a novel bi-convex fuzzy variation image segmentation method in~\cite{gong2015efficient}. Bi-convex fuzzy variation function finds the global optimal solution and reduces the computational cost.

Tran {\em et al.}~\cite{tran2014image} developed a fuzzy energy based active contour model with shape prior for image segmentation. The authors modified fuzzy energy function defined by Krinidis and Chatzis~\cite{fuzzy_krinidis_2009} in Eq.~(\ref{energy_function}). The modified energy function is 

\begin{eqnarray}
\begin{array}{l}
 F\left( {C,c_1,c_2,u, \hat{\psi}} \right) = \mu .L\left( C \right) + \lambda _1 \int\limits_\Omega  {\left[ {u\left( X \right)} \right]} ^m \left\| {I(X) - c_1 } \right\|^2 dX \\
  + \lambda _2 \int\limits_\Omega  {\left[ {1 - u\left( X \right)} \right]} ^m \left\| {I(X) - c_2 } \right\|^2 dX +\beta F_{shape}\left(u, \hat{\psi} \right)  \label{energy3_function},
 \end{array}
\end{eqnarray}
where $\beta > 0$ is a weighting factor and $F_{shape}\left(u, \hat{\psi}\right)$ is the shape prior.  The authors defined shape prior as 

\begin{eqnarray}
\begin{array}{l}
 F_{shape}\left( {u, \hat{\psi}} \right) = \int\limits_\Omega  {\left[ {u\left( X \right)} \right]} ^m \left({1 - \hat{\psi}(X)} \right) dX + \int\limits_\Omega  {\left[ {1 - u\left( X \right)} \right]} ^m \left(\hat{\psi}(X) \right) dX  \label{shape_function},
 \end{array}
\end{eqnarray}

where $\hat{\psi}(x) \in [0, 1]$ is the reference shape. $u$ as the shape indicated by a curve $C$ which is associated to the segmentation. $u$ can be interpreted as a binary image whose evolving shape $u$
value is $1$ for the pixels located inside $C$ and $0$ for the pixels outside $C$. The dissimilarity between $u$ and $\hat{\psi}$ can be expressed as $1-\hat{\psi}(X)$ for pixel inside $C$ and $\hat{\psi(X)}$ for pixel outside $C$. $u(x)$ is similarly defined as in Eq.~(\ref{energy_function}).  
The shape prior helps the model to properly segment images with background clutter and object occlusion. Limitation of this model is unavailability of shape information.

In the same direction, Pham {\em et al.}~\cite{pham2016shape}
presented a fuzzy energy-based active contour model for image segmentation with shape prior based on collaborative representation of training shapes. The authors defined new energy function as 

\begin{eqnarray}
\begin{array}{l}
 F\left( {C,c_1,c_2,u,S}\right) = \int\limits_\Omega  {\left[ {u\left( X \right)} \right]} ^m \left\| {I(X) - c_1 } \right\|^2 dX + \int\limits_\Omega  {\left[ {1 - u\left( X \right)} \right]} ^m  \left\| {I(X) - c_2 } \right\|^2 dX \\
 +\beta \left(\int\limits_\Omega(H(u(x)-T)-DS)^{2} dX+ \lambda \left\| S \right\|^2 \right) + \mu \left(\delta(u(x)-T)|\bigtriangledown(u(x)-T)|dX \right) \label{energy4_function},
\end{array}
\end{eqnarray}

where $S \in R$ is a coefficient vector and $DS$ is combination of shape dictionary $D$. $H(v)$ is heavy side function and $0<T<1$. $\beta$ and $\mu$ are positive weighting coefficients. Here, fuzzy energy consists of a data term and a shape prior term. The data term relies on image information to guide the evolution of the contour. The prior shape is represented as the combination of atoms in the shape dictionary based on collaborative representation. Meanwhile, the shape prior term constrains the contour evolution with respect to the prior shape to handle the background clutter and the object occlusion. This model can segment images with background clutter and object occlusion even when the training set includes shapes with large variation. In addition, this shape collaborative representation model also takes less computational time compared to shape sparse representation approach~\cite{tran2014image}.

Wu {\em et al.} proposed a novel fuzzy energy based active contour model with kernel metric for a robust and stable image segmentation~\cite{wu2015novel}. The author defined the energy function as 

\begin{eqnarray}
\begin{array}{l}
 F\left( {C,c_1,c_2,u}\right) = \mu \int\limits_\Omega \delta(u(X-0.5)) \left\| \bigtriangledown(X-0.5)\right\| dX \\
 + \lambda_1 \int\limits_\Omega (u(X))^m \left(1- K(I(X),c_1)\right) dX + \lambda_2 \int\limits_\Omega (1-u(X))^m \left(1- K(I(X),c_2)\right) dX \label{energy5_function},
\end{array}
\end{eqnarray}

where $K$ is Gaussian kernel function defined as $K(a,b) = exp\left(\frac{-(a-b)^2}{\sigma}\right)$ where $\sigma$ is the bandwidth of the kernel function. Incorporation of kernel metric in the energy function and the fuzziness of the energy evolve the contour very stably without the re-initialization. This model properly segment low contrast images. In the similar direction, Badshah and Ahmad~\cite{badshah2018segmentation} proposed a fuzzy energy based active contour model to segment images having multi-objects with varying intensities. The authors defined energy function as 

\begin{eqnarray}
\begin{array}{l}
 F\left( {C,c_1,c_2,u}\right) = \mu \int\limits_\Omega (u(X))^m \left(1- K(I(X),c_1)\right) dX \\ + \int\limits_\Omega (1-u(X))^m \left(1- K(I(X),c_2)\right) dX \label{energy6_function},
\end{array}
\end{eqnarray}

where $K$ is kernel function and $\mu$ is positive parameter. The authors considered Gaussian type radial basis kernel based on generalized average into their energy function. Instead of length term as in~\cite{fuzzy_krinidis_2009}, Gaussian smoothing is considered as regularizer term in their energy function. Experimentally they showed that it performs well for the complex images. 

\section{Local Fuzzy Energy based Active Contour Models} \label{local_fuzzy_active_contour_model}

Although, the global fuzzy energy based active models provide good segmentation results, however they fail to extract proper boundary of the object when images contain noise and intensity in-homogeneity. In such cases, the local energy term derived from the local image statistics is better to extract proper object boundary. Various methods~\cite{shyu2012fuzzy,tran2014zernike,fang2016localized} are invented to properly segment images under complex environment.          

Shyu {\em et al.}~\cite{shyu2012fuzzy} proposed a fuzzy energy based active contour model involving intensity distribution information of the image to segment them. The authors defined energy function as 

\begin{eqnarray}
\begin{array}{l}
 F_{local}\left( {u, h_{1}(X), h_{2}(X)}\right) = \lambda_1 \int\limits_\Omega \left(\int\limits G_{\sigma}(X-Y) \left\|I(Y)-h_{1}(X) \right\|^{2} \left[u(Y) \right]^{m} dY\right)dX \\ +
 \lambda_2 \int\limits_\Omega \left(\int\limits G_{\sigma}(X-Y) \left\|I(Y)-h_{2}(X) \right\|^{2} \left[u(Y) \right]^{m} dY\right)dX \label{energy7_function},
\end{array}
\end{eqnarray}

where $Y$ is pixel in the local region around pixel $X$, $I(Y)$ is the intensity of pixel $Y$, $G_\sigma(X-Y)$ is the Gaussian kernel function and $h_{1}(X)$ and $h_{2}(X)$ average intensity values of local region which are obtained by Gaussian mixture model based intensity distribution estimator operator. The intensity distributions are derived using a Gaussian mixture model based intensity distribution estimator before the curve evolution. This model provides good segmentation results for noisy images.
Tran {\em et al.}~\cite{tran2014zernike} proposed a fuzzy energy based active contour based on local image statistics similar to Eq.~(\ref{energy7_function}) to extract object boundaries. In this paper, the local image intensity distribution information is derived by Hueckel operator in the neighborhood of each image pixel. The parameters of Hueckel operator are estimated by a set of orthogonal Zernike moments. Here, the fuzzy membership function is considered to measure the association degree of each image pixel to the region outside and inside the contour. This model deals with images with intensity in-homogeneity. In similar direction, Fang {\em et al.}~\cite{fang2016localized} presented a novel fuzzy region-based active contour model for image segmentation. For each pixel, local patch energy is incorporated as fuzzy energy term for the curve evolution. Due to the local information, this method is robust to the noisy images. Most of the cases, the local fuzzy energy based methods produce better segmentation results than the global fuzzy energy based techniques.  

\section{Global and Local Fuzzy Energy based Active Contour Models} \label{global_local_fuzzy_active_contour_model}

Several algorithms~ \cite{shyu2012global,krinidis2012fuzzy,mondal2016robust,mondal2016robust_conference} have been developed to better segment images with noise, region in-homogeneity, etc. by incorporating both the local and global image information into the energy function. The local energy is generated 
based on local images statistics which can deal with images having high intensity in-homogeneity or non-homogeneity, noise and blurred boundary or discontinuous edges. While the global term is derived from the global image statistics to avoid unsatisfactory results
due to bad initialization. Combination of both local and global fuzzy energy based active contour models provide satisfactory segmentation results.

Shyu {\em et al.}~\cite{shyu2012global} introduced an energy function consisting of a local fuzzy energy and a global fuzzy energy terms to attract the active contour and stop it on the object boundary. They defined energy function as 

\begin{eqnarray}
\begin{array}{l}
 F\left( {C,c_1,c_2,f_1, f_2, u_1, u_2}\right) = \beta \sum_{i=1}^{2} \lambda_{i} \int\limits_\Omega\left(u_{i}(X)\right)^{m} \|I(X)-c_{i}\|^{2} dX \\ 
 + (1-\beta) \sum_{i=1}^{2} \lambda_{i} \int\limits_\Omega \left(\int\limits \left[u_{i}(Y) \right]^{m} G_{\sigma}(X-Y) \|I(Y)-f_{i}(X)\| dY \right)dX \label{energy8_function},
\end{array}
\end{eqnarray}
where $\sum_{i=1}^{2} \lambda_{i} \int\limits_\Omega\left(u_{i}(X)\right)^{m} \|I(X)-c_{i}\|^{2} dX$ is global fuzzy energy term and \\ $\sum_{i=1}^{2} \lambda_{i} \int\limits_\Omega \left(\int\limits \left[u_{i}(Y) \right]^{m} G_{\sigma}(X-Y) \|I(Y)-f_{i}(X)\| dY \right)dX$ is local fuzzy energy term. Similar to Eq.~(\ref{energy1_function}), $\lambda_{1}$, $\lambda_{2}>0$ are two fixed parameters, $c_{1}$ and $c_{2}$ are two constant that approximate the image intensities inside and outside of the contour ($C$), $u_{1}(X) = u(X) \in [0, 1]$ is the degree of membership of $I(X)$ to the inside of the contour $C$ and $m$ is  a weighting exponent on each fuzzy membership. $f_{1}(X)$ and $f_{2}(X)$ are two local functions using to approximate the intensity means of two local regions around the pixel $X$ inside and outside the contour $C$. $I(Y)$ represents the intensities of the pixels $Y$ which are in a local region centered at the pixel $X$ ($Y$ is a neighborhood of $X$). $G_{\sigma}$ is a Gaussian kernel. The membership function $u(Y)\in[0,1]$ is the belongingness of pixel $Y$ to the inside the local region centered at the pixel $X$ inside the contour $C$. $\beta$ is a balance constant ($0\leq \beta \leq 1$). The local energy term deals with the intensity in-homogeneity presents in the images. The global energy term avoids unsatisfying results due to unsuitable initial contour position. 

In this direction, Krinidis and Krinidis~\cite{krinidis2012fuzzy} proposed a robust fuzzy energy based active contour model using both global and local energy for image segmentation. The proposed energy function is defined as 

\begin{eqnarray}
\begin{array}{l}
 F\left( {C,c_1,c_2,u}\right) =  \int\limits_\Omega [u(X)]^{m} \left[ \|I(X)-c_{1} \|^{2} + \int\limits_\Omega w_{XY}[1-u(Y)]^{m} \|I(Y)-c_{1} \|^{2} dY \right]dX \\ +
 \int\limits_\Omega [1-u(X)]^{m} \left[ \|I(X)-c_{2} \|^{2} + \int\limits_\Omega w_{XY}[u(Y)]^{m} \|I(Y)-c_{2} \|^{2} dY \right]dX \label{energy9_function},
\end{array}
\end{eqnarray}

$w_{XY}$ incorporates the spatial dependency between pixel $X$ and its neighborhood pixels $Y$. $w_{XY}$ is defined as 

\begin{eqnarray}
\begin{array}{l}
w_{XY}=
     \begin{cases}
       \frac{1}{d_{XY}+1},  &\quad\text{if}\ X\ \text{and}\ Y \text{are neighbours} \\
        0, &\quad\text{otherwise.} \\ 
     \end{cases}
 \label{energy10_function},
\end{array}
\end{eqnarray}
where $d_{XY}$ is the distance between pixels $X$ and $Y$. Here, the local energy is derived based on spatial neighborhood of a pixel. This method provides better results than global fuzzy energy based active contour model~\cite{fuzzy_krinidis_2009}. 

Thieu {\em et al.} developed a fuzzy energy based active contour model using Gaussian distribution function in~\cite{thieu2015efficient}. They defined energy function as

\begin{eqnarray}
\begin{array}{l}
 F\left( {C,c_1,c_2, \xi_{1}^{2}, \xi_{2}^{2}, f_1, f_2, \sigma_{1}^{2}, \sigma_{2}^{2}, u}\right) =  
\mu|C|+\lambda \left[-\int\limits_\Omega log G(I(X)-c_{1}, \xi_{1})[u(X)]^{m}dX\right] \\+
\lambda \left[-\int\limits_\Omega log G(I(X)-c_{2}, \xi_{2})[1-u(X)]^{m}dX \right] \\ +
(1-\lambda) \left[-\int\limits_\Omega \left[ \int\limits_\Omega G(X-Y, \sigma) 
log G(I(X)-f_{1}(Y), \sigma_{1}) [u(X)]^{m} dY \right]dX \right] \\ +
(1-\lambda)\left[-\int\limits_\Omega \left[ \int\limits_\Omega G(X-Y, \sigma) log G(I(X)-f_{2}(Y), \sigma_{2}) [1-u(X)]^{m} dY \right]dX \right] \label{energy11_function}.
\end{array}
\end{eqnarray}

$\mu\geq 0$ controls the length $|C|$ of the contour $C$. $0<\lambda<1$ controls the influence of global and local terms in the energy function. $c_{1}$ ($c_{2}$) and $\xi_{1}$ ($\xi_{2}$) are respectively global mean and standard deviation of the Gaussian distribution inside (outside, respectively) the contour $C$.   
$f_{1}$ ($f_{2}$) and $\sigma_{1}$ ($\sigma_{2}$) are respectively the local mean and standard deviation of the Gaussian distribution inside (outside, respectively) the contour $C$. $Y$ is the neighbourhood pixels of $X$. The advantages of this model are as the local intensity allows to handle intensity in-homogeneity, while the global intensity information helps to segment images with noisy or smooth boundary. 

Mondal {\em et al.}~\cite{mondal2016robust,mondal2016robust_conference} proposed a robust fuzzy energy based active contour model using both global and local information. Both local spatial and gray level/color information are considered to calculate local energy. The energy function is defined as

\begin{eqnarray}
\begin{array}{l}
 F\left( {C,c_1,c_2, u}\right) =  \mu.L(C)+ \lambda_{1} \beta \int\limits_\Omega [u(X)]^{m} \|I(X)-c_{1} \|^{2} dX \\ + 
\lambda_{2} \beta \int\limits_\Omega [1-u(X)]^{m} \|I(X)-c_{2} \|^{2} dX \\ +
\lambda_{1} (1-\beta) \int\limits_\Omega [u(X)]^{m}\left[\int\limits_\Omega w_{XY}[1-u(Y)]^{m} \|I(Y)-c_{1} \|^{2} dY \right]dX \\ +
\lambda_{2} (1-\beta) \int\limits_\Omega [1-u(X)]^{m}\left[\int\limits_\Omega w_{XY}[u(Y)]^{m} \|I(Y)-c_{2} \|^{2} dY \right]dX \label{energy11_function}. 
\end{array}
\end{eqnarray}

The term $I(Y)$ represents the intensity of a pixel $Y$ in a neighborhood of a pixel $X$. $w_{XY}$ is the weight of $Y$th pixel in a local neighborhood of pixel $X$. $\lambda_{1}>0$ and $\lambda_{2}>0$ are two fixed parameters. $0\leq \beta \leq 1$ is a constant to control the influence of both the global energy and local energy. The weight $w_{XY}$ depends on both the local spatial constraint and the local gray/color constraint. For each pixel $X\in \Omega$, the local spatial constraint reflects the damping extent of neighbors with the spatial distances from the central pixel and is defined as

\begin{eqnarray}
\begin{array}{l}
w_{XY}^{SC} = \frac{1}{1+d_{XY}}
\end{array}
\end{eqnarray}
with $d_{XY}=\|X-Y\|$, where $Y\in \eta_{X}$ is spatial neighborhood (local window) of $X$. The spatial constraint makes the influence of pixels within the local window. It can be changed according to their distances from the central pixel. With the help of spatial constraint, more local information is incorporated in the proposed energy model. Whereas, the local gray/color constraint is defined as

\begin{eqnarray}
\begin{array}{l}
w_{XY}^{FC} = exp\left[\frac{-\|X-Y\|}{\sum_{Y\in \eta_{X}} \|Y-X\|^2} \right],
\end{array}
\end{eqnarray}
where $I(X)$ is gray/color value of the central pixel $X$ within a spatial local window and $I(Y)$ is the gray/color value of $Y$th pixels in the same window. $\eta_{X}$ is the neighborhood of pixel $X$. The denominator is a function of local density surrounding the central pixel and its value reflects the gray/color value homogeneity degree of that local window. When its value is small, the local window is more homogeneous and vice versa. This equation indicates that when the intensity value $I(Y)$ of the $Y$th neighbors of central pixel $X$ is close to $I(X)$, $w_{XY}^{FC}$ should be large and vice versa. The value of $w_{XY}$ can be changed with different gray/color values of the pixels over an image and thus it reflects the damping extent in the intensity/color values. Now, the weight $w_{XY}$ based on both the local spatial constraint and the local gray/color constraint is defined as

\begin{eqnarray}
\begin{array}{l}
w_{XY}^{FC} = w_{XY}^{SC}.w_{XY}^{FC}.
\end{array}
\end{eqnarray}

This model better deals with images having high intensity in-homogeneity or non-homogeneity, noise and blurred boundary or discontinuous edges due to incorporation of the local energy term in the energy function. The global energy term is used to avoid unsatisfactory results due to bad initialization. They applied their proposed model for tracking camouflaged object in~\cite{mondal2017partially}. Due to incorporation of local and global image information into the energy function, the global and local fuzzy energy based active contour models produce better segmentation results than the individual local and global fuzzy energy based models. 

\section{Experiments} \label{experiment}

\subsection{Evaluation Measures} \label{evaluation_measures}

Qualities of segmentation results obtained by any algorithm depend on nature of the considered images. Here, it may be noted that one segmentation algorithm may not be always good choice for all types of images. On the contrary, different segmentation algorithms may produce different results for a particular image~\cite{zhang2008image}. Therefore, performance evaluation of a segmentation algorithm is the necessary task. Qualitative comparison is the most common one. In this article, two measures namely Tanimoto coefficient/ Jacard error~\cite{fuzzy_krinidis_2009} and F-measure~\cite{fmeasure2008} are considered to analysis the performance of these existing algorithms. 

\paragraph{\textbf{Jacard Error:}}

It is defined as
\begin{eqnarray}
\begin{array}{l}
J(S_{E},S_{O})=1-\frac{\int\limits_{S_{E}\bigcap S_{O}} dX}{\int\limits_{S_{E} \bigcup S_{O}} dX}\label{jacard_index},
\end{array}
\end{eqnarray}
where $S_{E}$ and $S_{O}$ are obtained and desired (ground truth) segmentation, respectively. Lower value of $J(S_{E},S_{O})$ indicates better segmentation result.

\paragraph{\textbf{F-measure:}}

It is defined as
\begin{eqnarray}
\begin{array}{l}
F=\frac{2P_{r}R_{e}}{P_{r}+R_{e}}\label{f_measure},
\end{array}
\end{eqnarray}
where $P_{r}$ and $R_{e}$ are precision and recall, respectively and defined as
\begin{eqnarray*}
\begin{array}{l}
P_{r}=\frac{TP}{TP+FP};\ \
R_{e}=\frac{TP}{TP+FN}.
\end{array}
\end{eqnarray*}
Here, $TP$ is true positive, $FP$ is false positive and $FN$ is false negative. Higher value of $F$ highlights the better segmentation algorithm.

\subsection{Considered Images}

We consider several categories of $100$ images\footnote{\url{https://www2.eecs.berkeley.edu/Research/Projects/CS/vision/bsds/}}\footnote{\url{http://imageprocessingplace.com/DIP-3E/dip3e_book_images_downloads.htm}}\footnote{\url{http://www.robots.ox.ac.uk/~vgg/data/flowers/102/}} for our experiments. Images contain intensity in-homogeneity, low contrast, background clutter, etc. To analysis the robustness of the considered algorithms under various conditions, the original images are corrupted by adding Gaussian noise, salt \& pepper noise, mixture of Gaussian and salt \& pepper noise and Gaussian blur. Therefore, these corrupted images are also taken into consideration for the experiment to analysis the robustness of these algorithms on difficult noisy environment.     

\subsection{Considered Algorithms}

We consider seven state-of-the-art algorithms to analysis their performance on various images under the different conditions. The selected algorithms are (i) fuzzy energy-based active contours ({\sc FEAC})~\cite{fuzzy_krinidis_2009}, (ii) novel fuzzy active contour model with kernel metric for image segmentation ({\sc NFACMKM})~\cite{wu2015novel}, (iii) on segmentation of images having multi-regions using Gaussian type radial basis kernel in fuzzy sets framework ({\sc FACGK})~\cite{badshah2018segmentation}, (iv) localized patch-based fuzzy active contours for image segmentation ({\sc LPFAC})~\cite{fang2016localized}, (v) fuzzy distribution fitting energy-based active contours for image segmentation ({\sc FDFEAC})~\cite{shyu2012fuzzy}, (vi) global and local fuzzy energy-based active contours for image segmentation ({\sc GLFEAC})~\cite{shyu2012global} and (vii) robust global and local fuzzy energy based active contour for image segmentation ({\sc RGLFEAC})~\cite{mondal2016robust}.

\section{Results Analysis}

\subsection{Segmentation of Original Images}

\begin{figure}[h!]
\centerline{
\psfig{figure=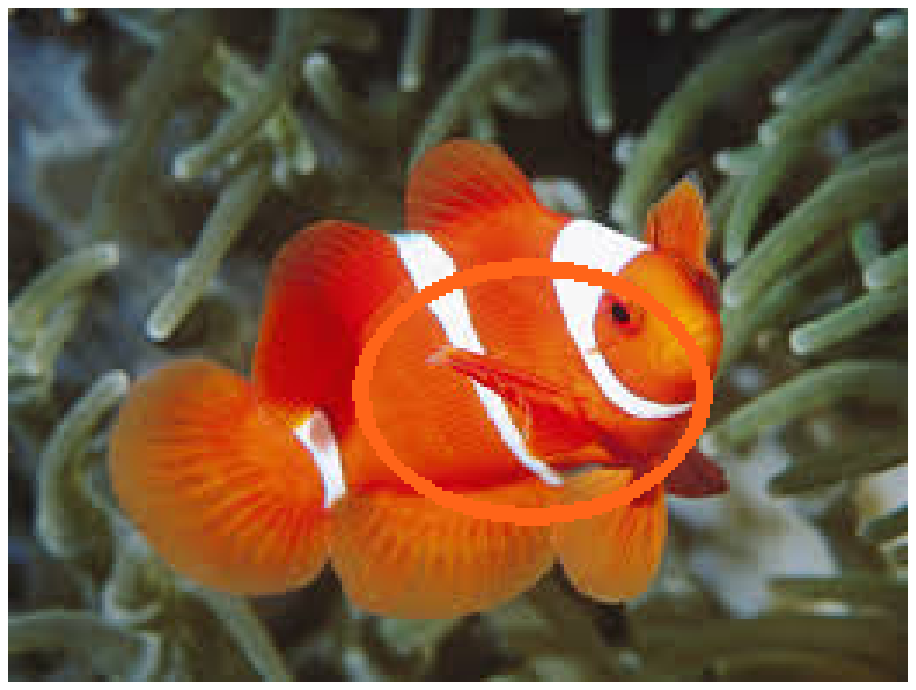,height=0.13\textwidth, width=0.2\textwidth}
\hspace{0.000001\textwidth}
\psfig{figure=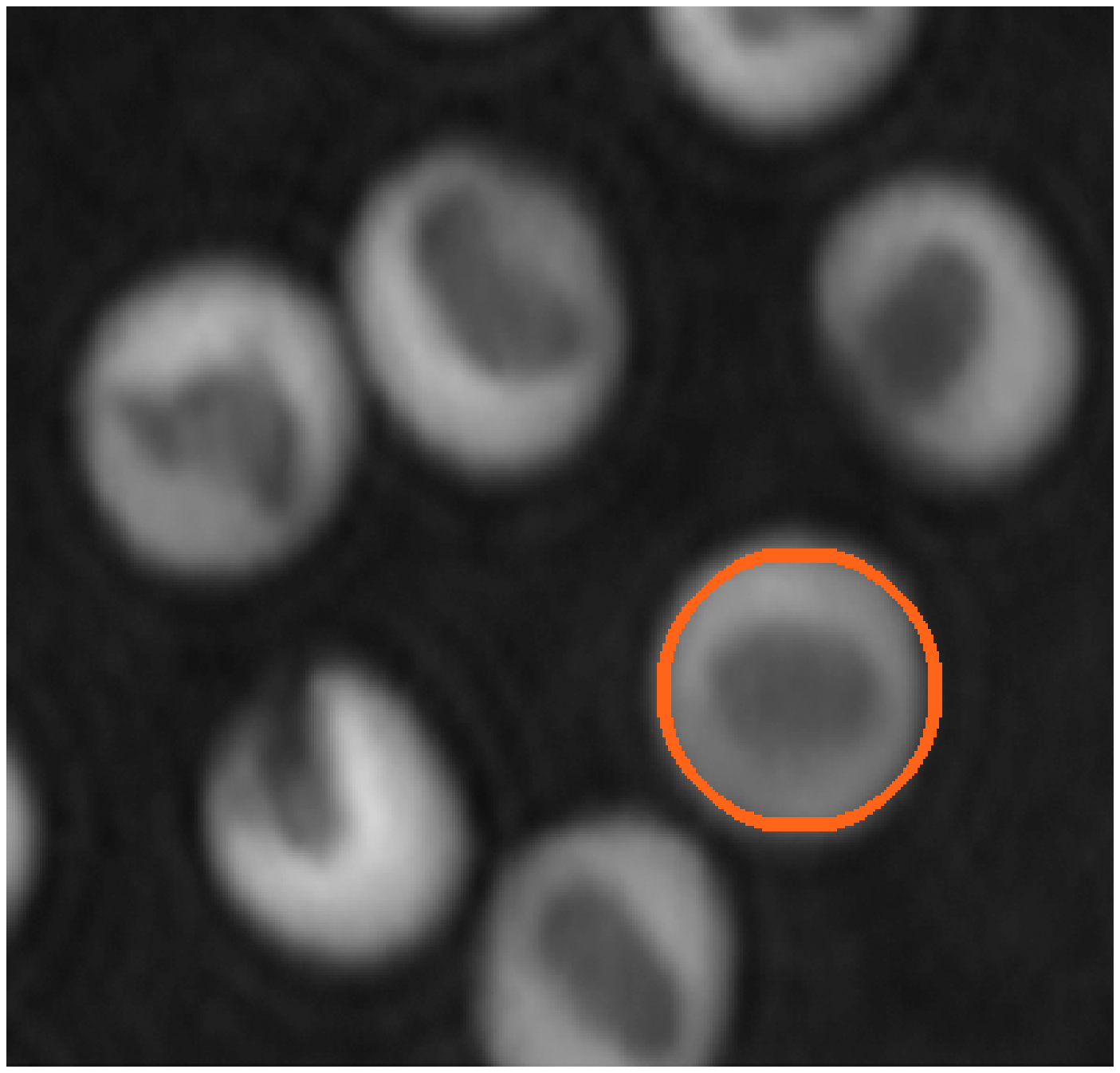,height=0.13\textwidth, width=0.2\textwidth}
\hspace{0.000001\textwidth}
\psfig{figure=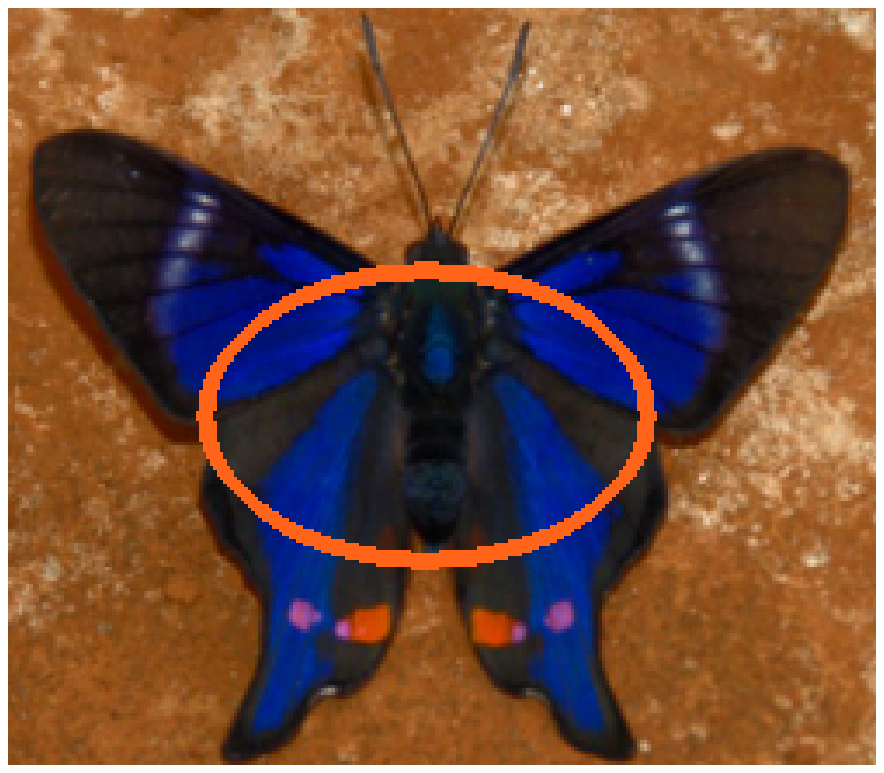,height=0.13\textwidth, width=0.2\textwidth}
\hspace{0.000001\textwidth}
\psfig{figure=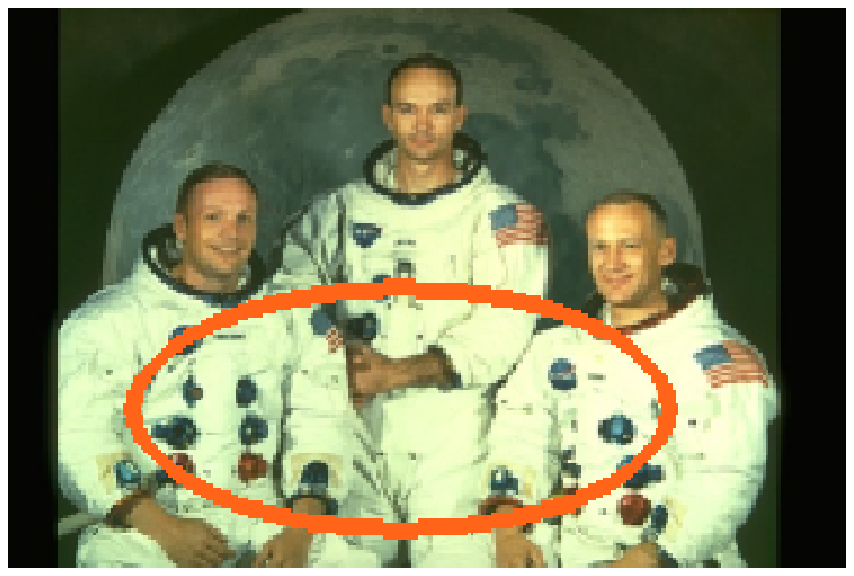,height=0.13\textwidth, width=0.2\textwidth}
\hspace{0.000001\textwidth}
\psfig{figure=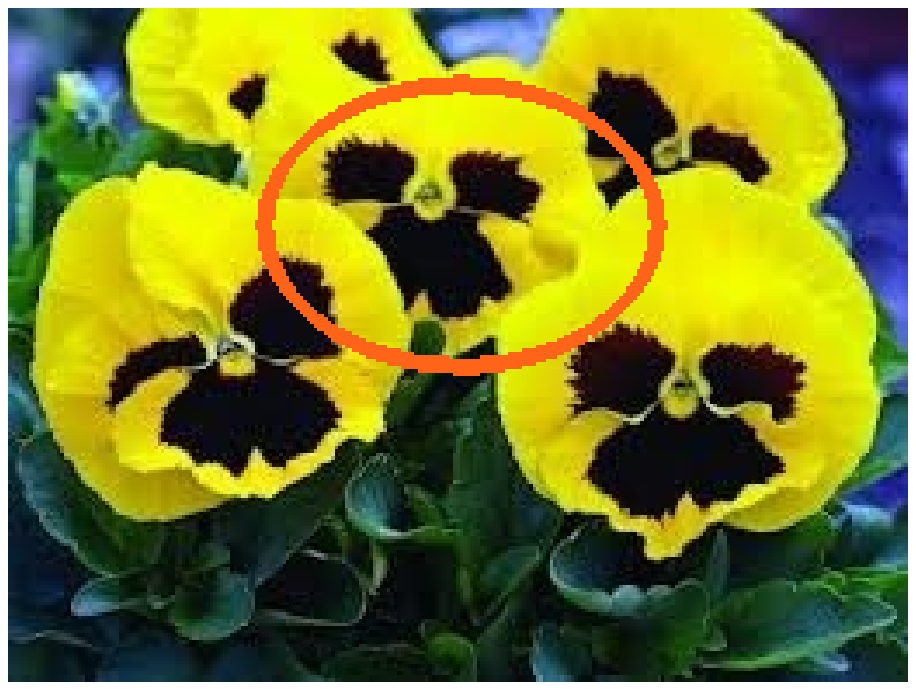,height=0.13\textwidth, width=0.2\textwidth}}
\vspace{0.000001\textwidth}
\centerline{
\psfig{figure=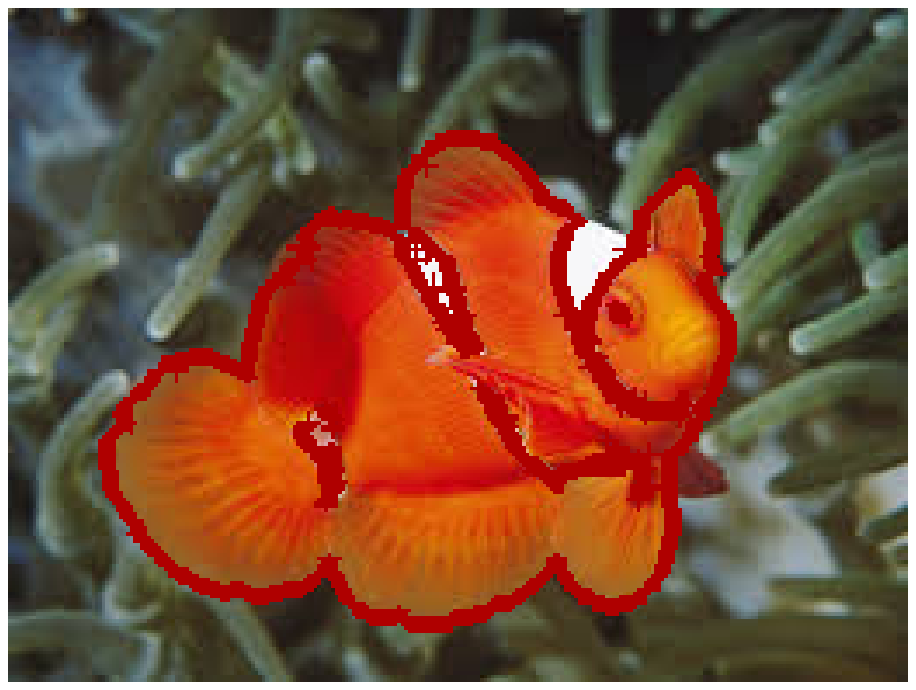,height=0.13\textwidth, width=0.2\textwidth}
\hspace{0.000001\textwidth}
\psfig{figure=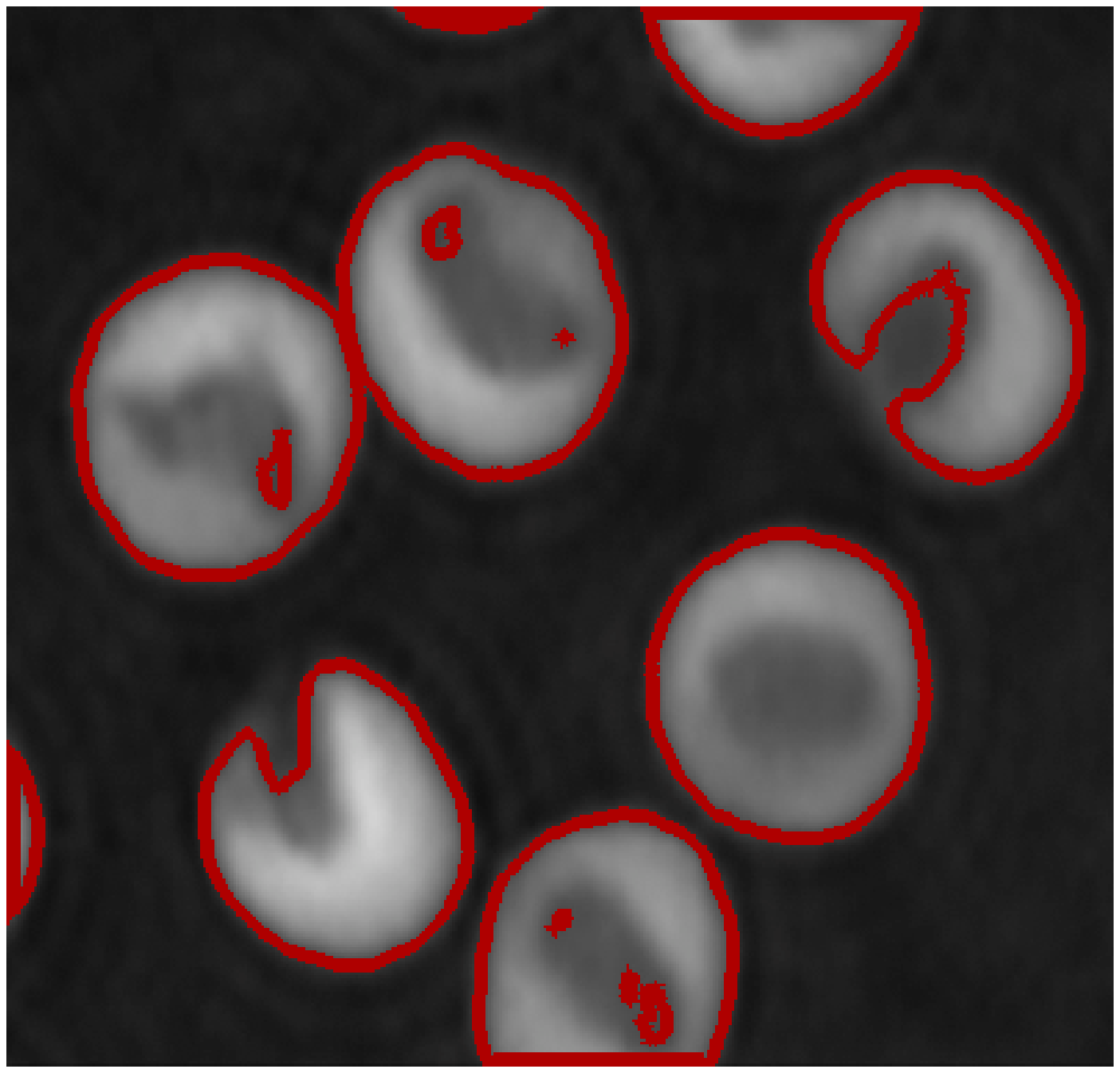,height=0.13\textwidth, width=0.2\textwidth}
\hspace{0.000001\textwidth}
\psfig{figure=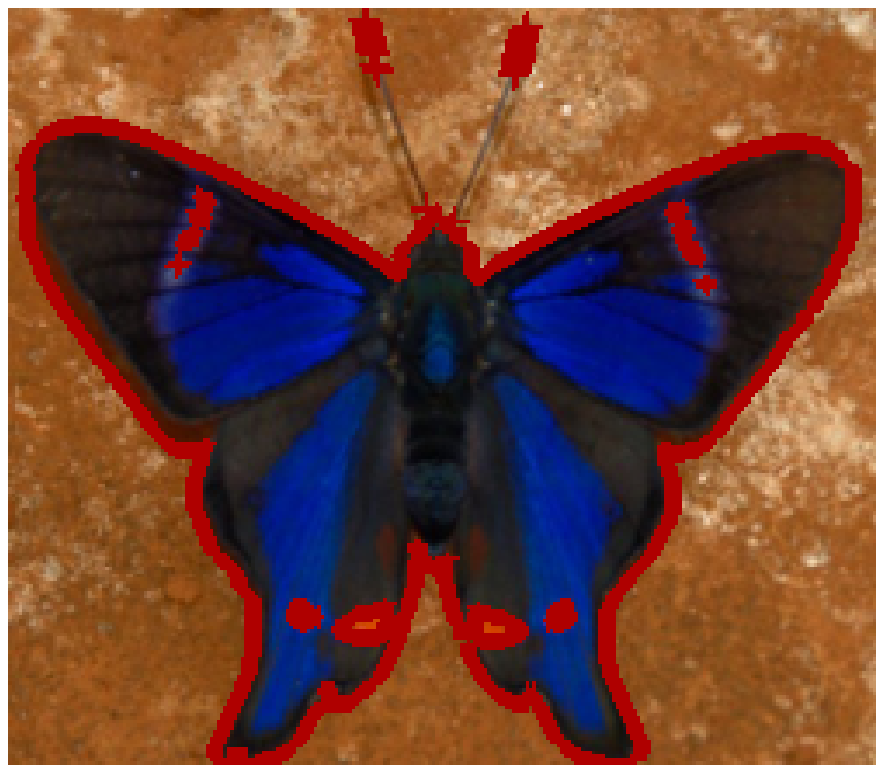,height=0.13\textwidth, width=0.2\textwidth}
\hspace{0.000001\textwidth}
\psfig{figure=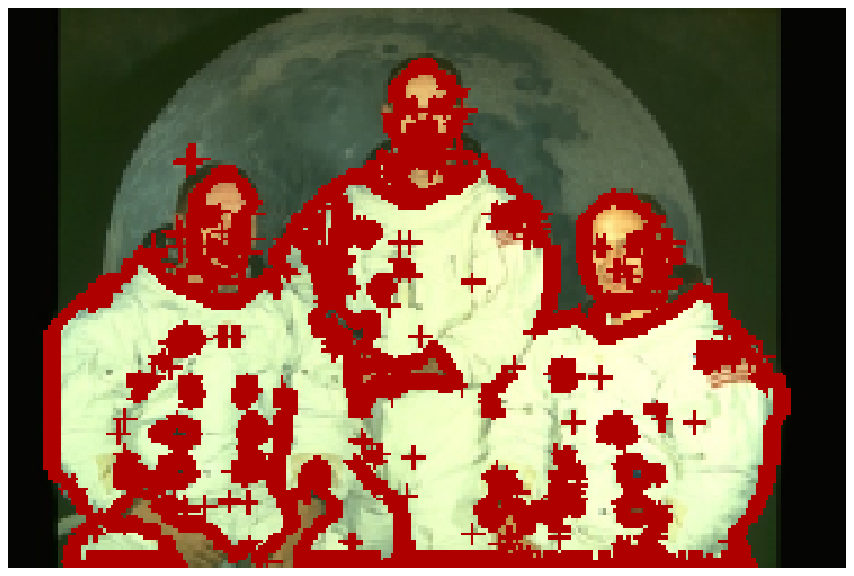,height=0.13\textwidth, width=0.2\textwidth}
\hspace{0.000001\textwidth}
\psfig{figure=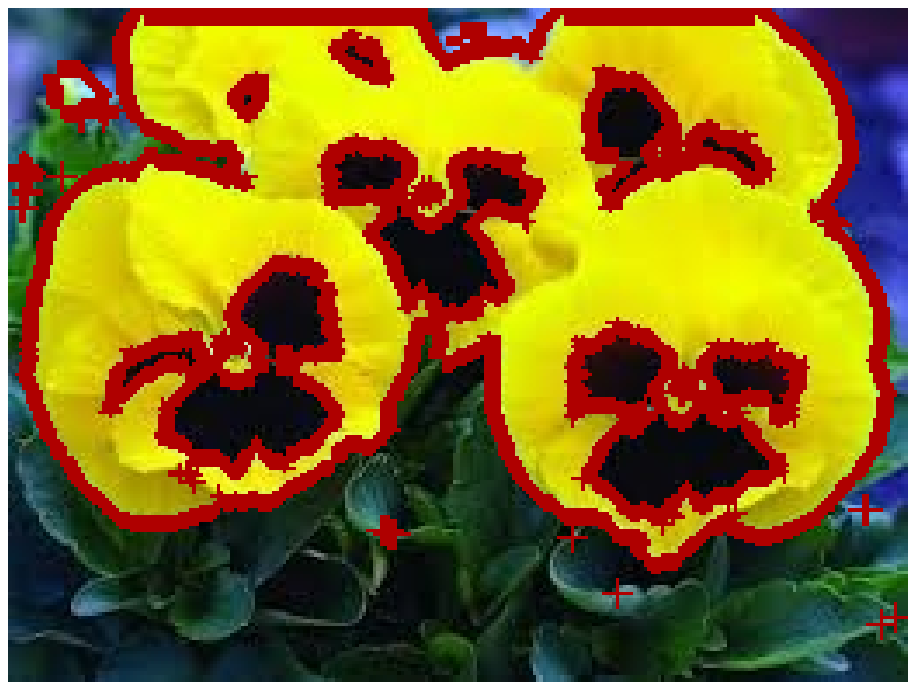,height=0.13\textwidth, width=0.2\textwidth}}
\vspace{0.000001\textwidth}
\centerline{
\psfig{figure=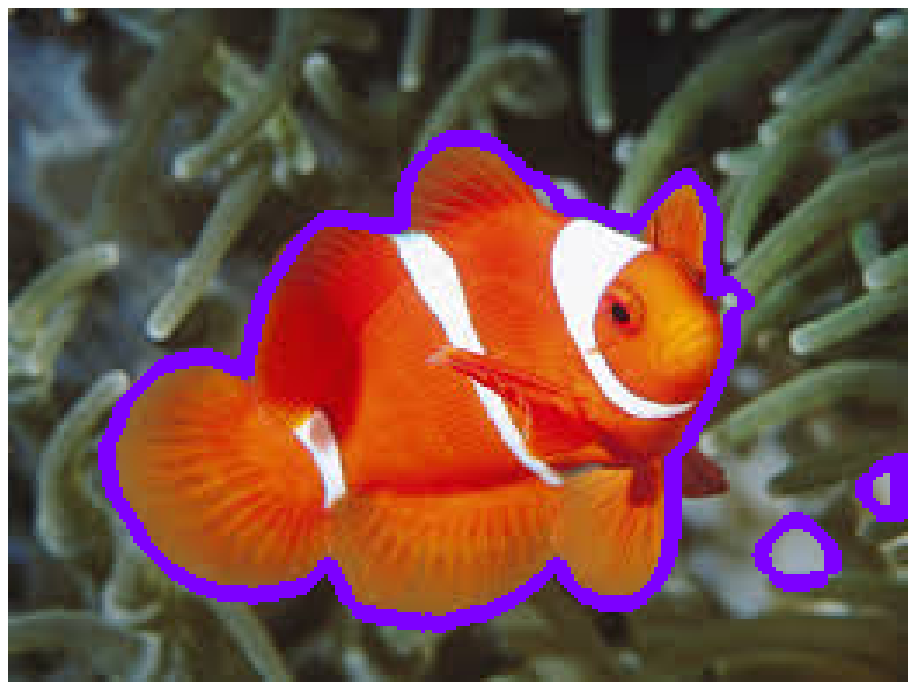,height=0.13\textwidth, width=0.2\textwidth}
\hspace{0.000001\textwidth}
\psfig{figure=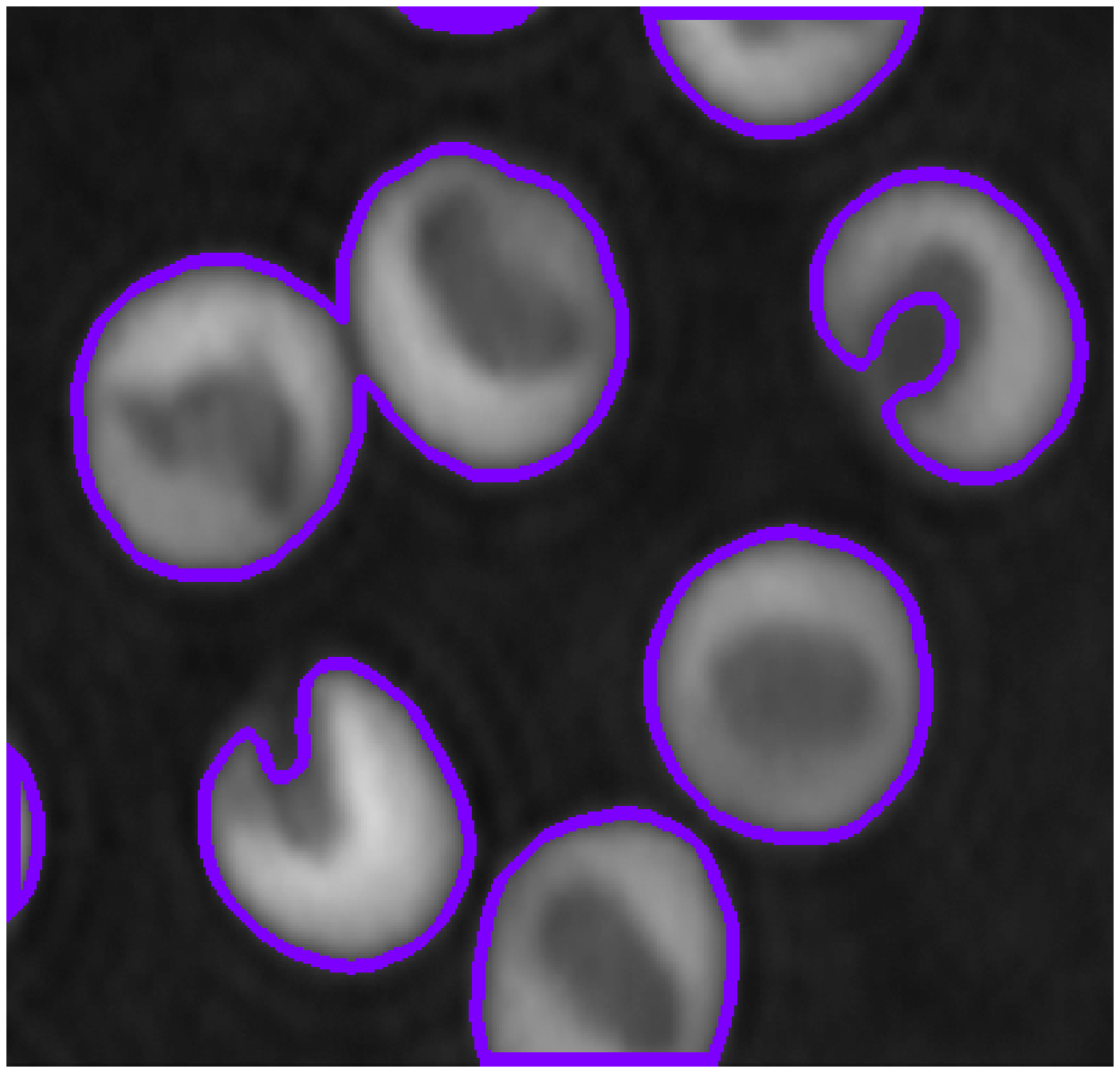,height=0.13\textwidth, width=0.2\textwidth}
\hspace{0.000001\textwidth}
\psfig{figure=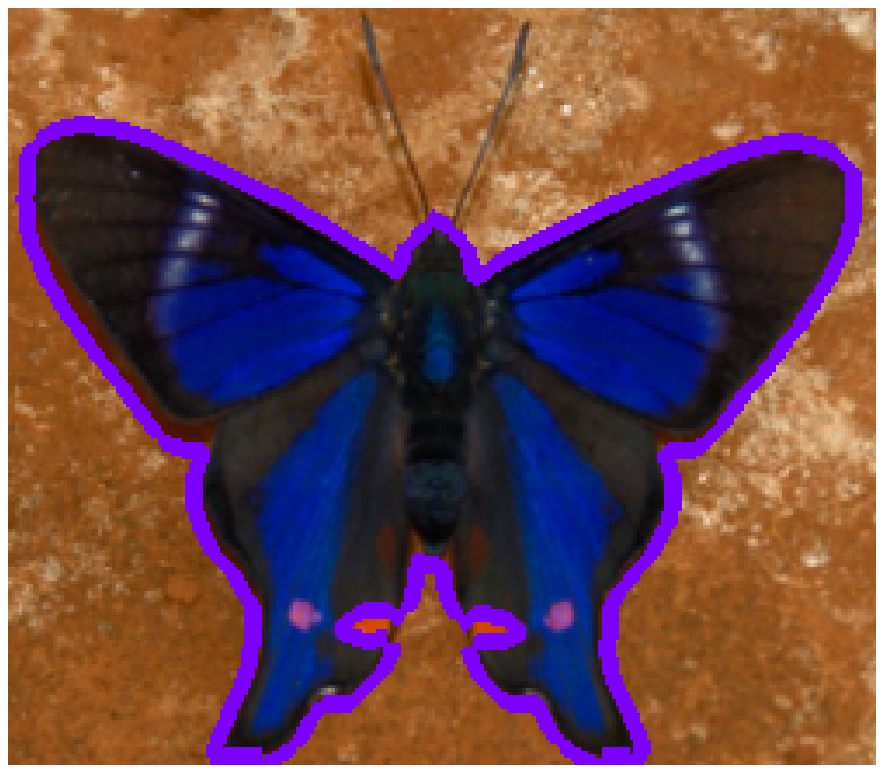,height=0.13\textwidth, width=0.2\textwidth}
\hspace{0.000001\textwidth}
\psfig{figure=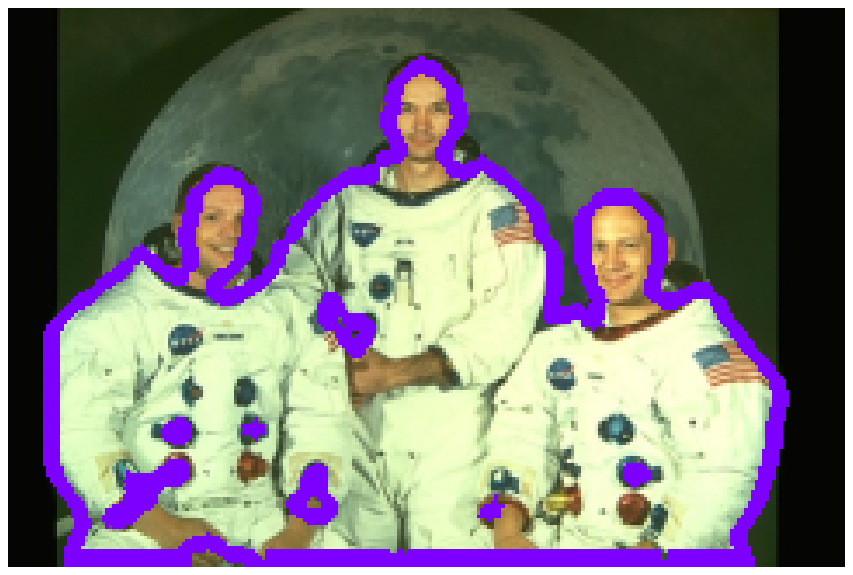,height=0.13\textwidth, width=0.2\textwidth}
\hspace{0.000001\textwidth}
\psfig{figure=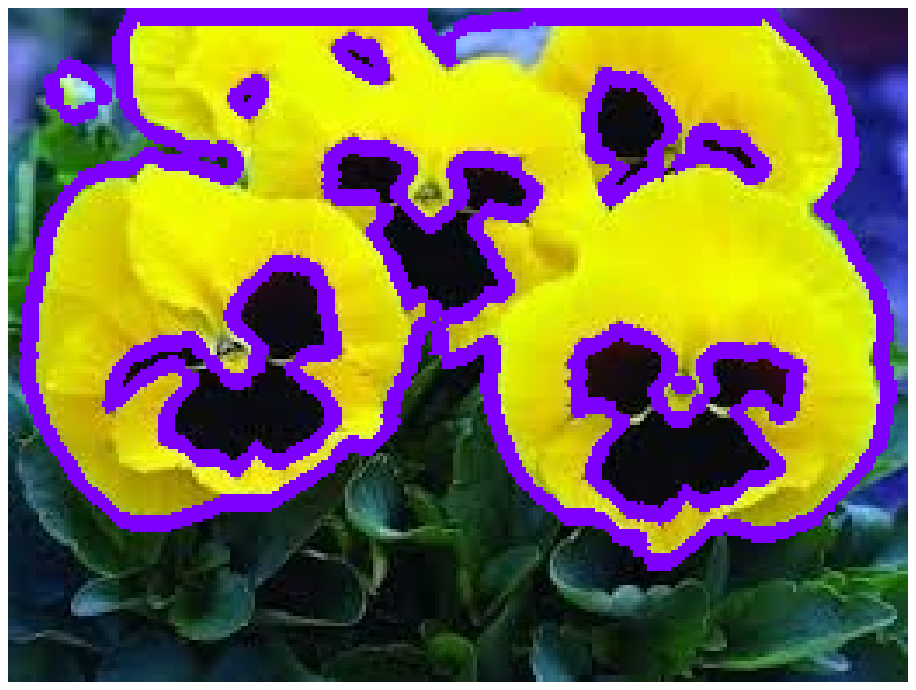,height=0.13\textwidth, width=0.2\textwidth}}
\vspace{0.000001\textwidth}
\centerline{
\psfig{figure=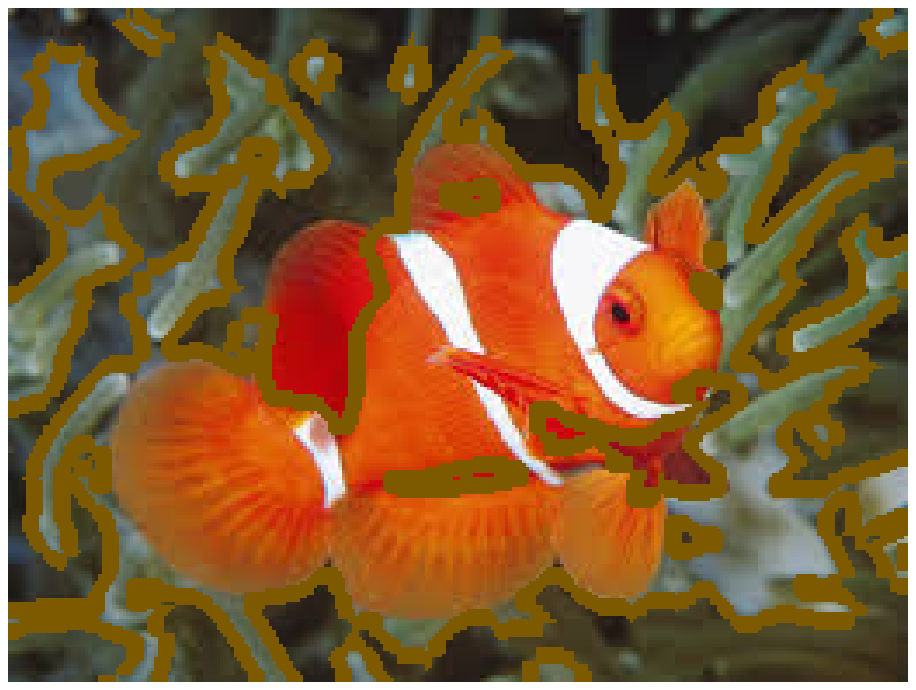,height=0.13\textwidth, width=0.2\textwidth}
\hspace{0.000001\textwidth}
\psfig{figure=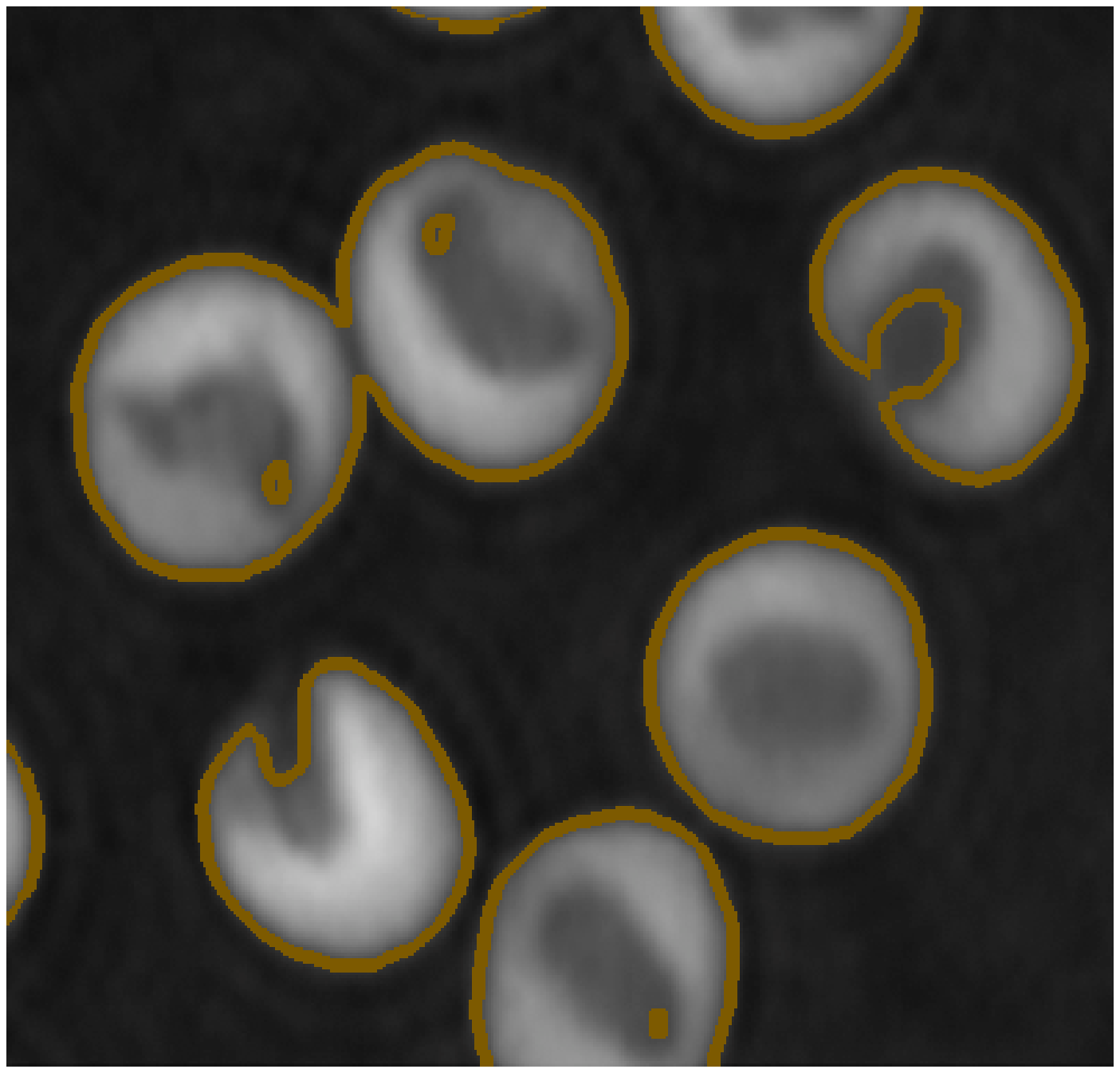,height=0.13\textwidth, width=0.2\textwidth}
\hspace{0.000001\textwidth}
\psfig{figure=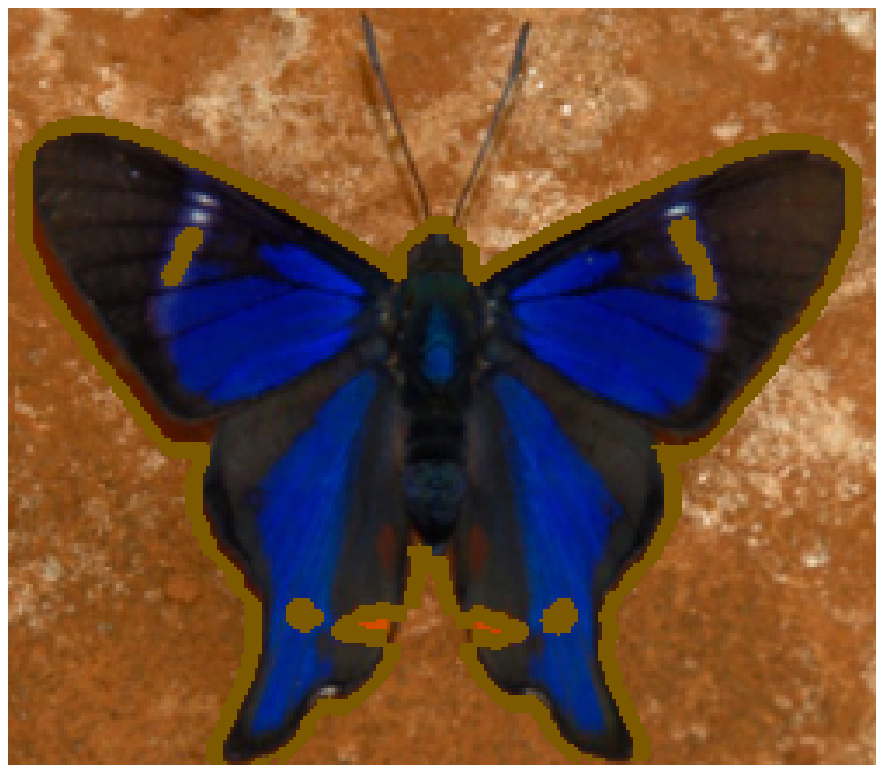,height=0.13\textwidth, width=0.2\textwidth}
\hspace{0.000001\textwidth}
\psfig{figure=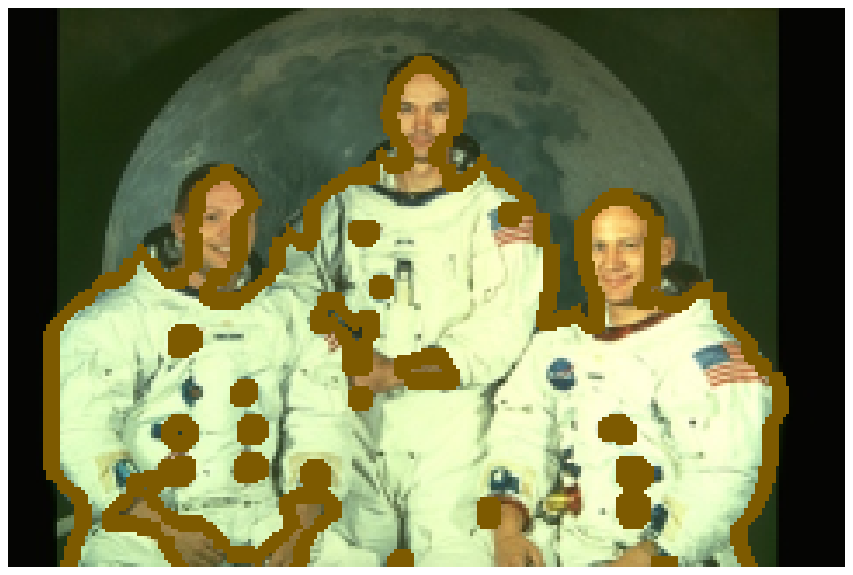,height=0.13\textwidth, width=0.2\textwidth}
\hspace{0.000001\textwidth}
\psfig{figure=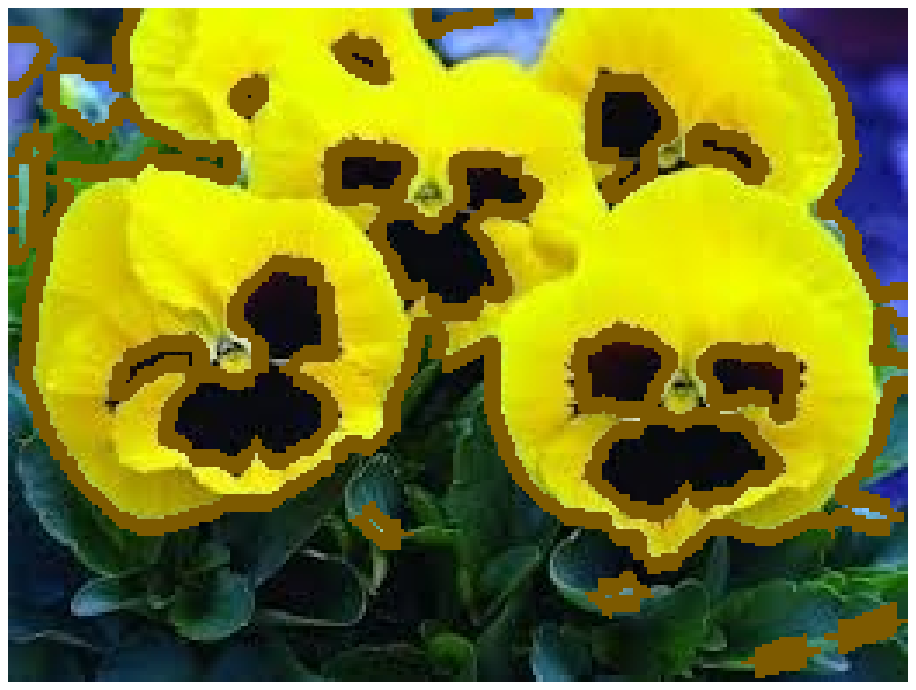,height=0.13\textwidth, width=0.2\textwidth}}
\vspace{0.000001\textwidth}
\centerline{
\psfig{figure=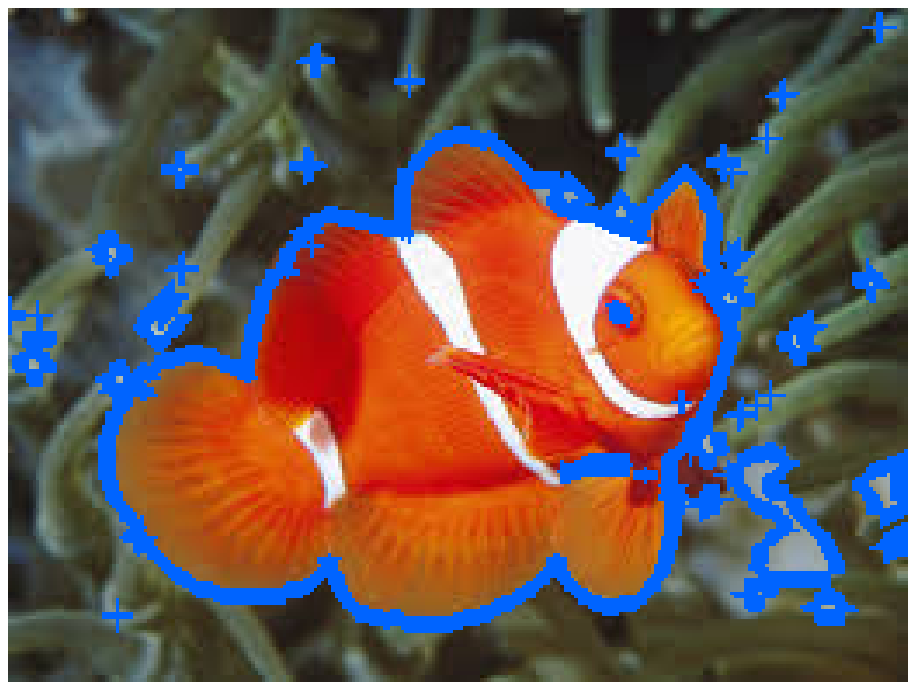,height=0.13\textwidth, width=0.2\textwidth}
\hspace{0.000001\textwidth}
\psfig{figure=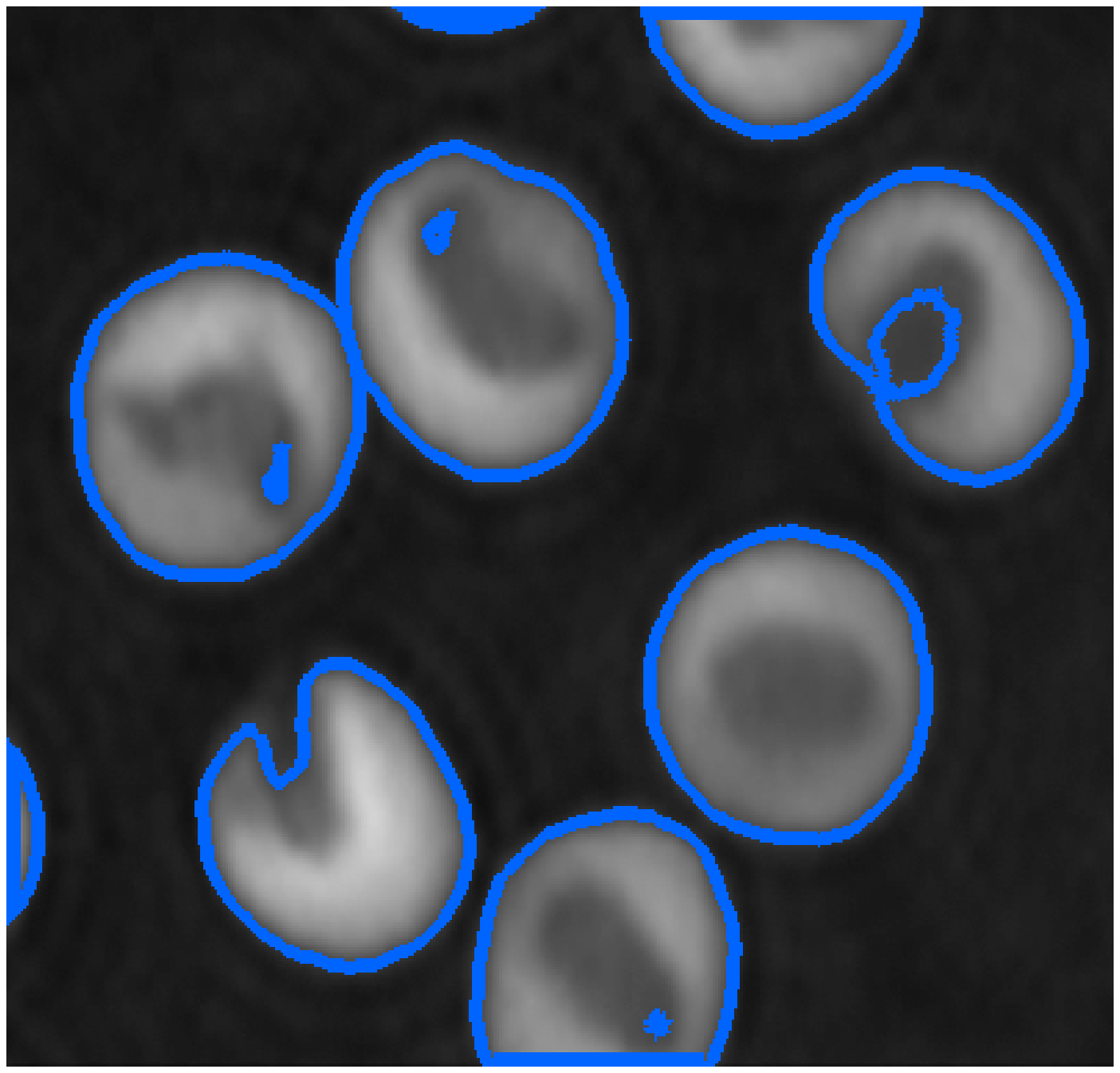,height=0.13\textwidth, width=0.2\textwidth}
\hspace{0.000001\textwidth}
\psfig{figure=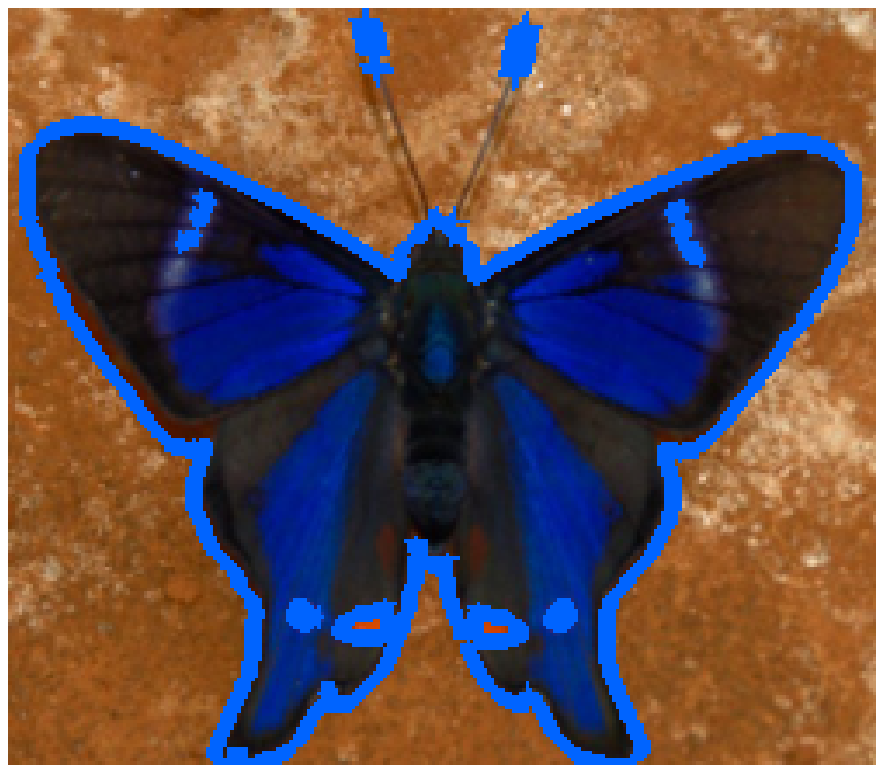,height=0.13\textwidth, width=0.2\textwidth}
\hspace{0.000001\textwidth}
\psfig{figure=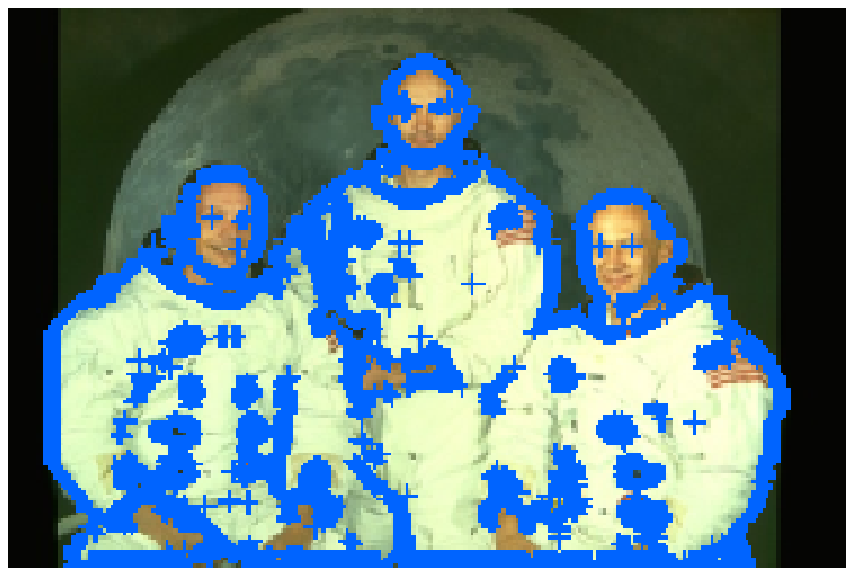,height=0.13\textwidth, width=0.2\textwidth}
\hspace{0.000001\textwidth}
\psfig{figure=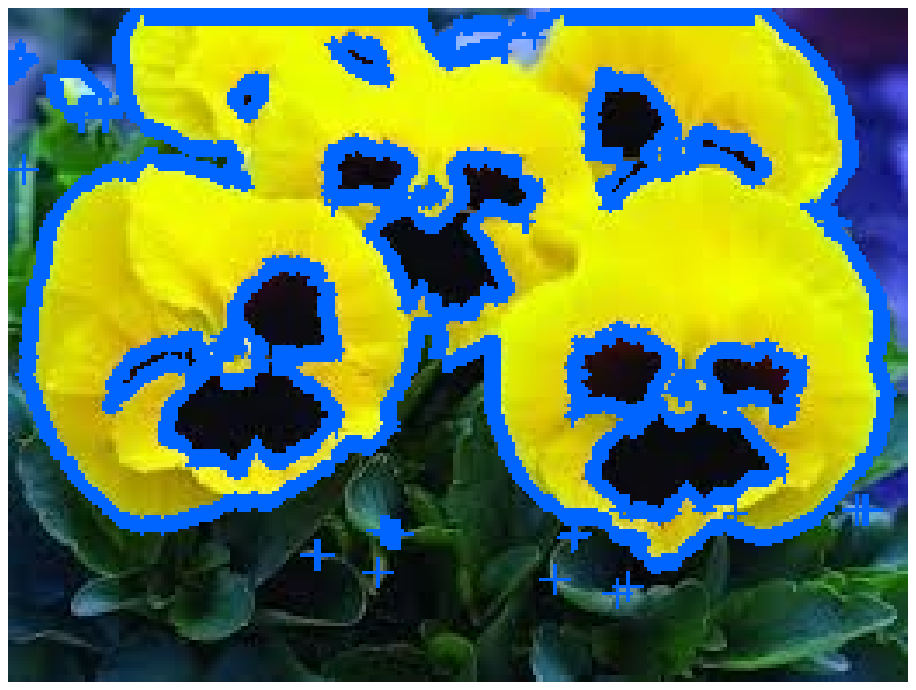,height=0.13\textwidth, width=0.2\textwidth}}
\vspace{0.000001\textwidth}
\centerline{
\psfig{figure=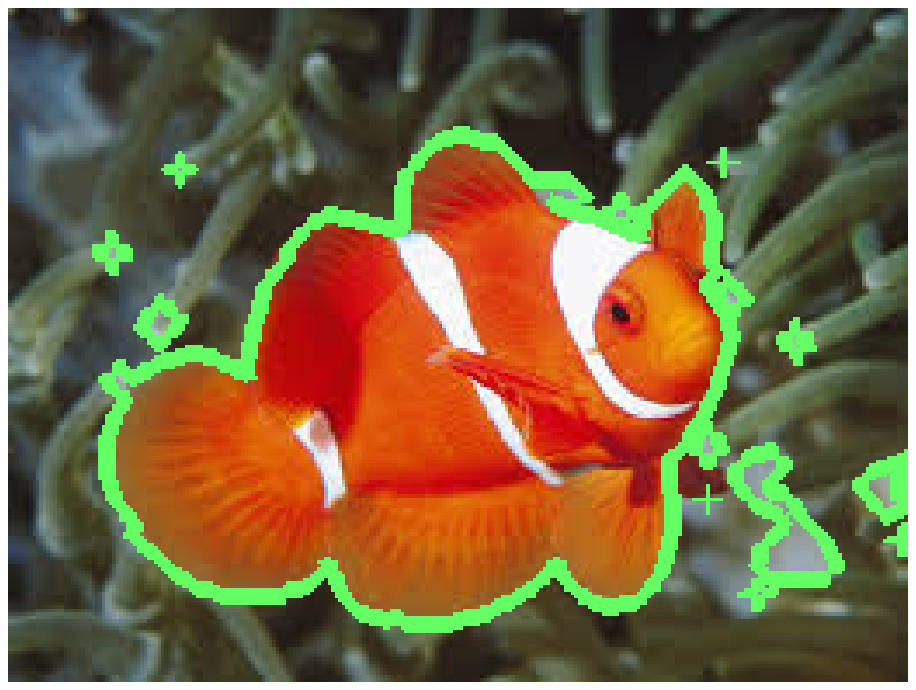,height=0.13\textwidth, width=0.2\textwidth}
\hspace{0.000001\textwidth}
\psfig{figure=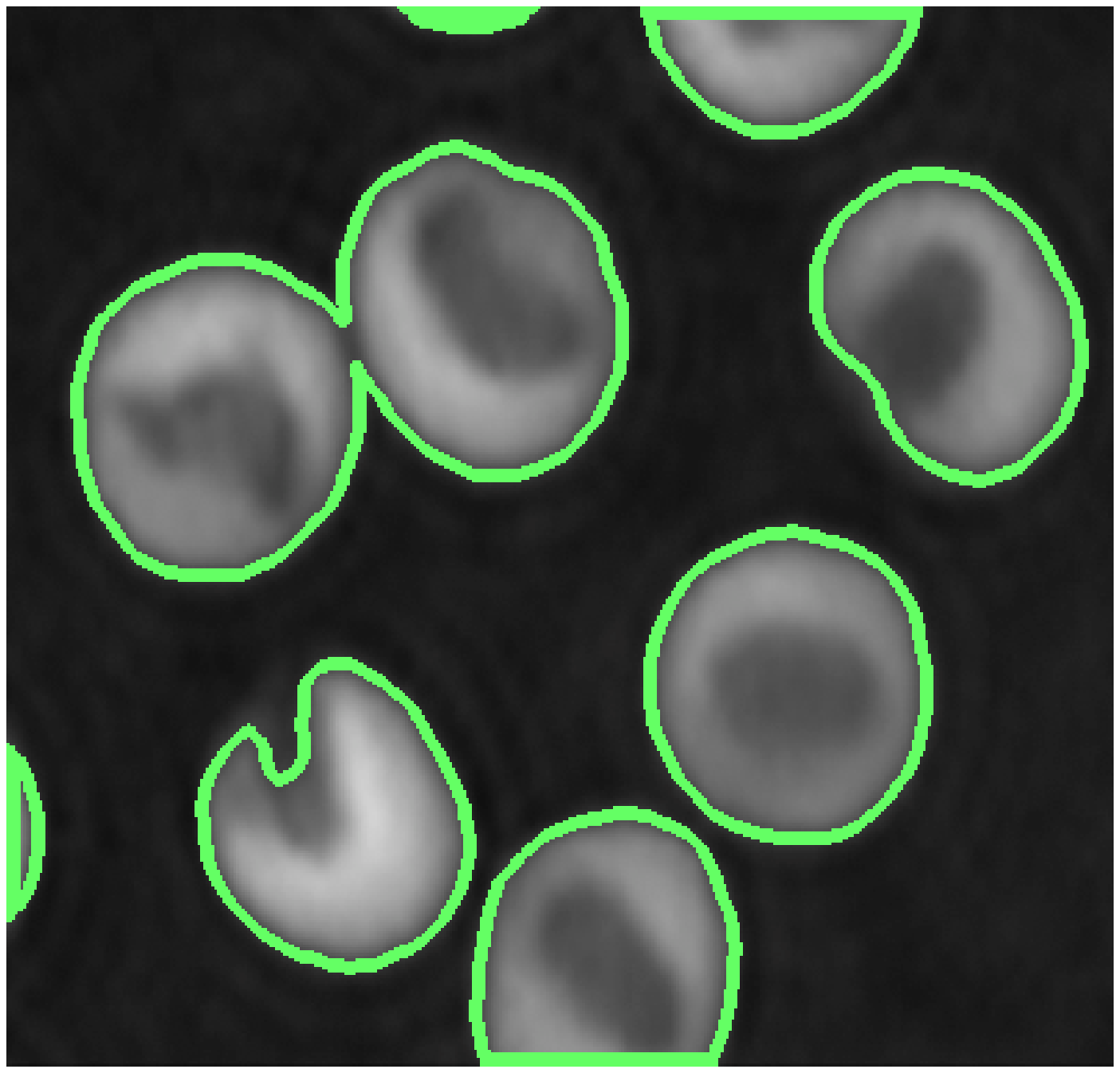,height=0.13\textwidth, width=0.2\textwidth}
\hspace{0.000001\textwidth}
\psfig{figure=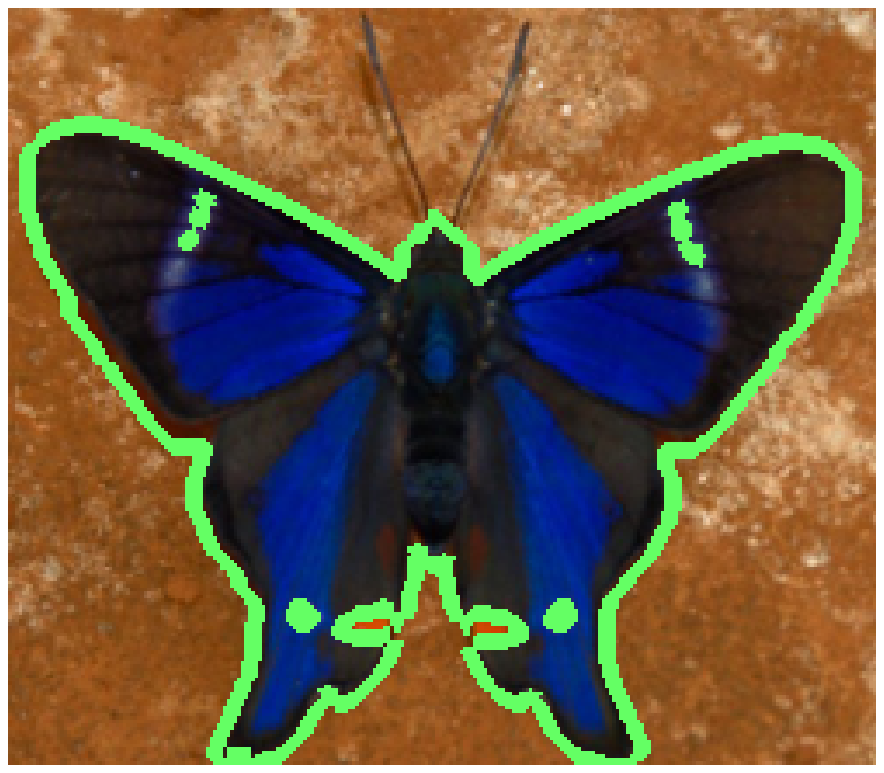,height=0.13\textwidth, width=0.2\textwidth}
\hspace{0.000001\textwidth}
\psfig{figure=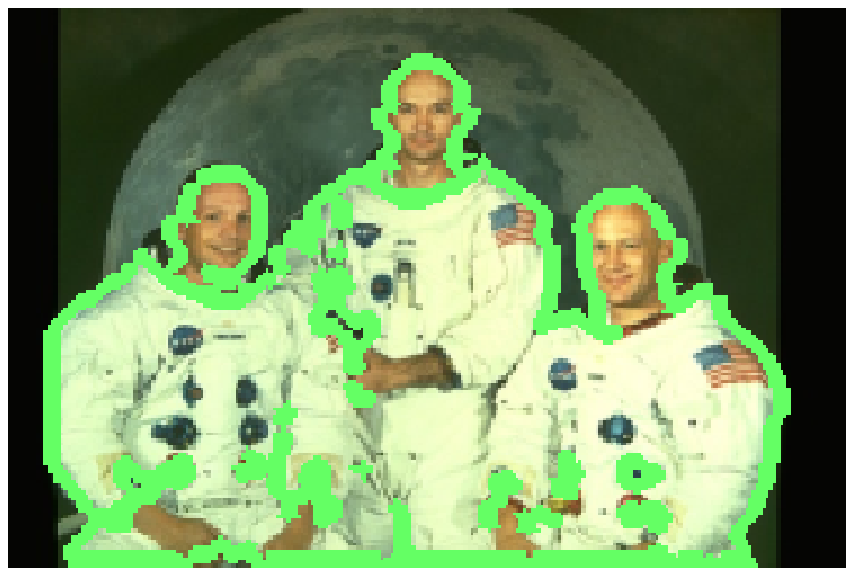,height=0.13\textwidth, width=0.2\textwidth}
\hspace{0.000001\textwidth}
\psfig{figure=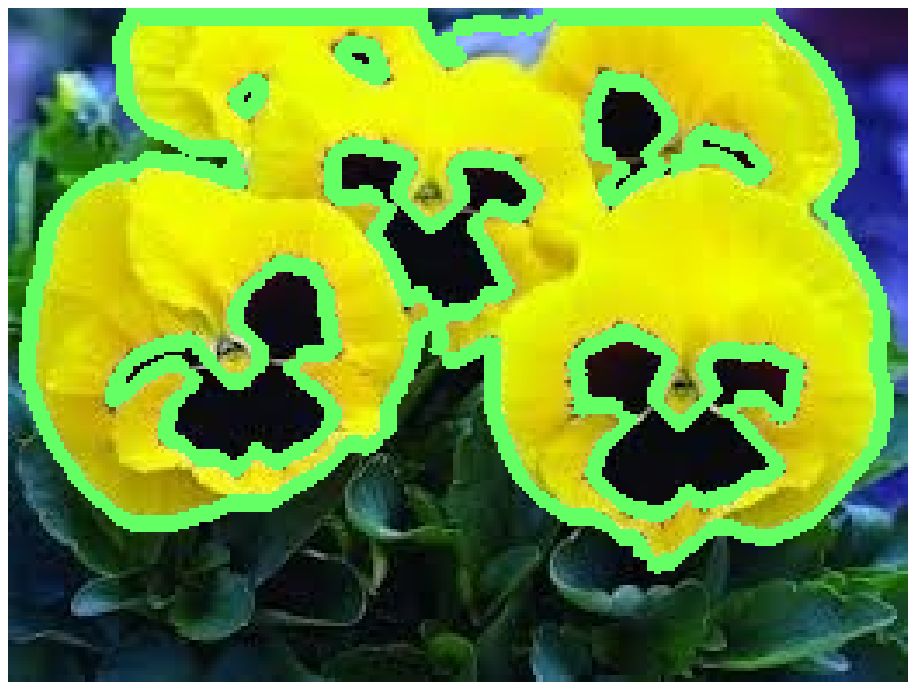,height=0.13\textwidth, width=0.2\textwidth}}
\vspace{0.000001\textwidth}
\centerline{
\psfig{figure=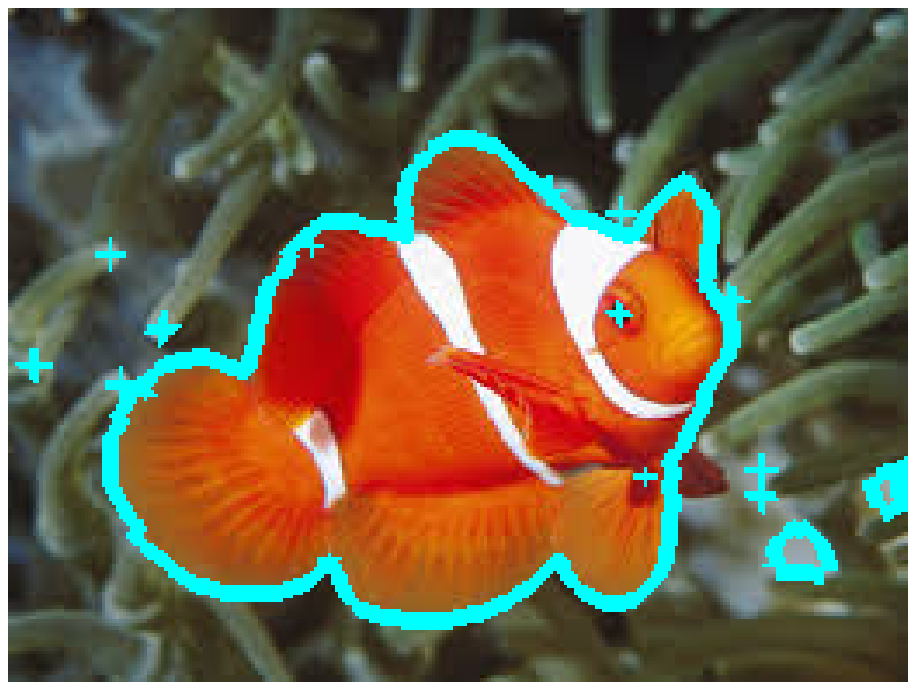,height=0.13\textwidth, width=0.2\textwidth}
\hspace{0.000001\textwidth}
\psfig{figure=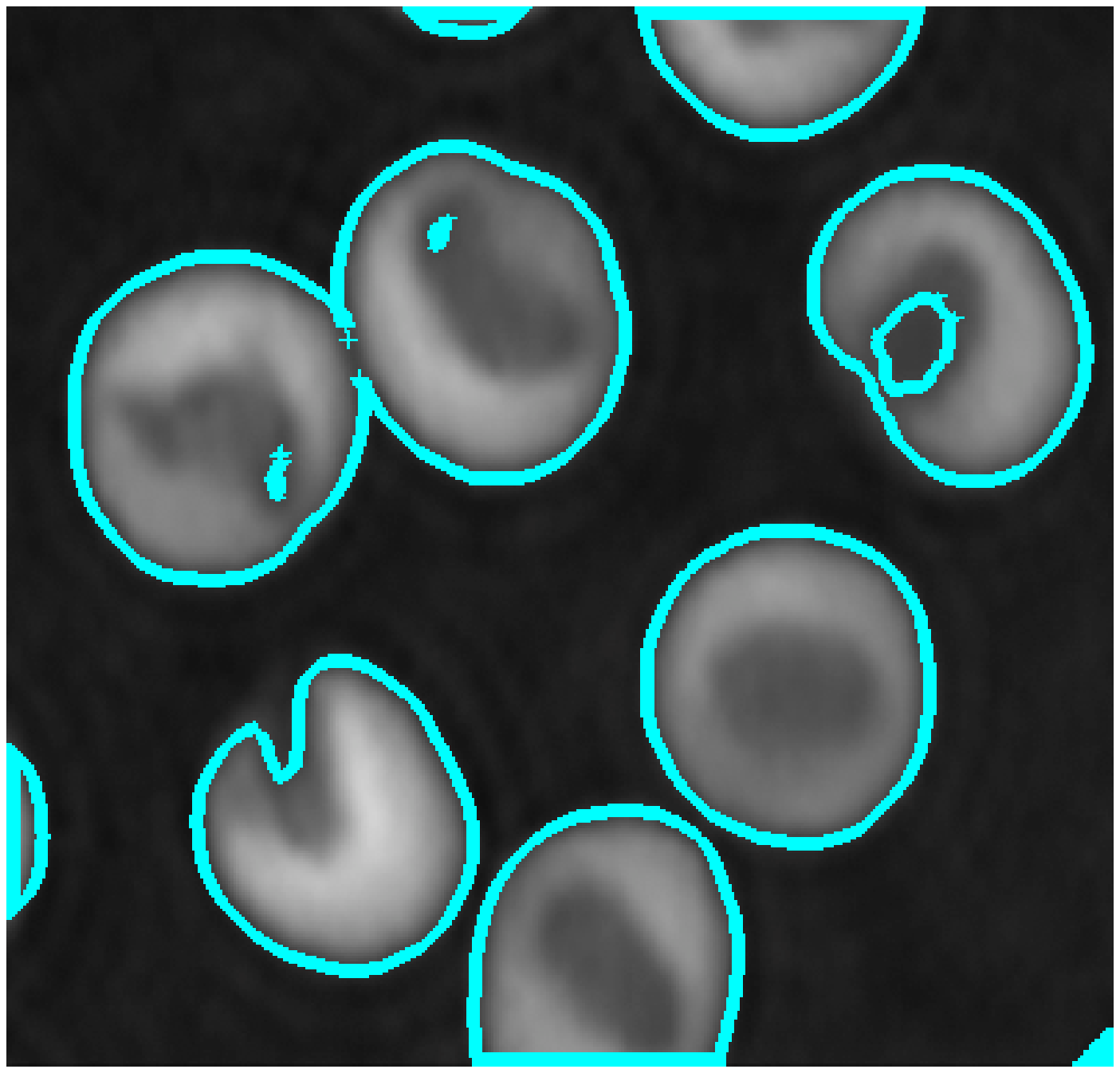,height=0.13\textwidth, width=0.2\textwidth}
\hspace{0.000001\textwidth}
\psfig{figure=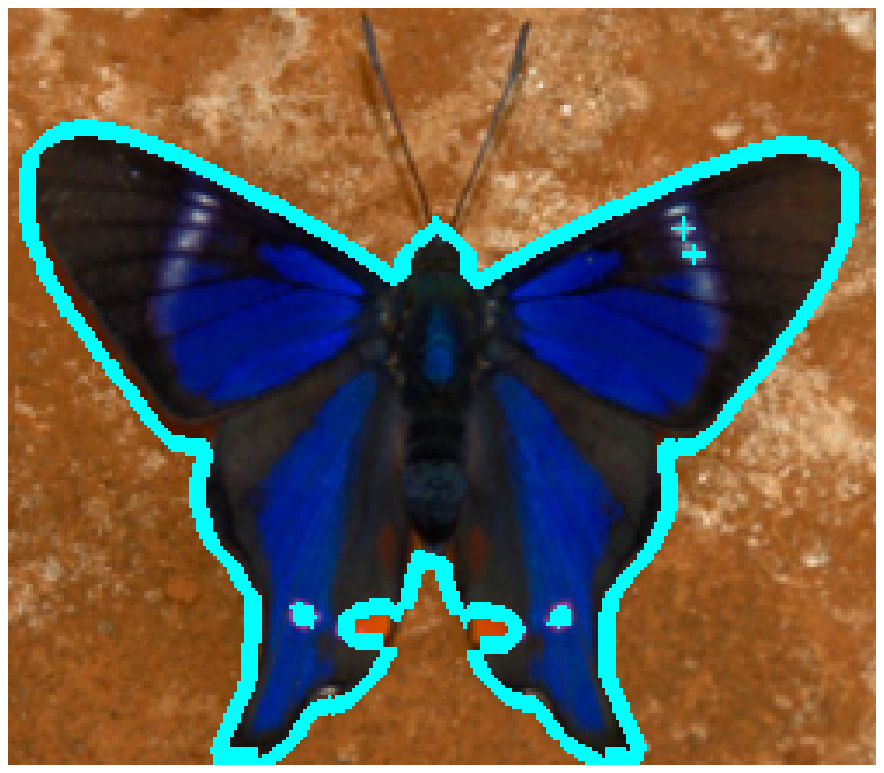,height=0.13\textwidth, width=0.2\textwidth}
\hspace{0.000001\textwidth}
\psfig{figure=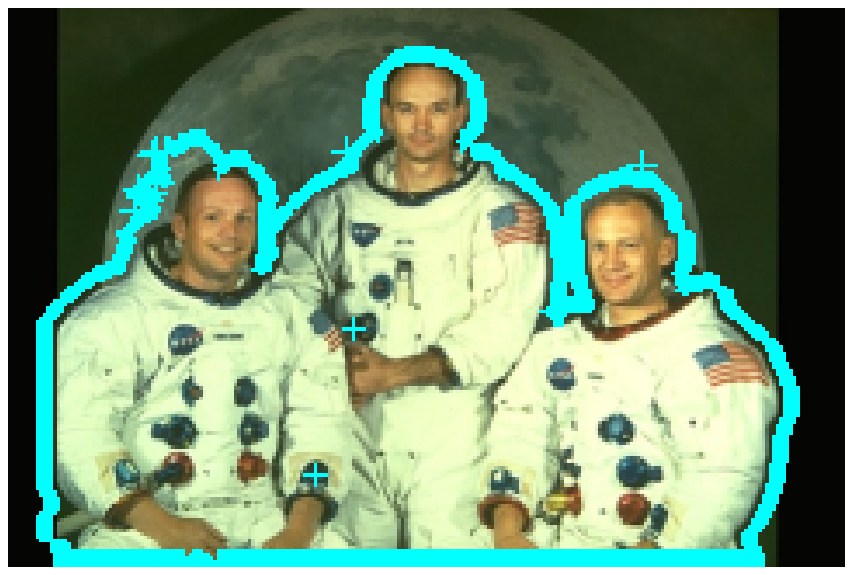,height=0.13\textwidth, width=0.2\textwidth}
\hspace{0.000001\textwidth}
\psfig{figure=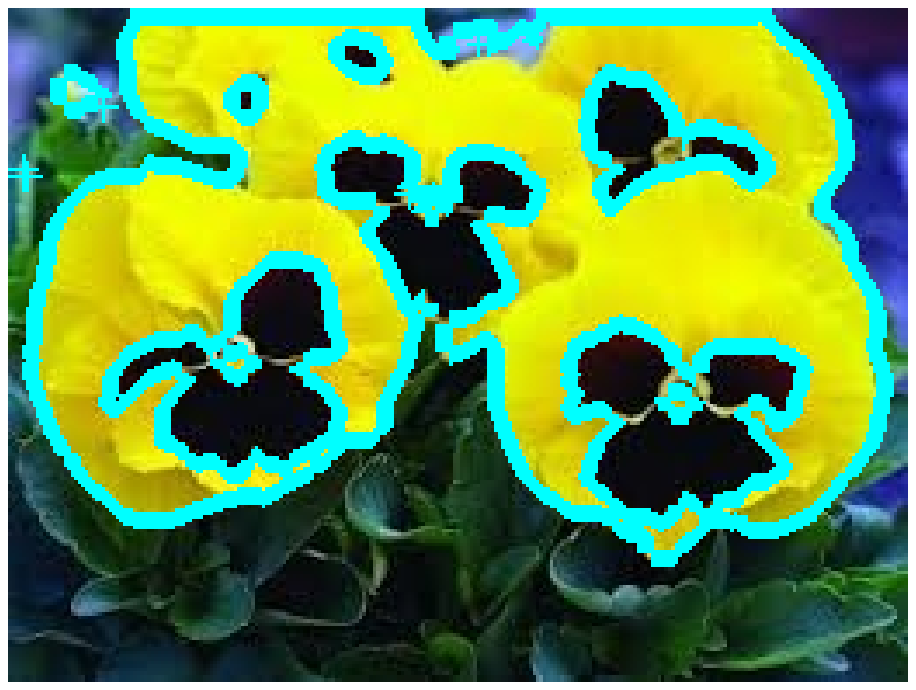,height=0.13\textwidth, width=0.2\textwidth}}
\vspace{0.000001\textwidth}
\centerline{
\psfig{figure=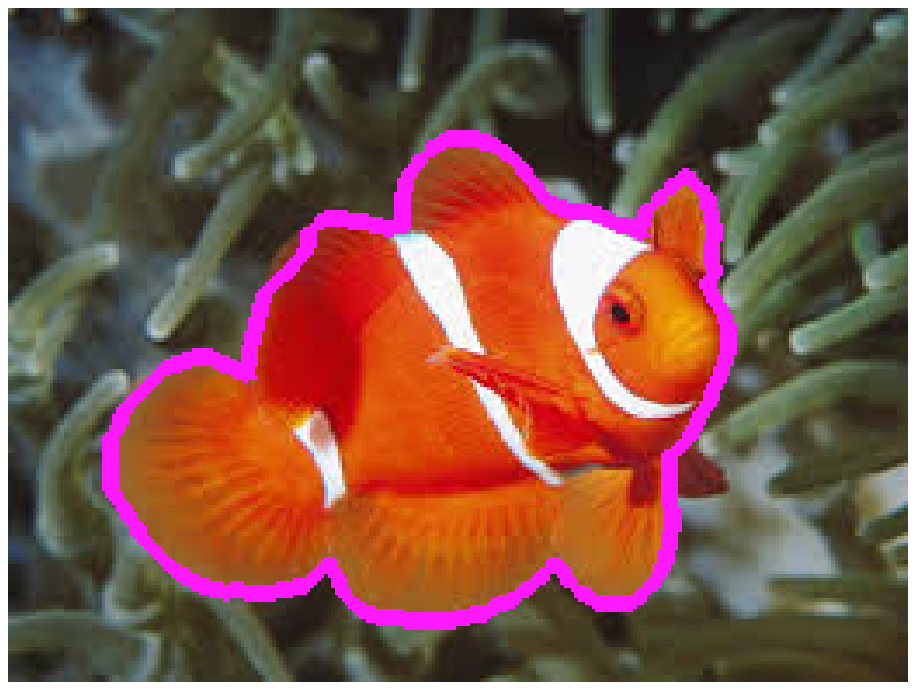,height=0.13\textwidth, width=0.2\textwidth}
\hspace{0.000001\textwidth}
\psfig{figure=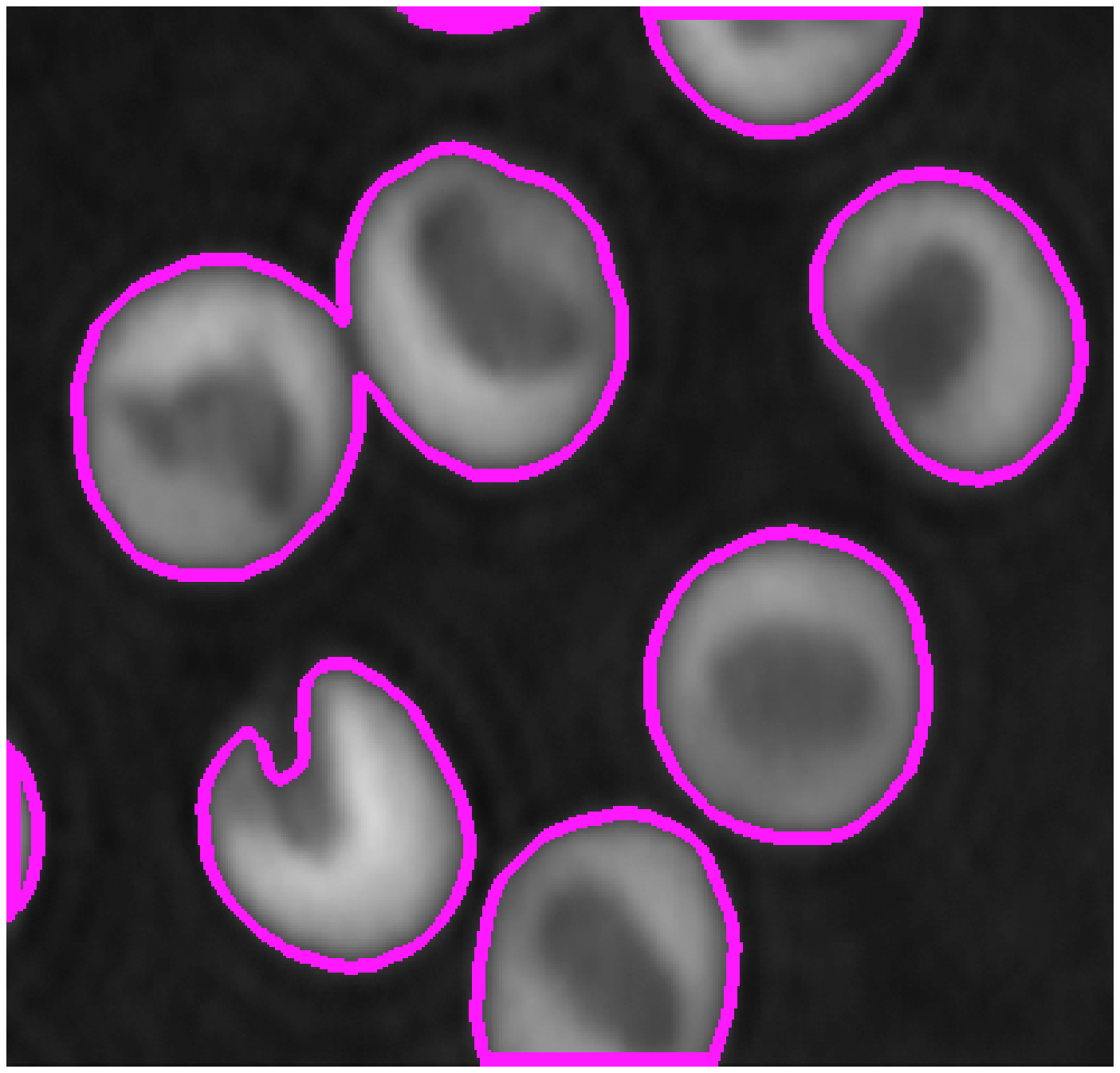,height=0.13\textwidth, width=0.2\textwidth}
\hspace{0.000001\textwidth}
\psfig{figure=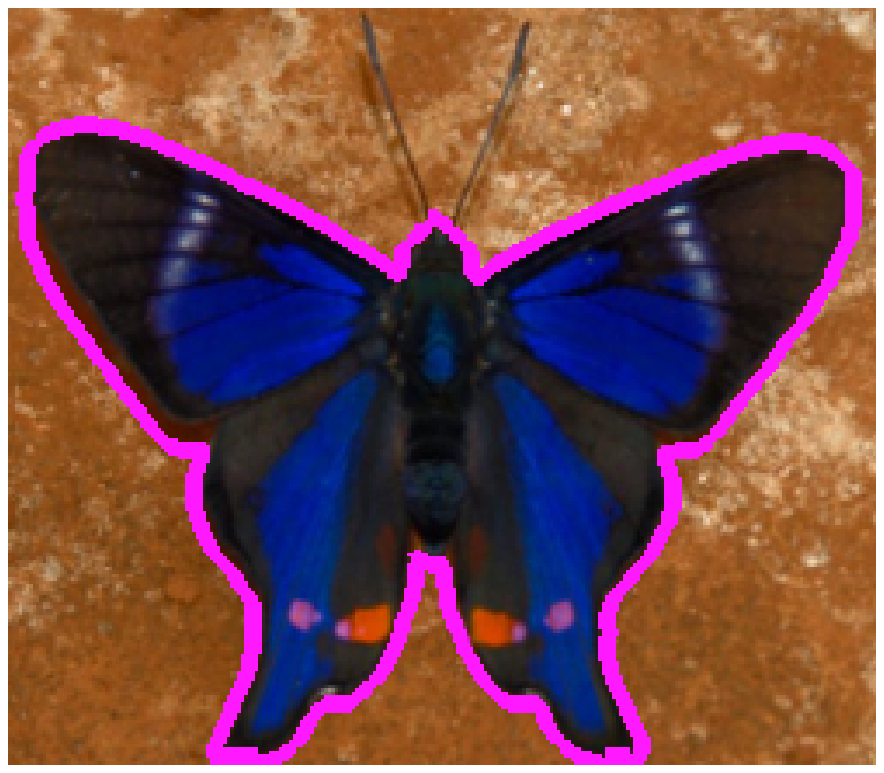,height=0.13\textwidth, width=0.2\textwidth}
\hspace{0.000001\textwidth}
\psfig{figure=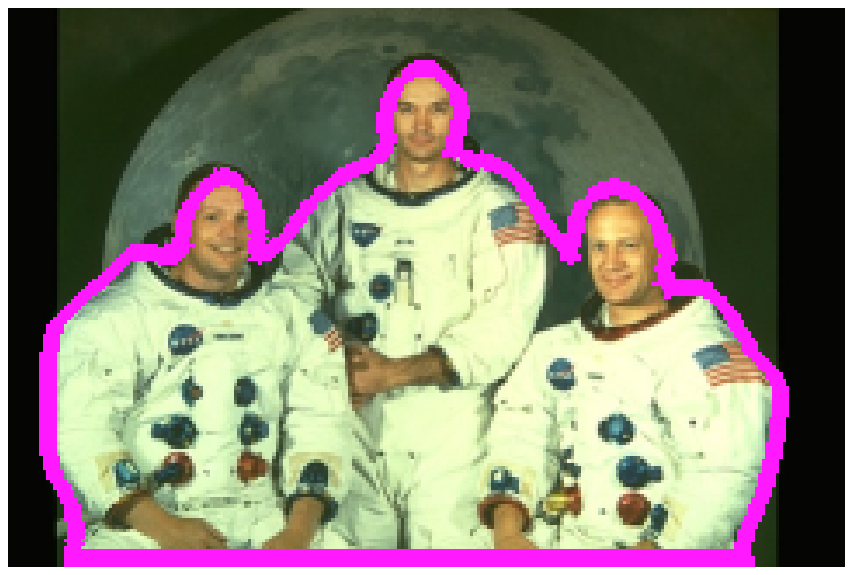,height=0.13\textwidth, width=0.2\textwidth}
\hspace{0.000001\textwidth}
\psfig{figure=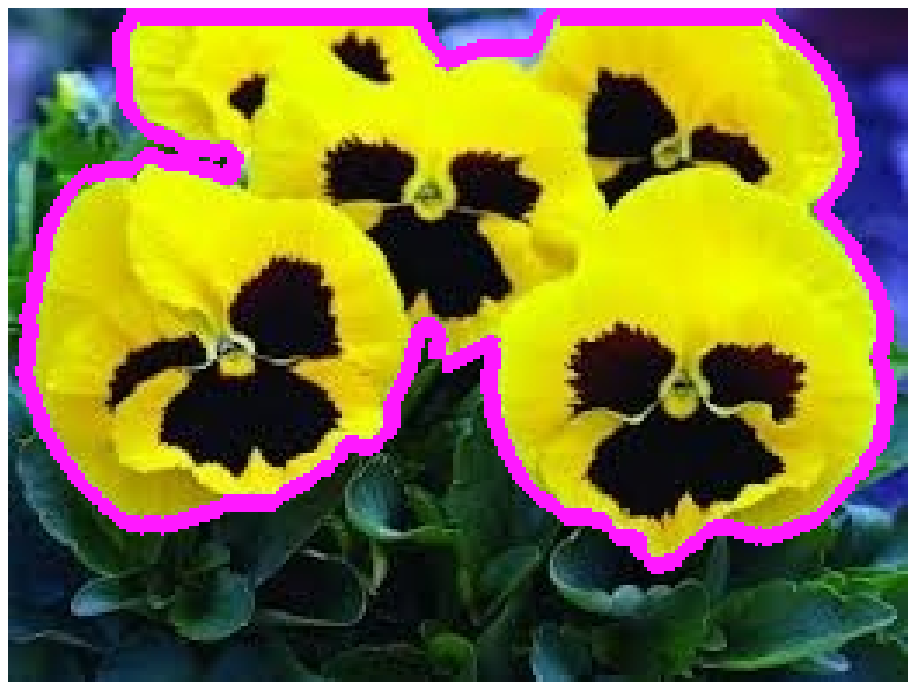,height=0.13\textwidth, width=0.2\textwidth}}
\centerline{(a)\hspace{0.17\textwidth} (b)\hspace{0.17\textwidth} (c)\hspace{0.17\textwidth} (d)\hspace{0.17\textwidth} (e)}
\caption{Visual comparison of segmentation results of (a) Fish, (b) Cell, (c) Butterfly, (d) 323016, (e) Yellow Pansy flower images using various techniques. \textbf{First row:} images with initial contours, \textbf{Second row:} segmentation result obtained by {\sc FEAC}, \textbf{Third row:} segmentation results obtained by {\sc NFACMKM}, \textbf{Fourth row:} segmentation results obtained by {\sc FACGK}, \textbf{Fifth row:} segmentation results obtained by {\sc LPFAC}, \textbf{Sixth row:} segmentation results obtained by {\sc FDFEAC}, \textbf{Seventh row:} segmentation results obtained by {\sc GLFEAC} and \textbf{Eighth row:} segmentation results obtained by {\sc RGLFEAC}.\label{figure_original_seg} }
\end{figure}

Figure~\ref{figure_original_seg} shows few sample segmentation results using various techniques. From this figure, it is observed that due to consideration of global information, {\sc FEAC} fails to properly segment images having local variations: background clutter, intensity in-homogeneity, etc (see second row of this figure). Due to incorporation of kernel metric into the energy function, both {\sc NFACMKM} and {\sc FACGK} produce better segmentation than {\sc FEAC}. Kernel function tries to smooth the local image variation and produces better segmentation results. However, they are also fail to properly segment the images with background clutter and high region in-homogeneity, etc. Local image information based techniques: {\sc LPFAC} and {\sc FDFEAC} also produce better results than {\sc FEAC}. Local image information in the energy term makes both {\sc LPFAC} and {\sc FDFEAC} techniques robust against background clutter and region in-homogeneity. However, they are worst than both {\sc NFACMKM} and {\sc FACGK}. Consideration of the global and local image information in both techniques: {\sc GLFEAC} and {\sc RGLFEAC} help to segment images properly. Moreover, {\sc RGLFEAC} segments images better than all other techniques. Table~\ref{table_original} displays the quantitative comparison of them. This table also highlights the similar finding.                

\begin{table*}[htp]
\begin{center}
\begin{tabular}{|l|l|l|l|l|l|l|l|}\hline
Measures & \multicolumn{7}{|l|}{Techniques } \\\cline{2-8}
& {\sc m1} & {\sc m2} & {\sc m3} & {\sc m4} & {\sc m5} & {\sc m6} & {\sc m7}\\\cline{1-8}
Ave. Jacard error & 0.217 & 0.148 & 0.201 & 0.251 & 0.288 & 0.222 & \textcolor[rgb]{0.00,1.00,0.00}{0.068} \\\hline
Ave. F-measure    & 0.876& 0.902 & 0.886& 0.849 &0.796 & 0.867& \textcolor[rgb]{1.00,0.00,0.00}{0.963}  \\\hline
\end{tabular}
\end{center}
\caption{Quantitative comparison among various techniques with respect to average Jacard error and average F-measure over $100$ original images. {\sc m1: feac}, {\sc m2: nfacmkm}, {\sc m3: facgk}, {\sc m4: lpfac}, {\sc m5: fdfeac}, {\sc m6: glfeac} and {\sc m7: rglfeac}. Green colored numeric value indicates least average Jacard error corresponds to the best segmentation result. Whereas, red colored numeric value indicates highest F-measure corresponds to the best segmentation result.\label{table_original}}
\end{table*}

\subsection{Segmentation of Blurred Images}

\begin{figure}
\centerline{
\psfig{figure=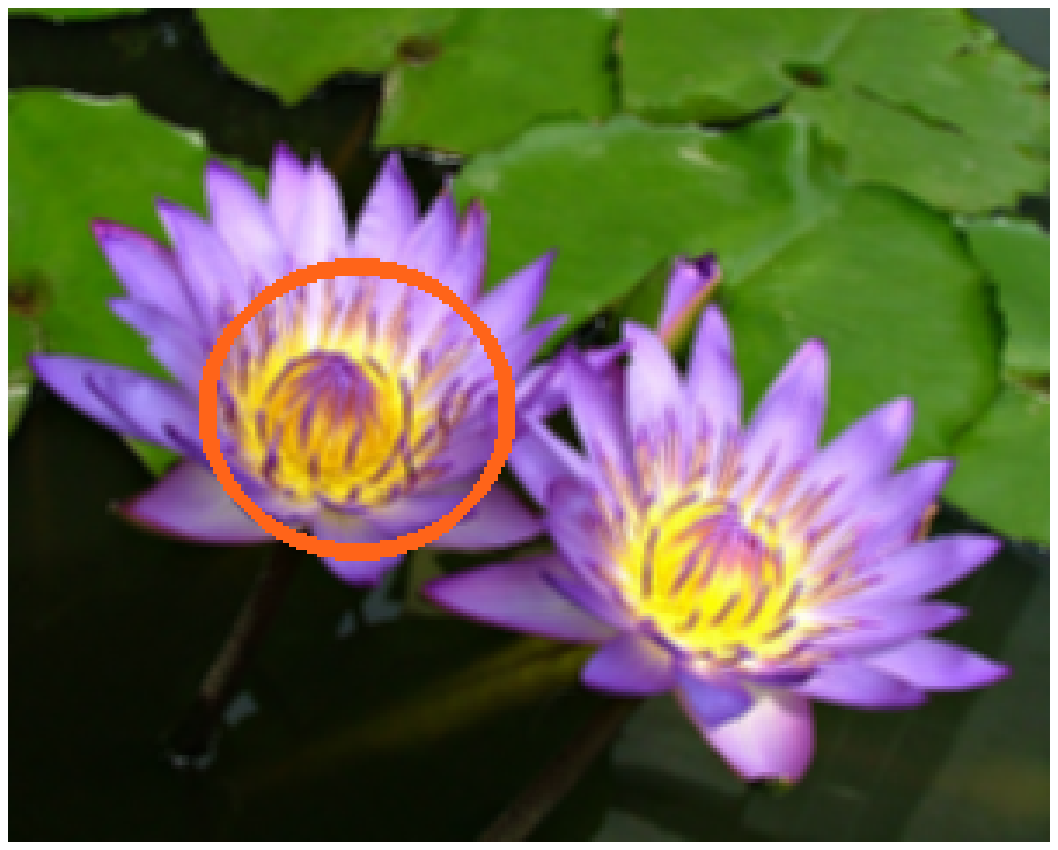,height=0.13\textwidth, width=0.2\textwidth}
\hspace{0.000001\textwidth}
\psfig{figure=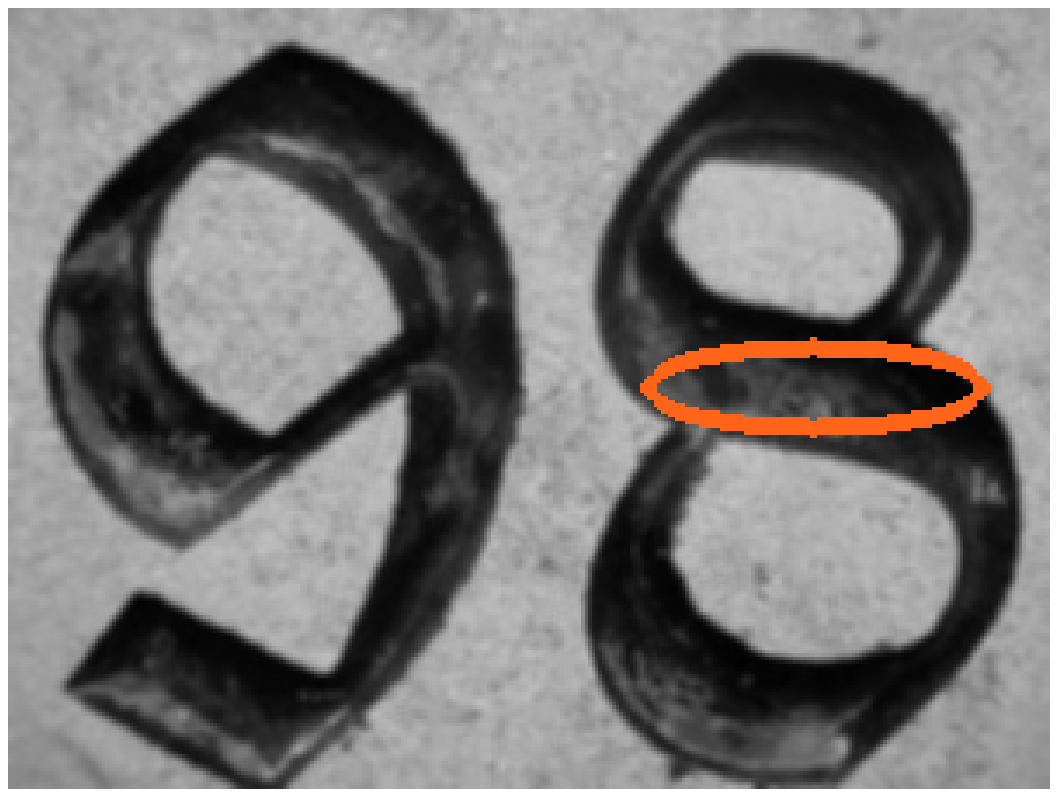,height=0.13\textwidth, width=0.2\textwidth}
\hspace{0.000001\textwidth}
\psfig{figure=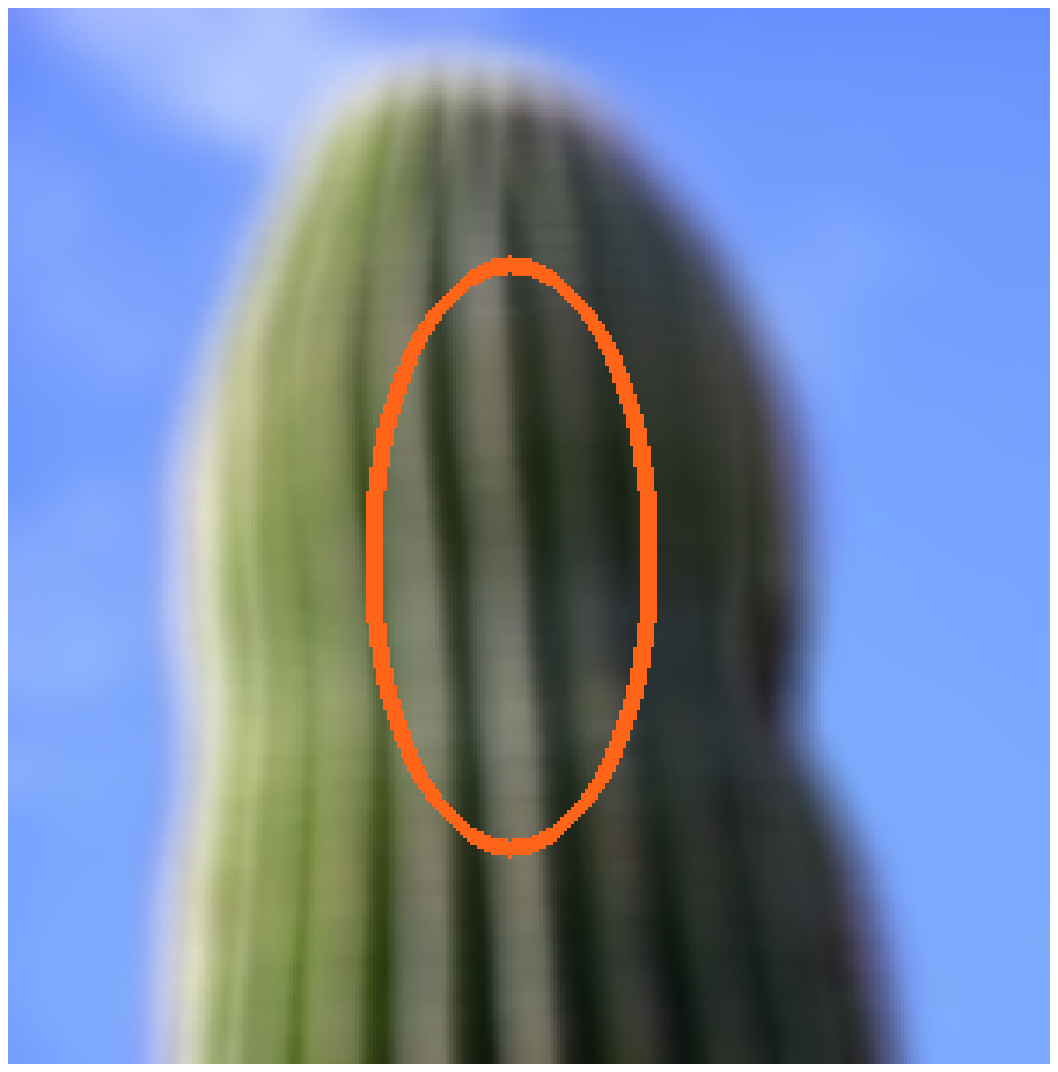,height=0.13\textwidth, width=0.2\textwidth}
\hspace{0.000001\textwidth}
\psfig{figure=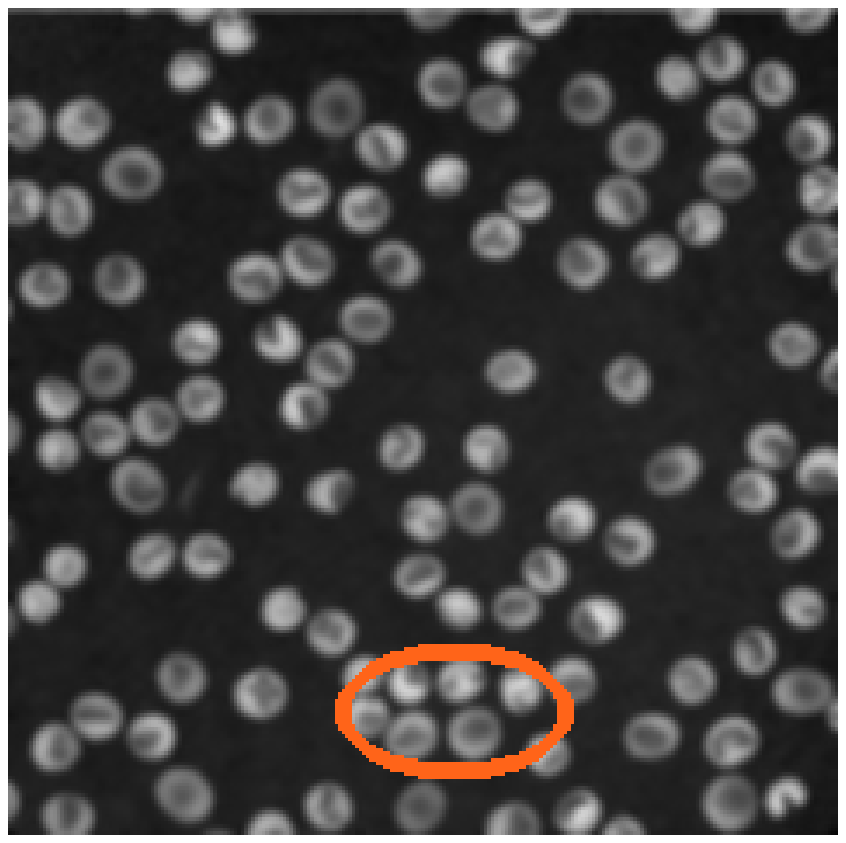,height=0.13\textwidth, width=0.2\textwidth}
\hspace{0.000001\textwidth}
\psfig{figure=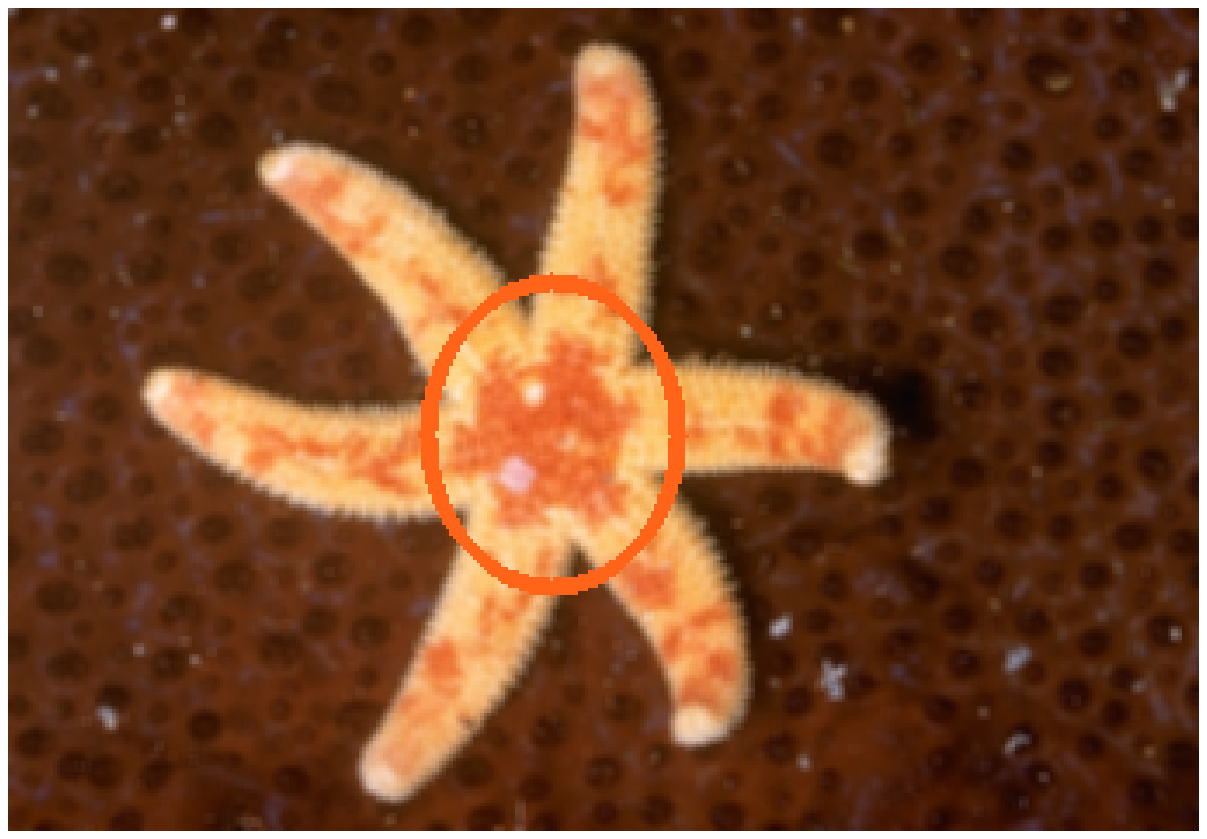,height=0.13\textwidth, width=0.2\textwidth}}
\vspace{0.000001\textwidth}
\centerline{
\psfig{figure=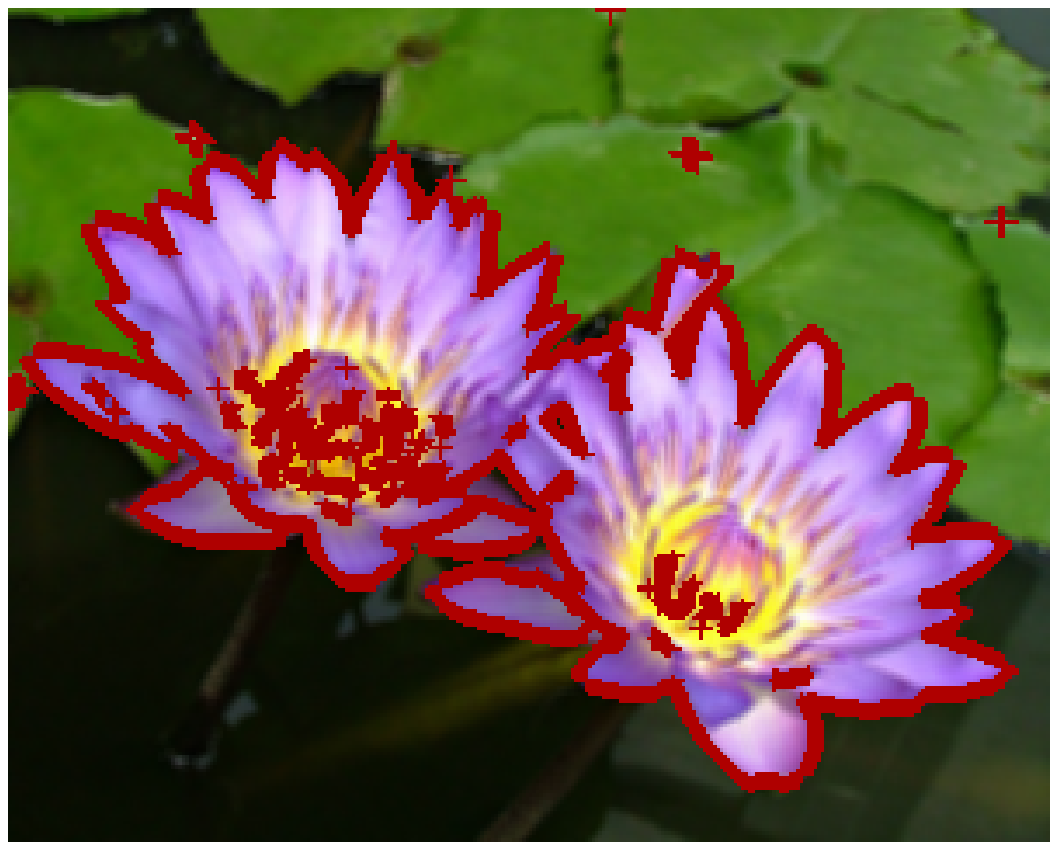,height=0.13\textwidth, width=0.2\textwidth}
\hspace{0.000001\textwidth}
\psfig{figure=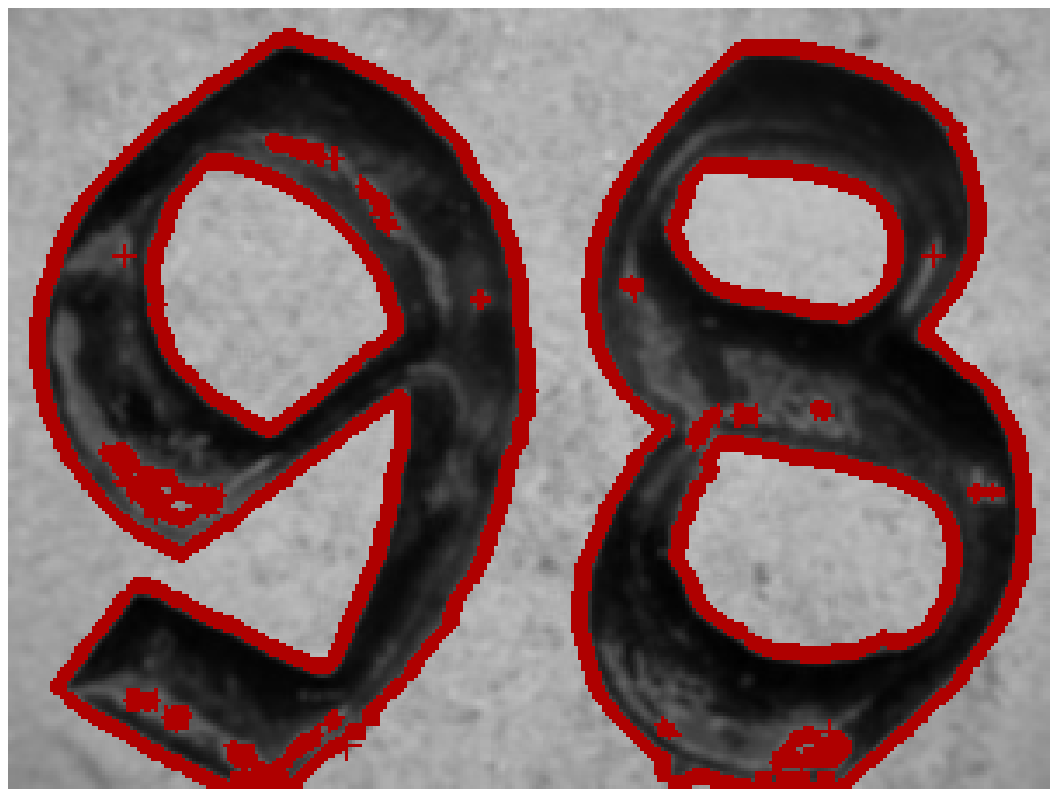,height=0.13\textwidth, width=0.2\textwidth}
\hspace{0.000001\textwidth}
\psfig{figure=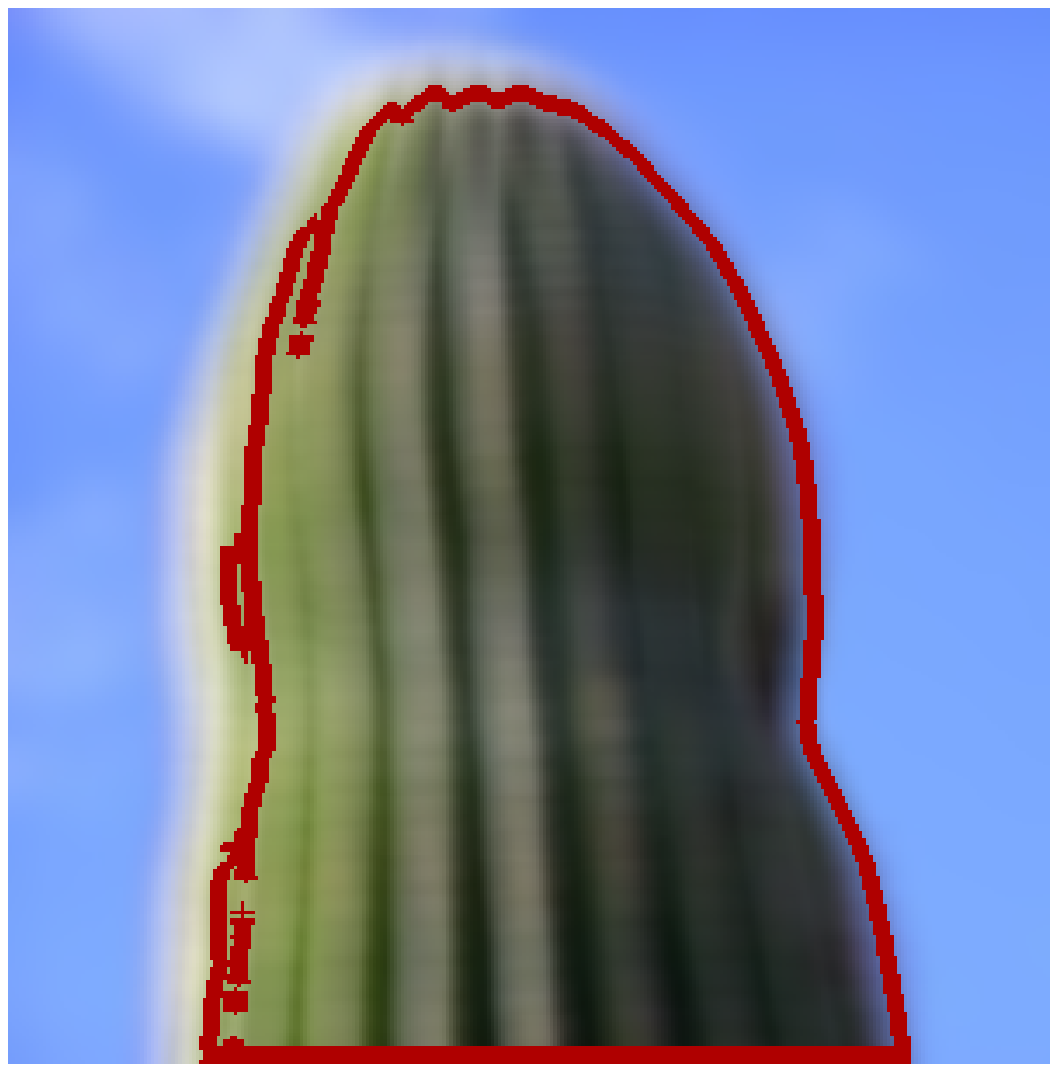,height=0.13\textwidth, width=0.2\textwidth}
\hspace{0.000001\textwidth}
\psfig{figure=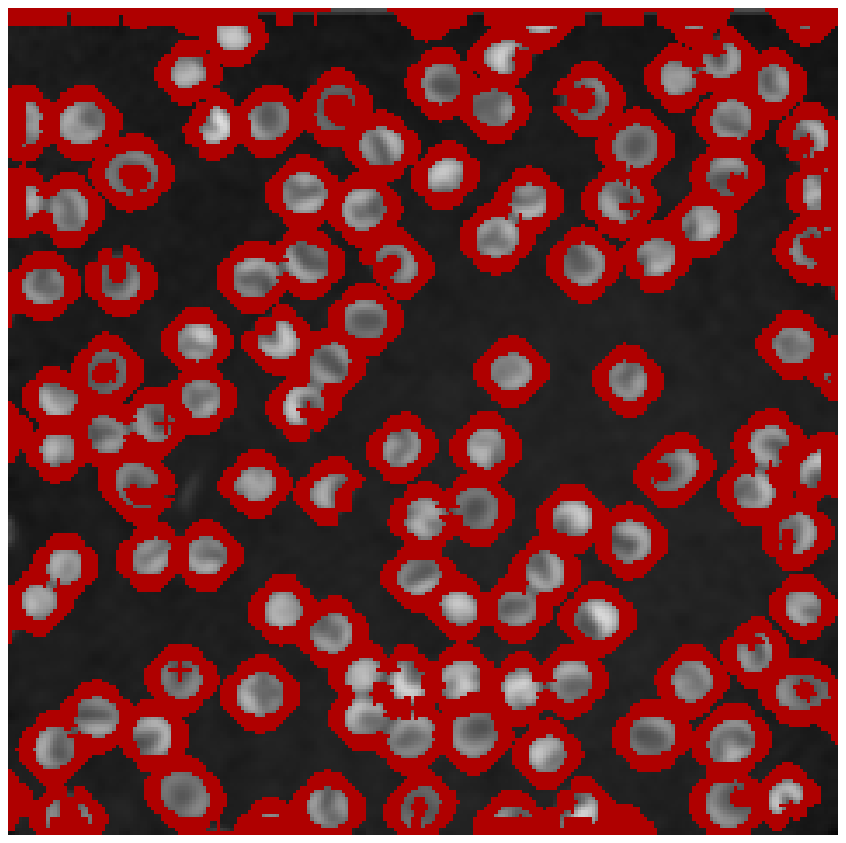,height=0.13\textwidth, width=0.2\textwidth}
\hspace{0.000001\textwidth}
\psfig{figure=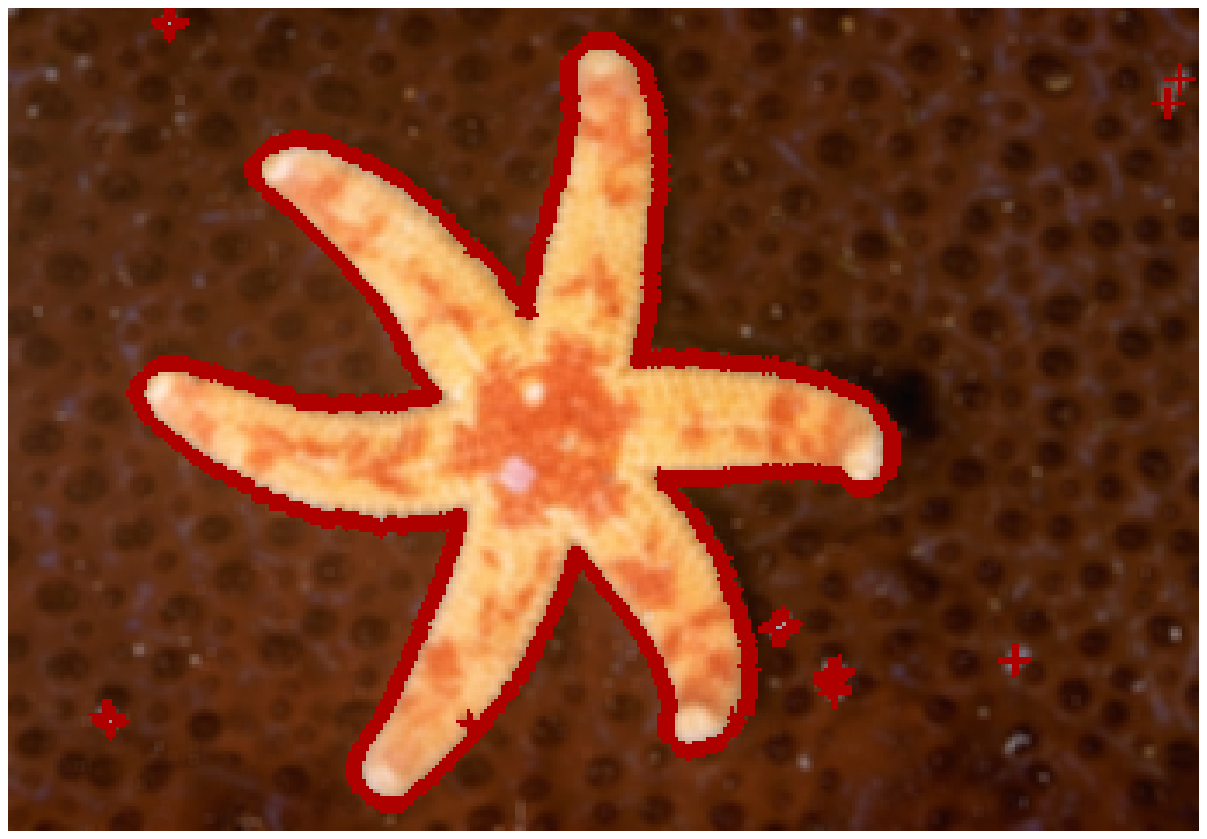,height=0.13\textwidth, width=0.2\textwidth}}
\vspace{0.000001\textwidth}
\centerline{
\psfig{figure=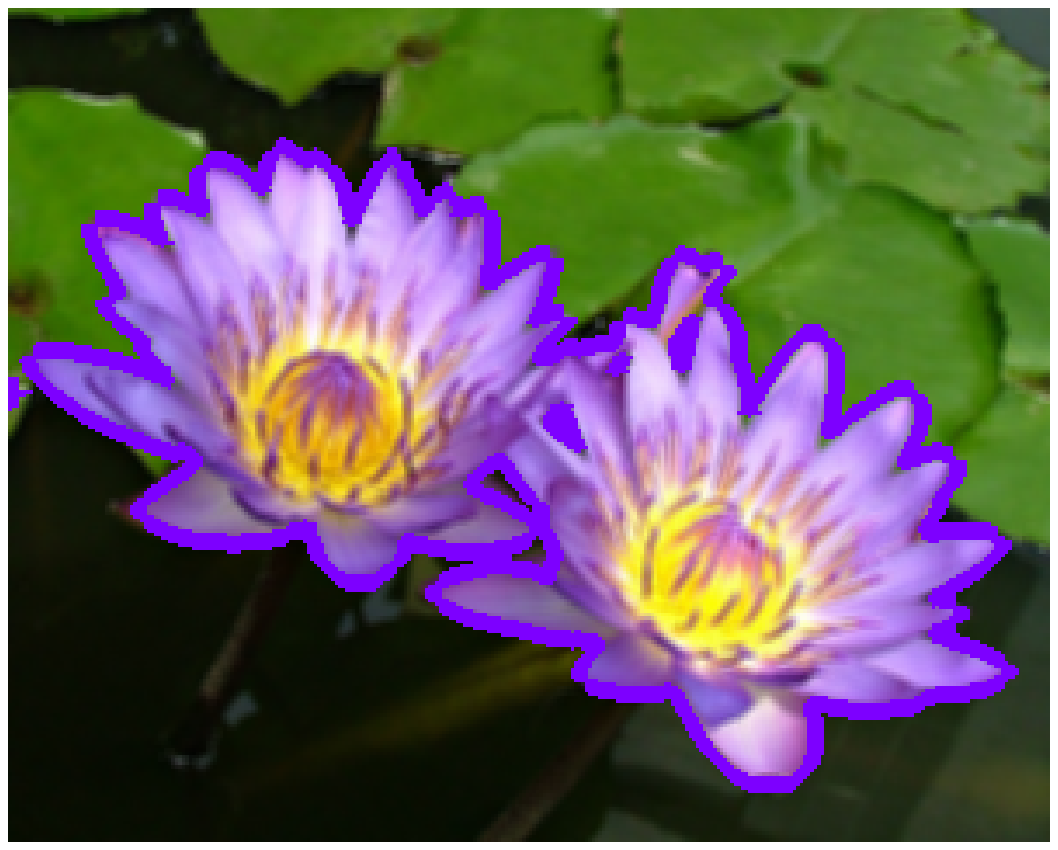,height=0.13\textwidth, width=0.2\textwidth}
\hspace{0.000001\textwidth}
\psfig{figure=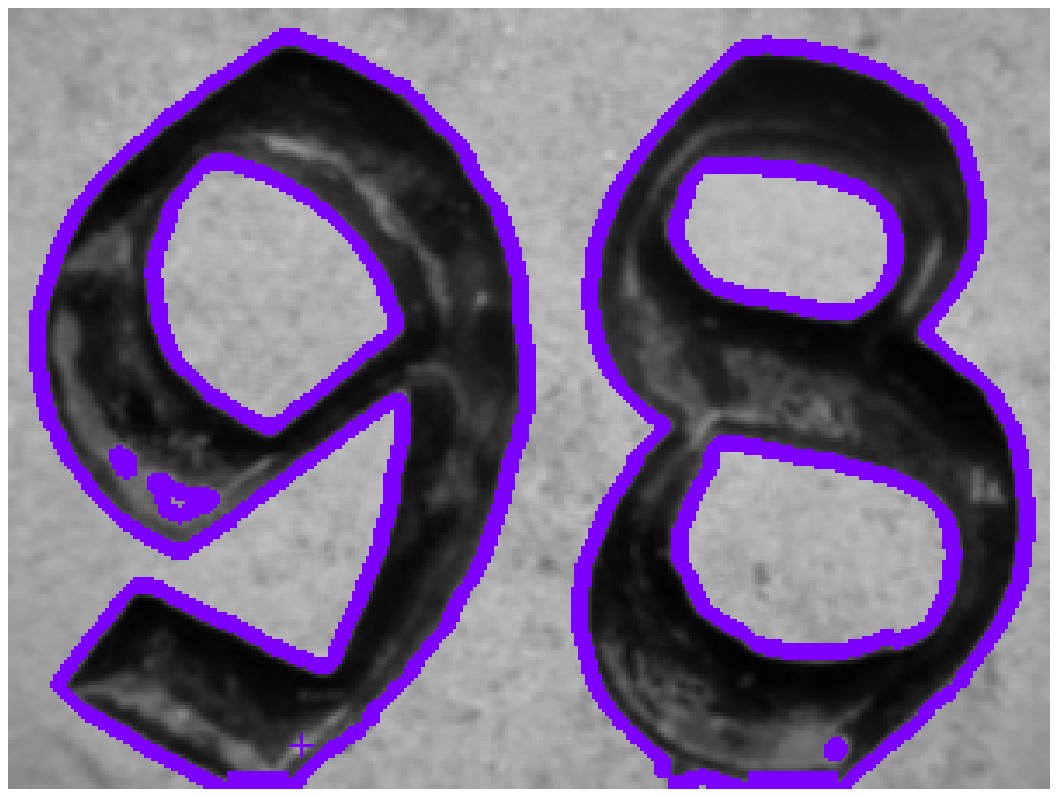,height=0.13\textwidth, width=0.2\textwidth}
\hspace{0.000001\textwidth}
\psfig{figure=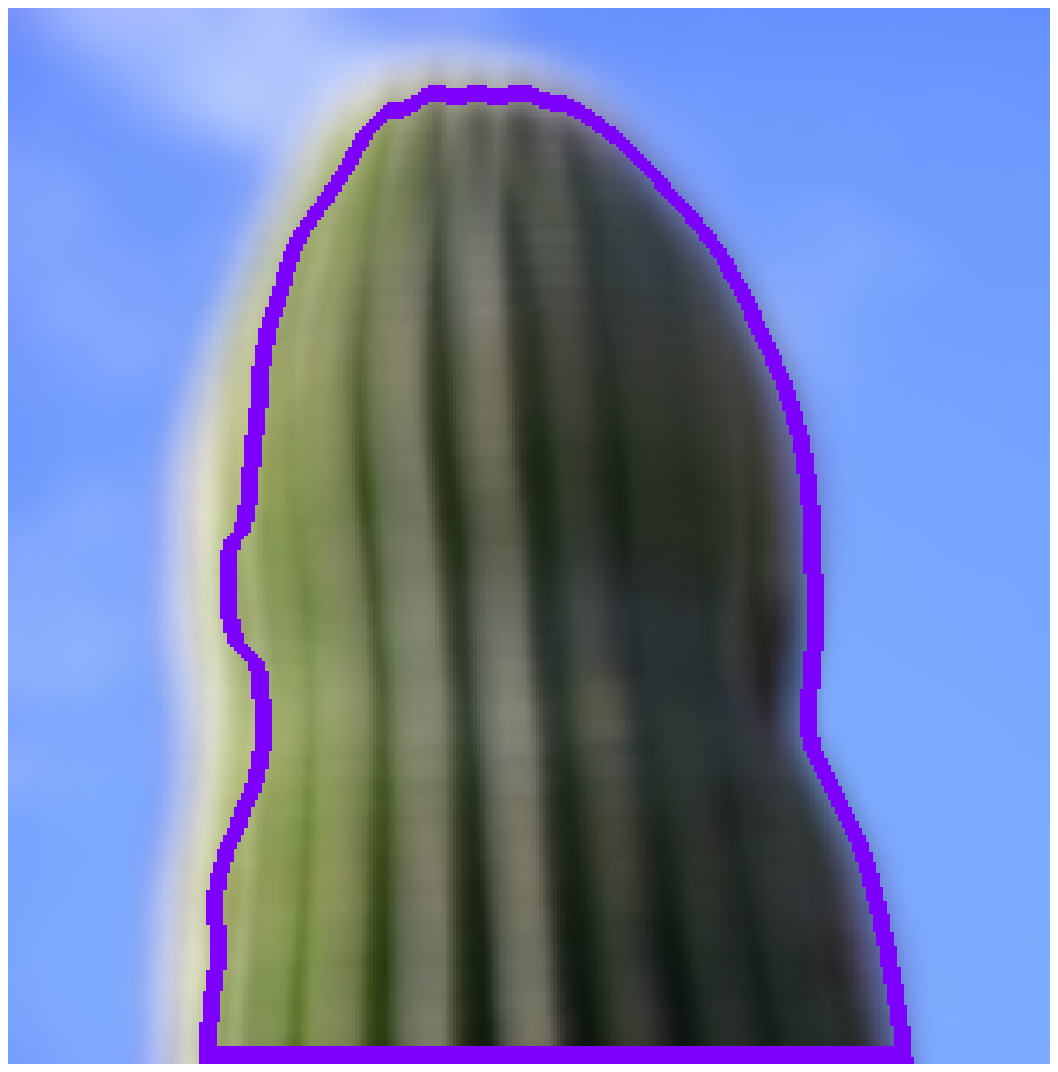,height=0.13\textwidth, width=0.2\textwidth}
\hspace{0.000001\textwidth}
\psfig{figure=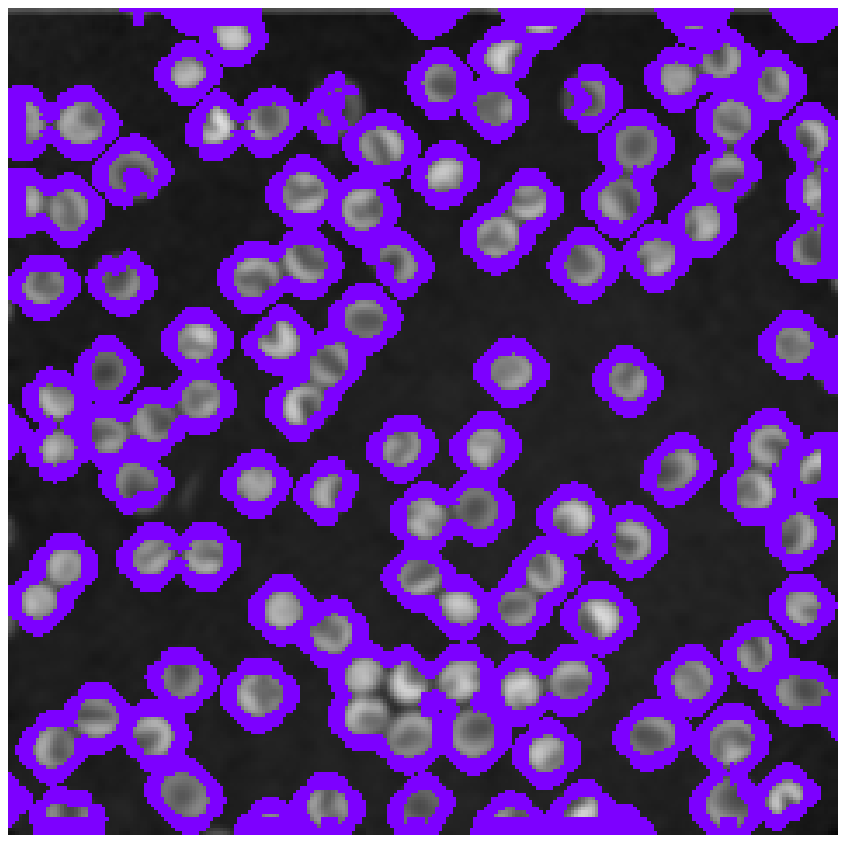,height=0.13\textwidth, width=0.2\textwidth}
\hspace{0.000001\textwidth}
\psfig{figure=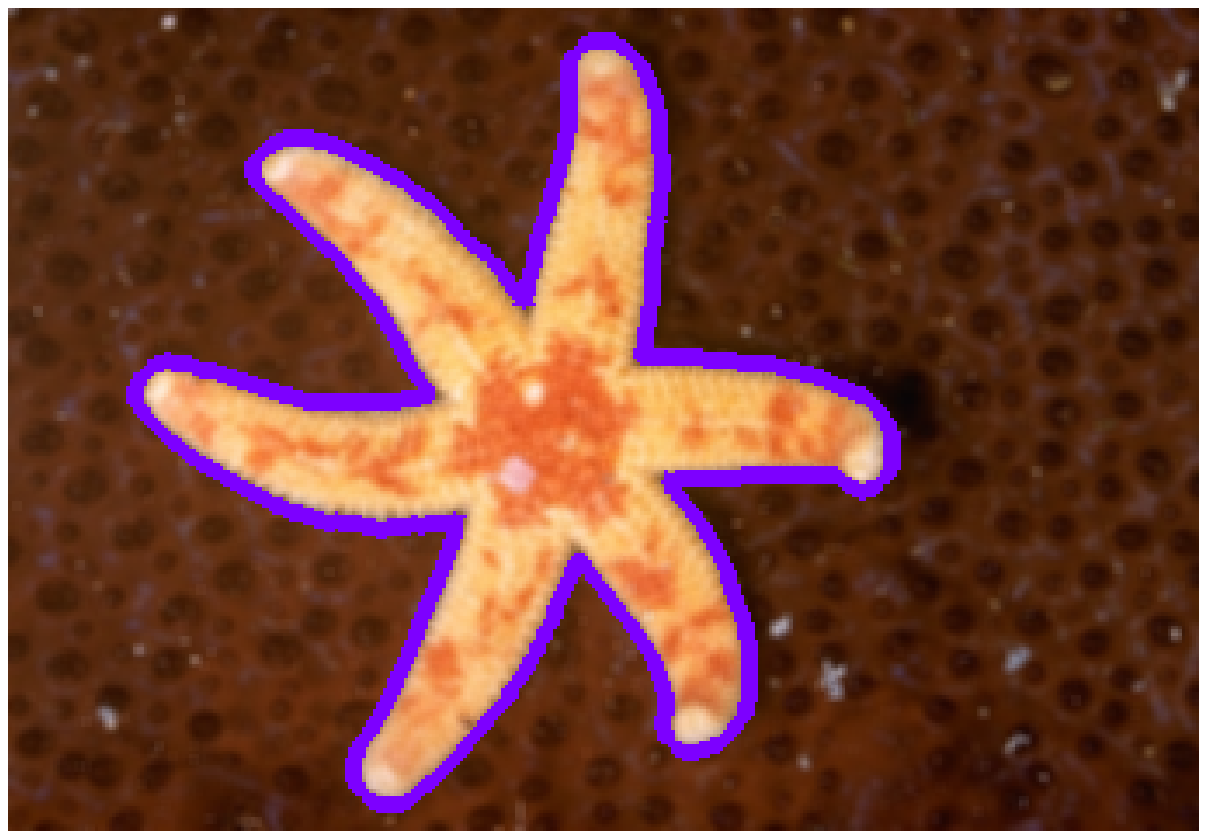,height=0.13\textwidth, width=0.2\textwidth}}
\vspace{0.000001\textwidth}
\centerline{
\psfig{figure=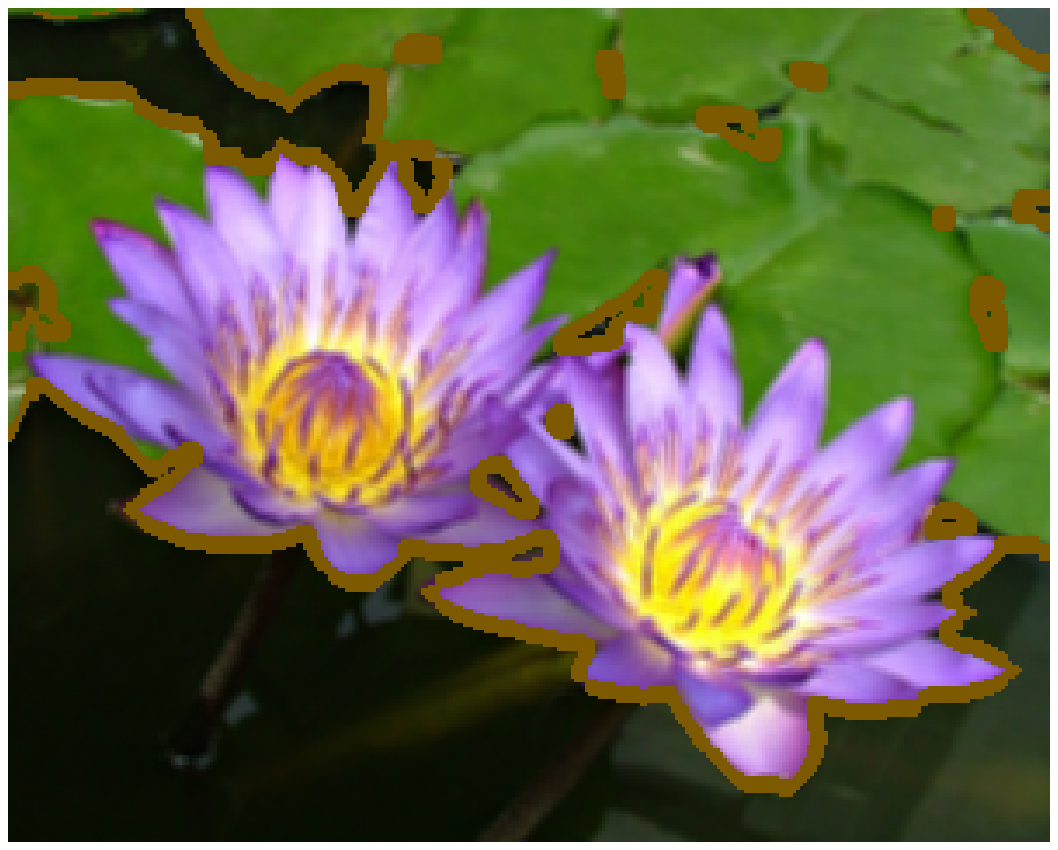,height=0.13\textwidth, width=0.2\textwidth}
\hspace{0.000001\textwidth}
\psfig{figure=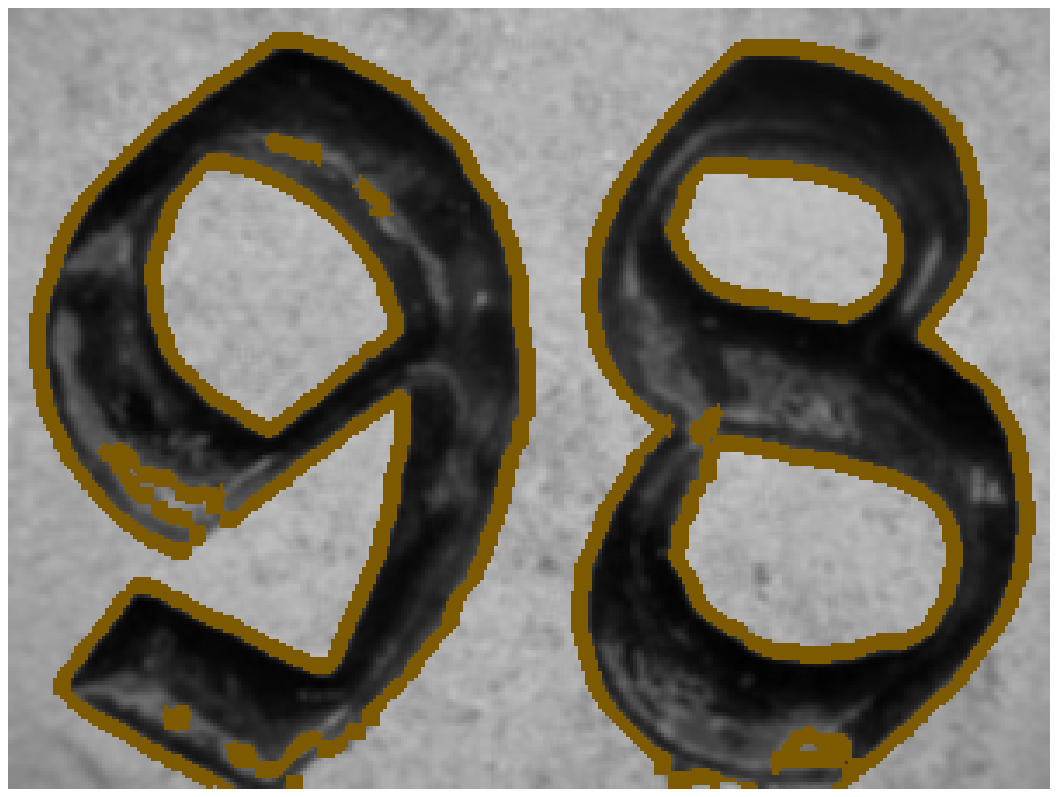,height=0.13\textwidth, width=0.2\textwidth}
\hspace{0.000001\textwidth}
\psfig{figure=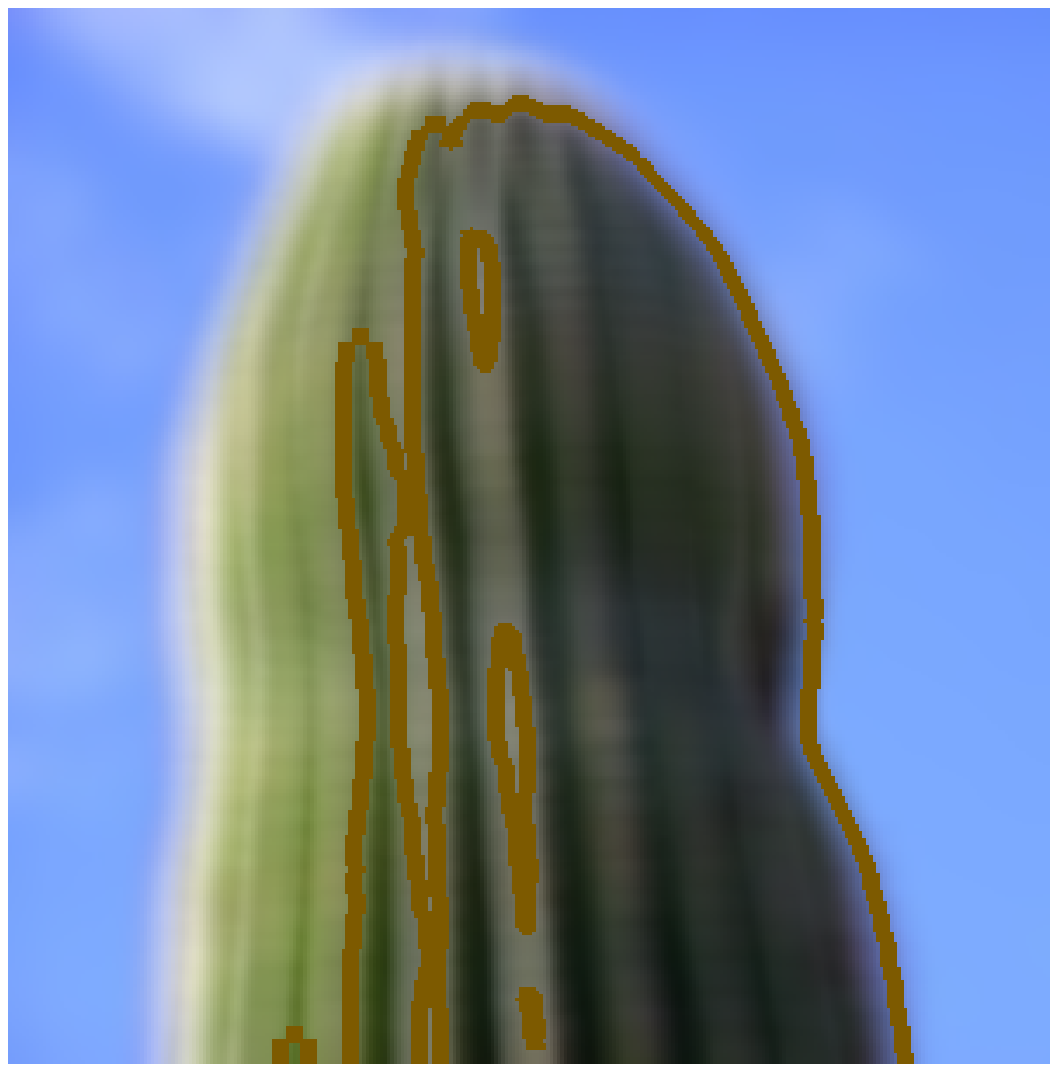,height=0.13\textwidth, width=0.2\textwidth}
\hspace{0.000001\textwidth}
\psfig{figure=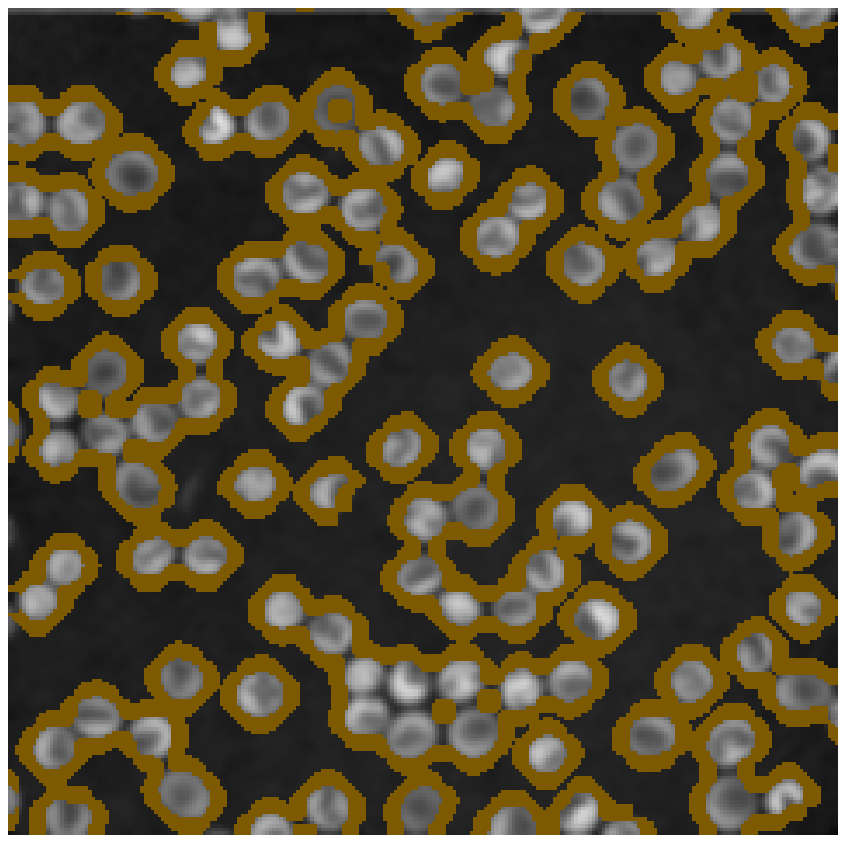,height=0.13\textwidth, width=0.2\textwidth}
\hspace{0.000001\textwidth}
\psfig{figure=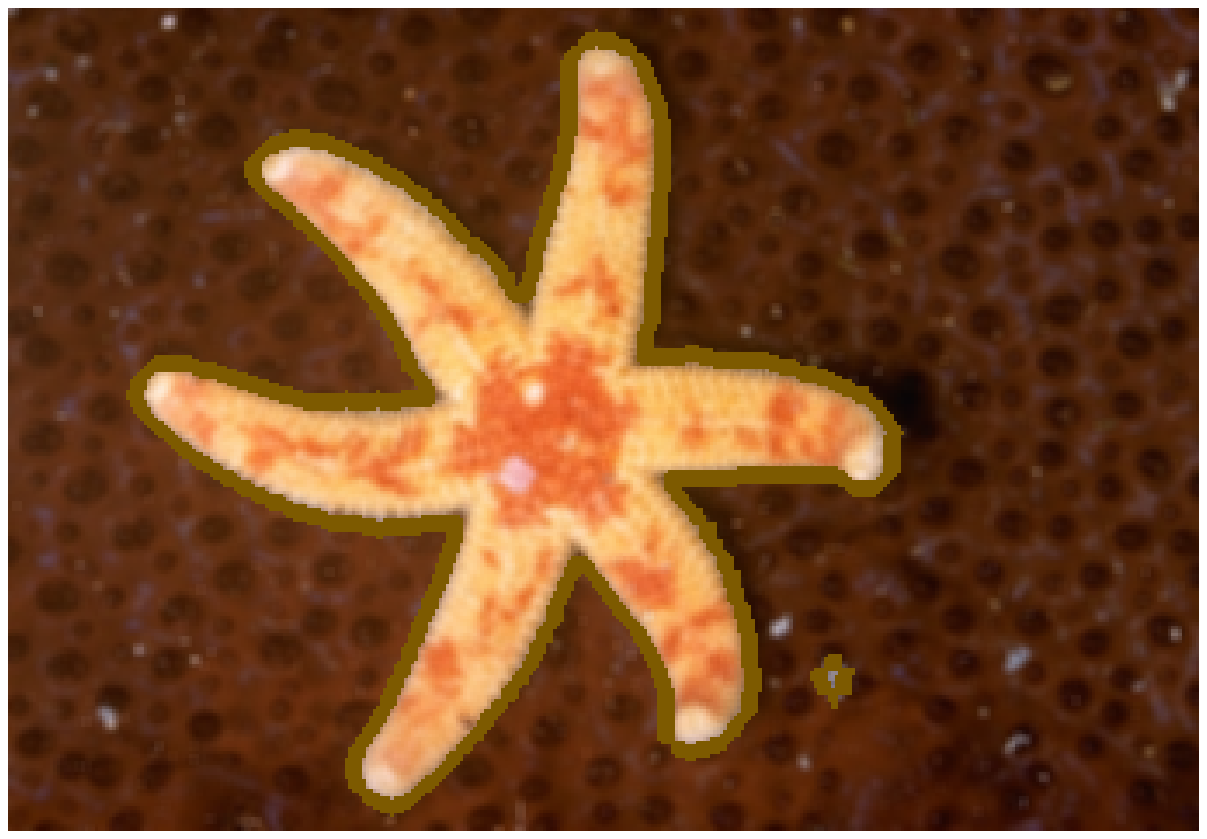,height=0.13\textwidth, width=0.2\textwidth}}
\vspace{0.000001\textwidth}
\centerline{
\psfig{figure=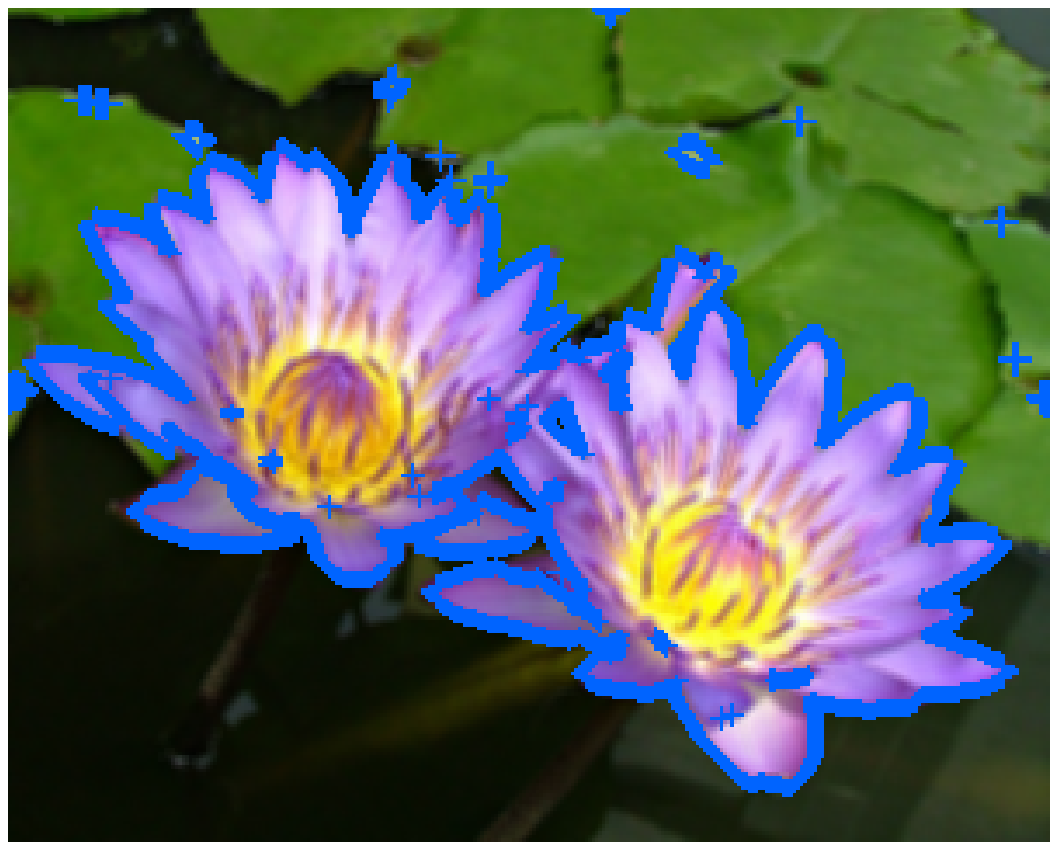,height=0.13\textwidth, width=0.2\textwidth}
\hspace{0.000001\textwidth}
\psfig{figure=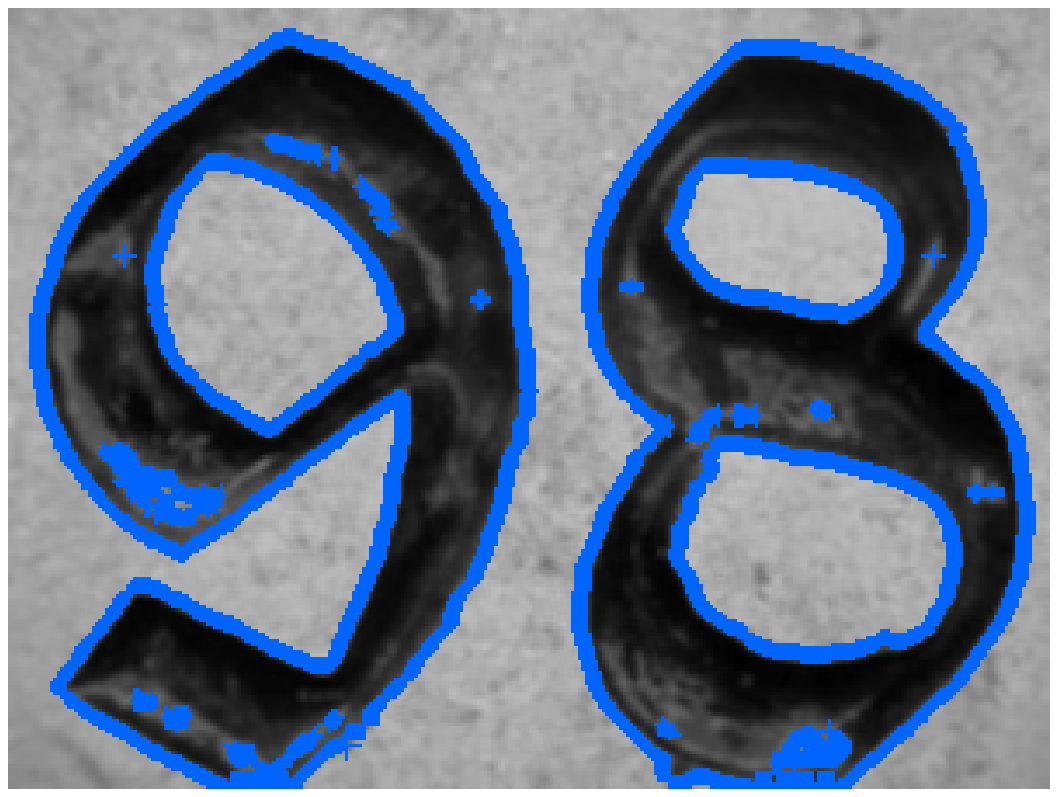,height=0.13\textwidth, width=0.2\textwidth}
\hspace{0.000001\textwidth}
\psfig{figure=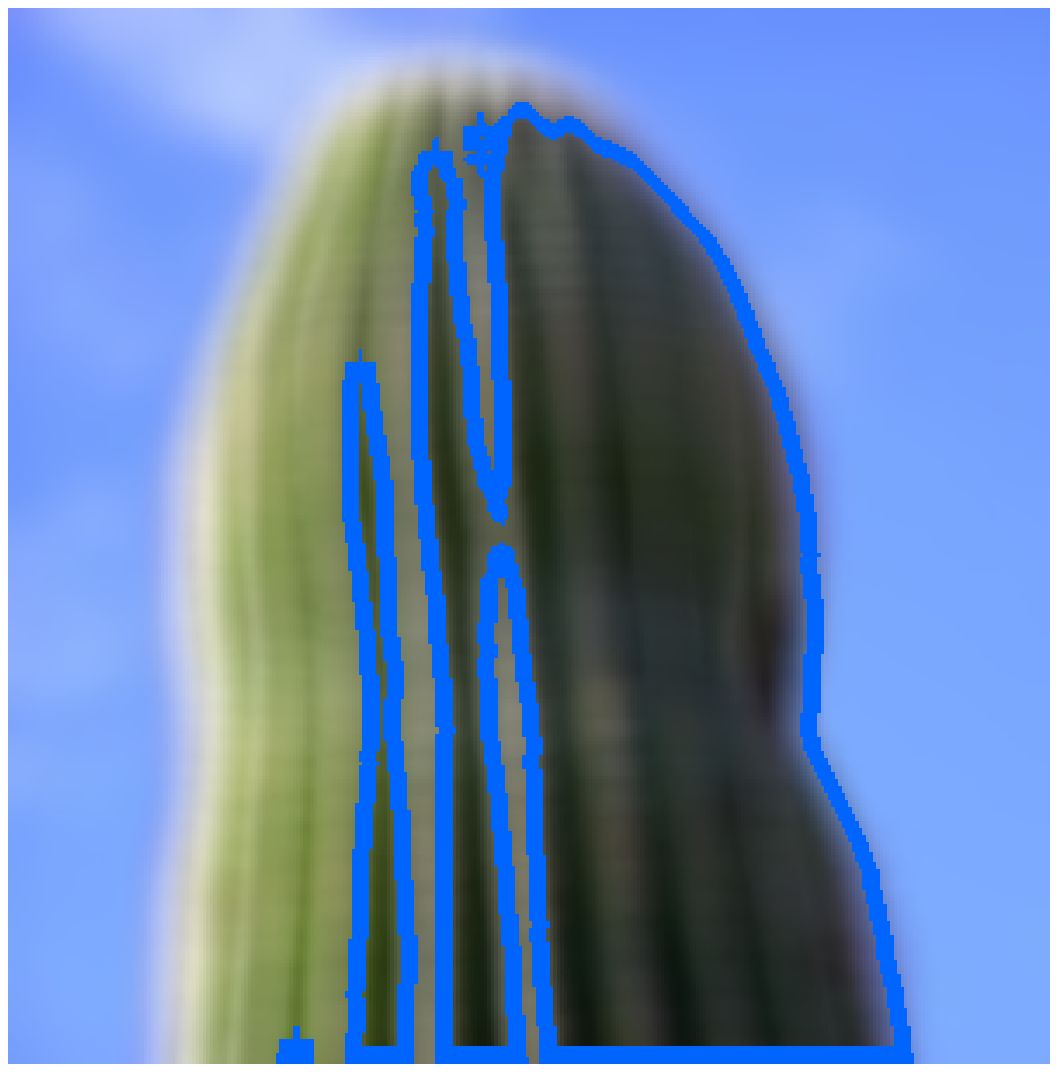,height=0.13\textwidth, width=0.2\textwidth}
\hspace{0.000001\textwidth}
\psfig{figure=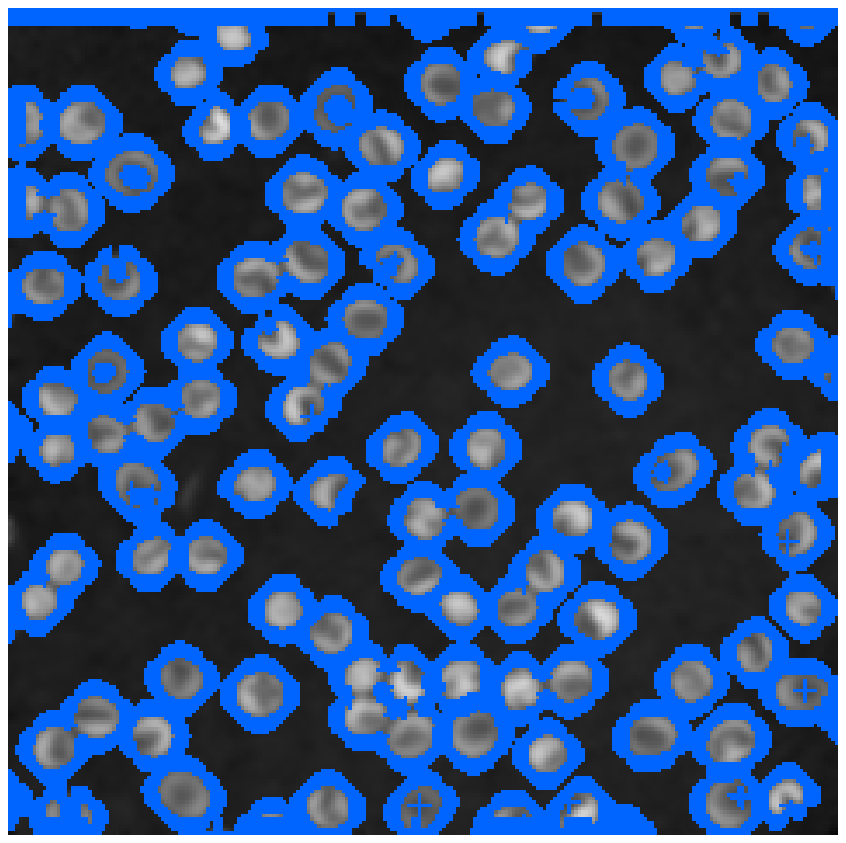,height=0.13\textwidth, width=0.2\textwidth}
\hspace{0.000001\textwidth}
\psfig{figure=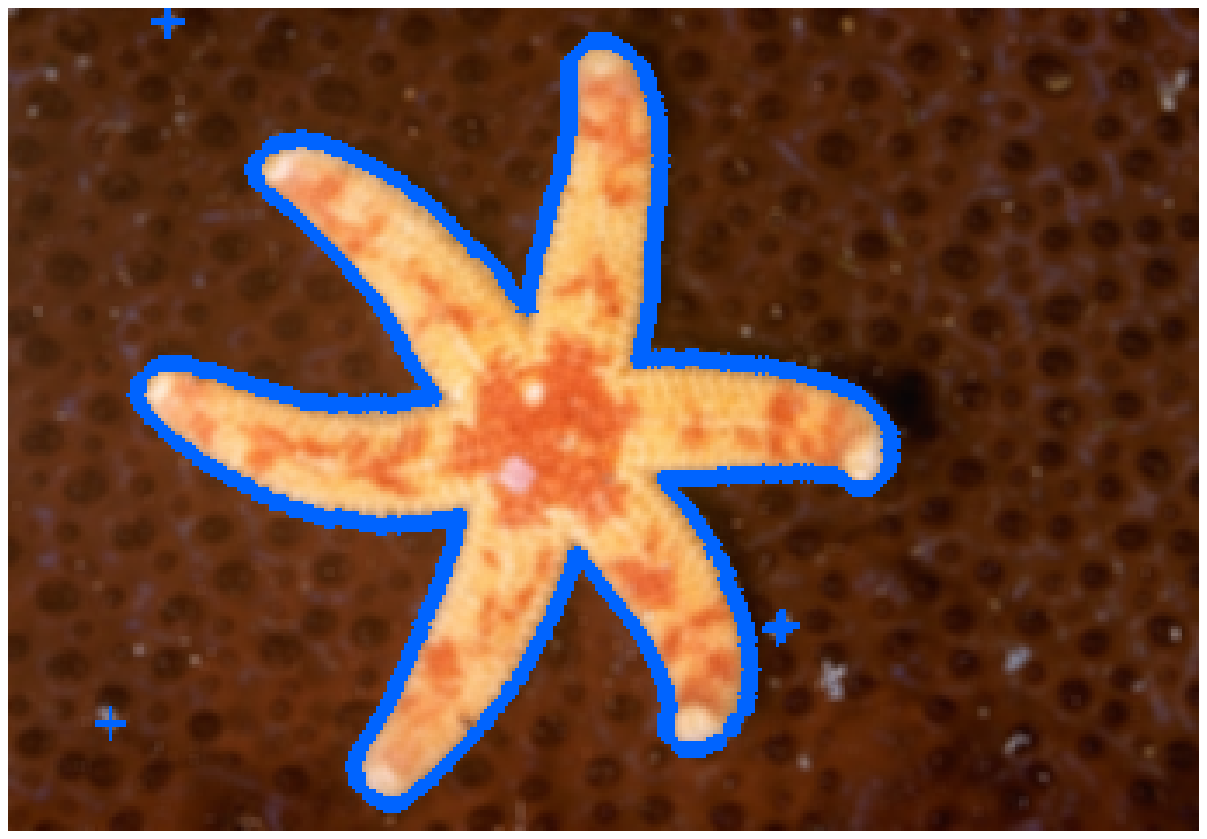,height=0.13\textwidth, width=0.2\textwidth}}
\vspace{0.000001\textwidth}
\centerline{
\psfig{figure=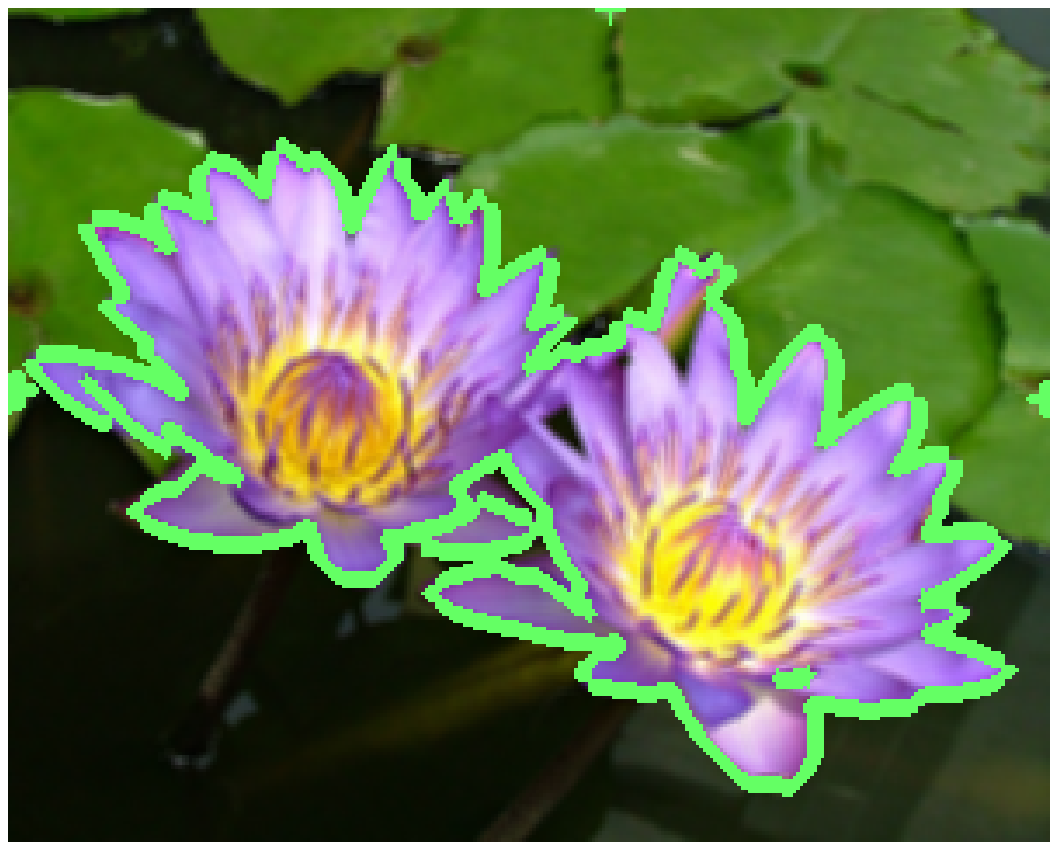,height=0.13\textwidth, width=0.2\textwidth}
\hspace{0.000001\textwidth}
\psfig{figure=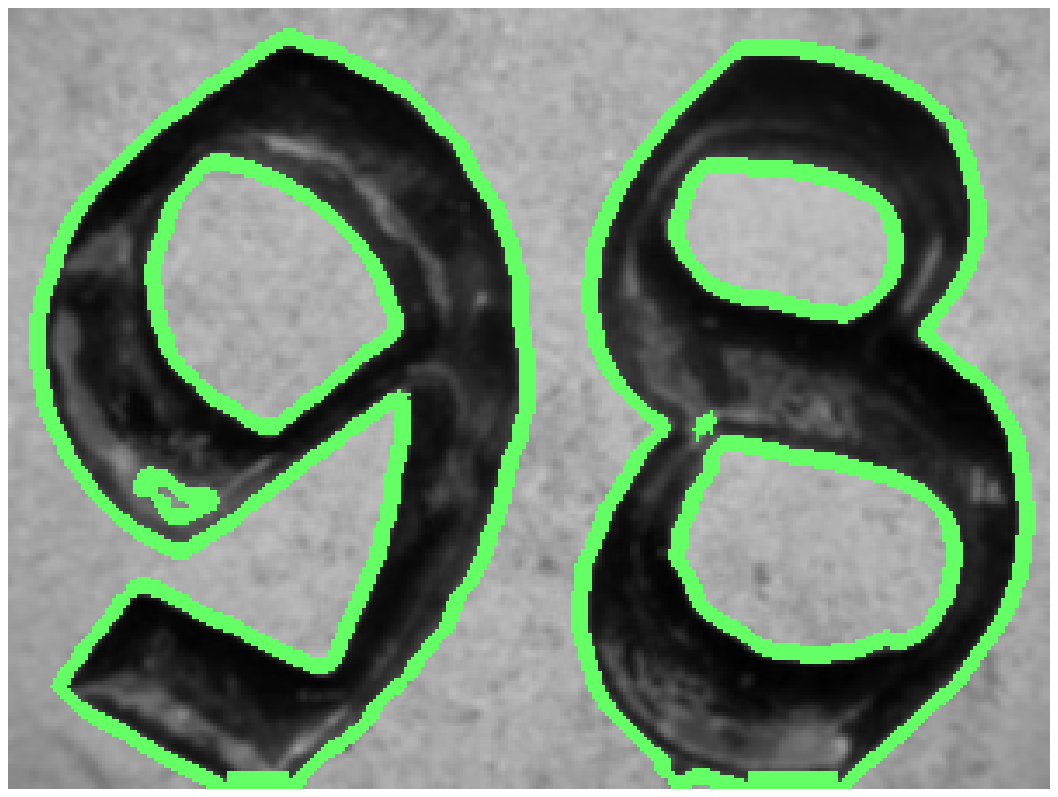,height=0.13\textwidth, width=0.2\textwidth}
\hspace{0.000001\textwidth}
\psfig{figure=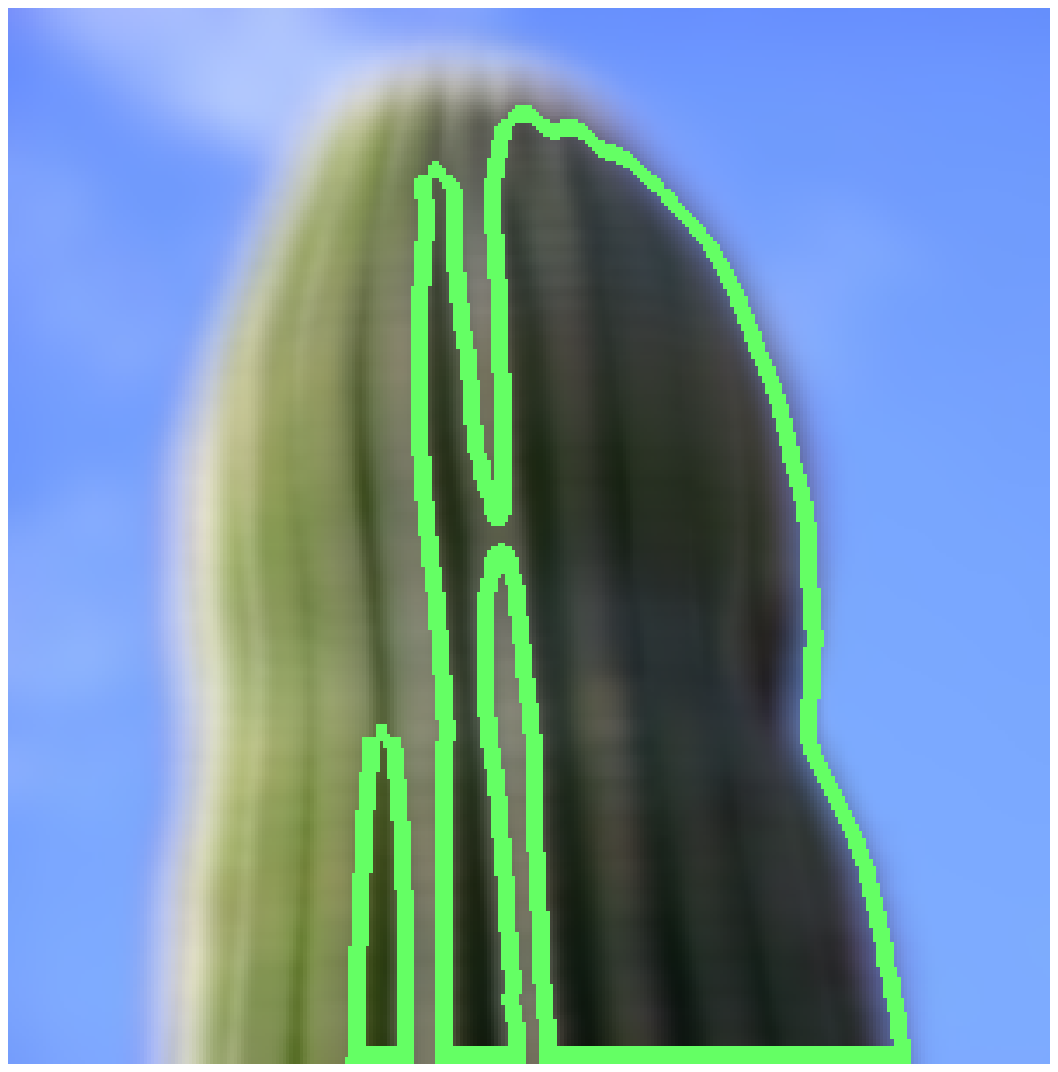,height=0.13\textwidth, width=0.2\textwidth}
\hspace{0.000001\textwidth}
\psfig{figure=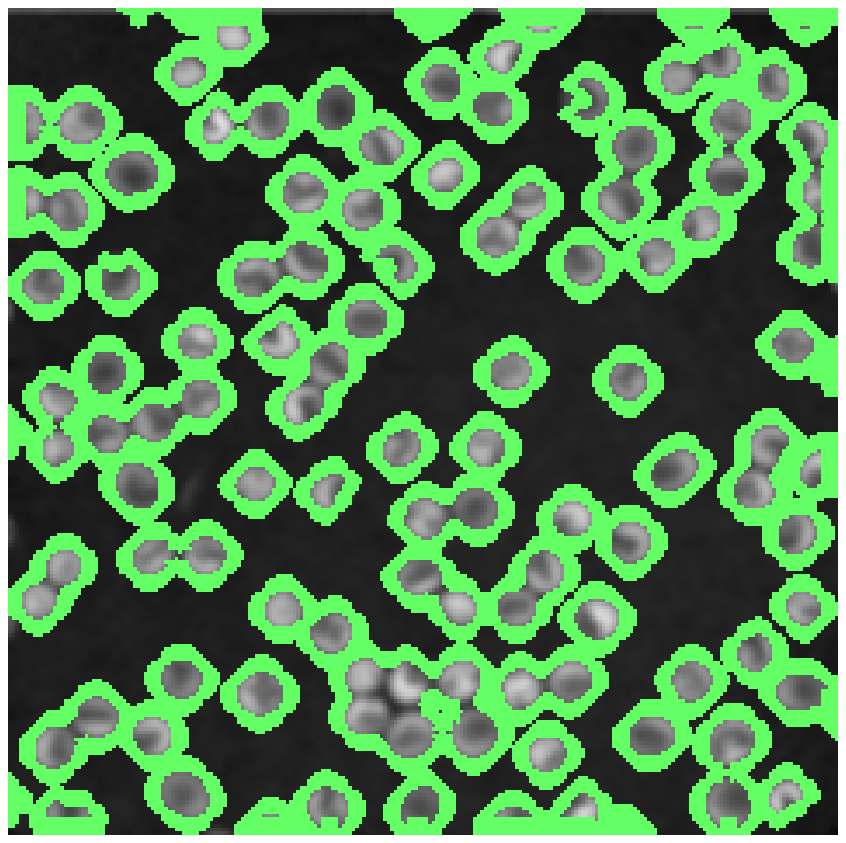,height=0.13\textwidth, width=0.2\textwidth}
\hspace{0.000001\textwidth}
\psfig{figure=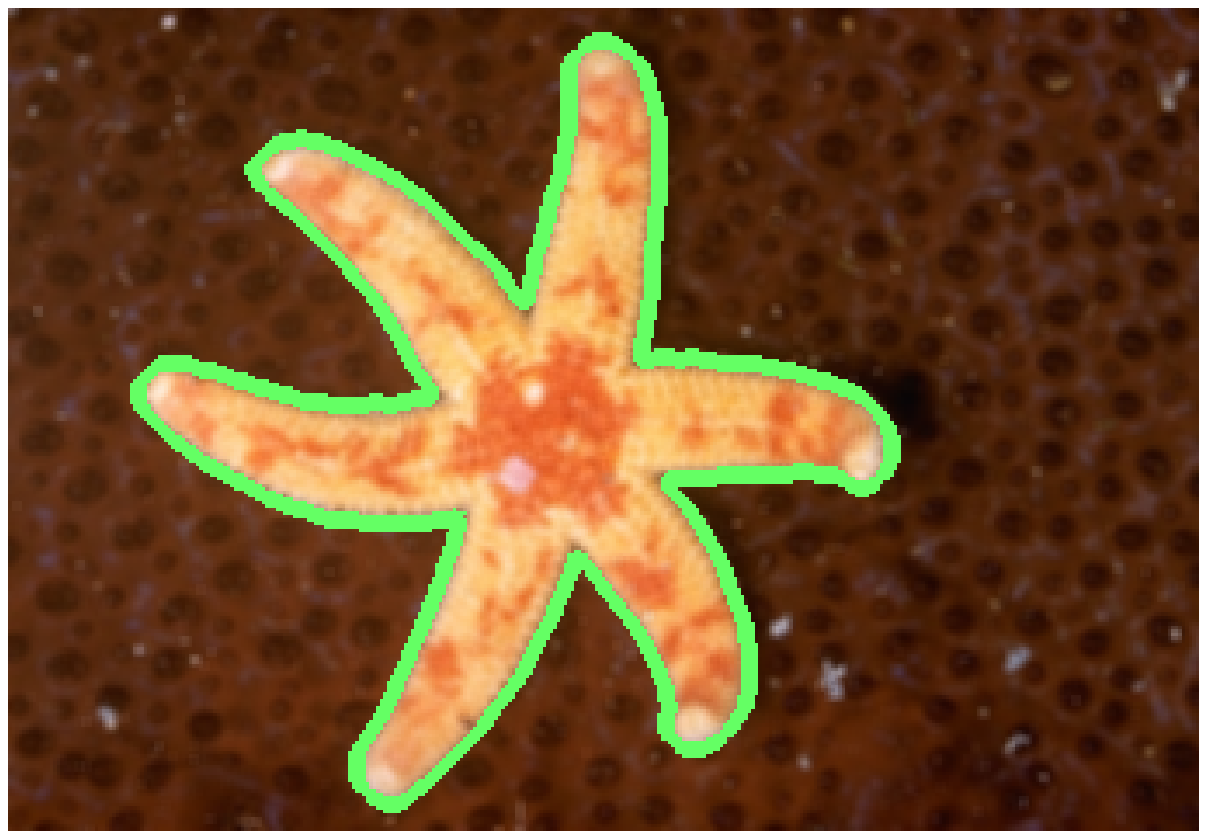,height=0.13\textwidth, width=0.2\textwidth}}
\vspace{0.000001\textwidth}
\centerline{
\psfig{figure=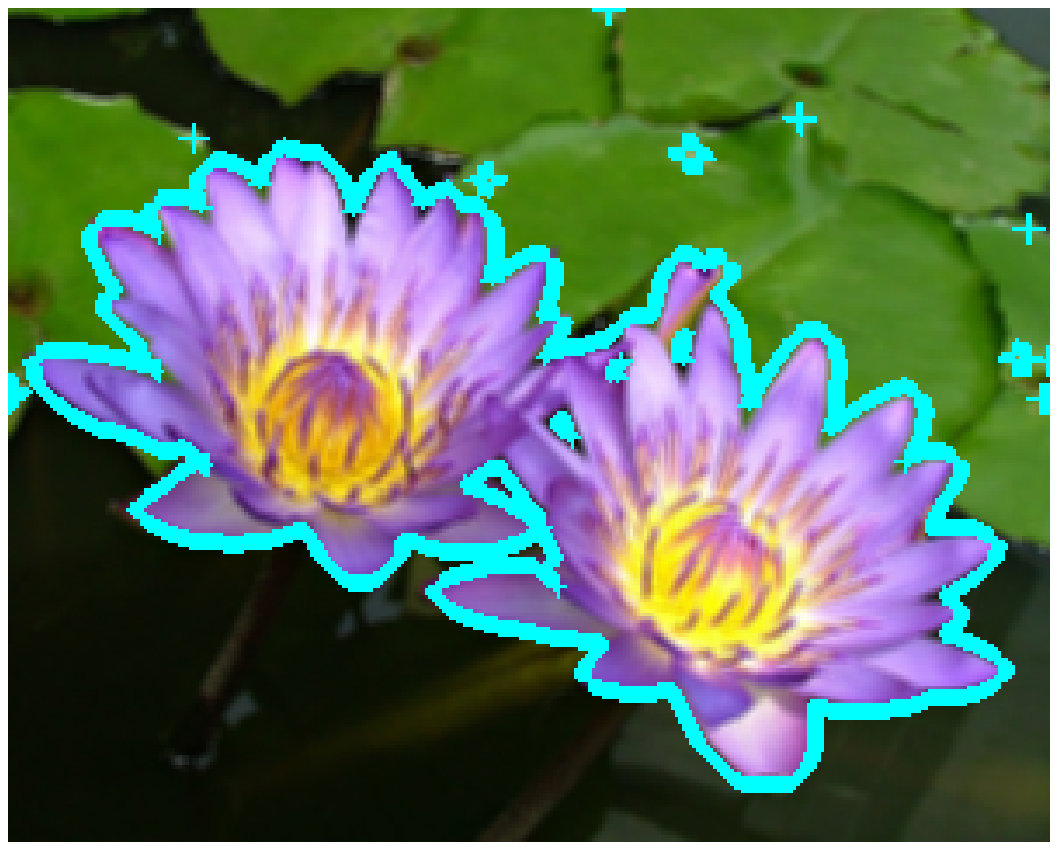,height=0.13\textwidth, width=0.2\textwidth}
\hspace{0.000001\textwidth}
\psfig{figure=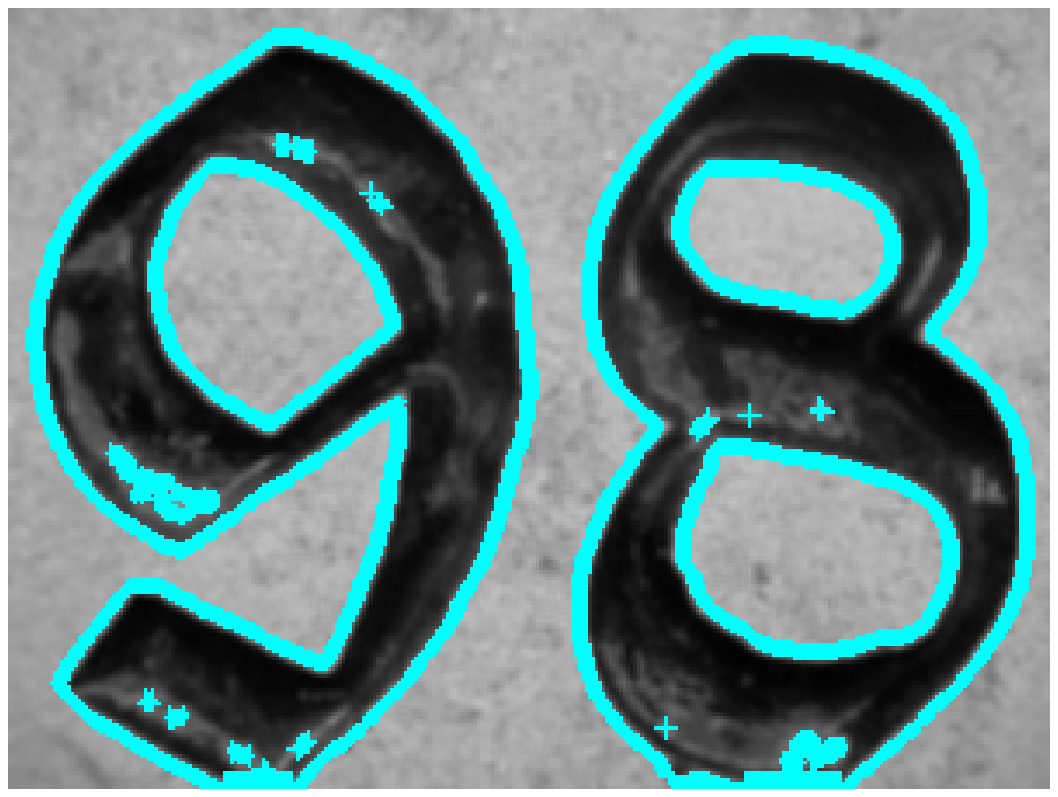,height=0.13\textwidth, width=0.2\textwidth}
\hspace{0.000001\textwidth}
\psfig{figure=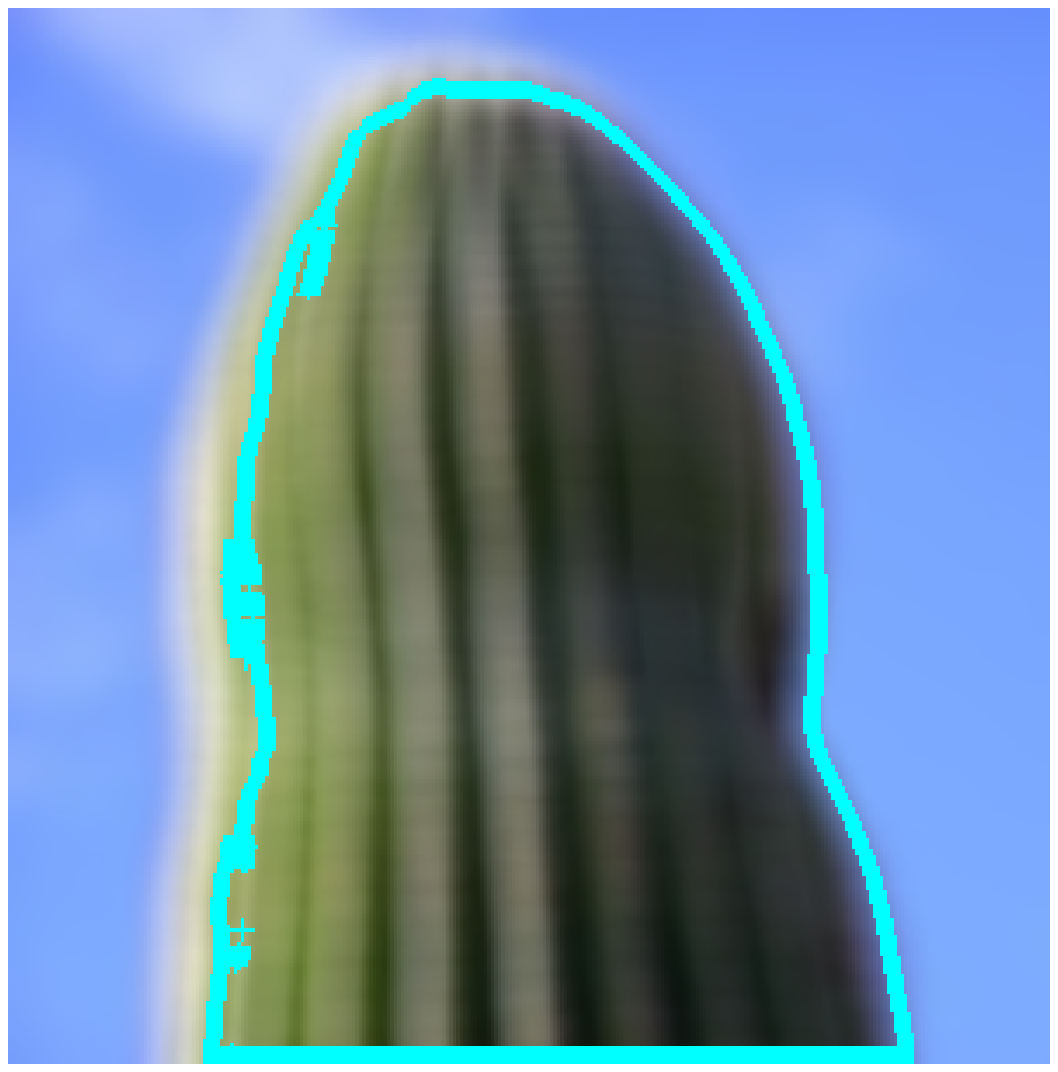,height=0.13\textwidth, width=0.2\textwidth}
\hspace{0.000001\textwidth}
\psfig{figure=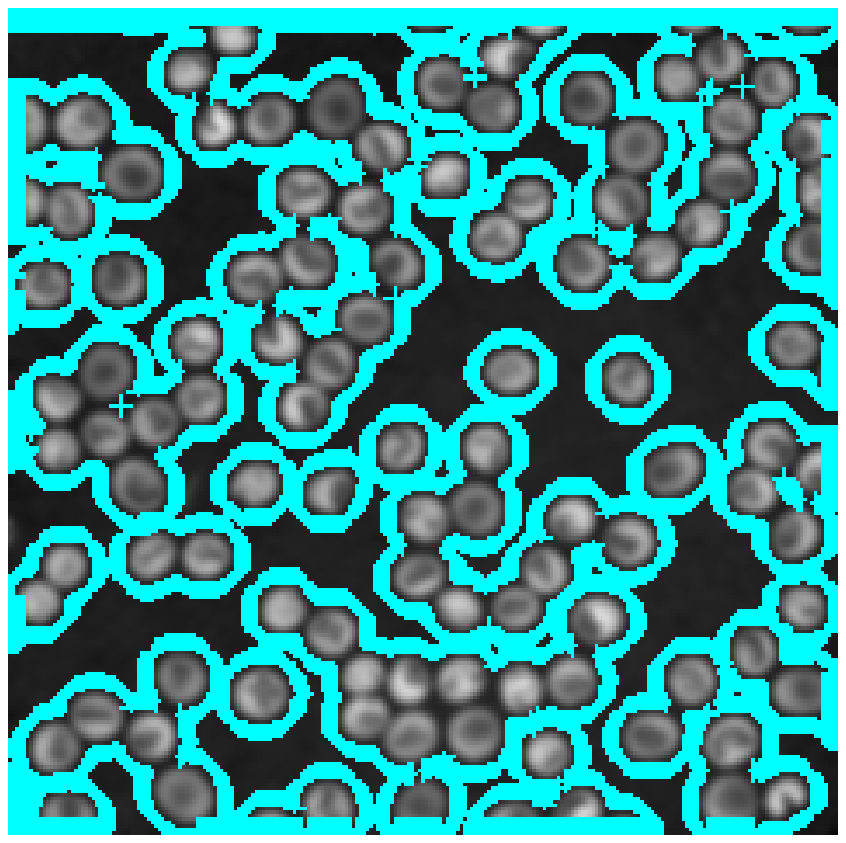,height=0.13\textwidth, width=0.2\textwidth}
\hspace{0.000001\textwidth}
\psfig{figure=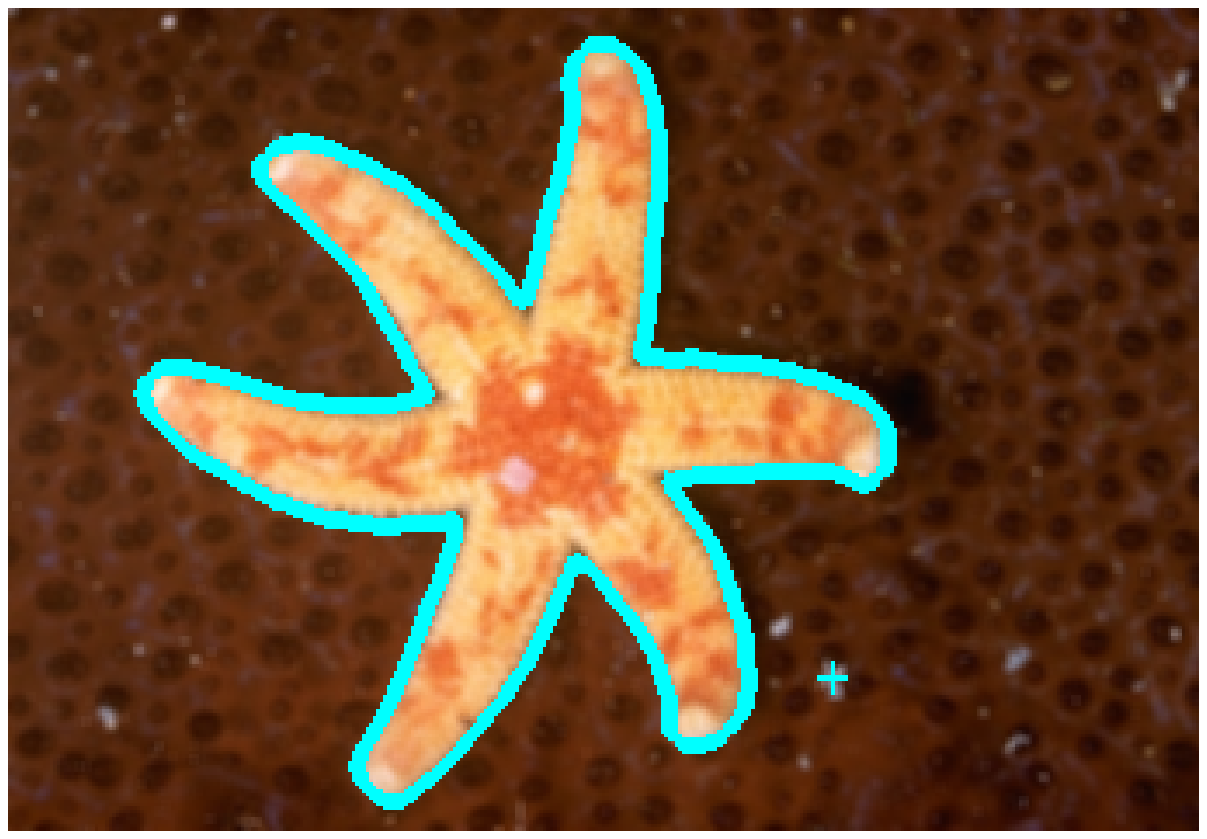,height=0.13\textwidth, width=0.2\textwidth}}
\vspace{0.000001\textwidth}
\centerline{
\psfig{figure=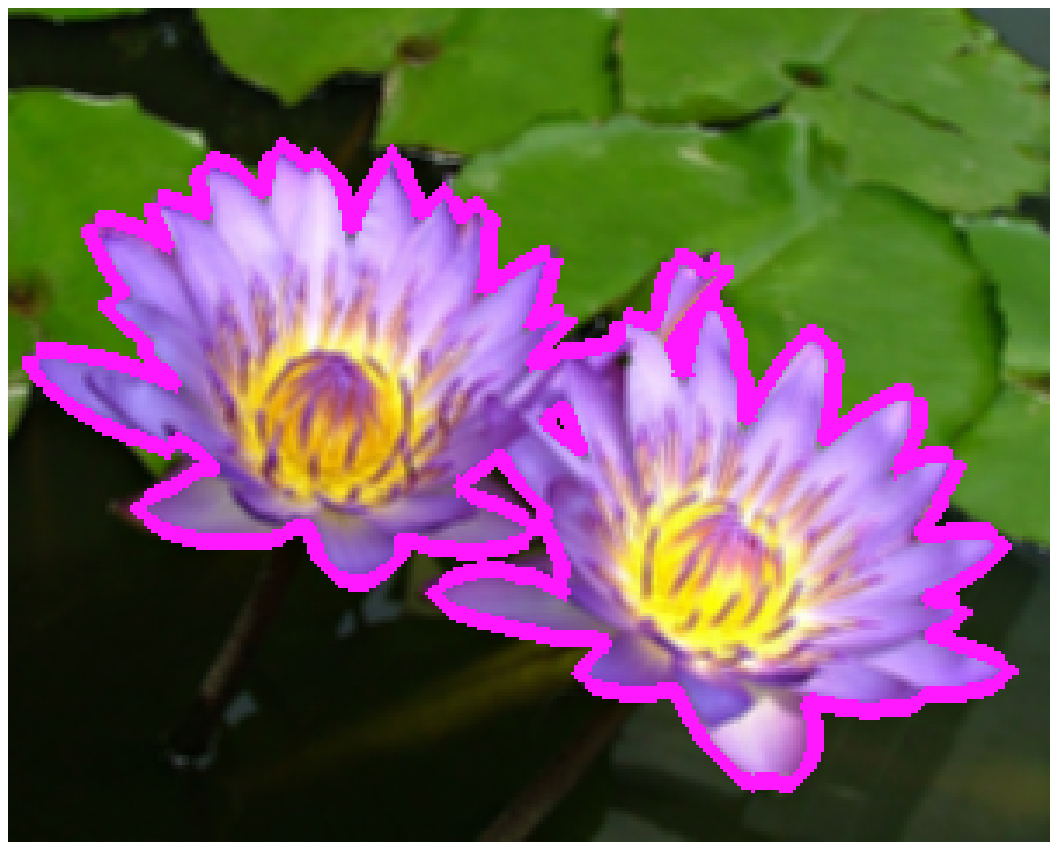,height=0.13\textwidth, width=0.2\textwidth}
\hspace{0.000001\textwidth}
\psfig{figure=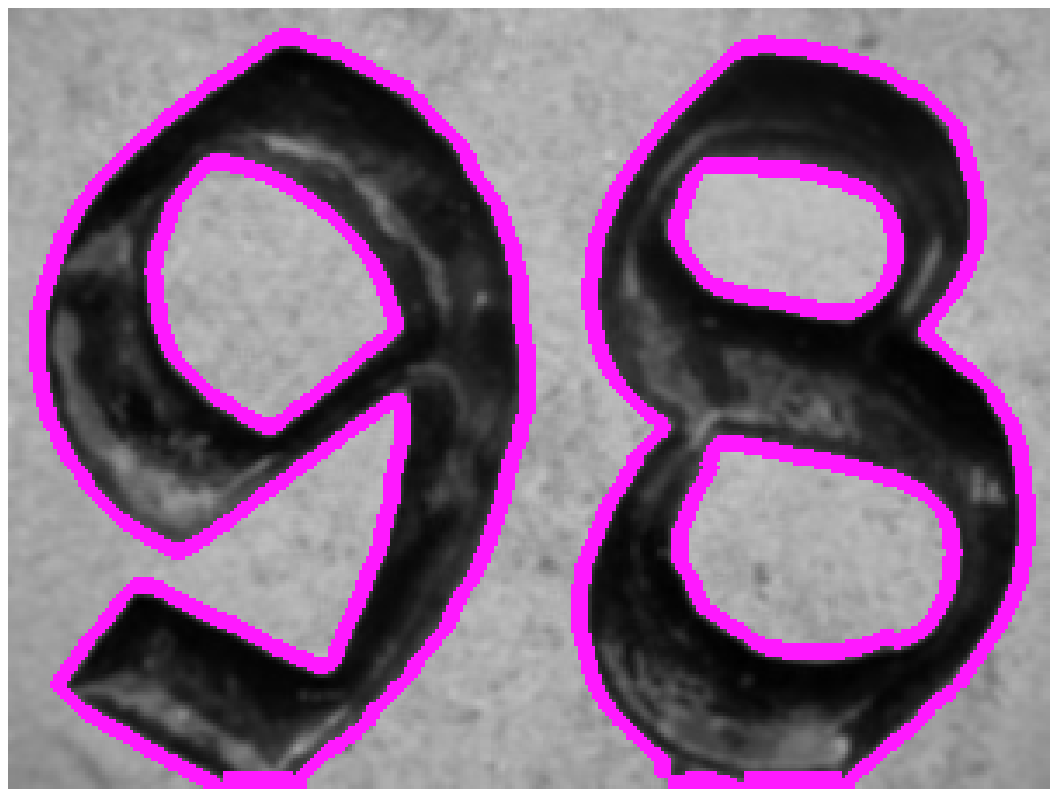,height=0.13\textwidth, width=0.2\textwidth}
\hspace{0.000001\textwidth}
\psfig{figure=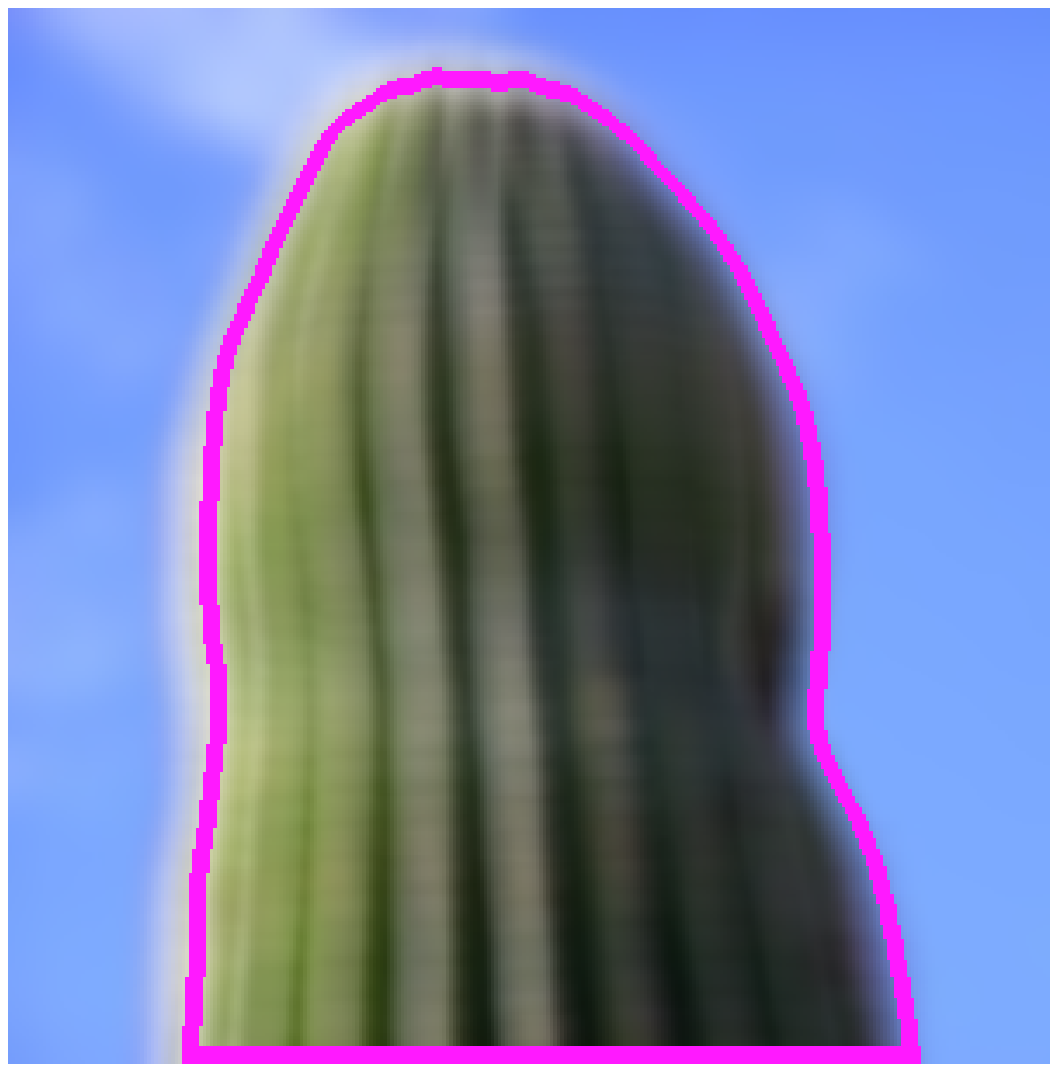,height=0.13\textwidth, width=0.2\textwidth}
\hspace{0.000001\textwidth}
\psfig{figure=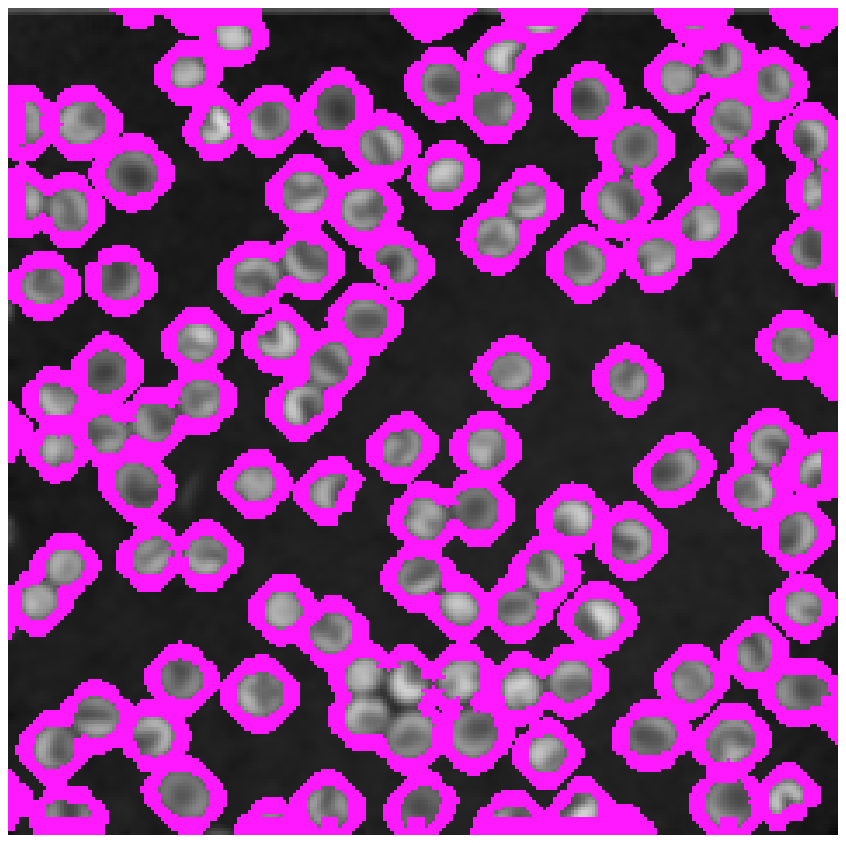,height=0.13\textwidth, width=0.2\textwidth}
\hspace{0.000001\textwidth}
\psfig{figure=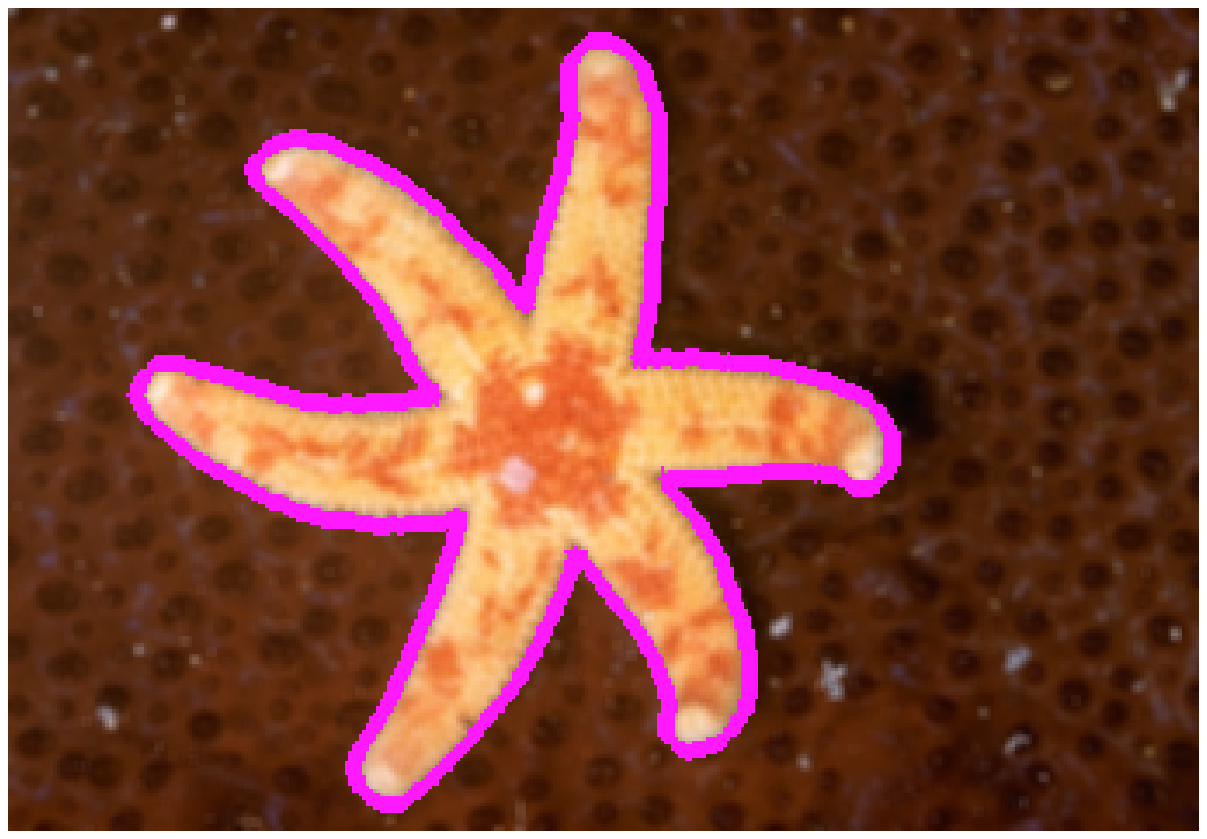,height=0.13\textwidth, width=0.2\textwidth}}
\centerline{(a)\hspace{0.17\textwidth} (b)\hspace{0.17\textwidth} (c)\hspace{0.17\textwidth} (d)\hspace{0.17\textwidth} (e)}
\caption{Visual comparison of segmentation results of blurred (a) Lily flower image with Gaussian function ($r_{s} = 21$, $\sigma = 50$), (b) 98 image with Gaussian function ($r_{s} = 11$, $\sigma = 50$), (c) Sharp image with Gaussian function ($r_{s} = 17$, $\sigma = 50$), (d) Cell4 image with Gaussian function ($r_{s} = 20$, $\sigma = 50$), (e) Starfish image with Gaussian function ($r_{s} = 20$, $\sigma = 50$) using various techniques. \textbf{First row:} images with initial contours, \textbf{Second row:} segmentation results obtained by {\sc FEAC}, \textbf{Third row:} segmentation results obtained by {\sc NFACMKM}, \textbf{Fourth row:} segmentation results obtained by {\sc FACGK}, \textbf{Fifth row:} segmentation results obtained by {\sc LPFAC}, \textbf{Sixth row:} segmentation results obtained by {\sc FDFEAC}, \textbf{Seventh row:} segmentation results obtained by {\sc GLFEAC} and \textbf{Eighth row:} segmentation results obtained by {\sc RGLFEAC}.\label{figure_blur_seg} }
\end{figure}

To analyze the performance of the existing algorithms under complex environment, blurred versions of the
original images are also considered in this experiment. Different Gaussian functions with varying radius ($r_{s}$) and 
sigma ($\sigma$) are used to blur the original images. Figure~\ref{figure_blur_seg} displays the segmentation results of blurred images. Since, Gaussian blur functions smooth the images, all these methods obtain better segmentation results than their segmentation results of corresponding original images. From this figure, it is observed that excepting RGLFEAC, all other methods produce small background patches as objects for those images having high intensity in-homogeneity. However, local and global information help {\sc RGLFEAC} to properly segment images having high region in-homogeneity.    

\begin{table}[htp]
\begin{center}
\begin{tabular}{|l|l|l|l|l|l|l|l|}\hline
Measures & \multicolumn{7}{|l|}{Techniques } \\\cline{2-8}
& {\sc m1} & {\sc m2} & {\sc m3} & {\sc m4} & {\sc m5} & {\sc m6} & {\sc m7}\\\cline{1-8}
Ave. Jacard error & 0.185 & 0.167 & 0.172 & 0.203 & 0.274 & 0.195 & \textcolor[rgb]{0.00,1.00,0.00}{0.066} \\\hline
Ave. F-measure    & 0.897& 0.906 & 0.905 & 0.884 &0.812 & 0.888& \textcolor[rgb]{1.00,0.00,0.00}{0.965}  \\\hline
\end{tabular}
\end{center}
\caption{Quantitative comparison among various techniques with respect to average Jacard error and average F-measure over $100$ blurred images with different Gaussian function. {\sc m1: feac}, {\sc m2: nfacmkm}, {\sc m3: facgk}, {\sc m4: lpfac}, {\sc m5: fdfeac}, {\sc m6: glfeac} and {\sc m7: rglfeac}. Green colored numeric value indicates least average Jacard error corresponds to the best segmentation result. Whereas, red colored numeric value indicates highest F-measure corresponds to the best segmentation result. \label{table_blur}}
\end{table}

\begin{figure}
\centerline{
\psfig{figure=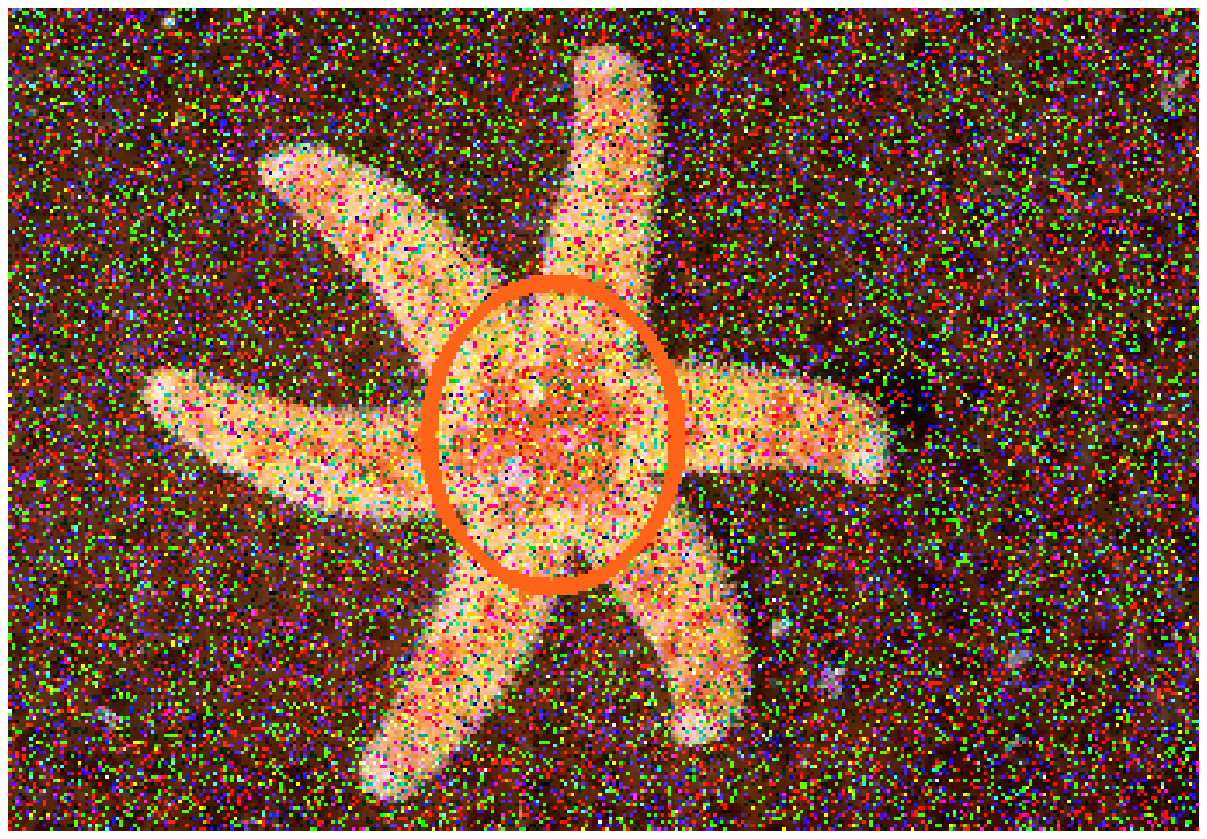, height=0.13\textwidth, width=0.2\textwidth}
\hspace{0.000001\textwidth}
\psfig{figure=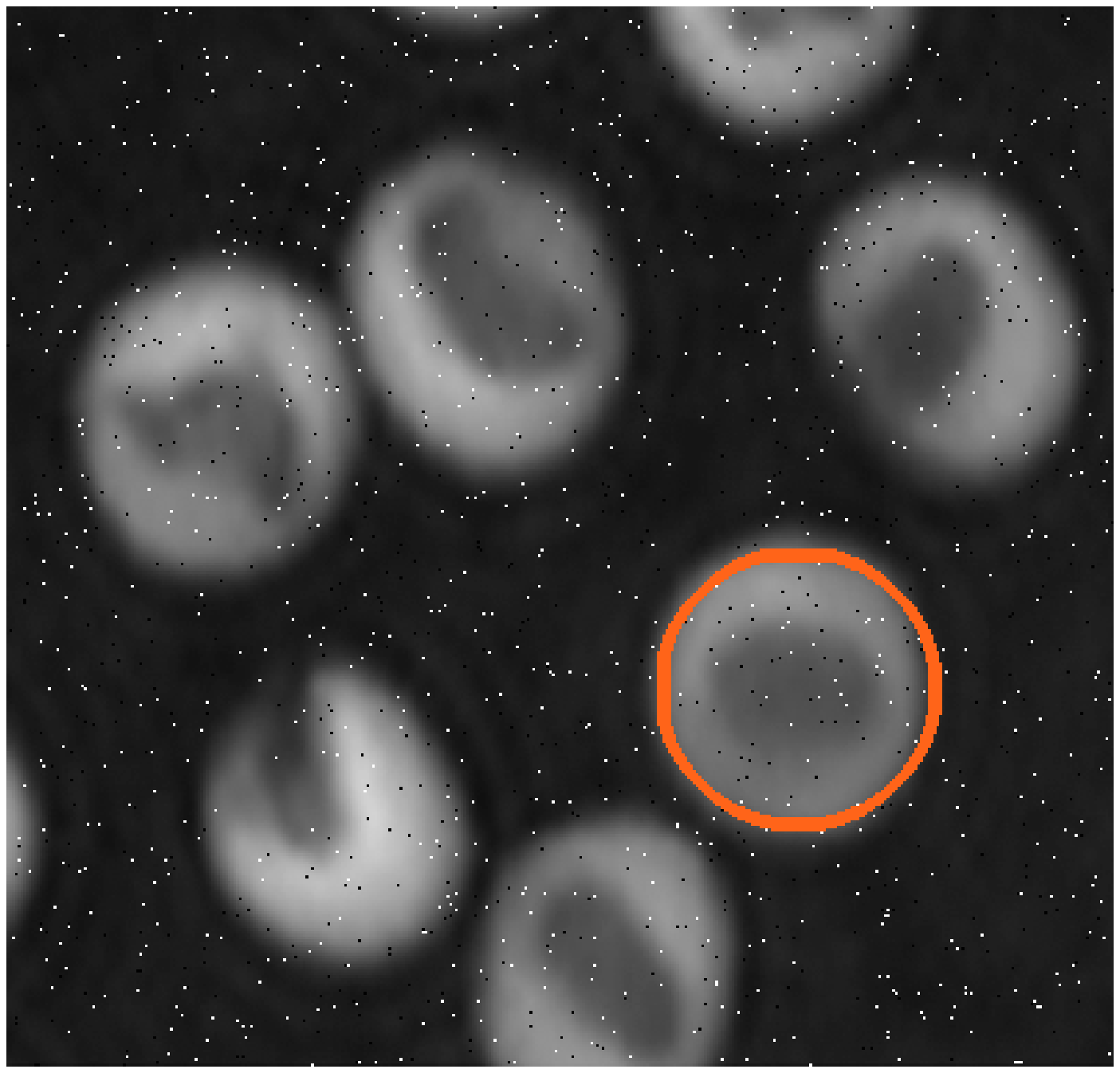,height=0.13\textwidth, width=0.2\textwidth}
\hspace{0.000001\textwidth}
\psfig{figure=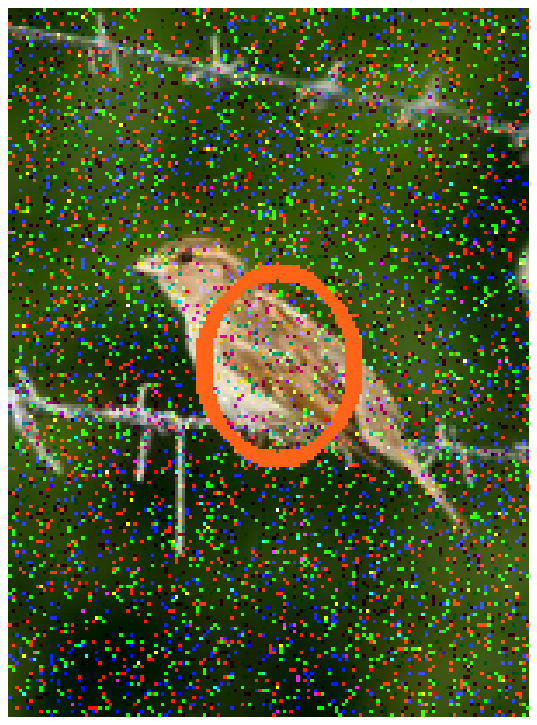,height=0.13\textwidth, width=0.2\textwidth}
\hspace{0.000001\textwidth}
\psfig{figure=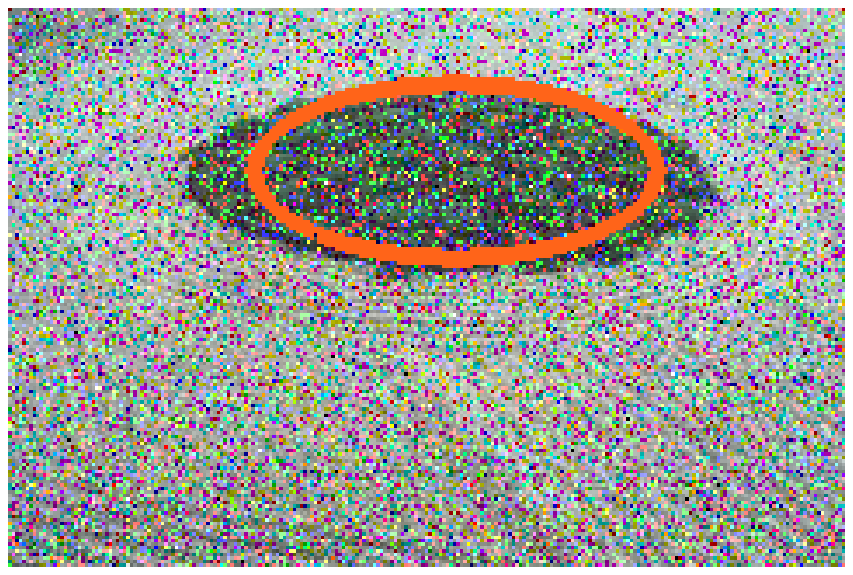,height=0.13\textwidth, width=0.2\textwidth}
\hspace{0.000001\textwidth}
\psfig{figure=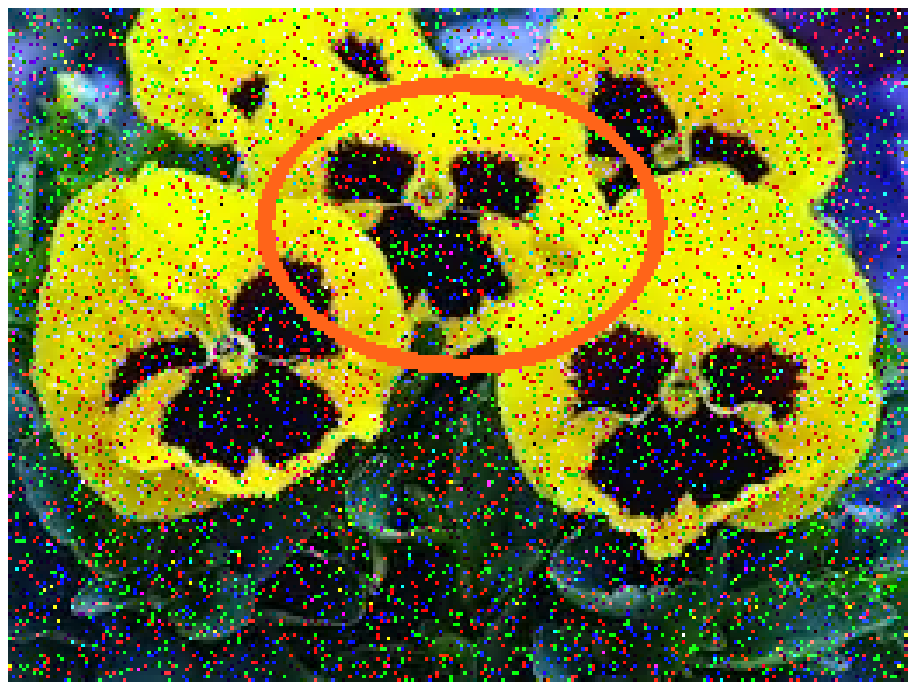,height=0.13\textwidth, width=0.2\textwidth}}
\vspace{0.000001\textwidth}
\centerline{
\psfig{figure=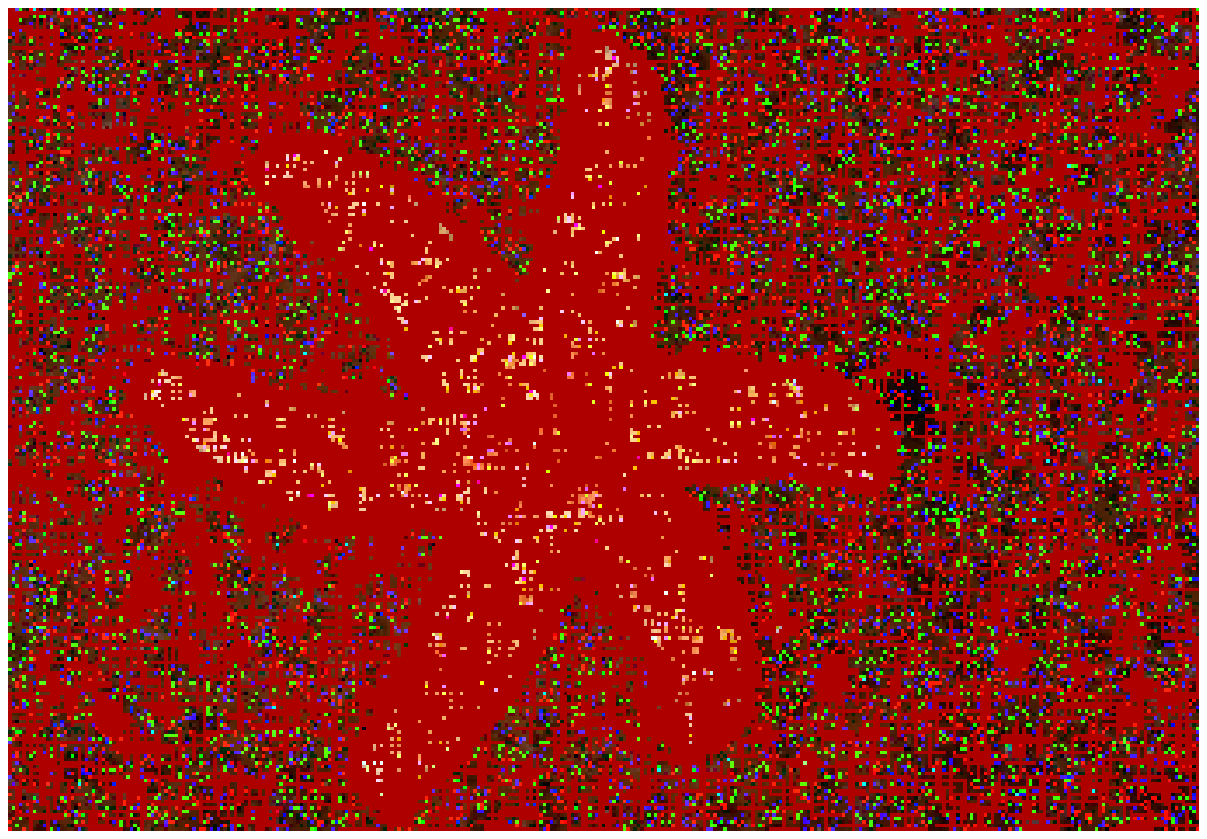,height=0.13\textwidth, width=0.2\textwidth}
\hspace{0.000001\textwidth}
\psfig{figure=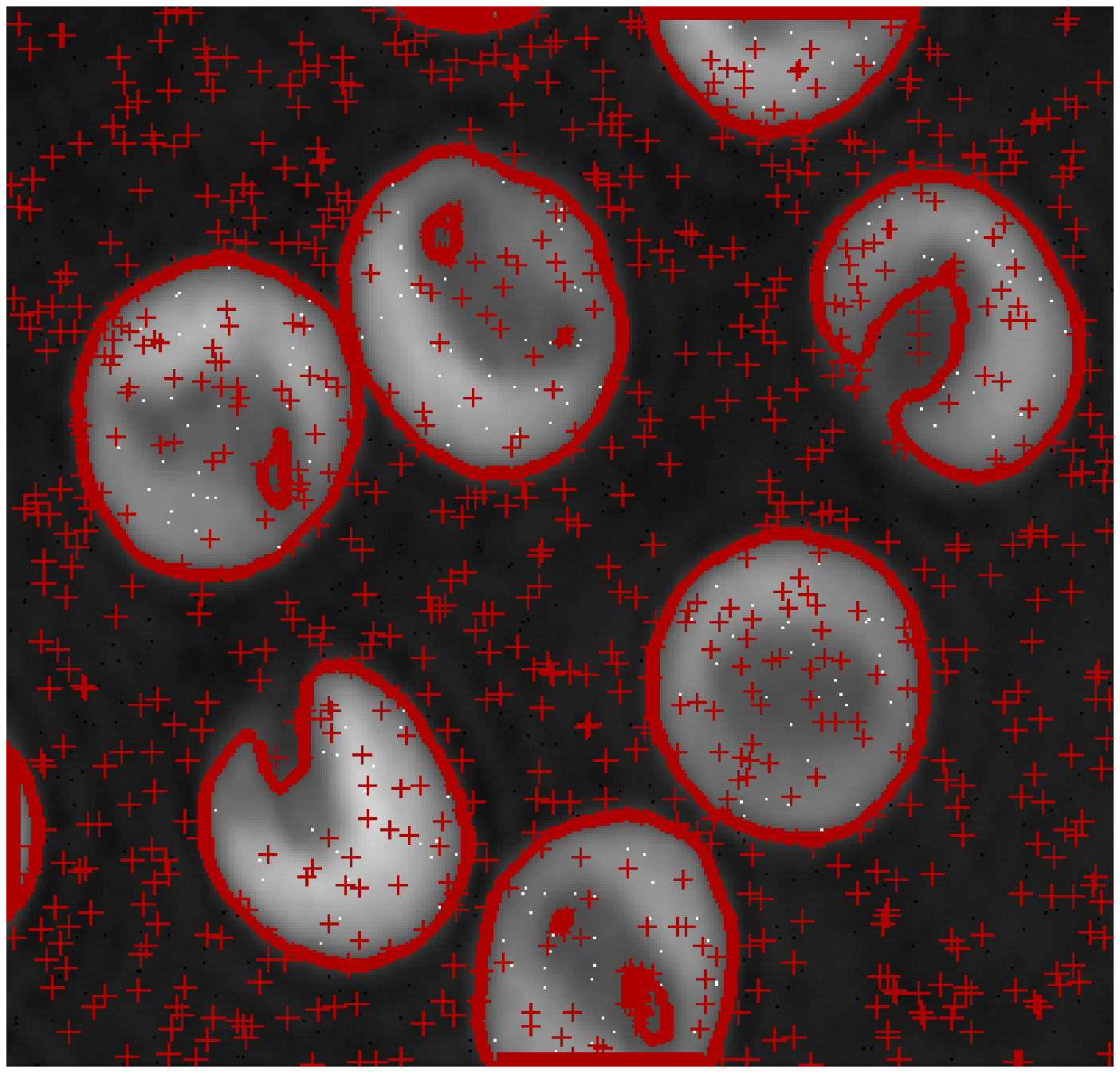,height=0.13\textwidth, width=0.2\textwidth}
\hspace{0.000001\textwidth}
\psfig{figure=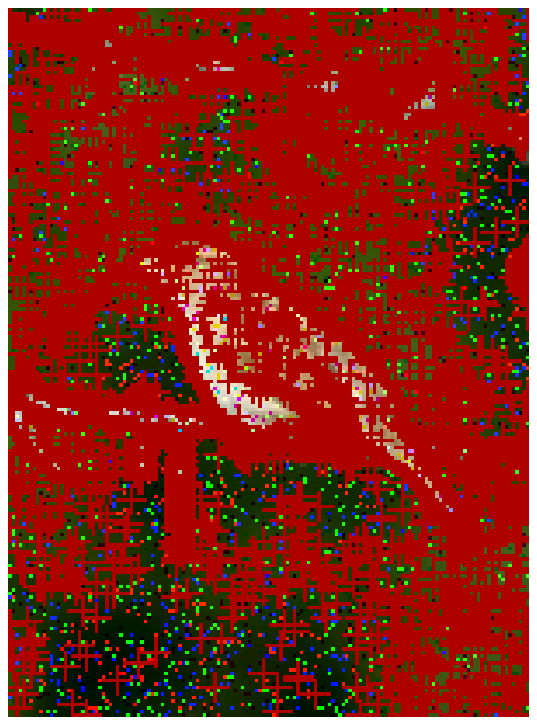,height=0.13\textwidth, width=0.2\textwidth}
\hspace{0.000001\textwidth}
\psfig{figure=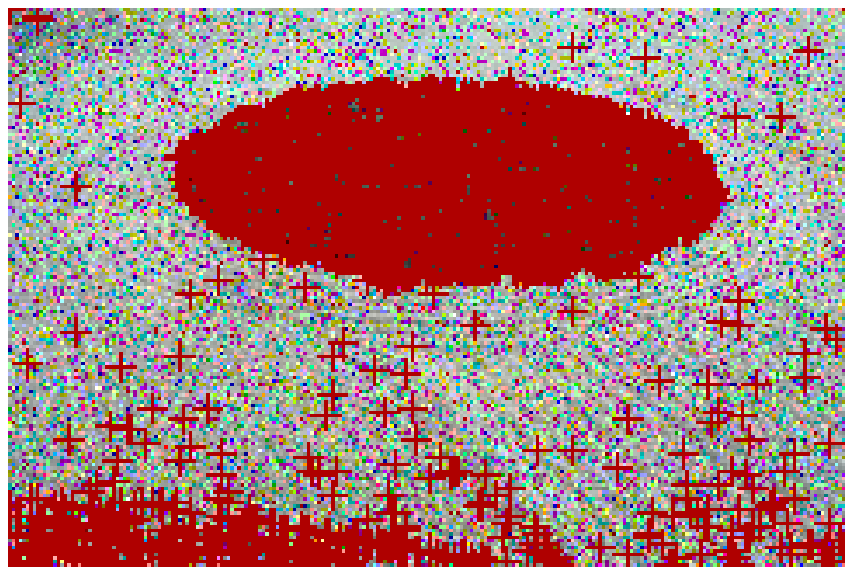,height=0.13\textwidth, width=0.2\textwidth}
\hspace{0.000001\textwidth}
\psfig{figure=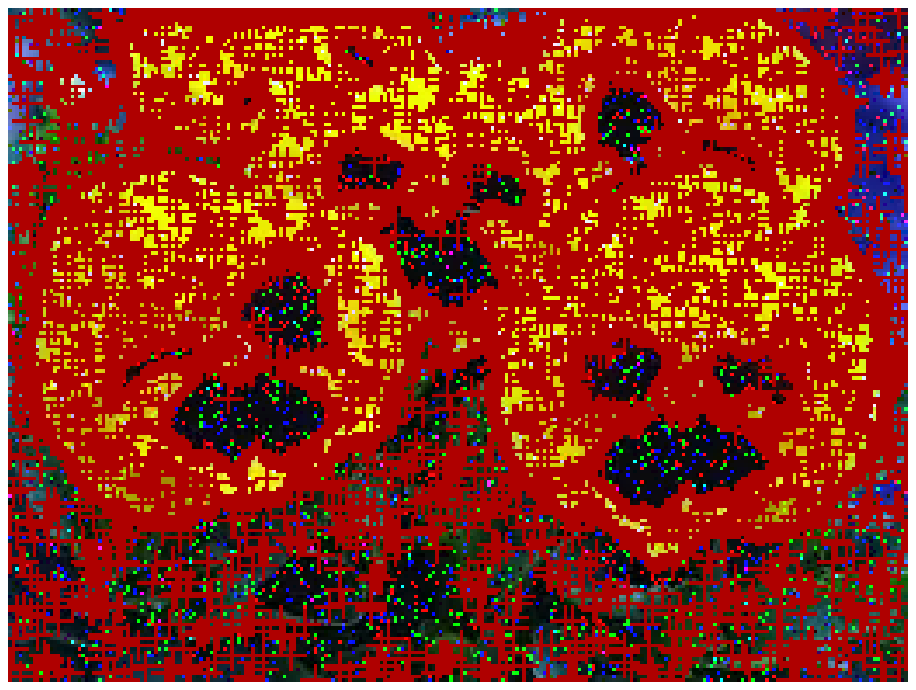,height=0.13\textwidth, width=0.2\textwidth}}
\vspace{0.000001\textwidth}
\centerline{
\psfig{figure=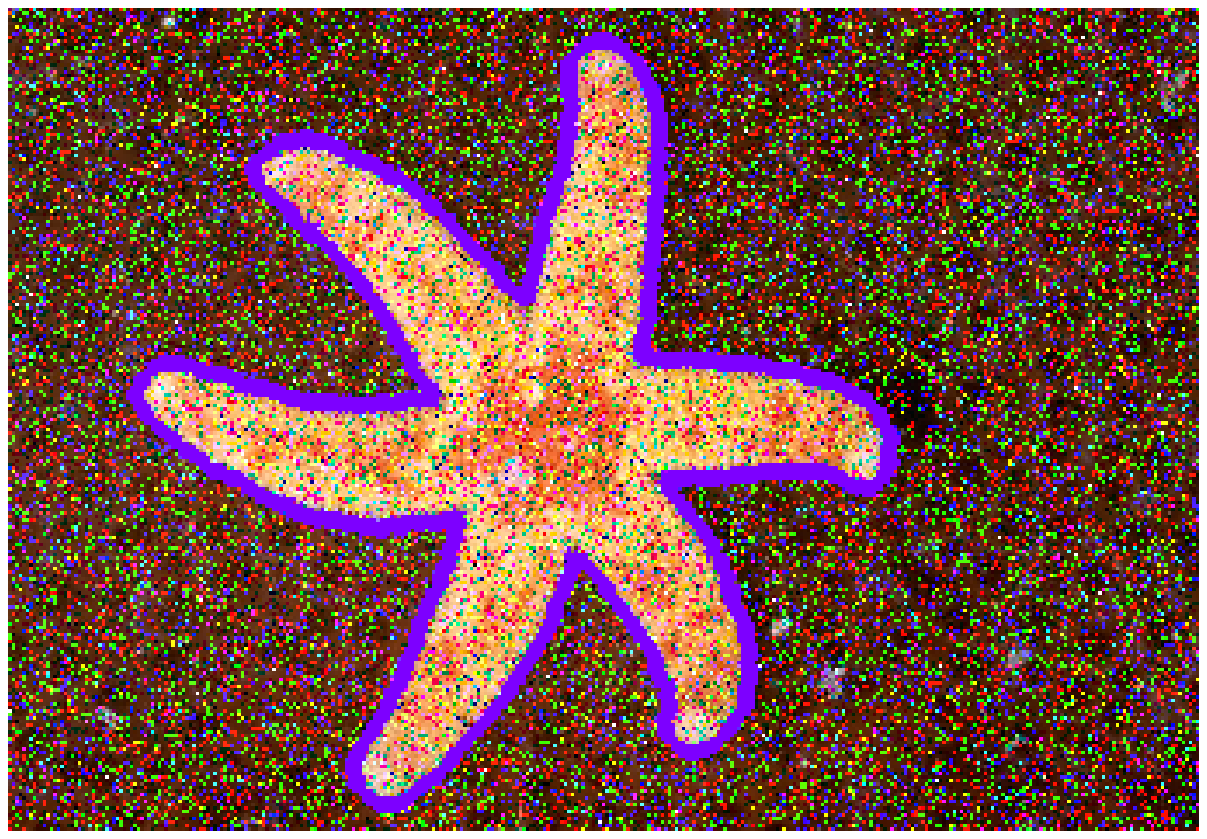,height=0.13\textwidth, width=0.2\textwidth}
\hspace{0.000001\textwidth}
\psfig{figure=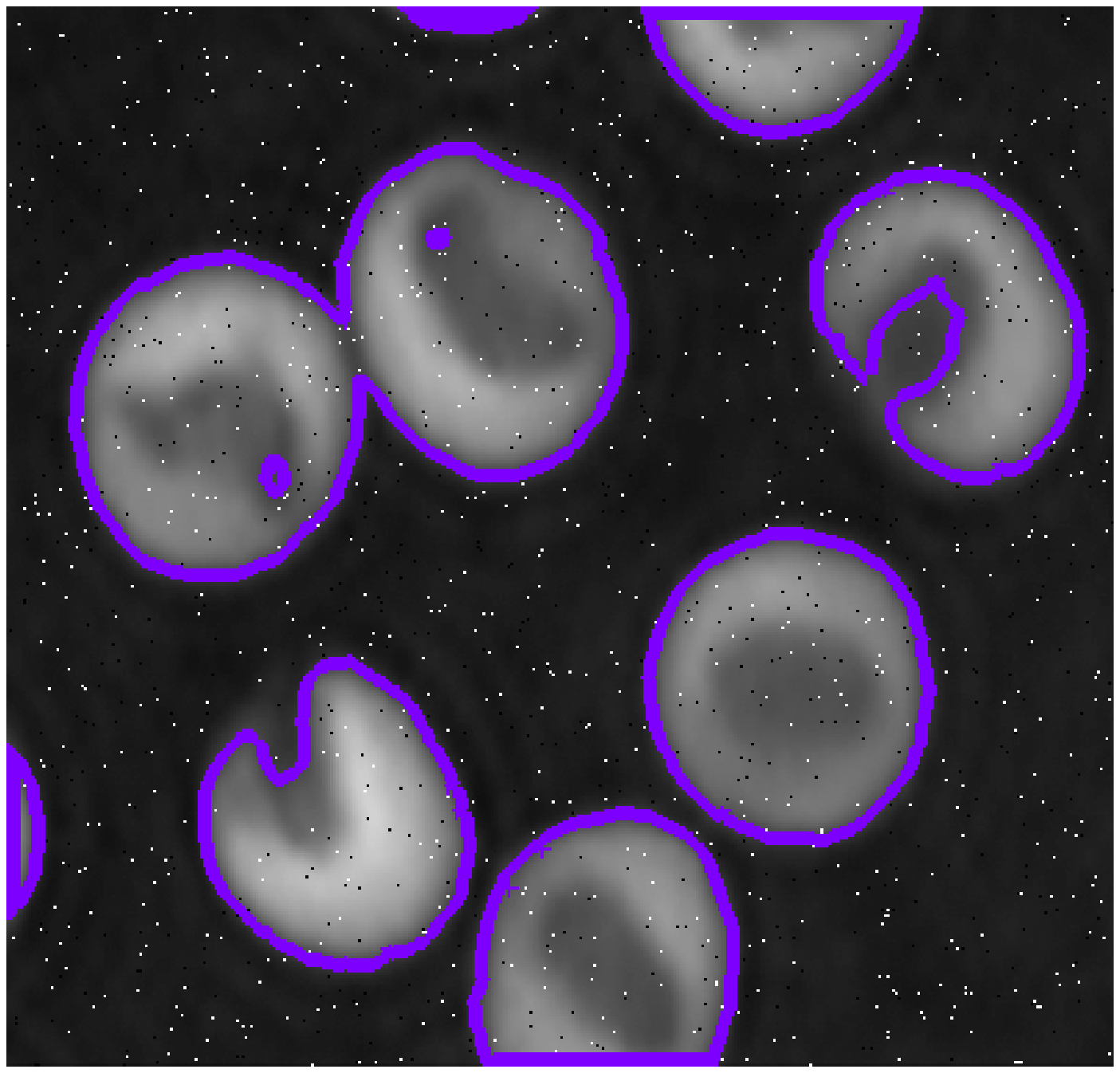,height=0.13\textwidth, width=0.2\textwidth}
\hspace{0.000001\textwidth}
\psfig{figure=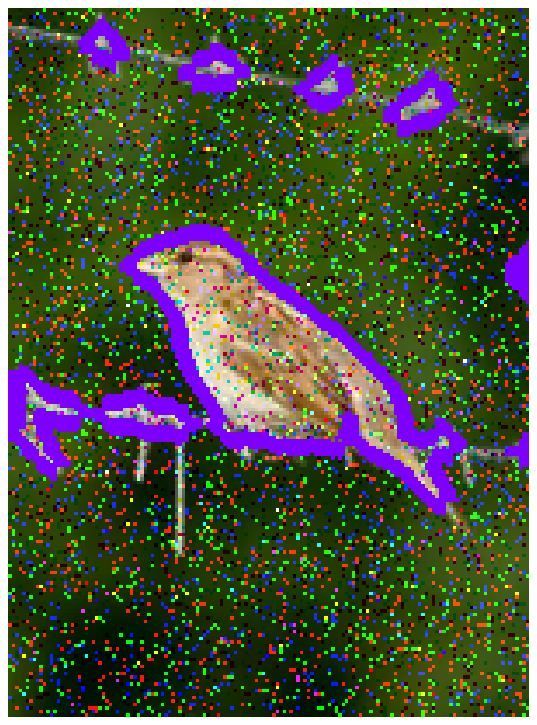,height=0.13\textwidth, width=0.2\textwidth}
\hspace{0.000001\textwidth}
\psfig{figure=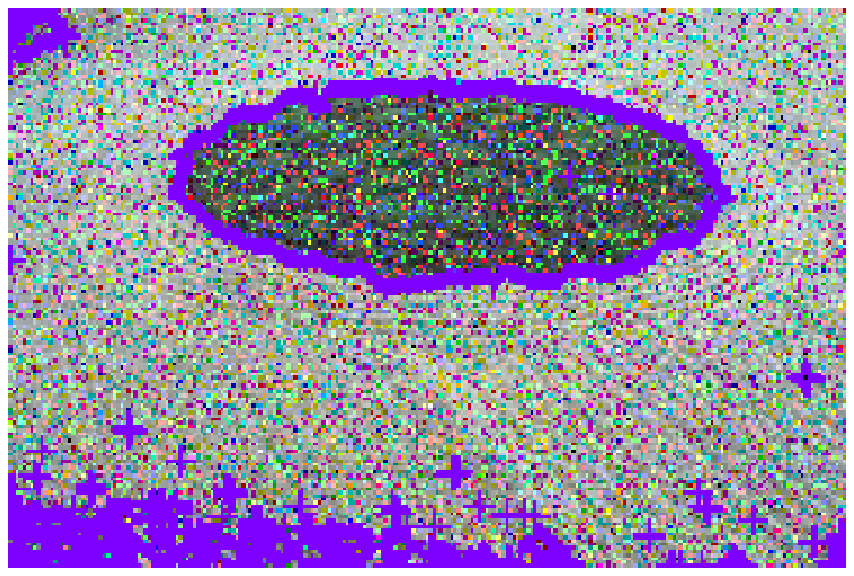,height=0.13\textwidth, width=0.2\textwidth}
\hspace{0.000001\textwidth}
\psfig{figure=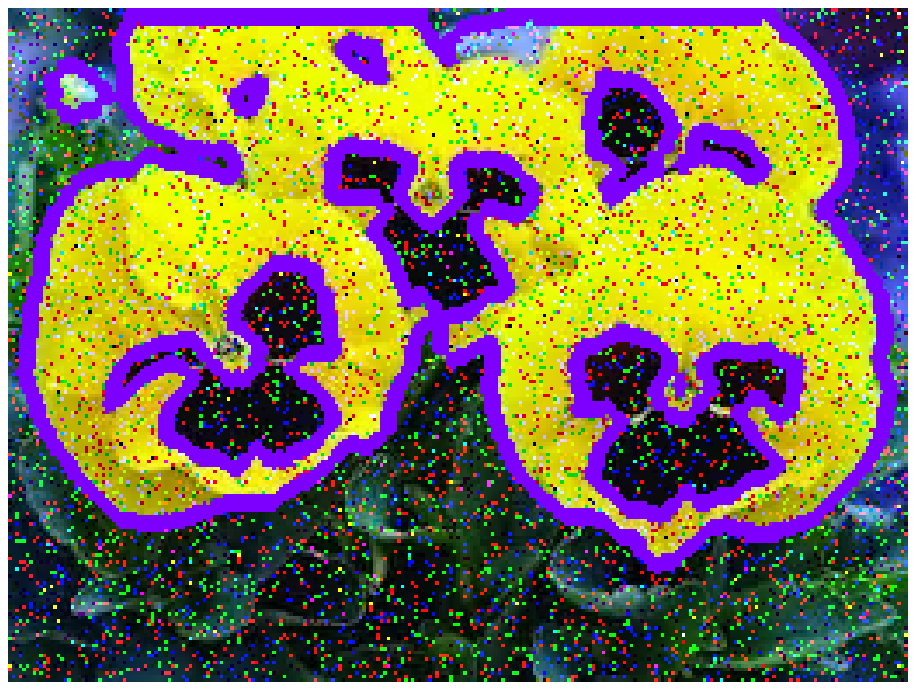,height=0.13\textwidth, width=0.2\textwidth}}
\vspace{0.000001\textwidth}
\centerline{
\psfig{figure=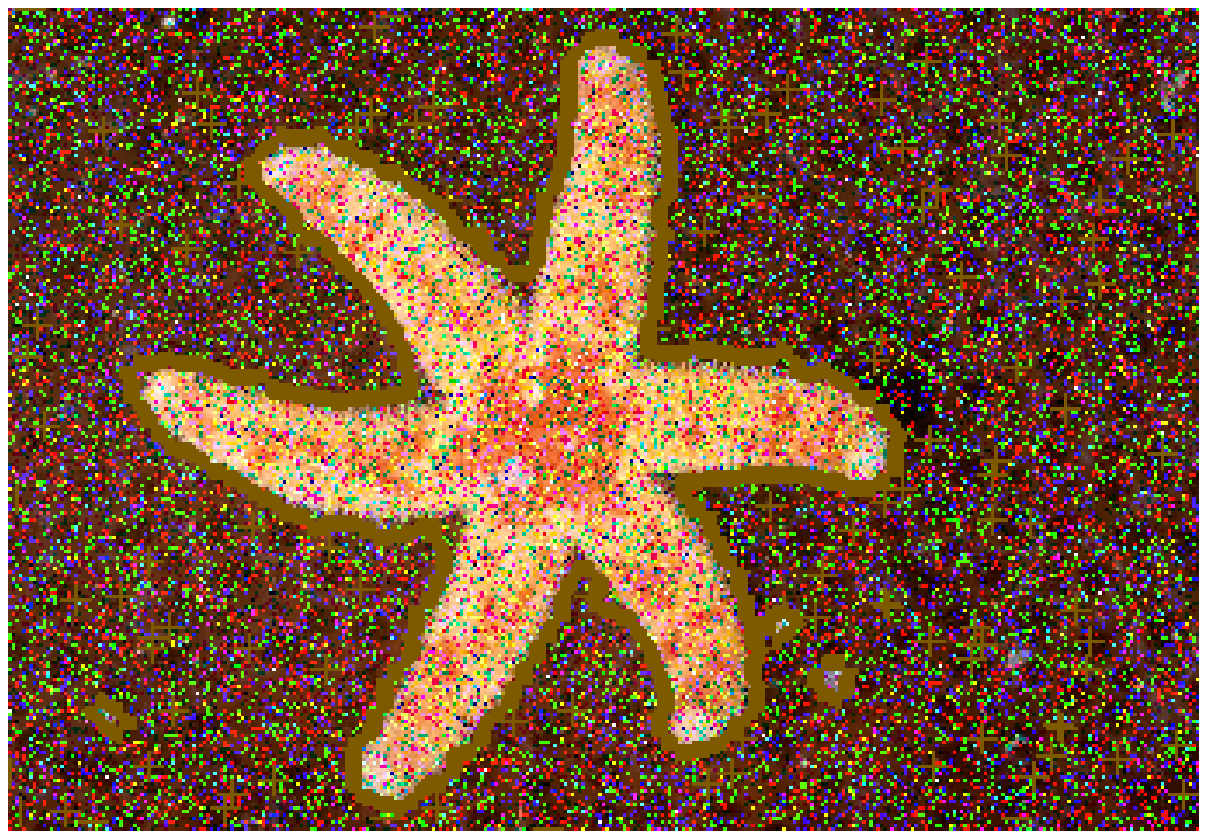,height=0.13\textwidth, width=0.2\textwidth}
\hspace{0.000001\textwidth}
\psfig{figure=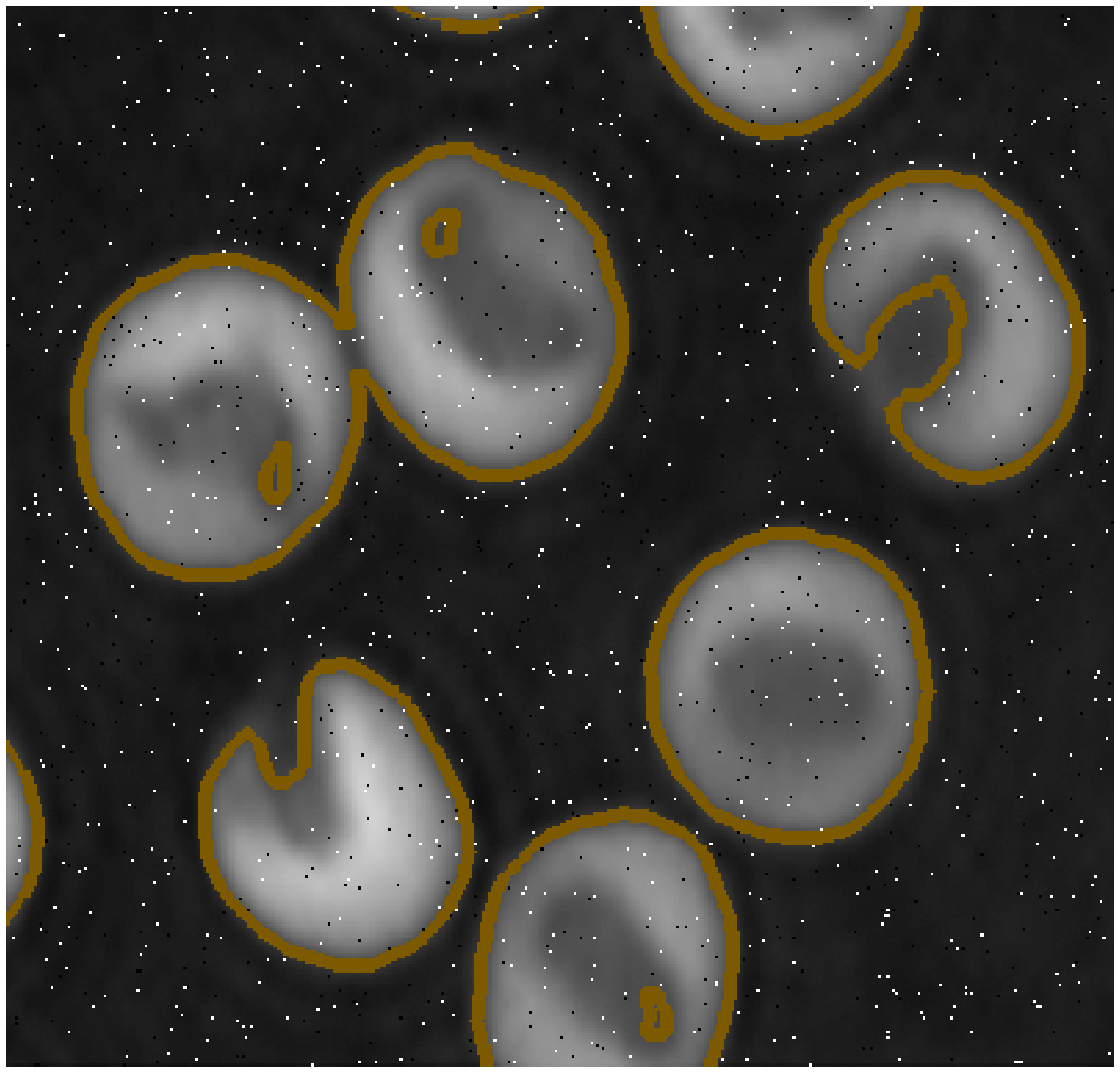,height=0.13\textwidth, width=0.2\textwidth}
\hspace{0.000001\textwidth}
\psfig{figure=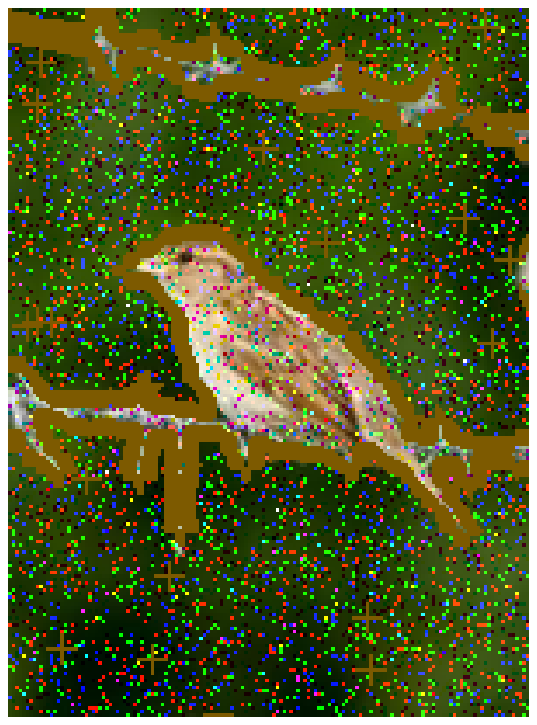,height=0.13\textwidth, width=0.2\textwidth}
\hspace{0.000001\textwidth}
\psfig{figure=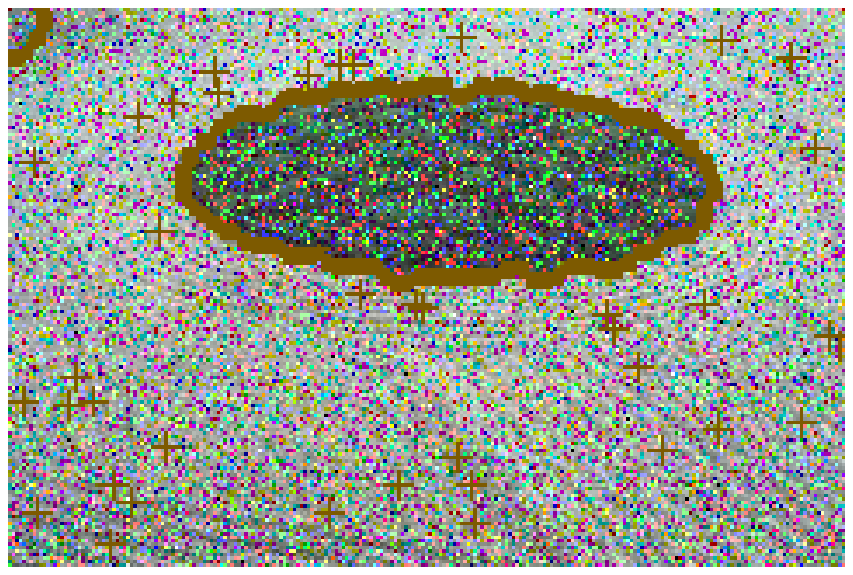,height=0.13\textwidth, width=0.2\textwidth}
\hspace{0.000001\textwidth}
\psfig{figure=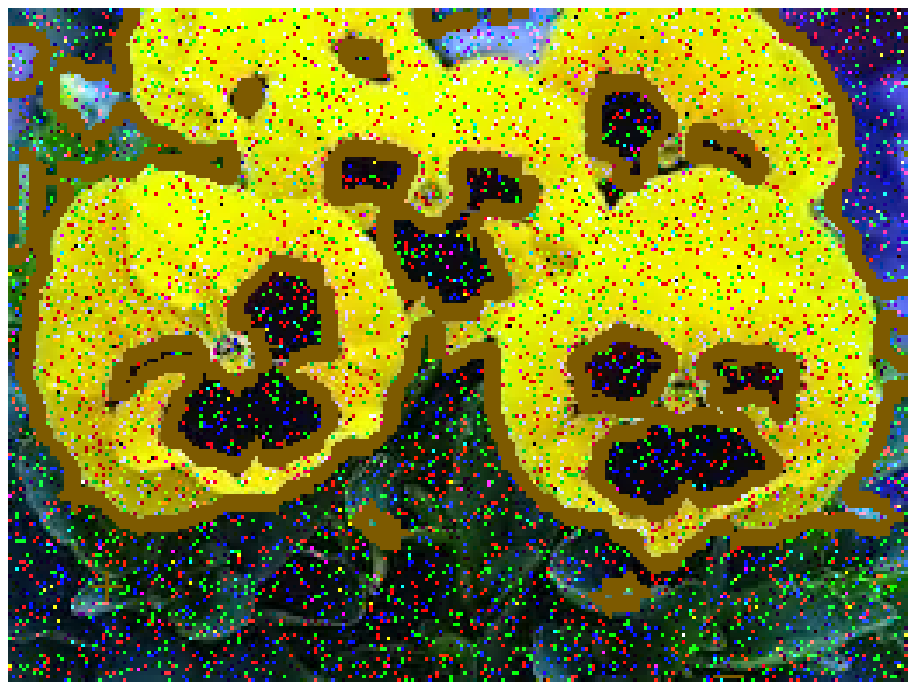,height=0.13\textwidth, width=0.2\textwidth}}
\vspace{0.000001\textwidth}
\centerline{
\psfig{figure=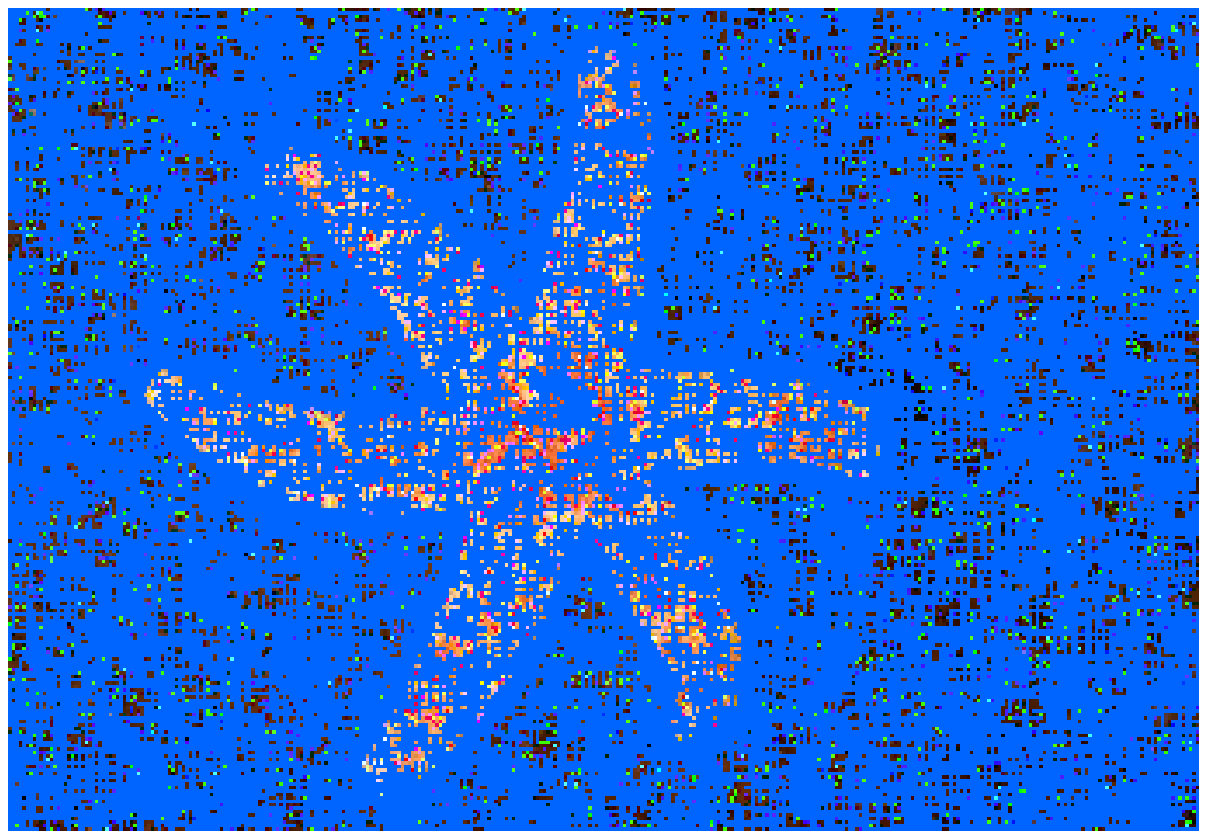,height=0.13\textwidth, width=0.2\textwidth}
\hspace{0.000001\textwidth}
\psfig{figure=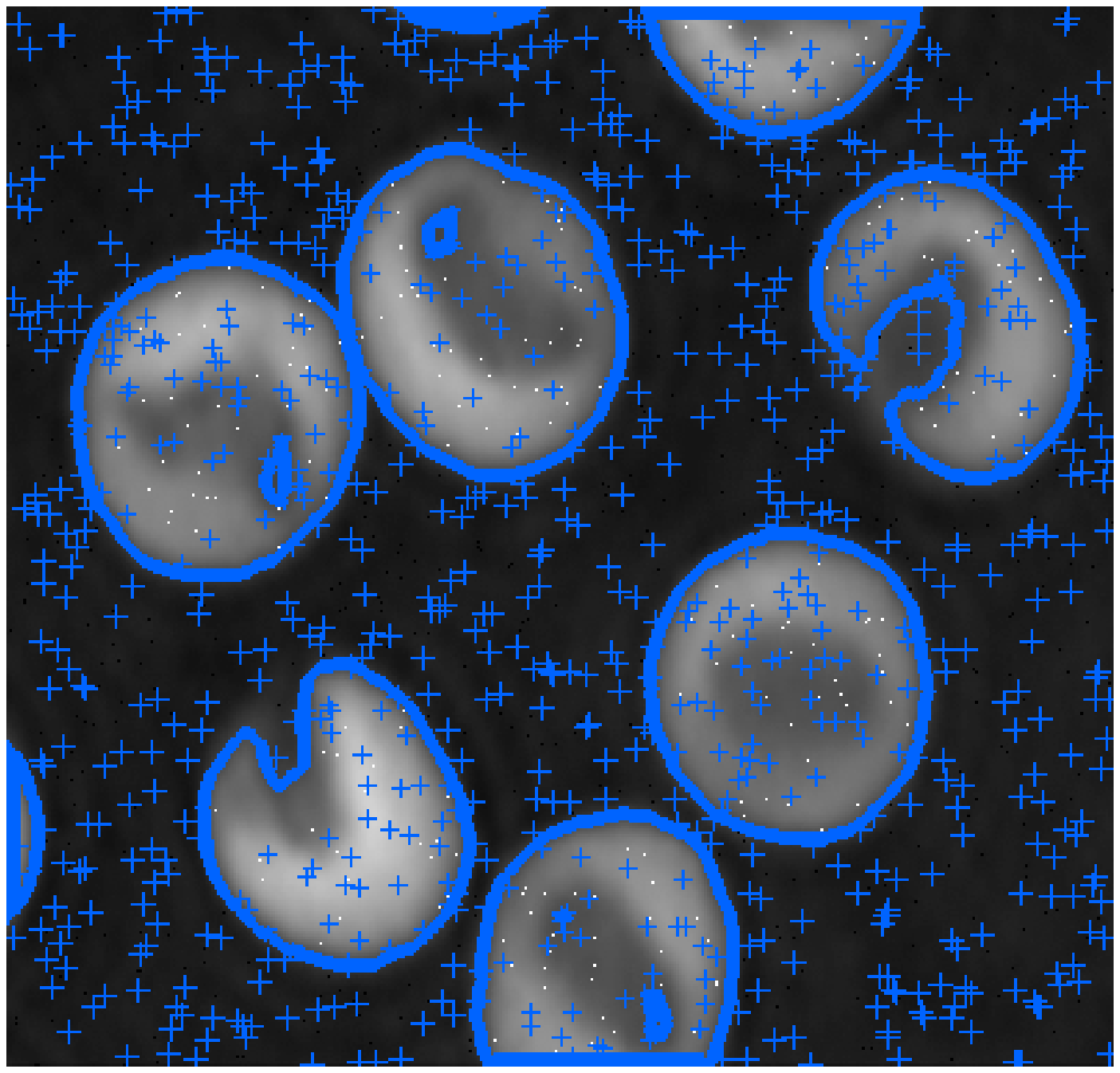,height=0.13\textwidth, width=0.2\textwidth}
\hspace{0.000001\textwidth}
\psfig{figure=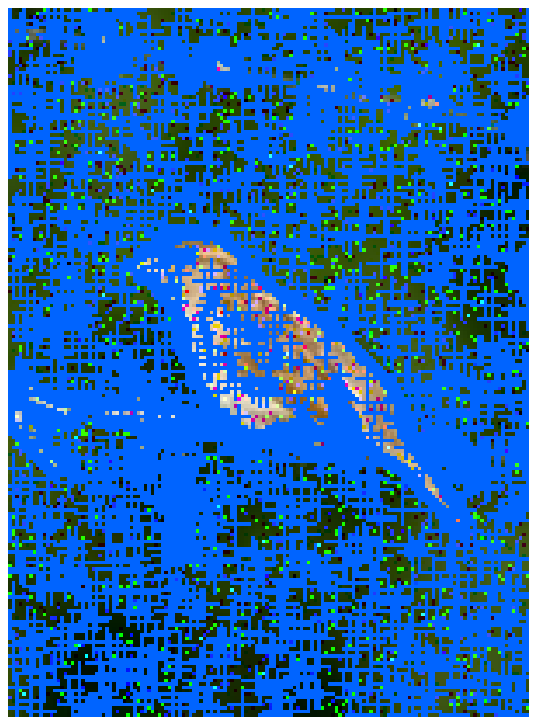,height=0.13\textwidth, width=0.2\textwidth}
\hspace{0.000001\textwidth}
\psfig{figure=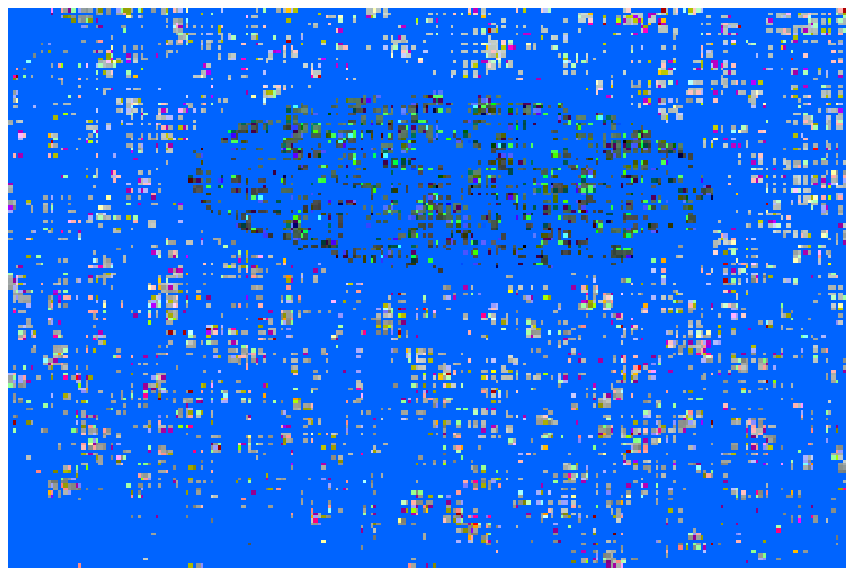,height=0.13\textwidth, width=0.2\textwidth}
\hspace{0.000001\textwidth}
\psfig{figure=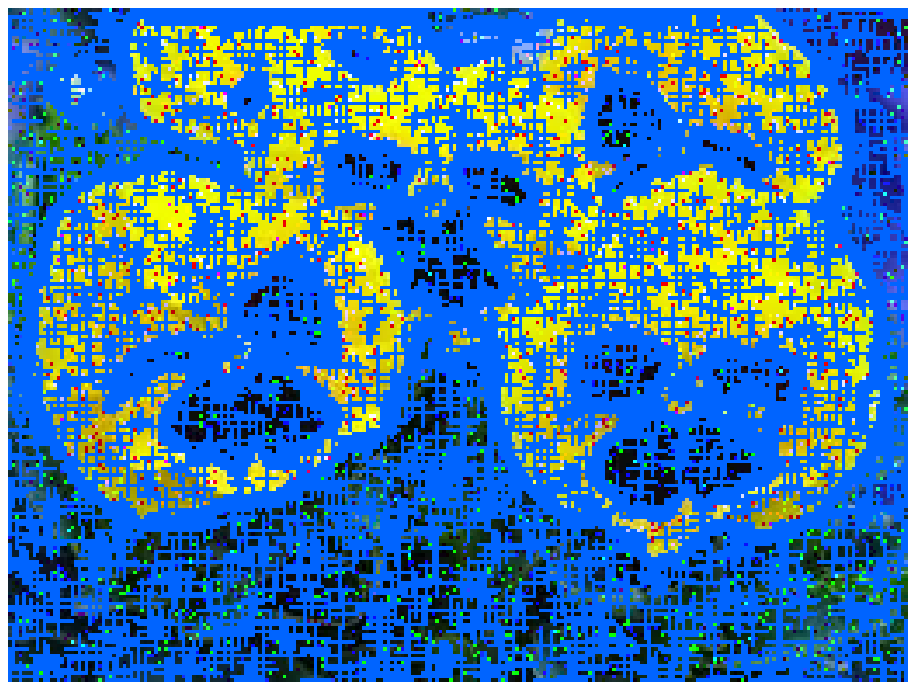,height=0.13\textwidth, width=0.2\textwidth}}
\vspace{0.000001\textwidth}
\centerline{
\psfig{figure=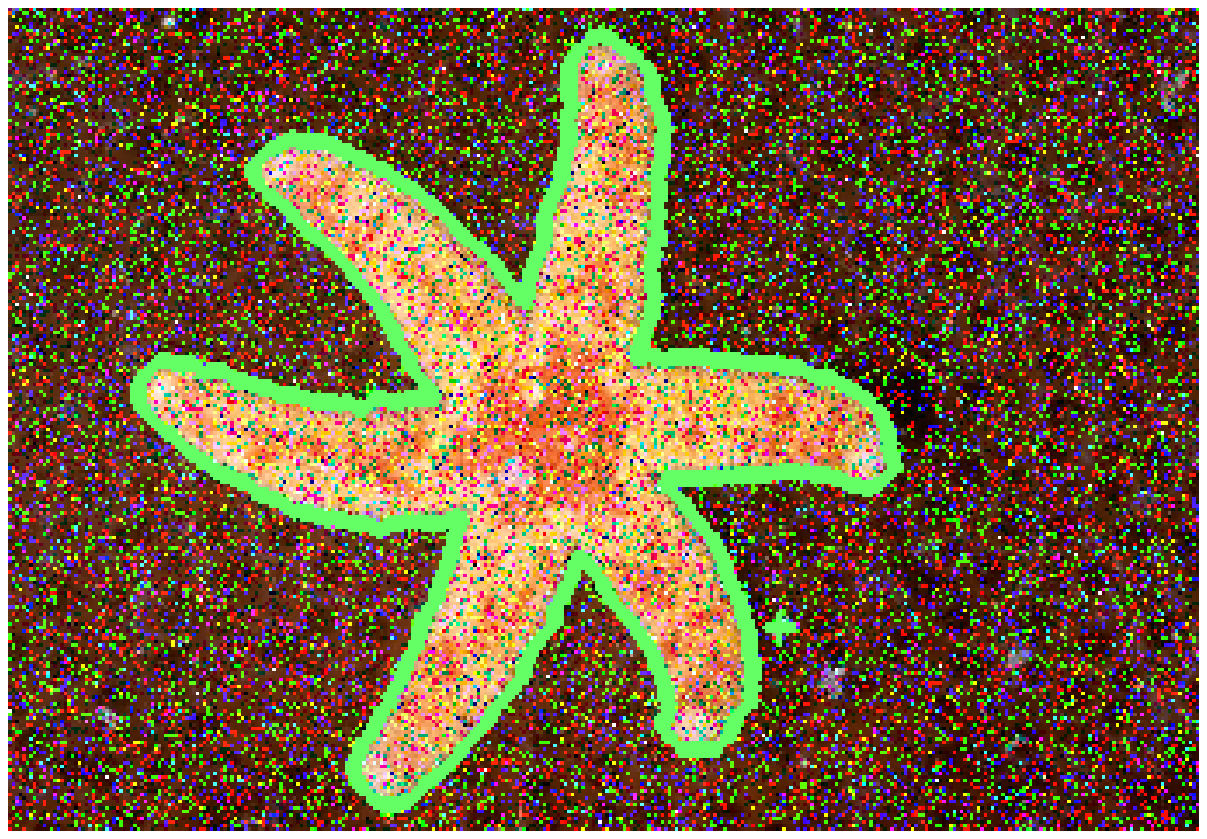,height=0.13\textwidth, width=0.2\textwidth}
\hspace{0.000001\textwidth}
\psfig{figure=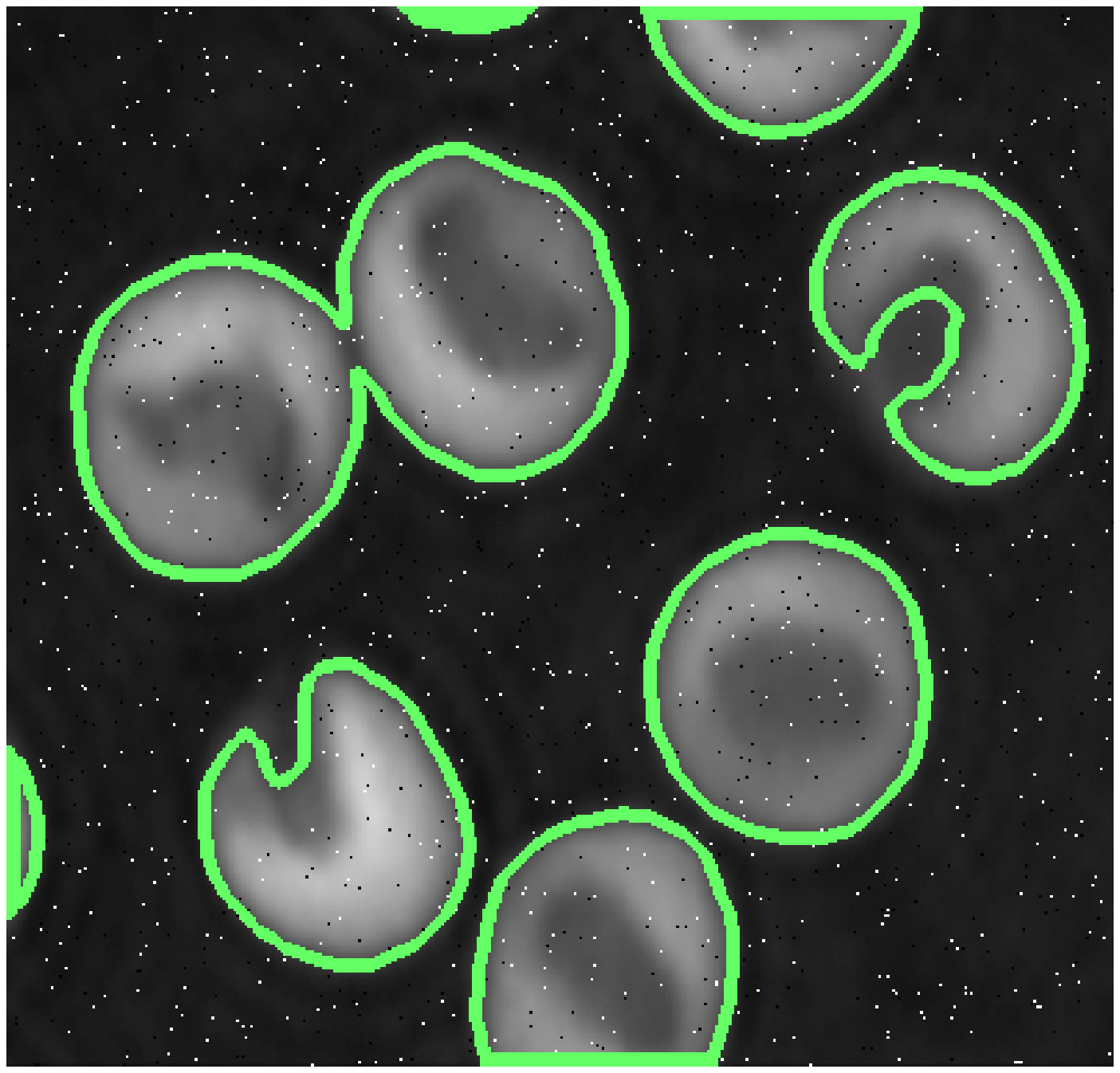,height=0.13\textwidth, width=0.2\textwidth}
\hspace{0.000001\textwidth}
\psfig{figure=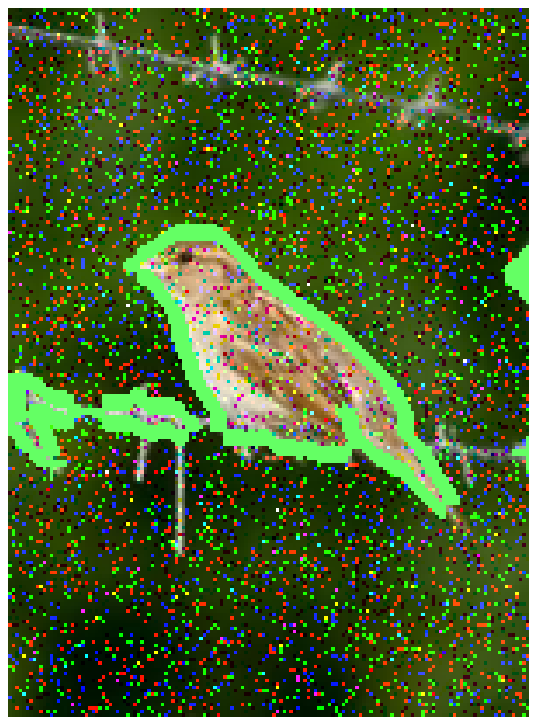,height=0.13\textwidth, width=0.2\textwidth}
\hspace{0.000001\textwidth}
\psfig{figure=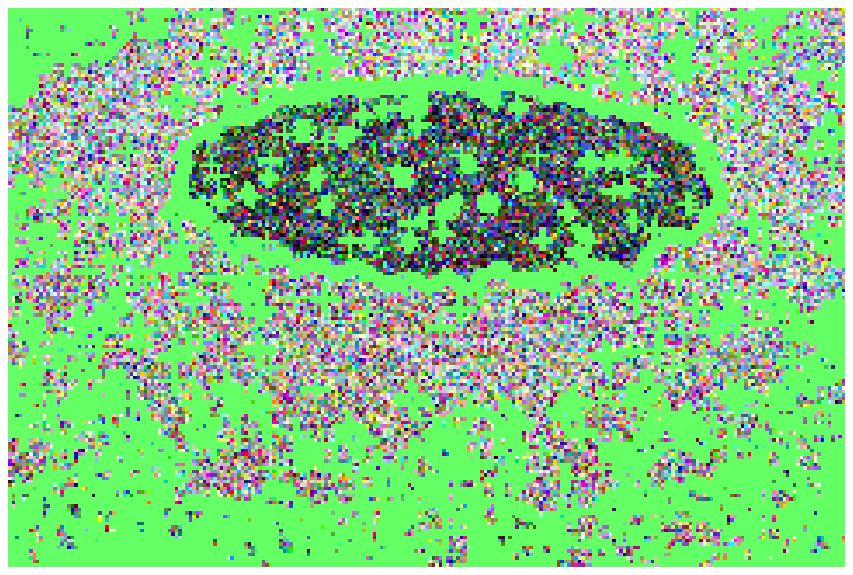,height=0.13\textwidth, width=0.2\textwidth}
\hspace{0.000001\textwidth}
\psfig{figure=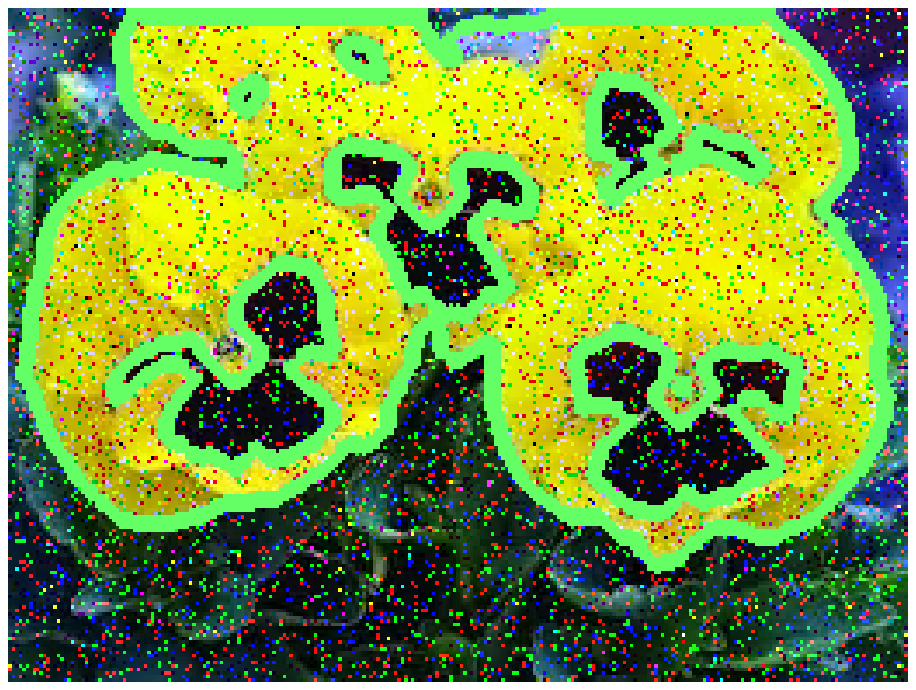,height=0.13\textwidth, width=0.2\textwidth}}
\vspace{0.000001\textwidth}
\centerline{
\psfig{figure=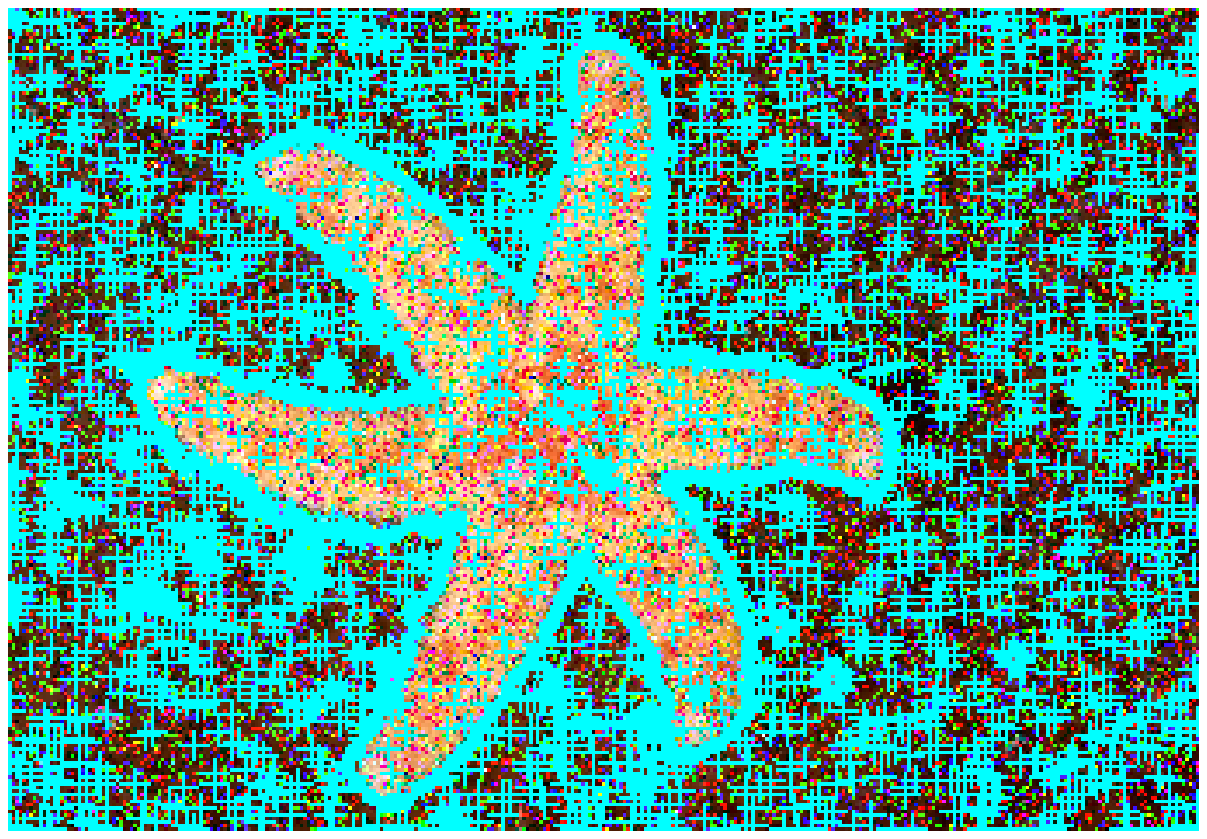,height=0.13\textwidth, width=0.2\textwidth}
\hspace{0.000001\textwidth}
\psfig{figure=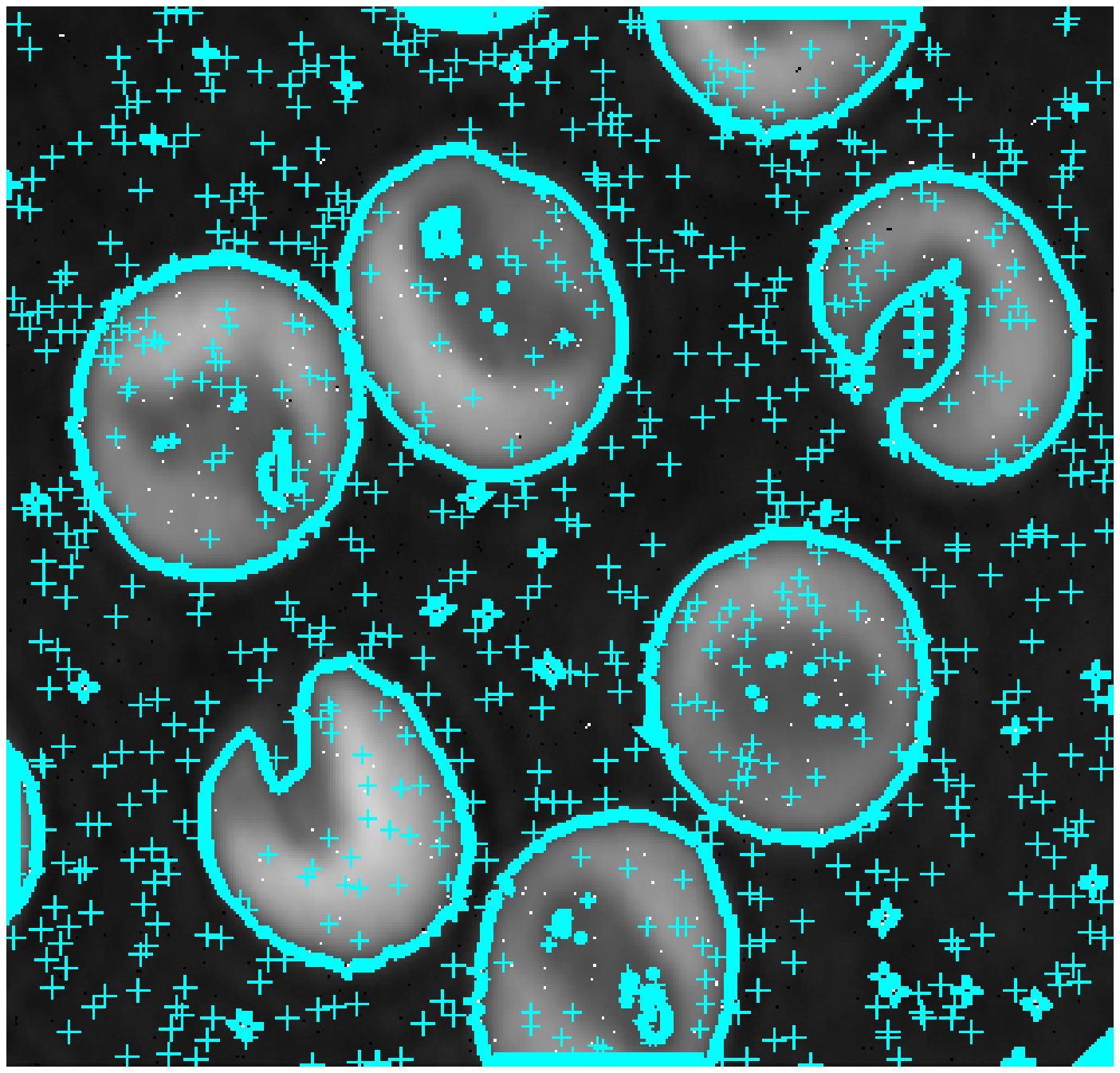,height=0.13\textwidth, width=0.2\textwidth}
\hspace{0.000001\textwidth}
\psfig{figure=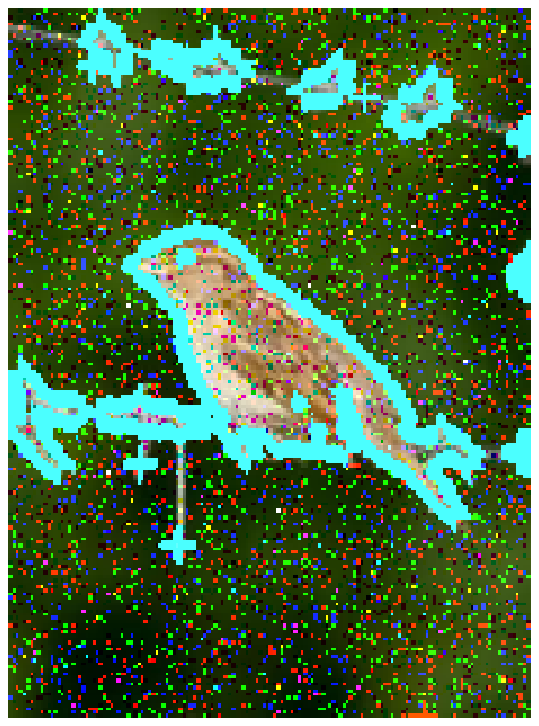,height=0.13\textwidth, width=0.2\textwidth}
\hspace{0.000001\textwidth}
\psfig{figure=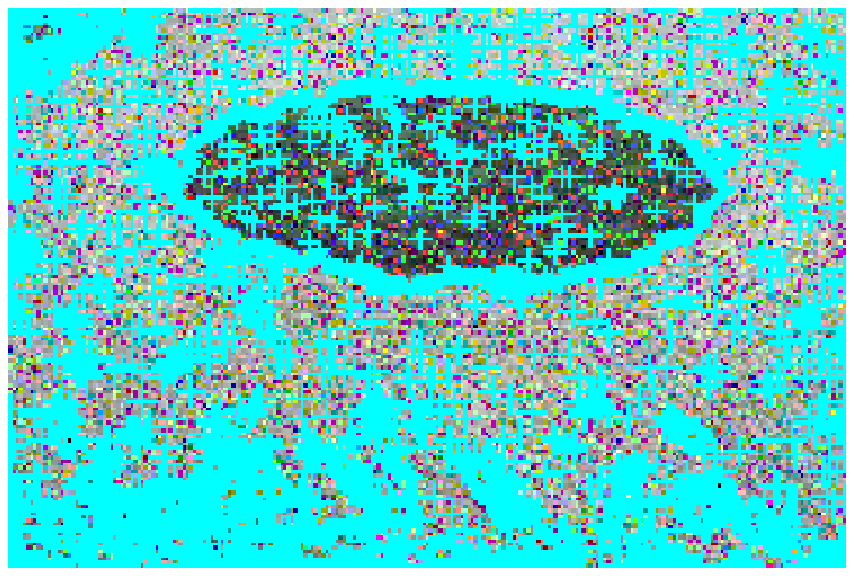,height=0.13\textwidth, width=0.2\textwidth}
\hspace{0.000001\textwidth}
\psfig{figure=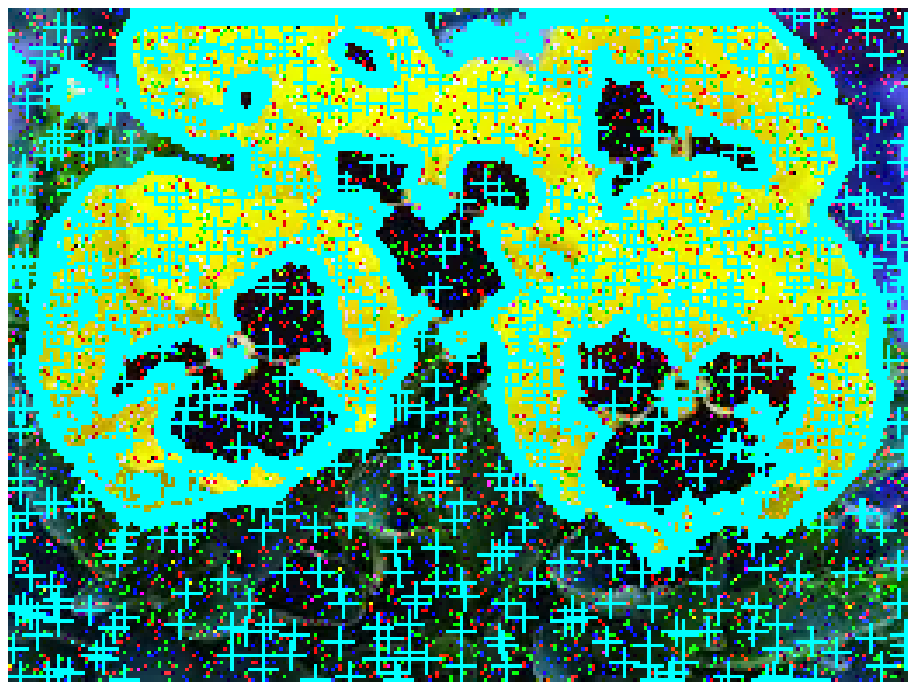,height=0.13\textwidth, width=0.2\textwidth}}
\vspace{0.000001\textwidth}
\centerline{
\psfig{figure=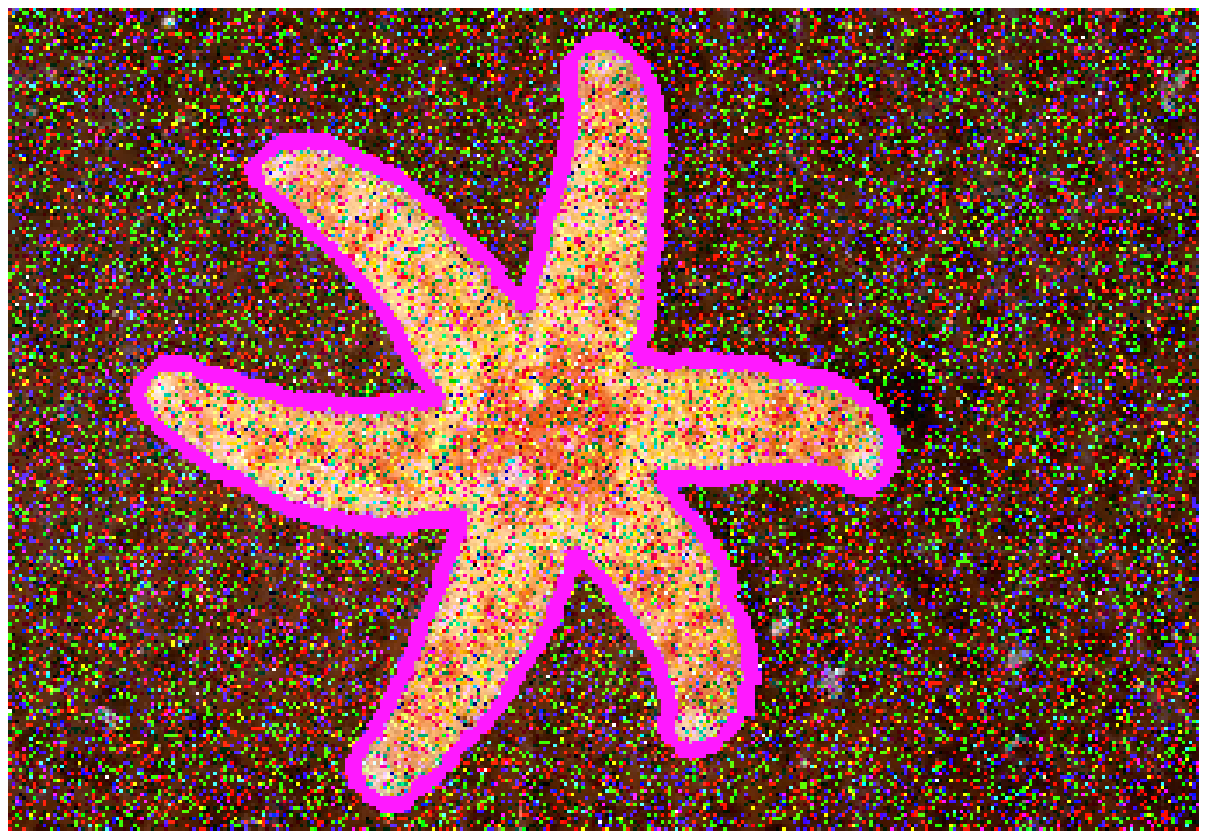,height=0.13\textwidth, width=0.2\textwidth}
\hspace{0.000001\textwidth}
\psfig{figure=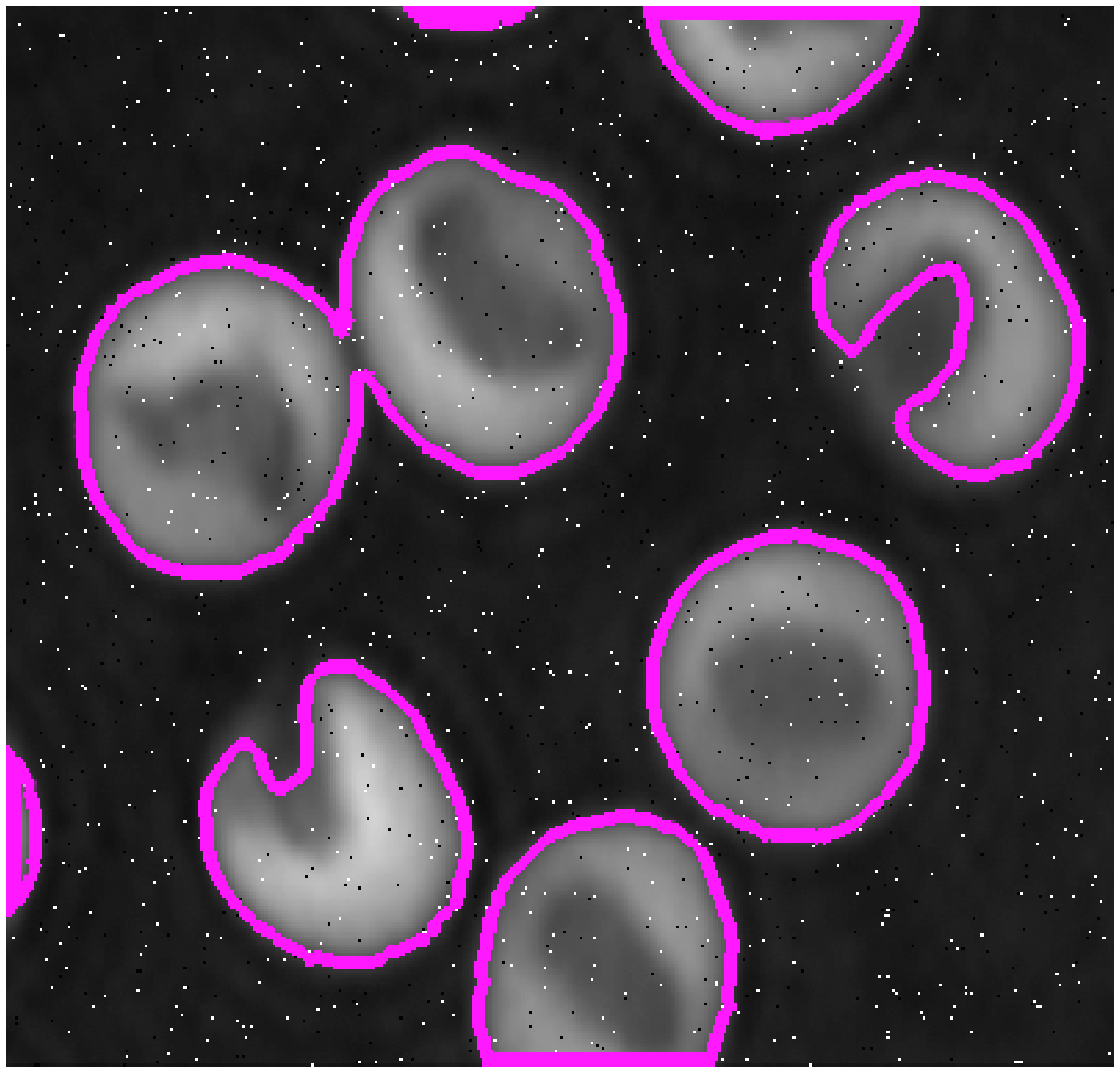,height=0.13\textwidth, width=0.2\textwidth}
\hspace{0.000001\textwidth}
\psfig{figure=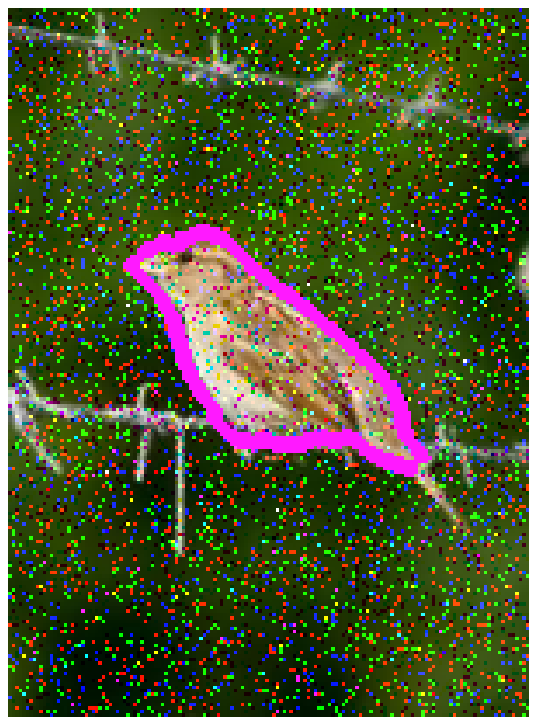,height=0.13\textwidth, width=0.2\textwidth}
\hspace{0.000001\textwidth}
\psfig{figure=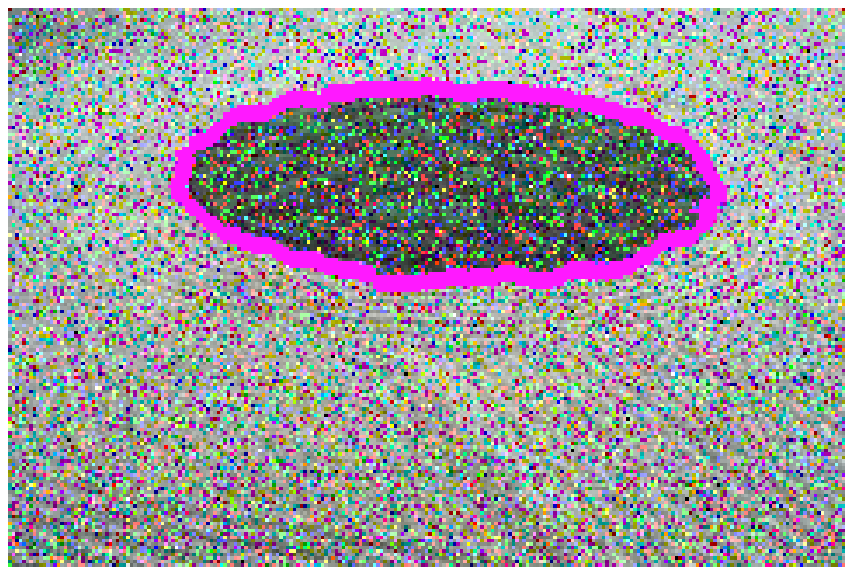,height=0.13\textwidth, width=0.2\textwidth}
\hspace{0.000001\textwidth}
\psfig{figure=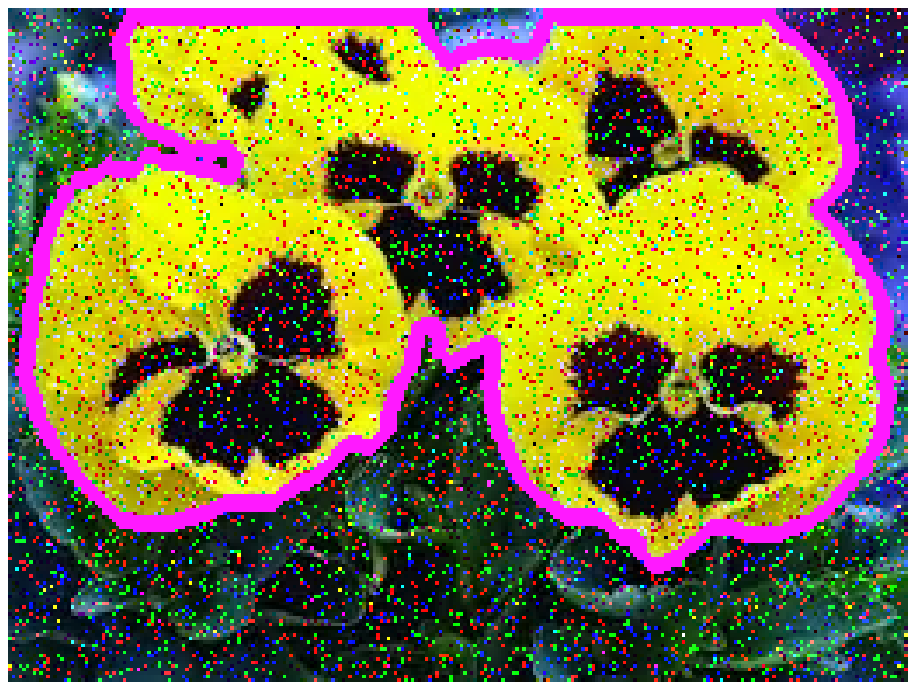,height=0.13\textwidth, width=0.2\textwidth}}
\centerline{(a)\hspace{0.17\textwidth} (b)\hspace{0.17\textwidth} (c)\hspace{0.17\textwidth} (d)\hspace{0.17\textwidth} (e)}
\caption{Visual comparison of segmentation results of corrupted (a) Starfish image with salt \& pepper noise (with probability $0.2$), (b) Cell5 image with salt \& pepper noise (with probability $0.01$), (c) Bird image with salt \& pepper noise (with probability $0.1$), (d) 86016 image with salt \& pepper noise (with probability $0.2$), (e) Yellow Pansy image with salt \& pepper noise (with probability $0.1$) using various techniques. \textbf{First row:} images with initial contours, \textbf{Second row:} segmentation results obtained by {\sc FEAC}, \textbf{Third row:} segmentation results obtained by {\sc NFACMKM}, \textbf{Fourth row:} segmentation results obtained  by {\sc FACGK}, \textbf{Fifth row:} segmentation results obtained by {\sc LPFAC}, \textbf{Sixth row:} segmentation results obtained by {\sc FDFEAC}, \textbf{Seventh row:} segmentation results obtained by {\sc GLFEAC} and \textbf{Eighth row:} segmentation results obtained by {\sc RGLFEAC}.\label{figure_sp_seg} }
\end{figure}

\subsection{Segmentation of Corrupted Images by Slat \& Pepper Noise}

To analyze the robustness of these considered algorithms, we conducted an experiment to segment images which are corrupted by various amount of salt \& pepper noise. Figure~\ref{figure_sp_seg} shows the segmentation results of distorted images using different algorithms. From this figure, it is observed that methods: {\sc FEAC}, {\sc LPFAC}, {\sc FDFEAC} and {\sc GLFEAC} fail to properly segment images due to noise. However, the kernel based methods: {\sc NFACMKM} and {\sc FACGK} are able to segment images along with small background patches. The local energy term defined based on region homogeneity in {\sc RGLFEAC} helps to properly segment noisy images. Table~\ref{table_salt_pepper} displays statistics of all these methods on segmentation of noisy images. This table also highlights the similar observation like visual results.                

\begin{table}[htp]
\begin{center}
\begin{tabular}{|l|l|l|l|l|l|l|l|}\hline
Measures & \multicolumn{7}{|l|}{Techniques } \\\cline{2-8}
& {\sc m1} & {\sc m2} & {\sc m3} & {\sc m4} & {\sc m5} & {\sc m6} & {\sc m7}\\\cline{1-8}
Ave. Jacard error & 0.278 & 0.175 & 0.194 & 0.276 & 0.278 & 0.357 & \textcolor[rgb]{0.00,1.00,0.00}{0.071} \\\hline
Ave. F-measure    & 0.839& 0.902 & 0.890 & 0.839 &0.816 & 0.781& \textcolor[rgb]{1.00,0.00,0.00}{0.961}  \\\hline
\end{tabular}
\end{center}
\caption{Quantitative comparison among various techniques with respect to average Jacard error and average F-measure over $100$ corrupted images with salt \& pepper noise. {\sc m1: feac}, {\sc m2: nfacmkm}, {\sc m3: facgk}, {\sc m4: lpfac}, {\sc m5: fdfeac}, {\sc m6: glfeac} and {\sc m7: rglfeac}. Green colored numeric value indicates least average Jacard error corresponds to the best segmentation result. Whereas, red colored numeric value indicates highest F-measure corresponds to the best segmentation result. \label{table_salt_pepper}}
\end{table}

\begin{figure}
\centerline{
\psfig{figure=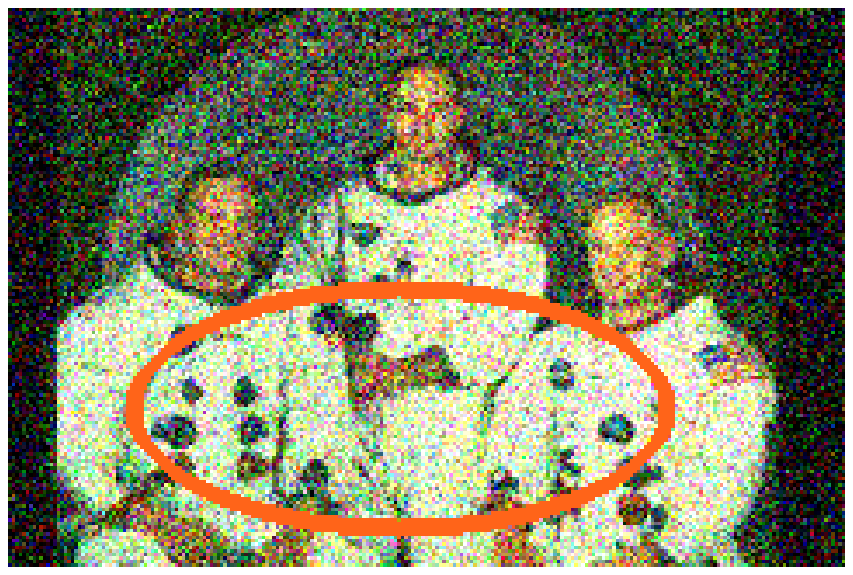,height=0.13\textwidth, width=0.2\textwidth}
\hspace{0.000001\textwidth}
\psfig{figure=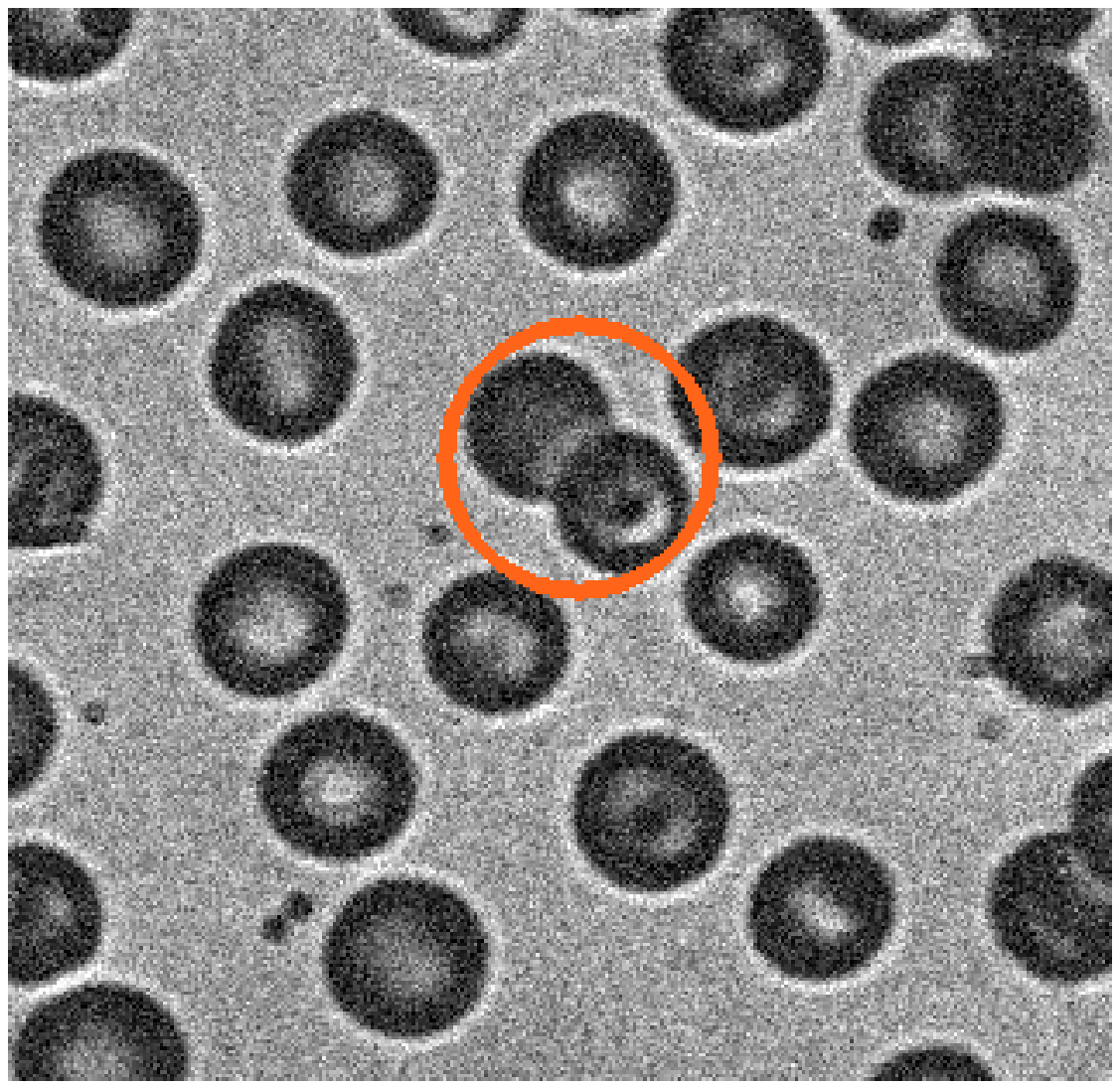,height=0.13\textwidth, width=0.2\textwidth}
\hspace{0.000001\textwidth}
\psfig{figure=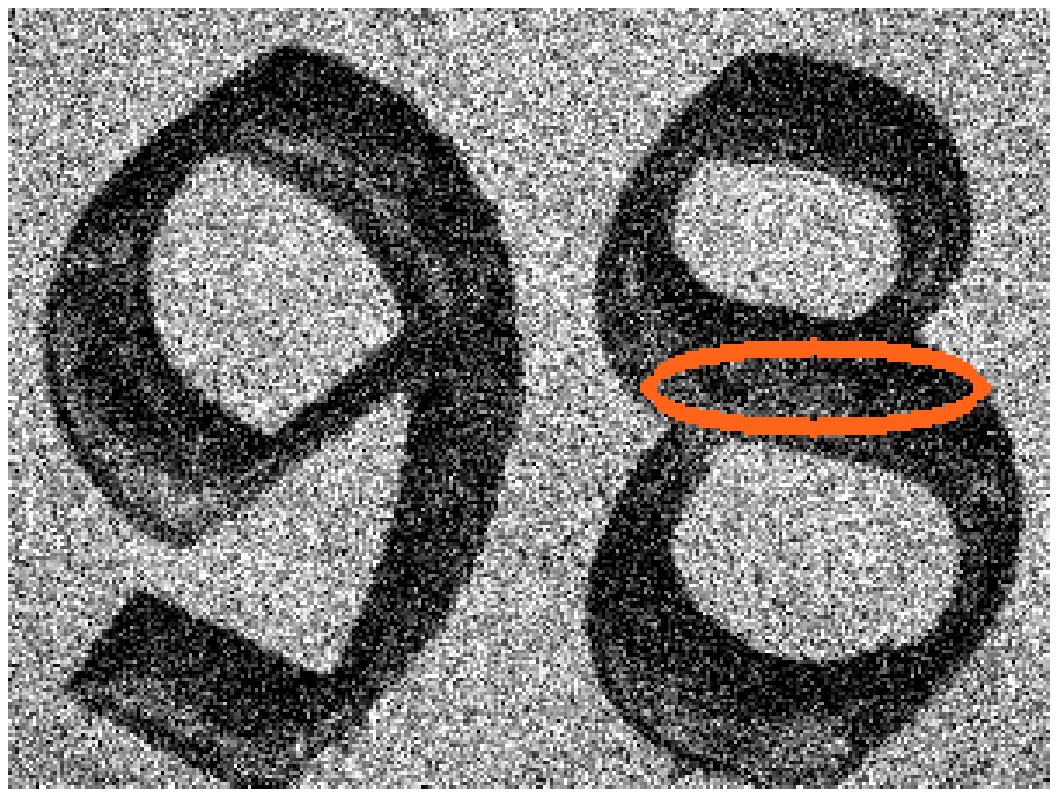,height=0.13\textwidth, width=0.2\textwidth}
\hspace{0.000001\textwidth}
\psfig{figure=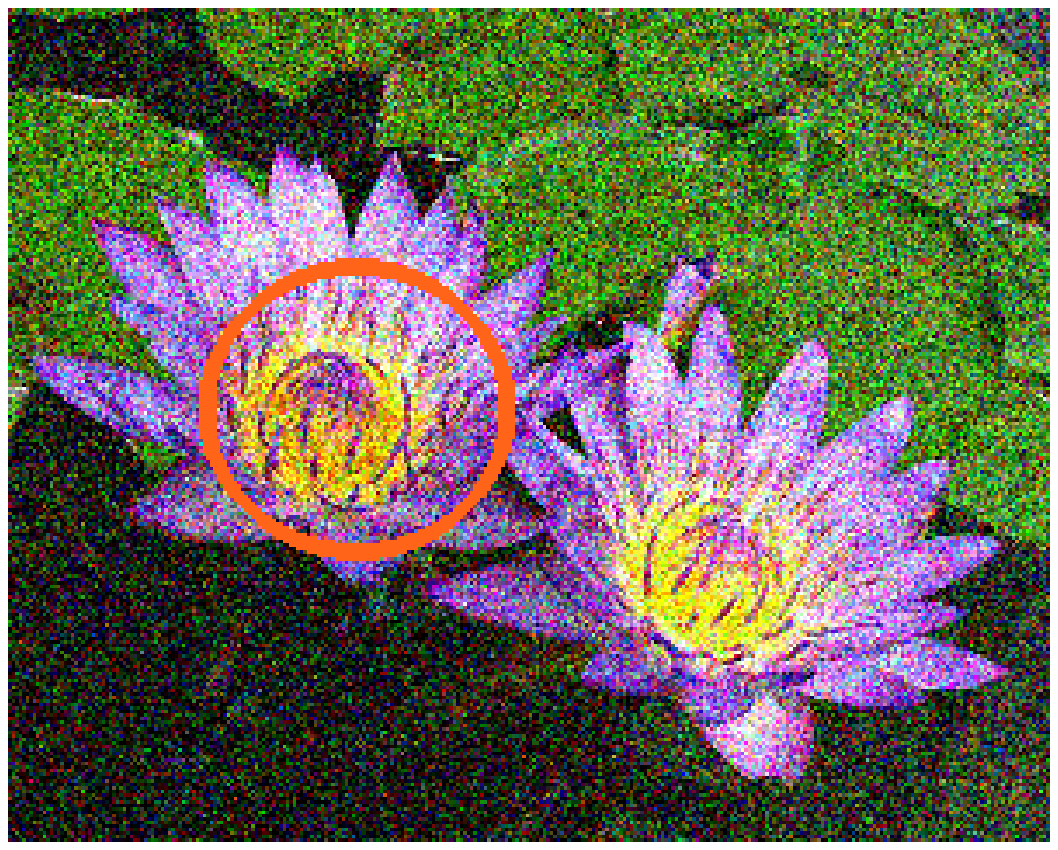,height=0.13\textwidth, width=0.2\textwidth}
\hspace{0.000001\textwidth}
\psfig{figure=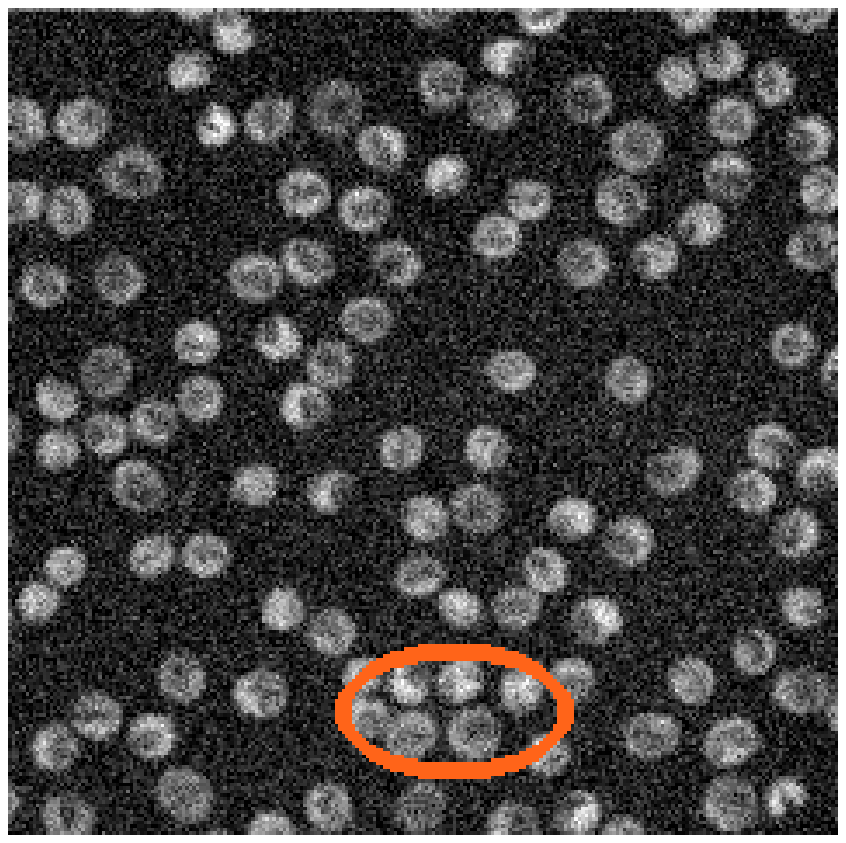,height=0.13\textwidth, width=0.2\textwidth}}
\vspace{0.000001\textwidth}
\centerline{
\psfig{figure=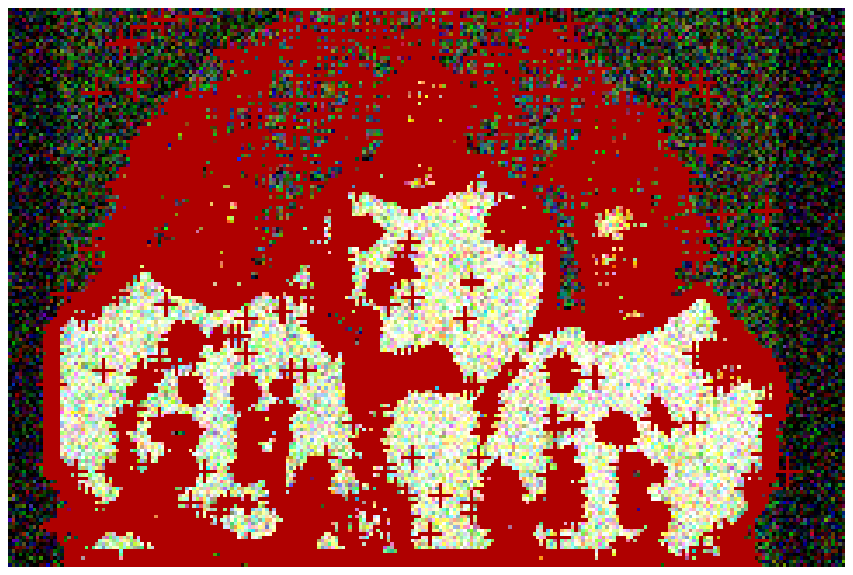,height=0.13\textwidth, width=0.2\textwidth}
\hspace{0.000001\textwidth}
\psfig{figure=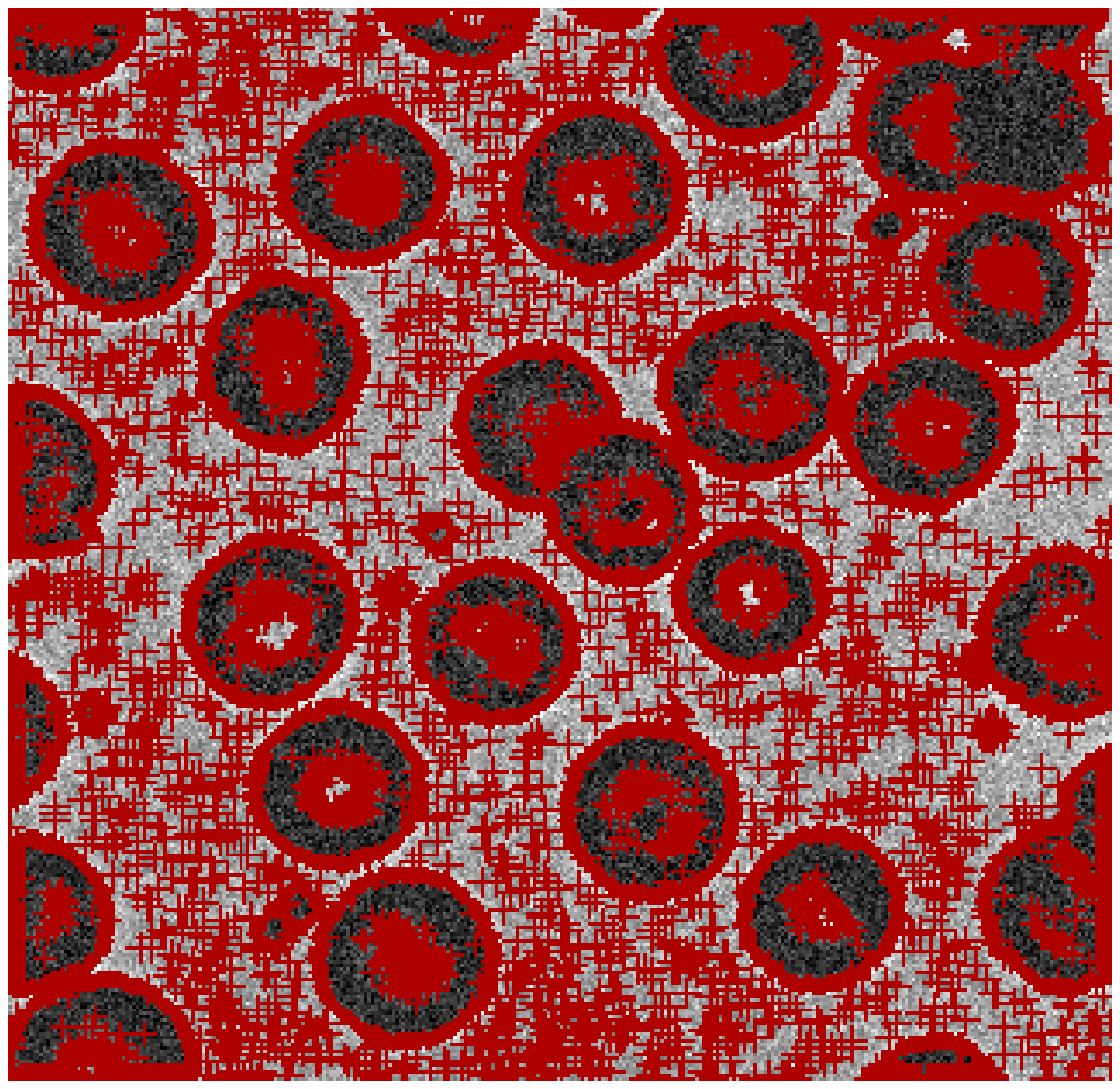,height=0.13\textwidth, width=0.2\textwidth}
\hspace{0.000001\textwidth}
\psfig{figure=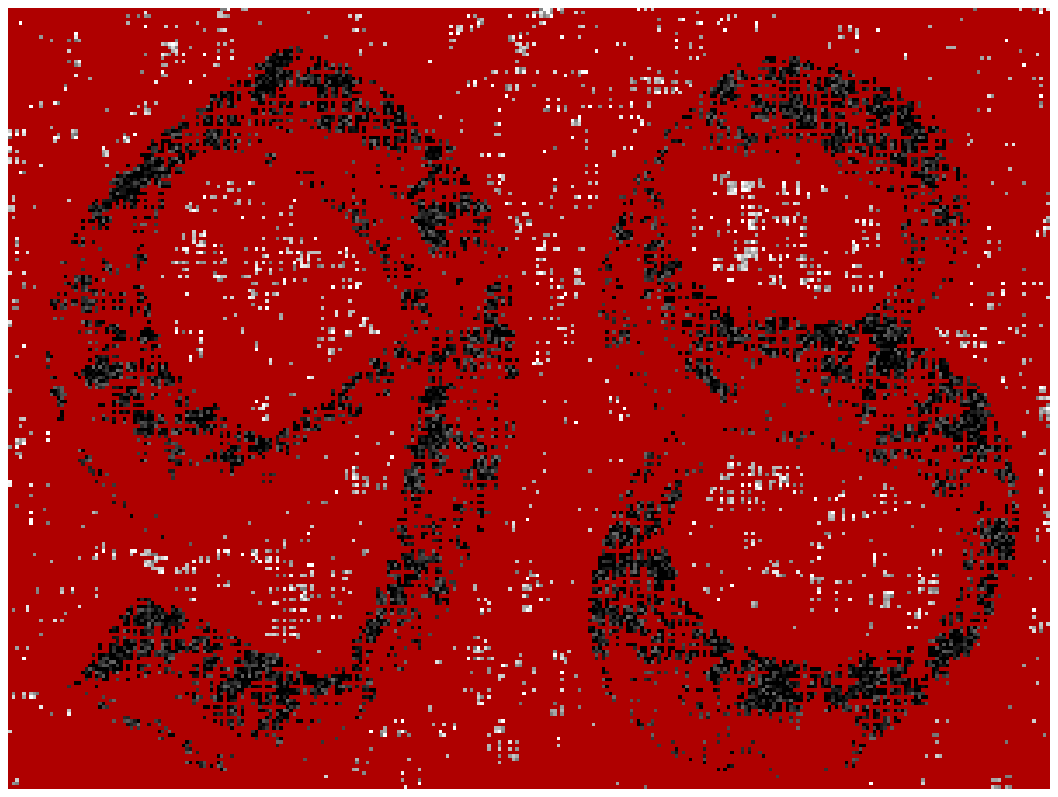,height=0.13\textwidth, width=0.2\textwidth}
\hspace{0.000001\textwidth}
\psfig{figure=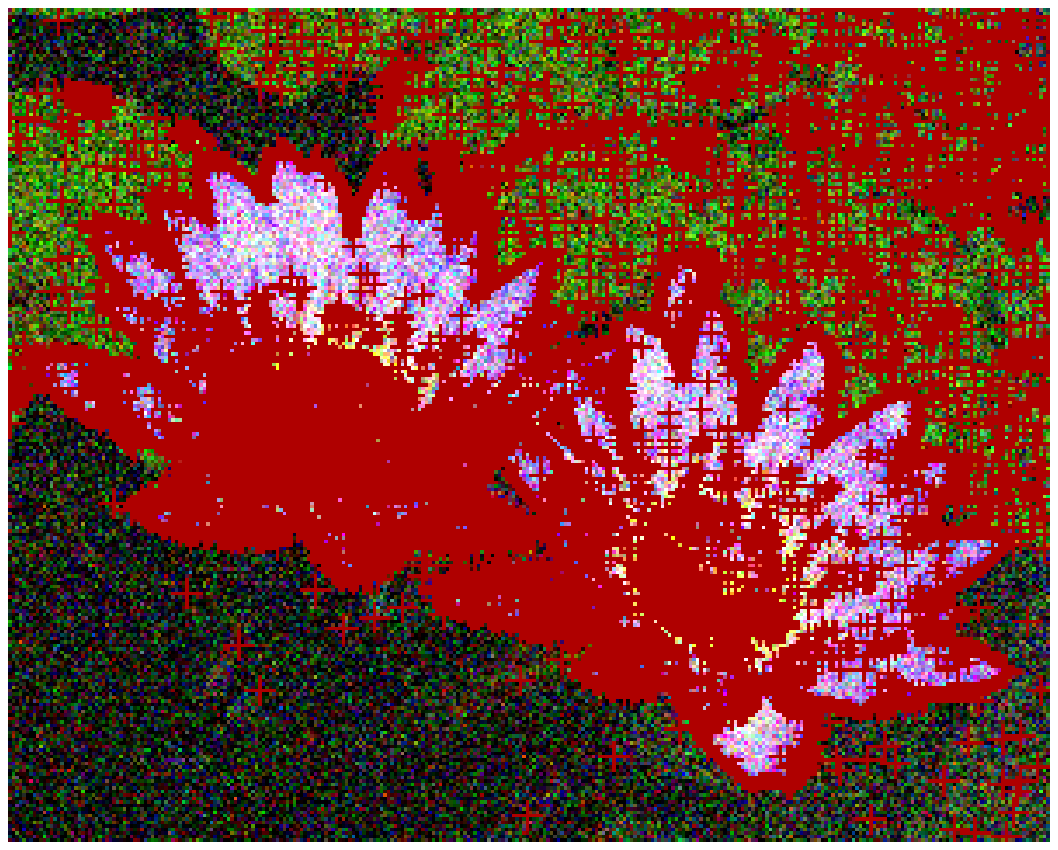,height=0.13\textwidth, width=0.2\textwidth}
\hspace{0.000001\textwidth}
\psfig{figure=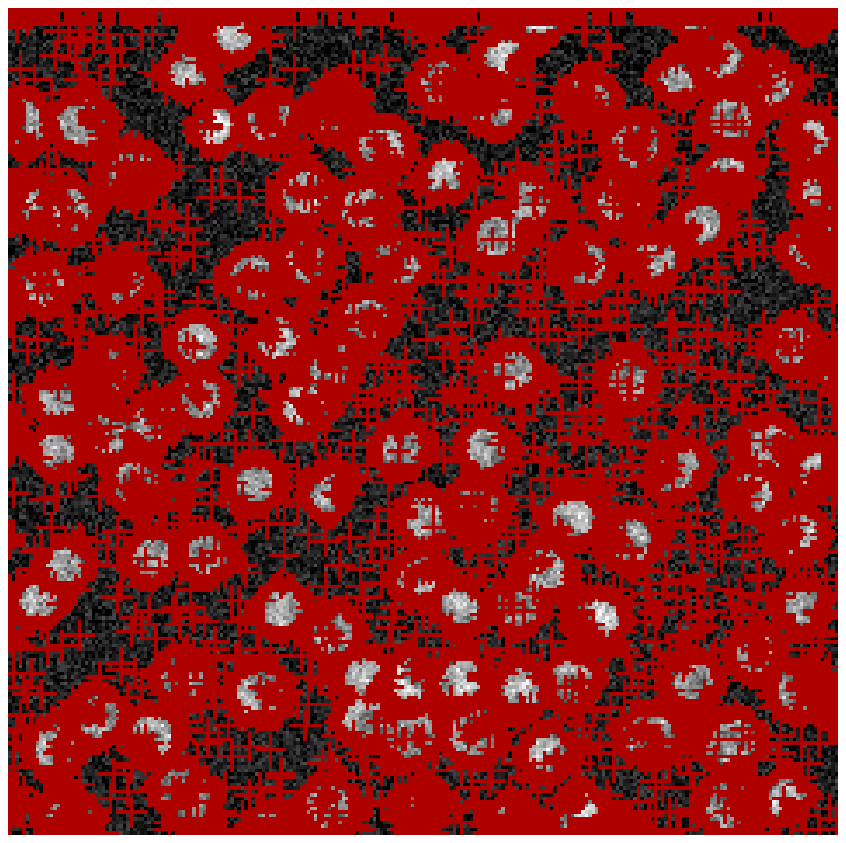,height=0.13\textwidth, width=0.2\textwidth}}
\vspace{0.000001\textwidth}
\centerline{
\psfig{figure=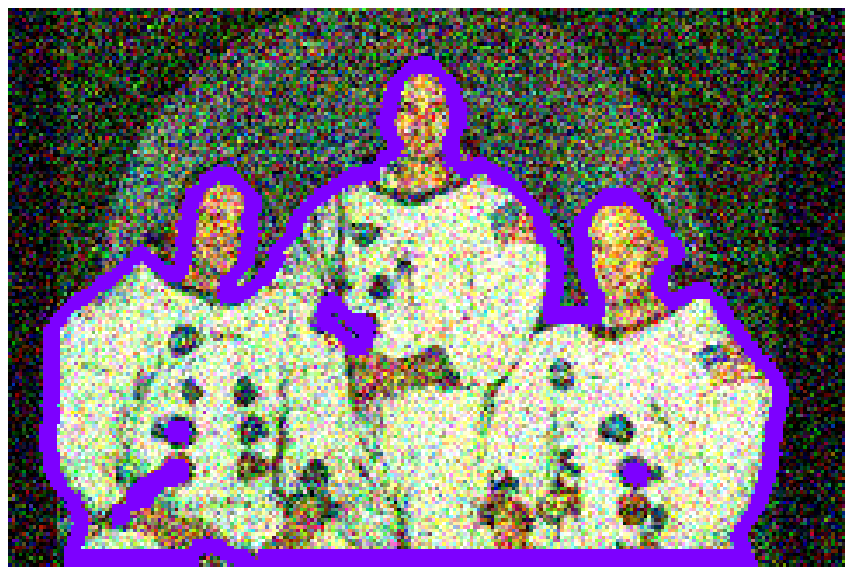,height=0.13\textwidth, width=0.2\textwidth}
\hspace{0.000001\textwidth}
\psfig{figure=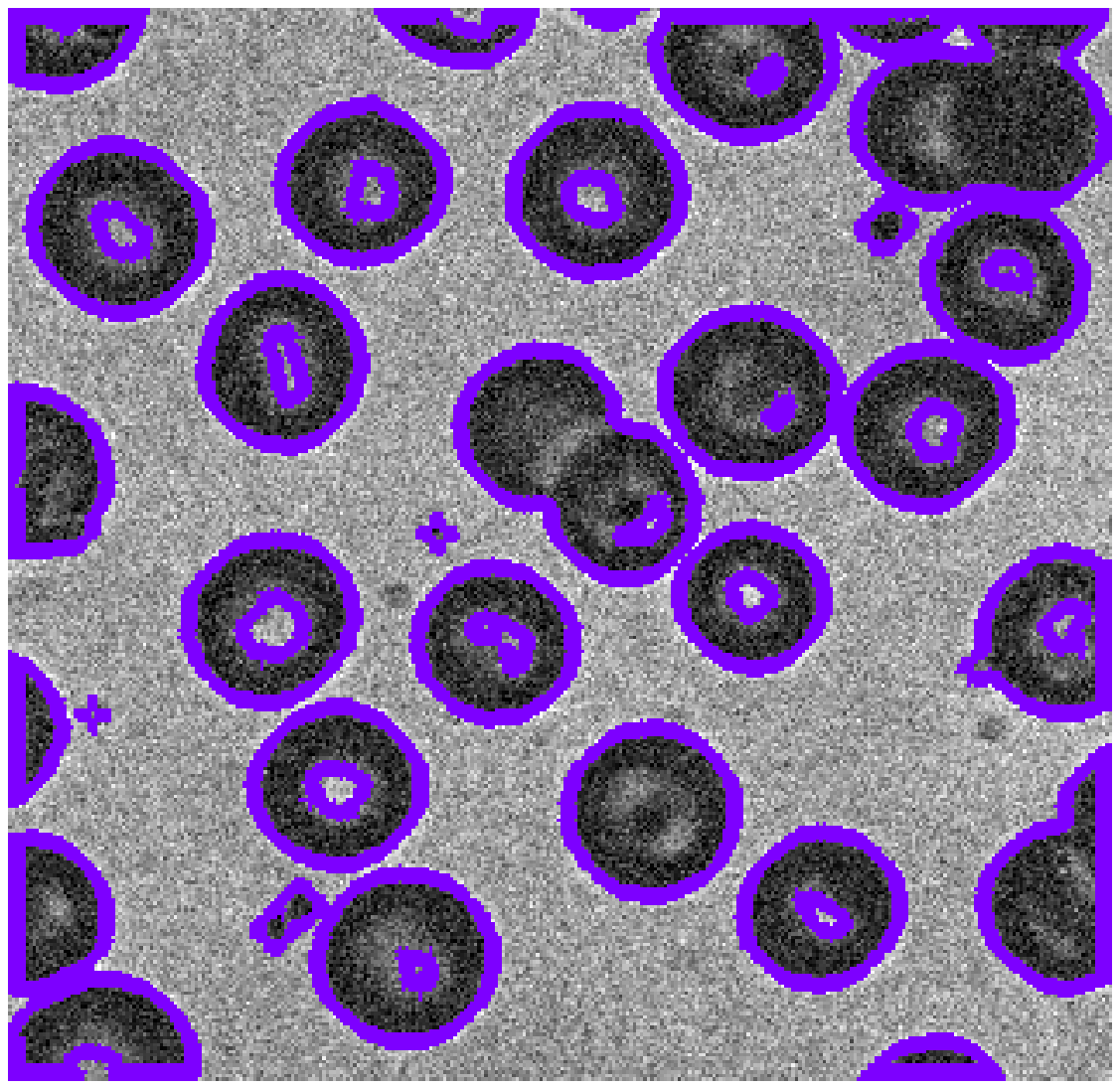,height=0.13\textwidth, width=0.2\textwidth}
\hspace{0.000001\textwidth}
\psfig{figure=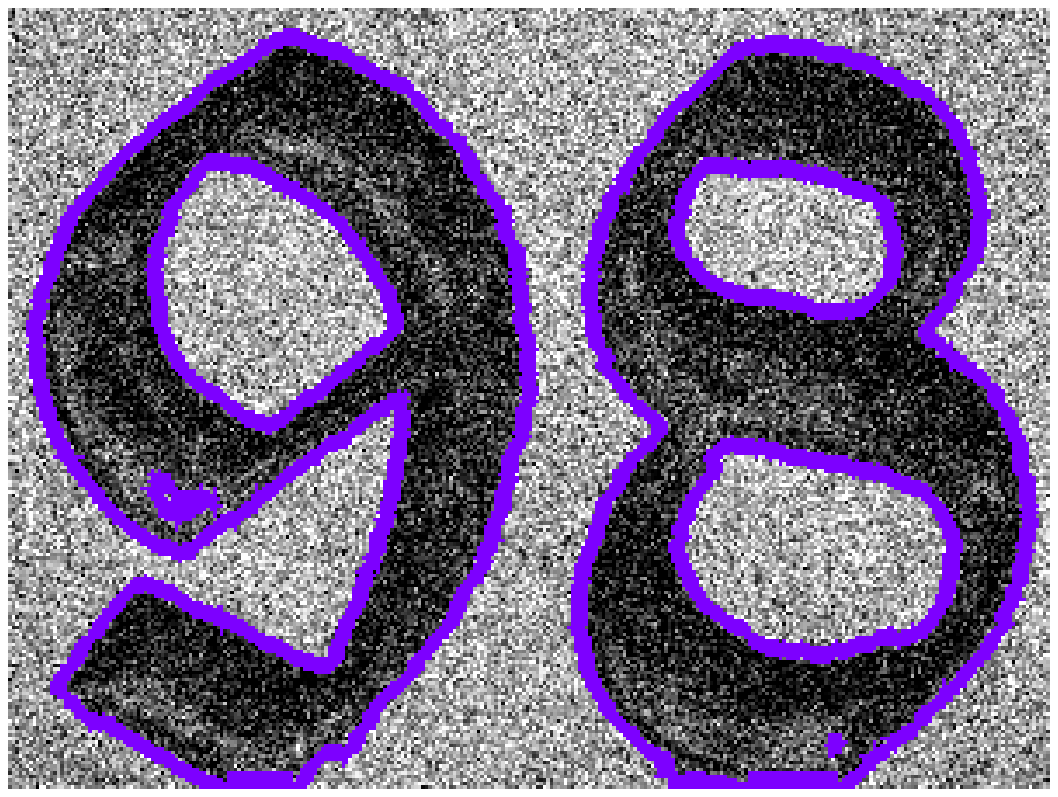,height=0.13\textwidth, width=0.2\textwidth}
\hspace{0.000001\textwidth}
\psfig{figure=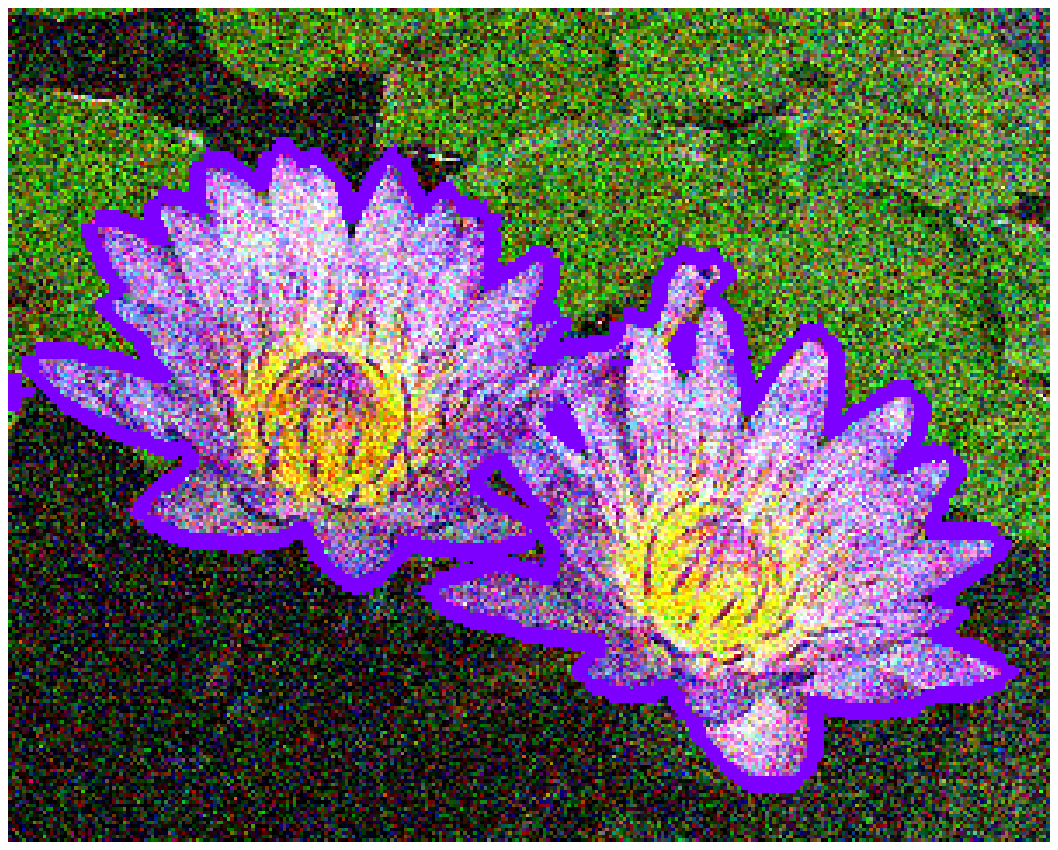,height=0.13\textwidth, width=0.2\textwidth}
\hspace{0.000001\textwidth}
\psfig{figure=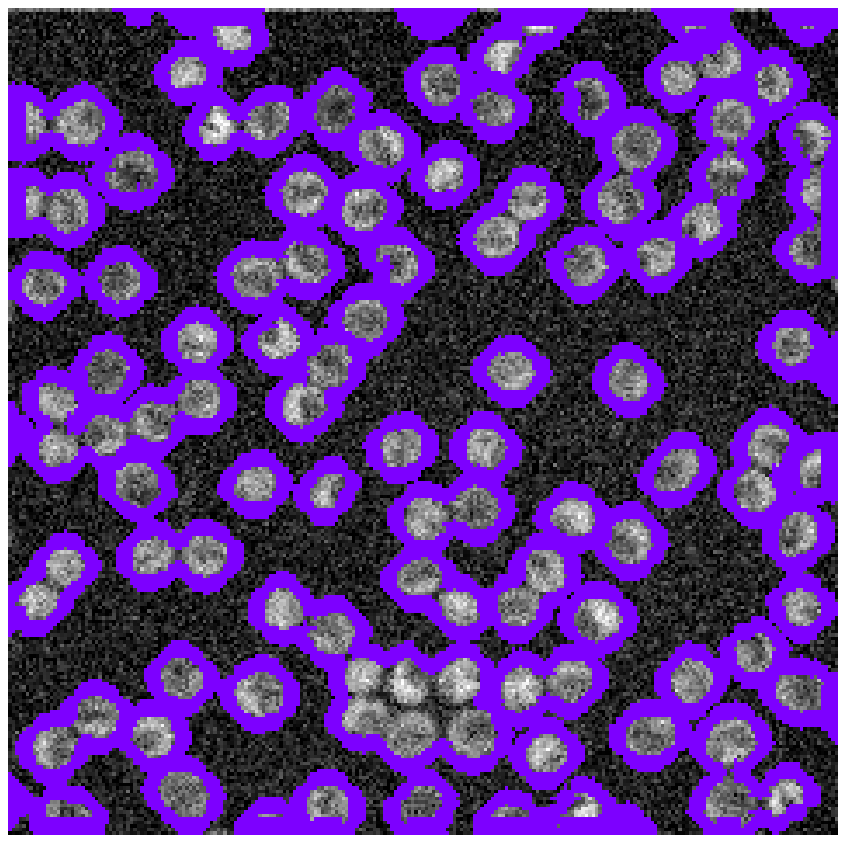,height=0.13\textwidth, width=0.2\textwidth}}
\vspace{0.000001\textwidth}
\centerline{
\psfig{figure=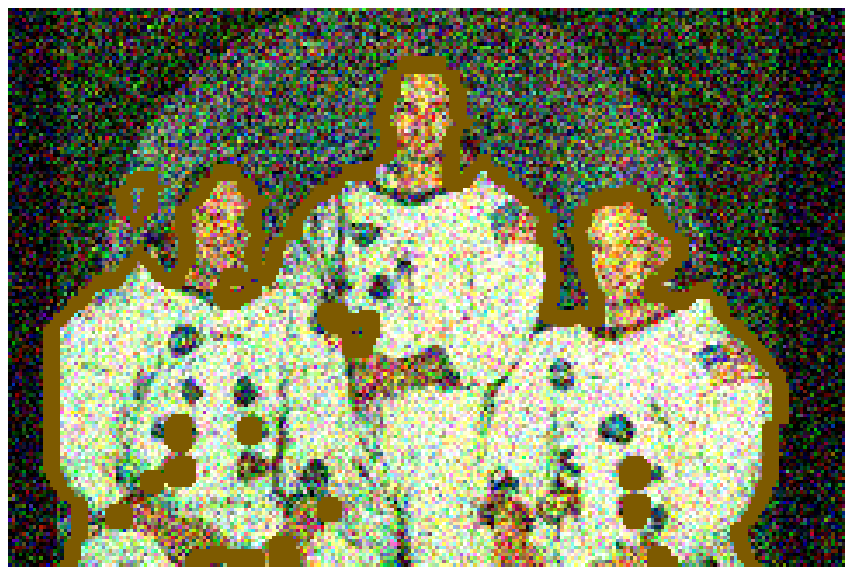,height=0.13\textwidth, width=0.2\textwidth}
\hspace{0.000001\textwidth}
\psfig{figure=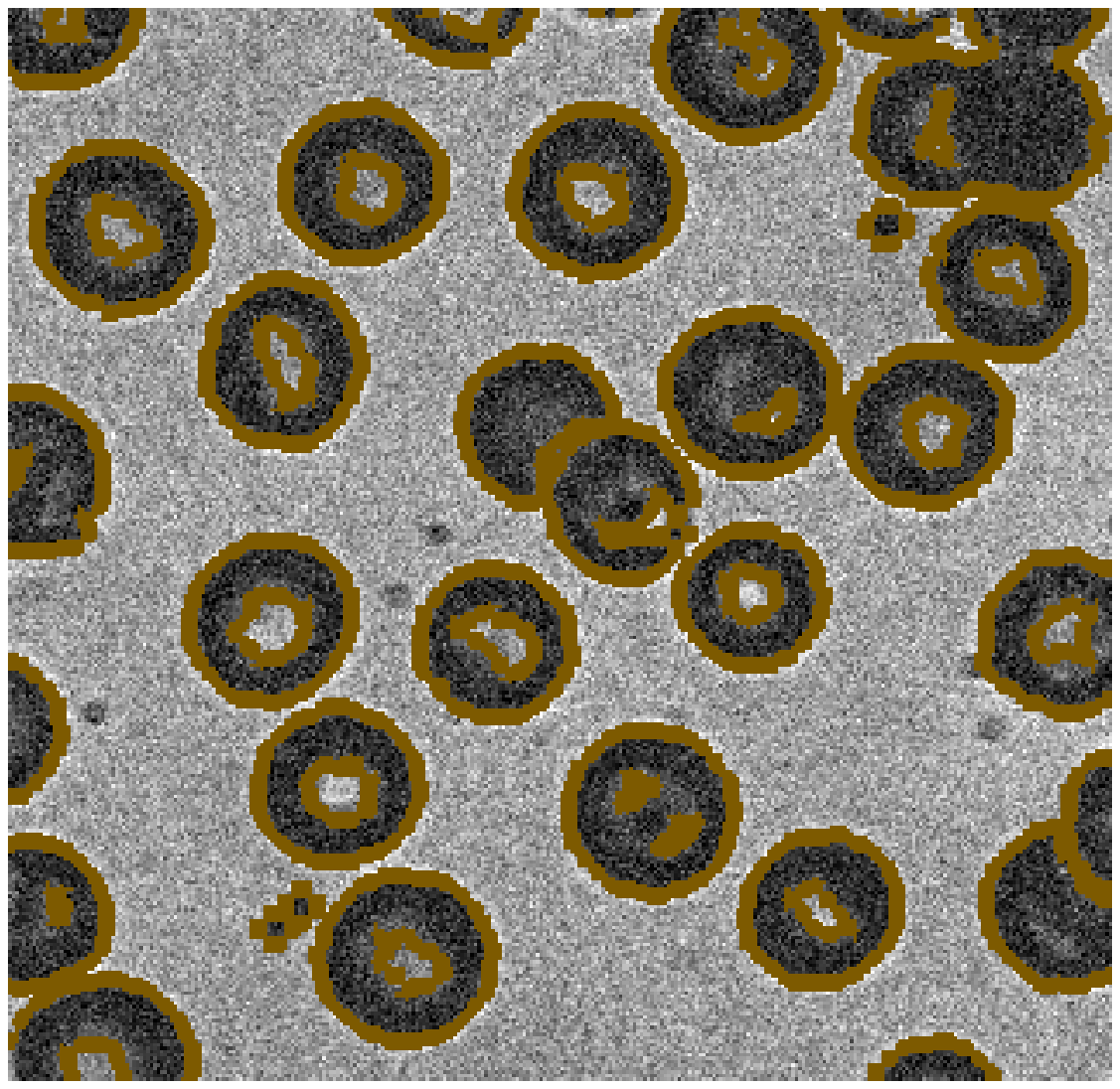,height=0.13\textwidth, width=0.2\textwidth}
\hspace{0.000001\textwidth}
\psfig{figure=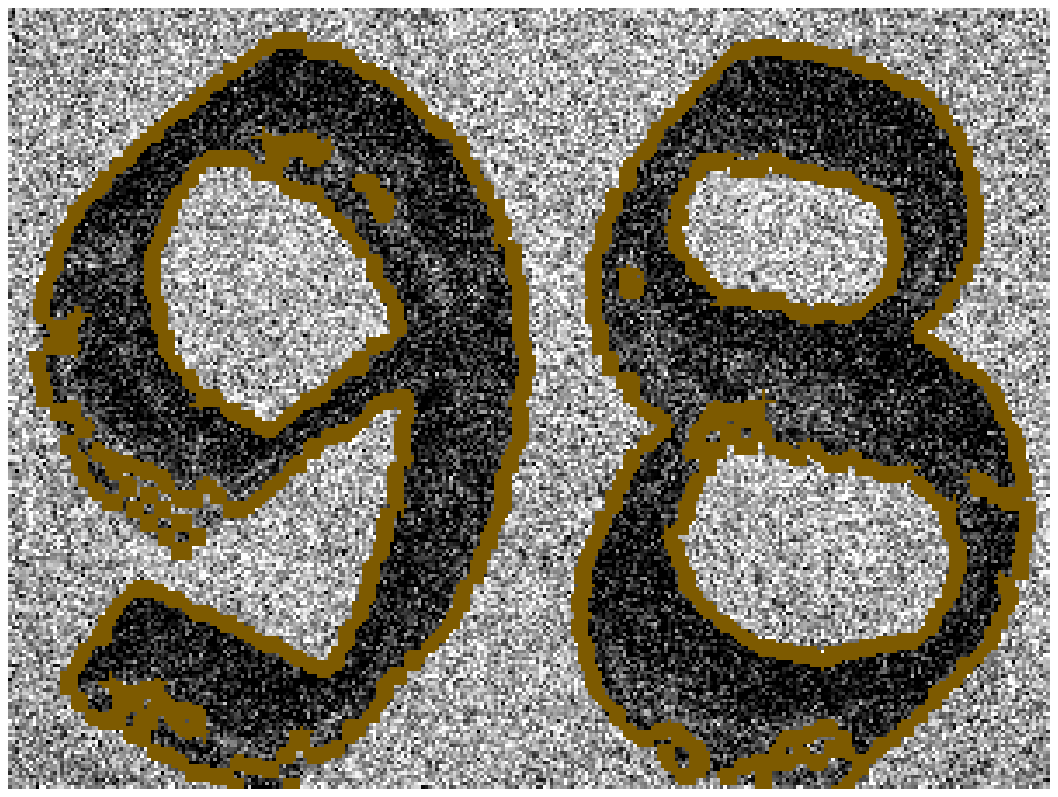,height=0.13\textwidth, width=0.2\textwidth}
\hspace{0.000001\textwidth}
\psfig{figure=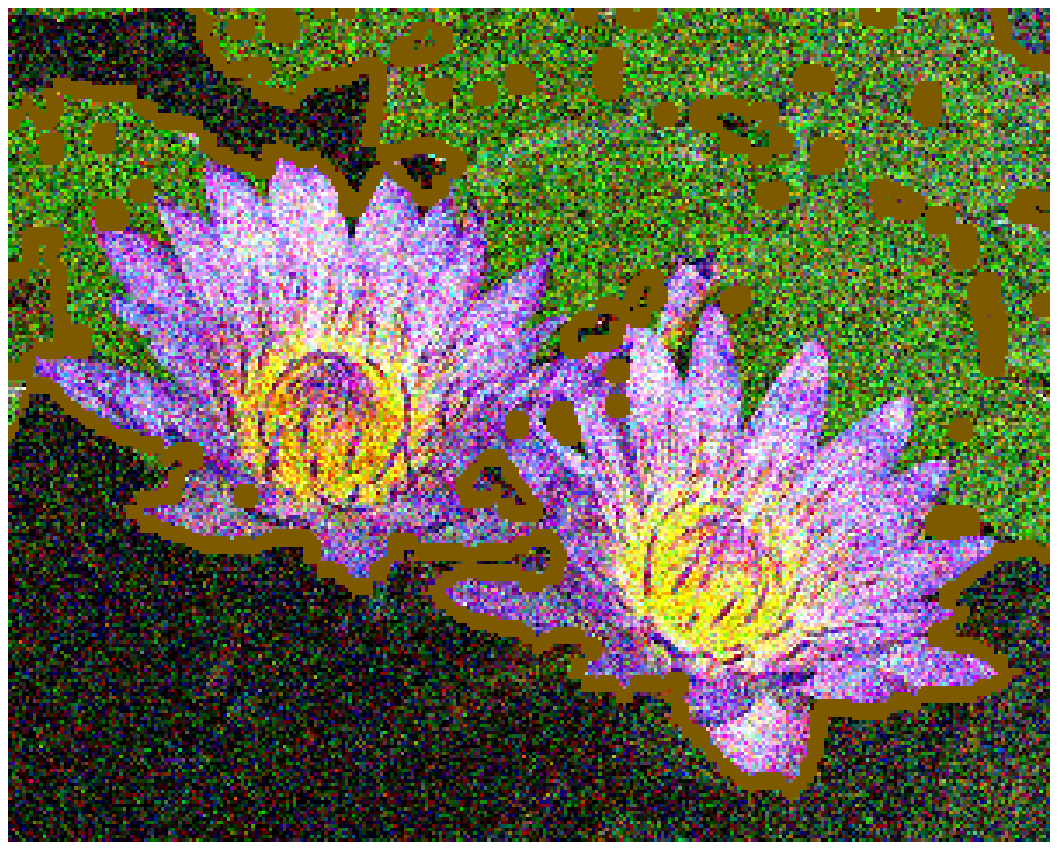,height=0.13\textwidth, width=0.2\textwidth}
\hspace{0.000001\textwidth}
\psfig{figure=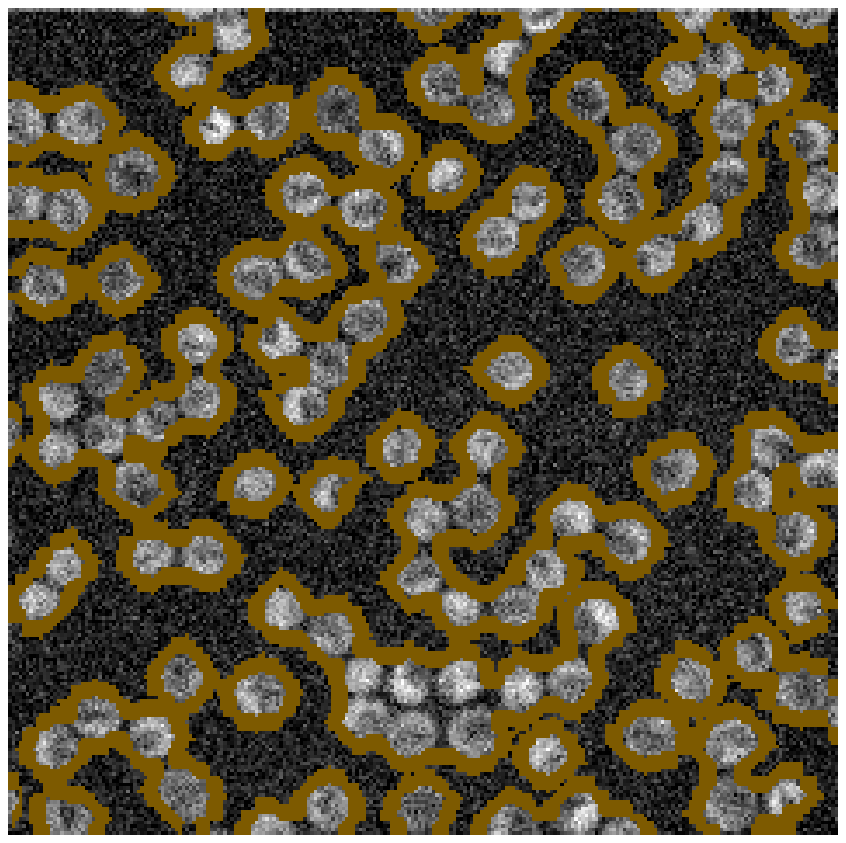,height=0.13\textwidth, width=0.2\textwidth}}
\vspace{0.000001\textwidth}
\centerline{
\psfig{figure=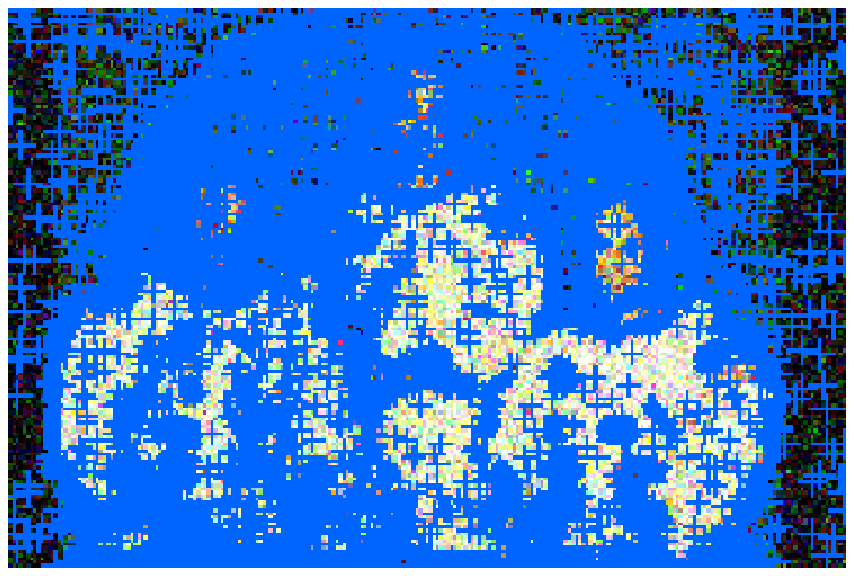,height=0.13\textwidth, width=0.2\textwidth}
\hspace{0.000001\textwidth}
\psfig{figure=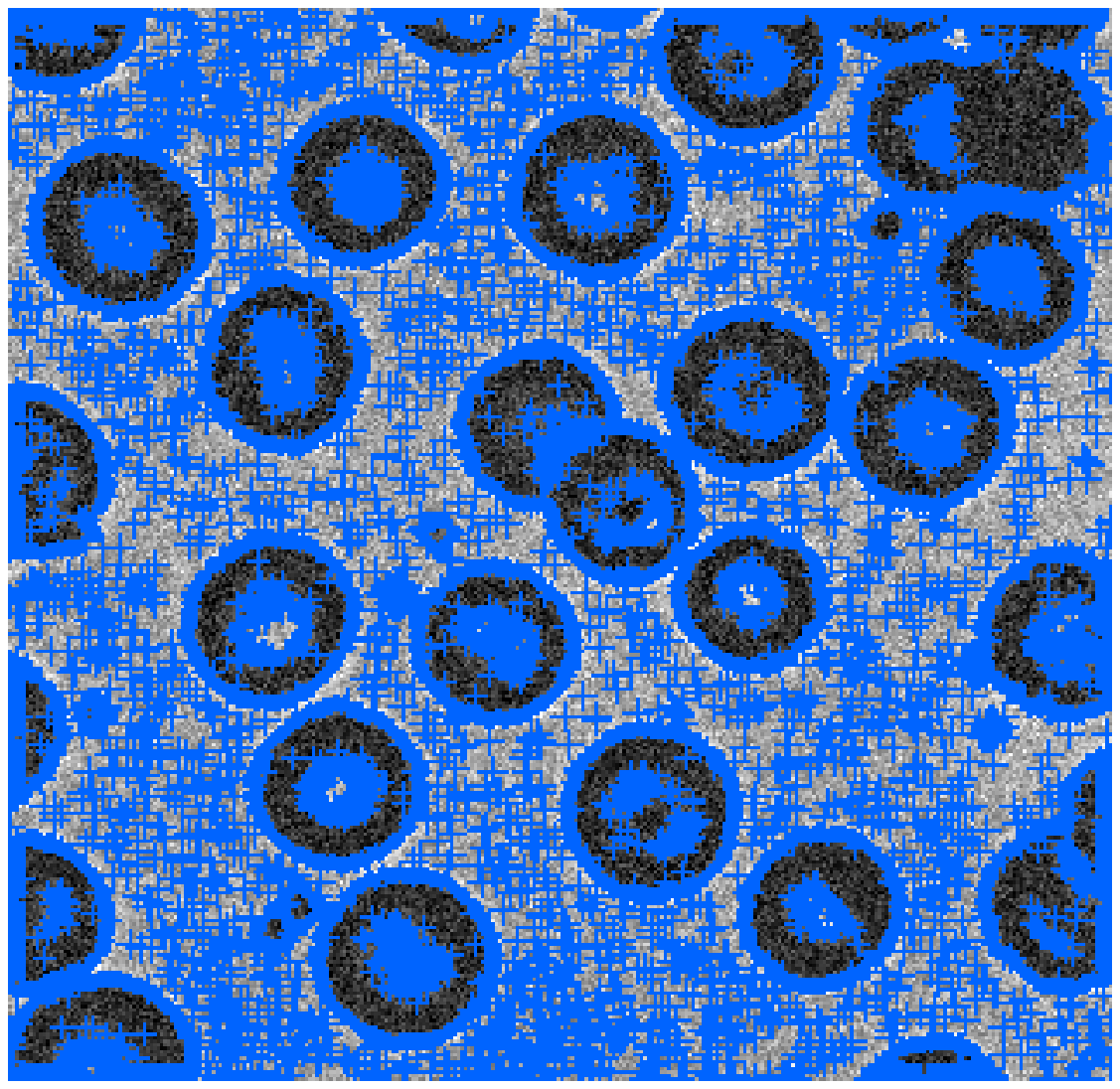,height=0.13\textwidth, width=0.2\textwidth}
\hspace{0.000001\textwidth}
\psfig{figure=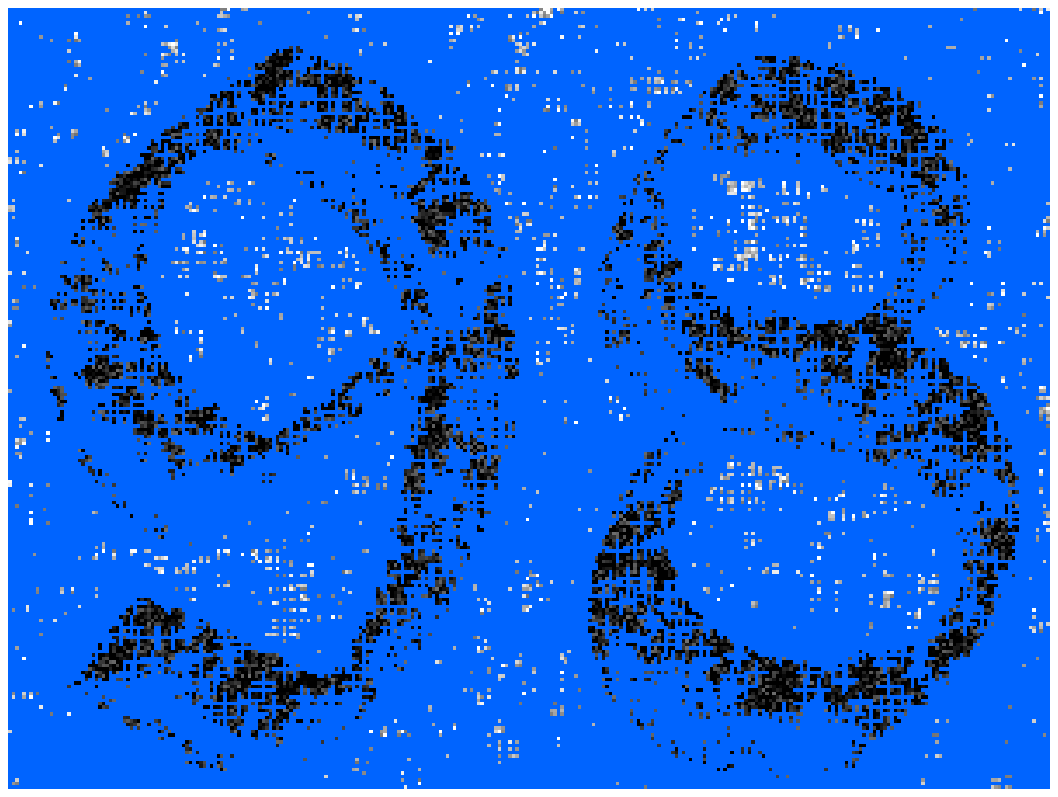,height=0.13\textwidth, width=0.2\textwidth}
\hspace{0.000001\textwidth}
\psfig{figure=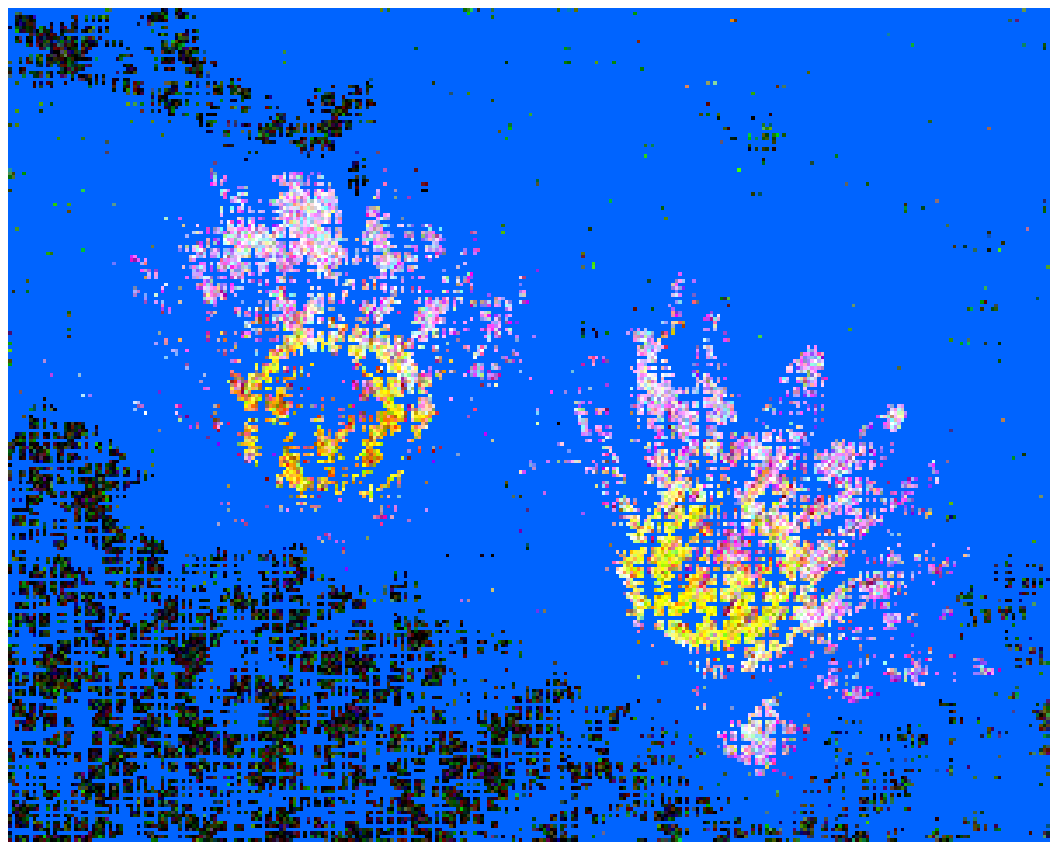,height=0.13\textwidth, width=0.2\textwidth}
\hspace{0.000001\textwidth}
\psfig{figure=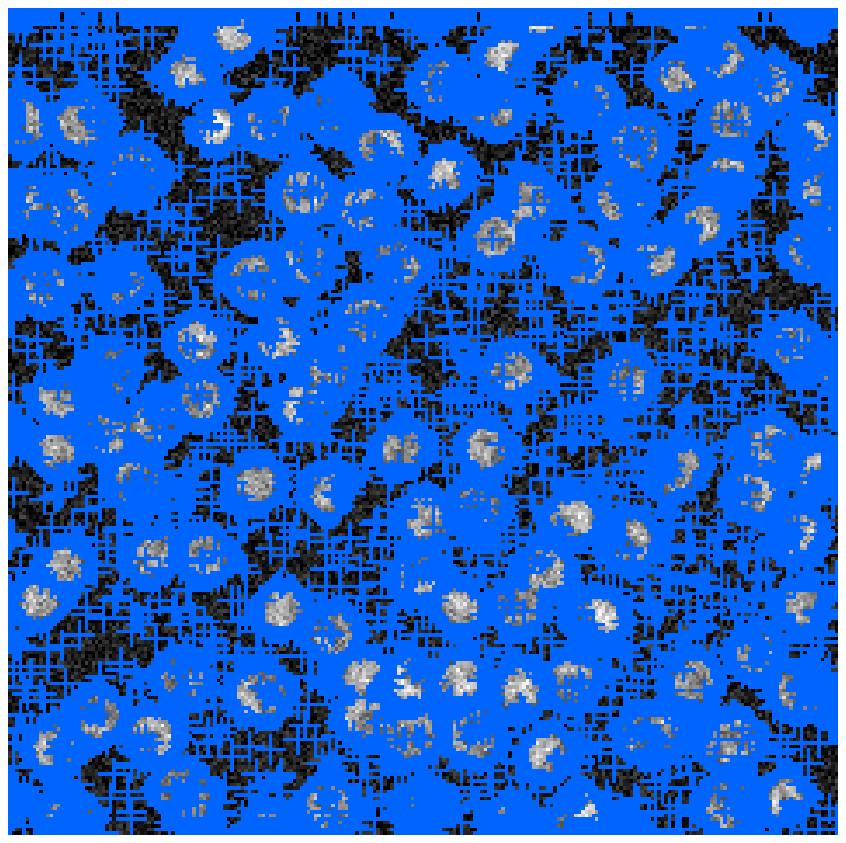,height=0.13\textwidth, width=0.2\textwidth}}
\vspace{0.000001\textwidth}
\centerline{
\psfig{figure=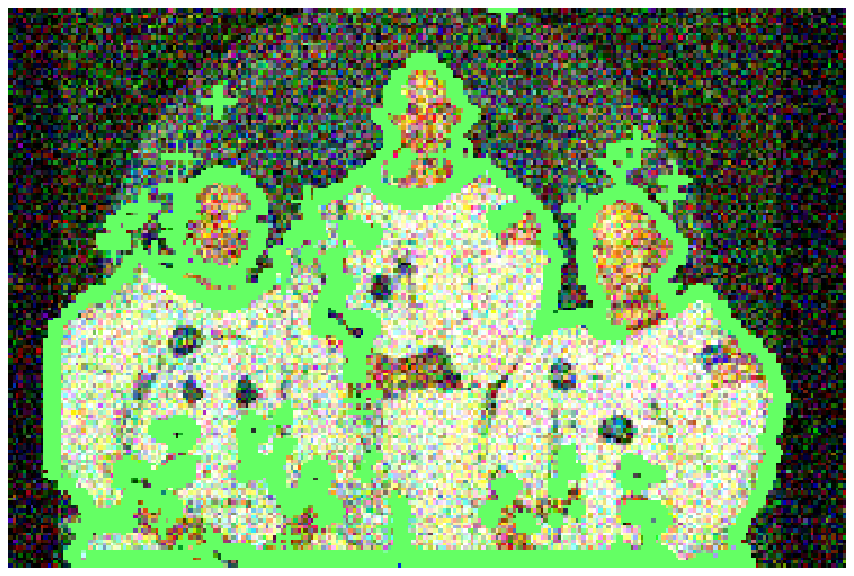,height=0.13\textwidth, width=0.2\textwidth}
\hspace{0.000001\textwidth}
\psfig{figure=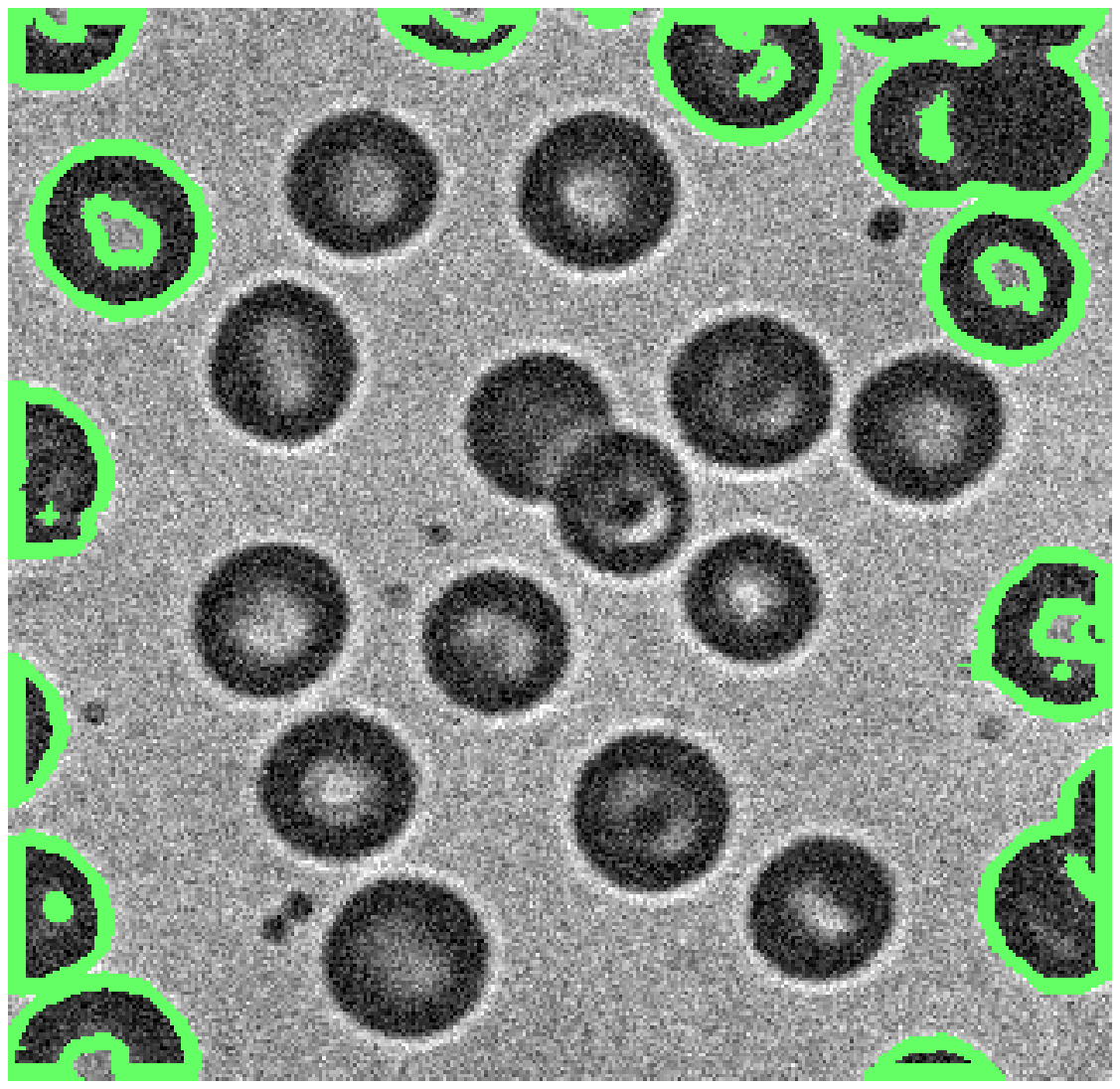,height=0.13\textwidth, width=0.2\textwidth}
\hspace{0.000001\textwidth}
\psfig{figure=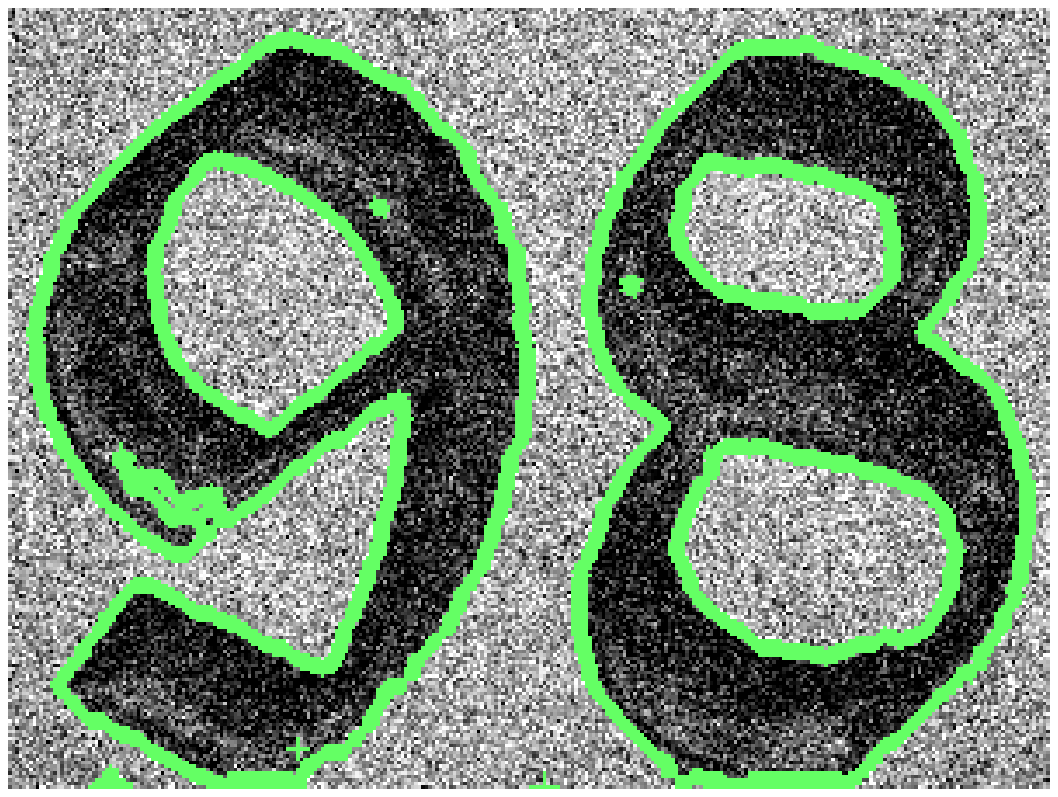,height=0.13\textwidth, width=0.2\textwidth}
\hspace{0.000001\textwidth}
\psfig{figure=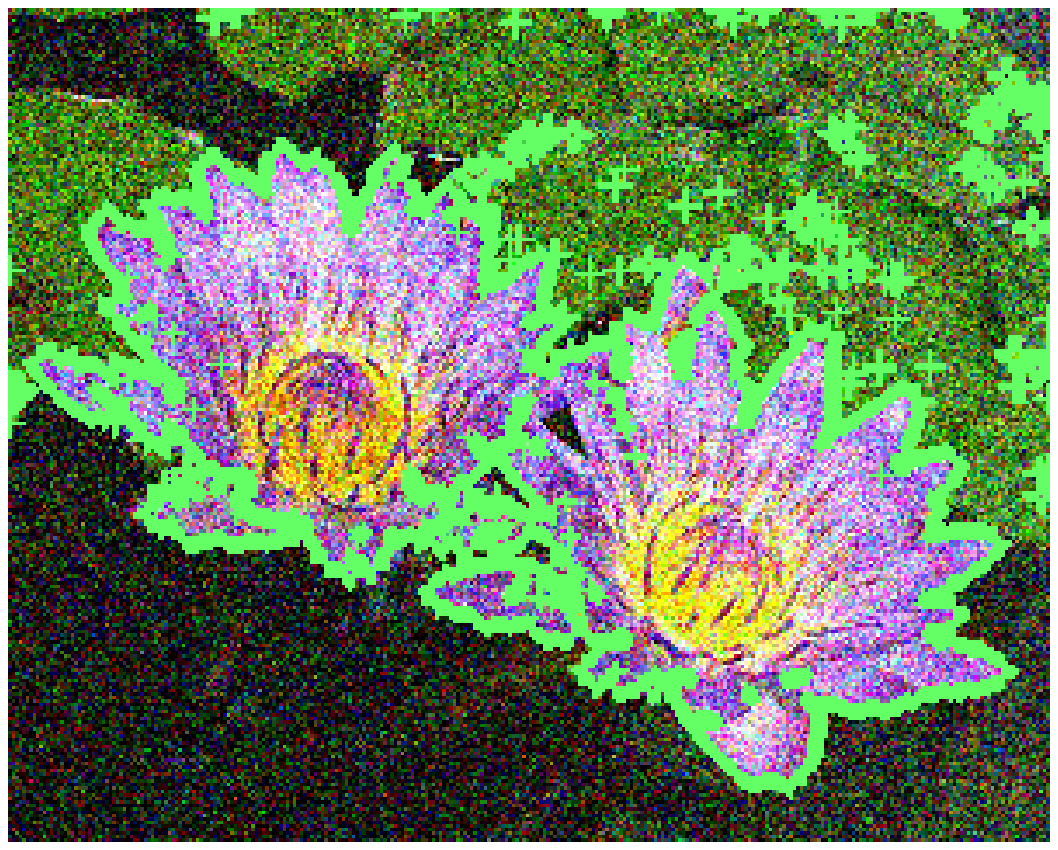,height=0.13\textwidth, width=0.2\textwidth}
\hspace{0.000001\textwidth}
\psfig{figure=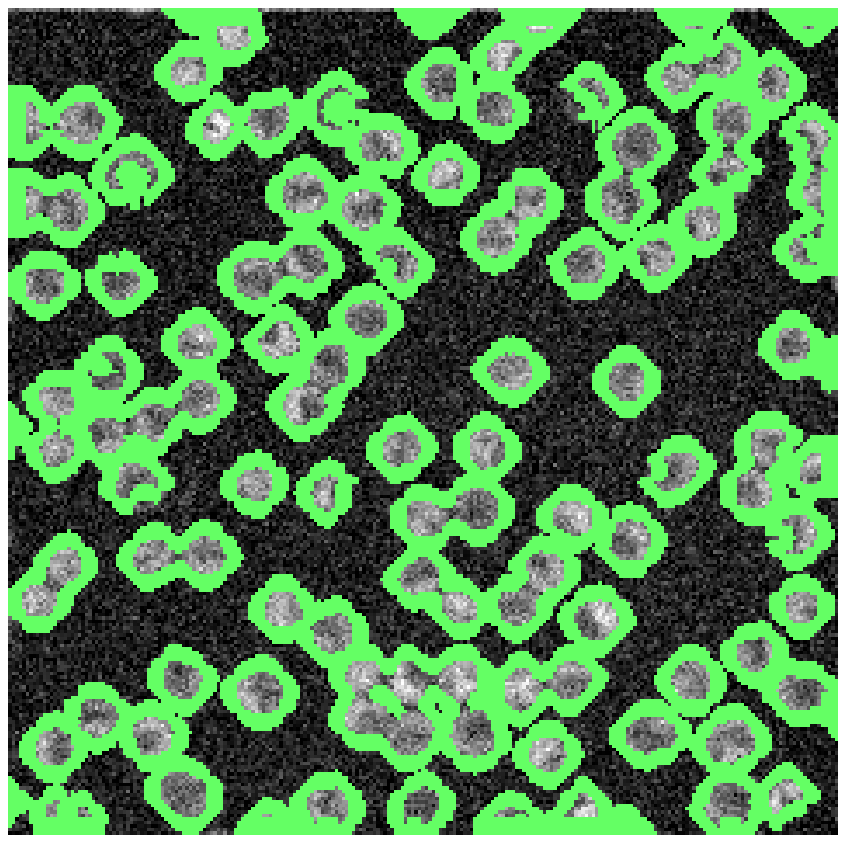,height=0.13\textwidth, width=0.2\textwidth}}
\vspace{0.000001\textwidth}
\centerline{
\psfig{figure=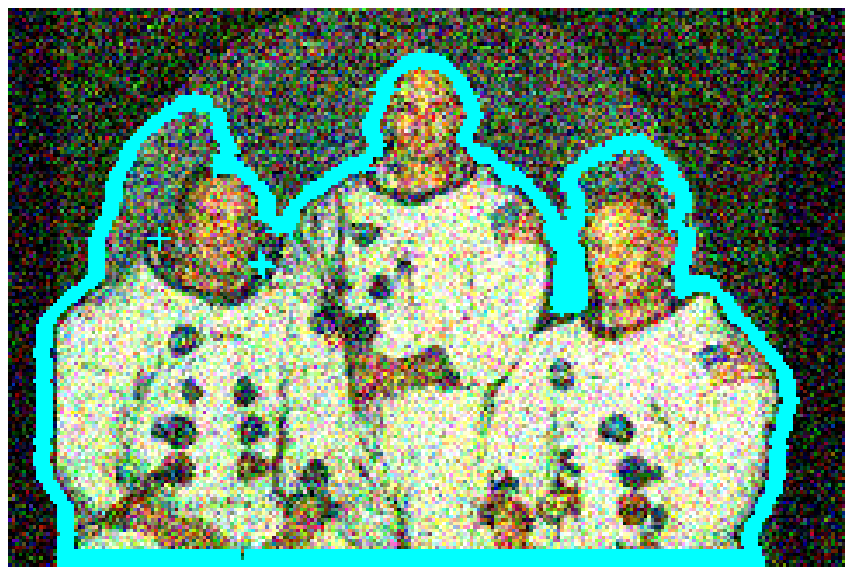,height=0.13\textwidth, width=0.2\textwidth}
\hspace{0.000001\textwidth}
\psfig{figure=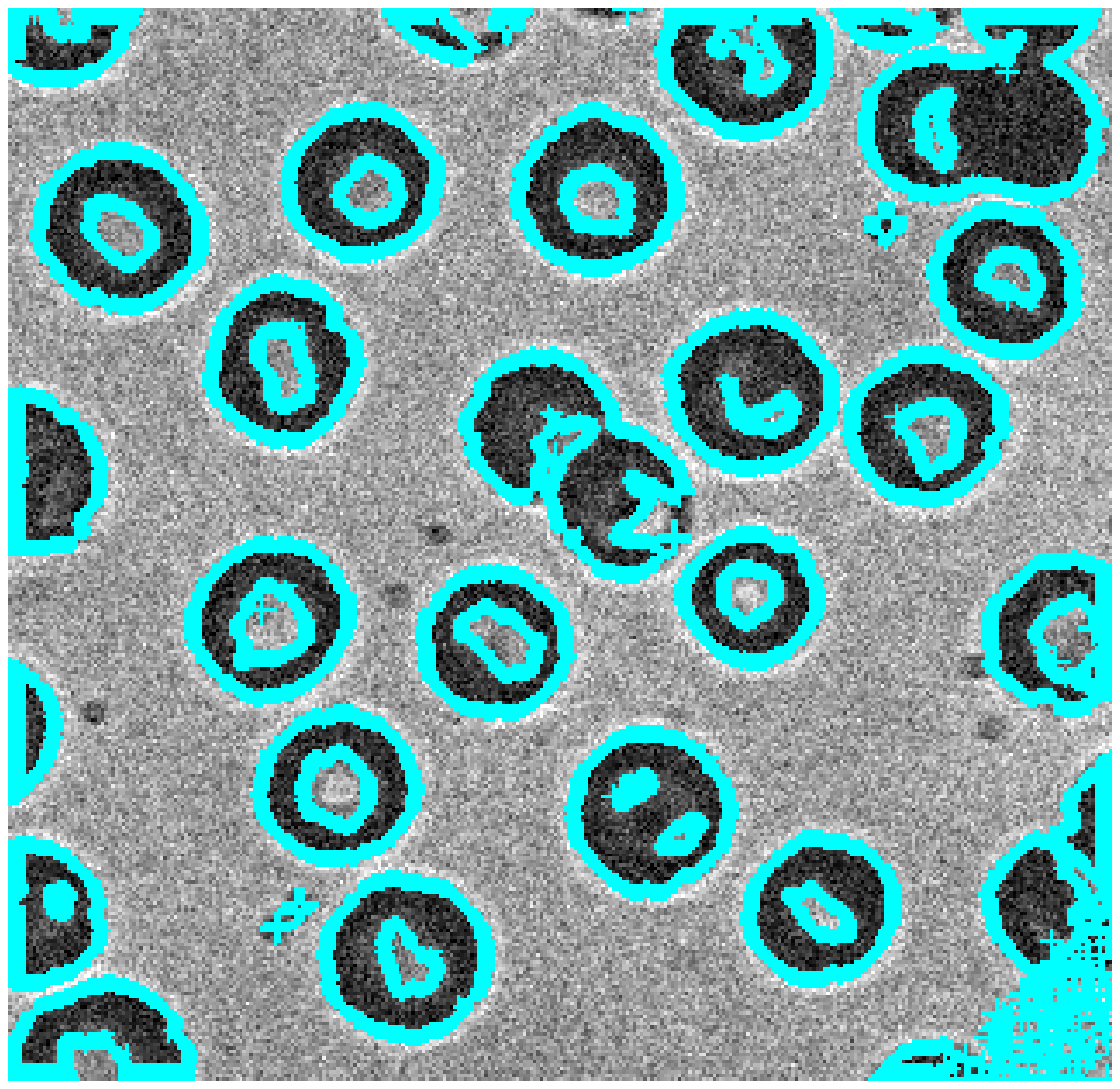,height=0.13\textwidth, width=0.2\textwidth}
\hspace{0.000001\textwidth}
\psfig{figure=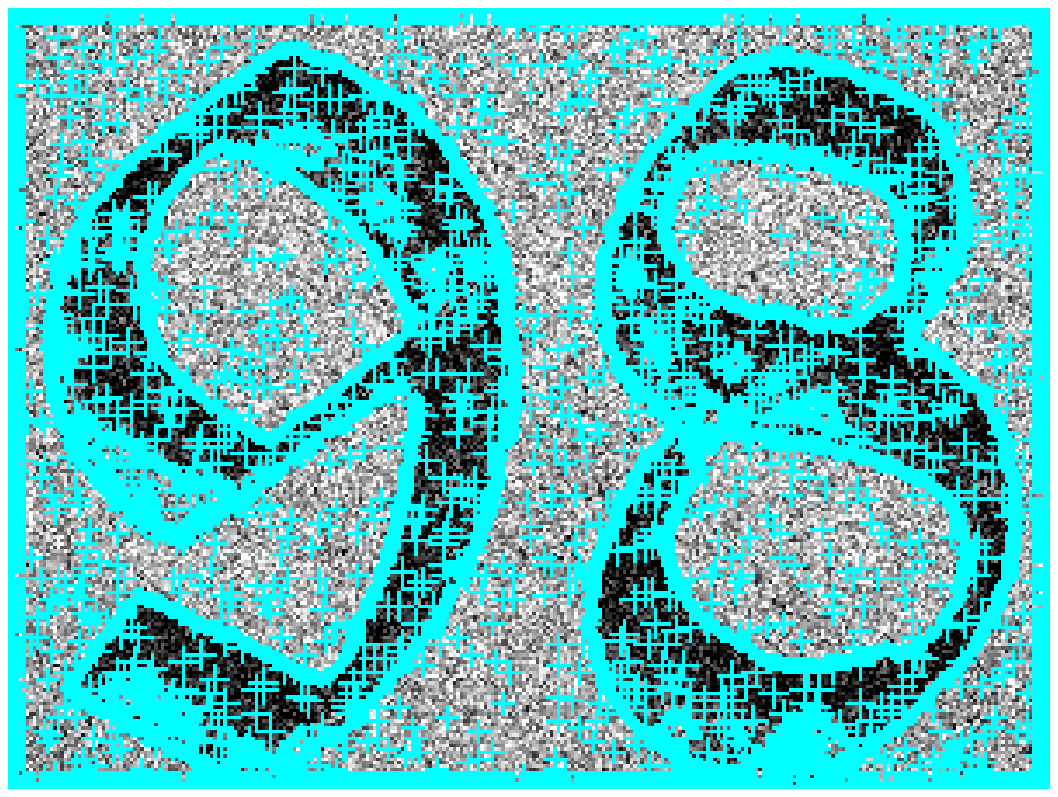,height=0.13\textwidth, width=0.2\textwidth}
\hspace{0.000001\textwidth}
\psfig{figure=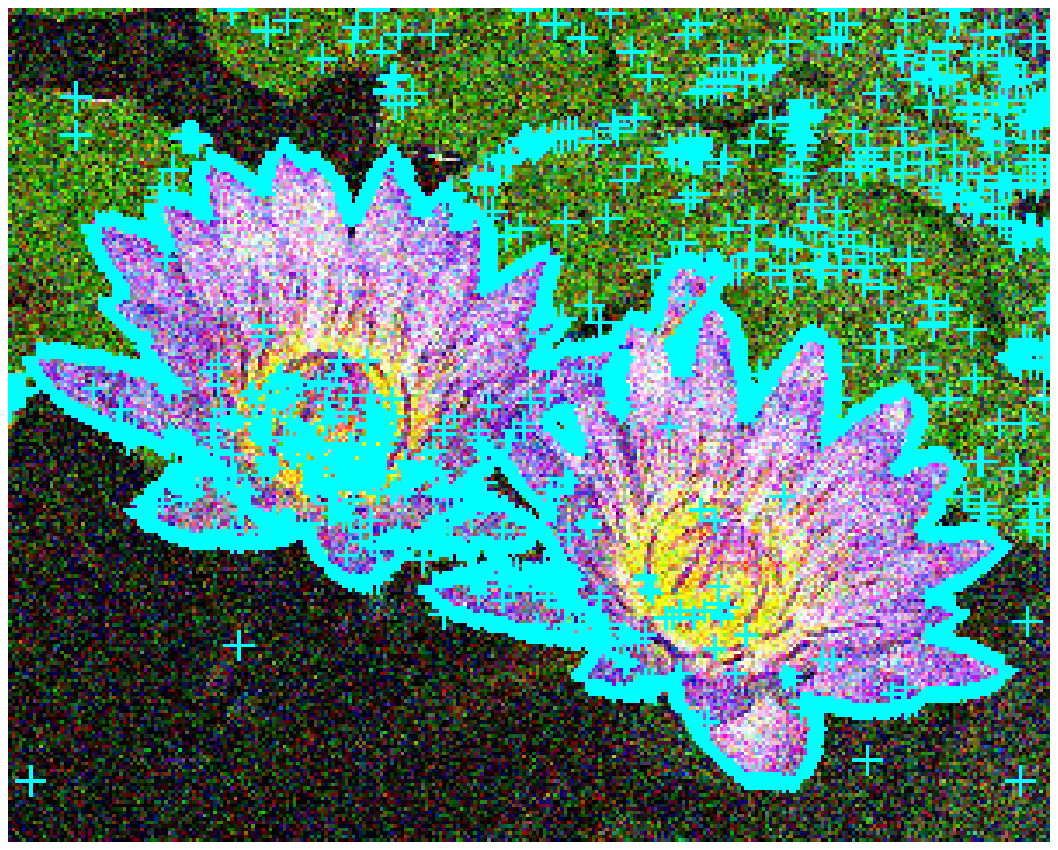,height=0.13\textwidth, width=0.2\textwidth}
\hspace{0.000001\textwidth}
\psfig{figure=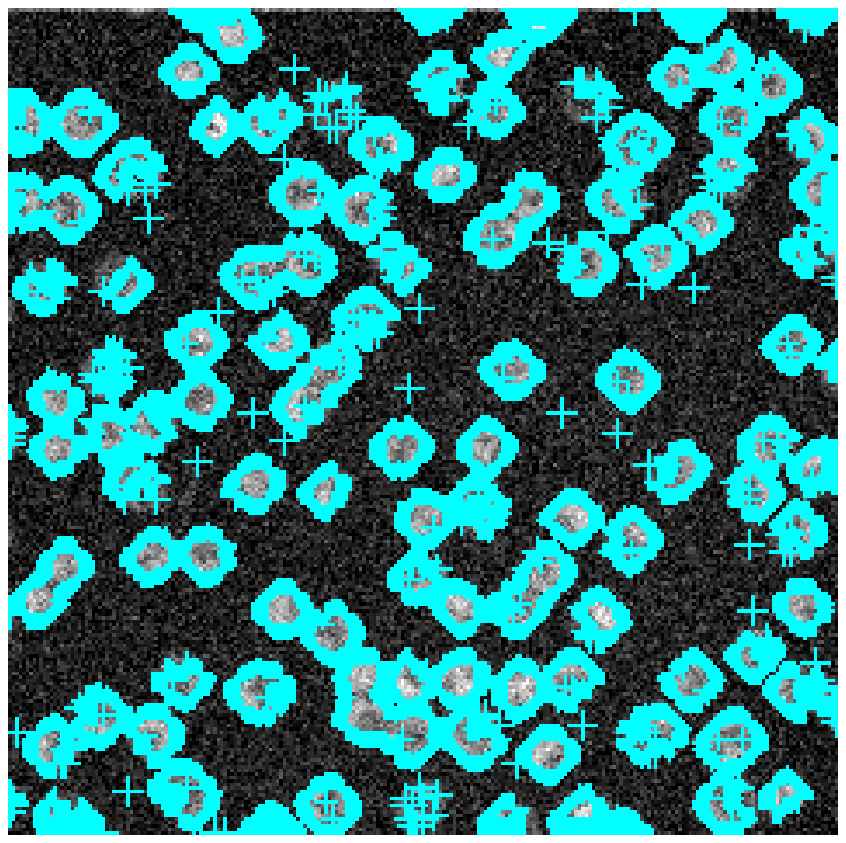,height=0.13\textwidth, width=0.2\textwidth}}
\vspace{0.000001\textwidth}
\centerline{
\psfig{figure=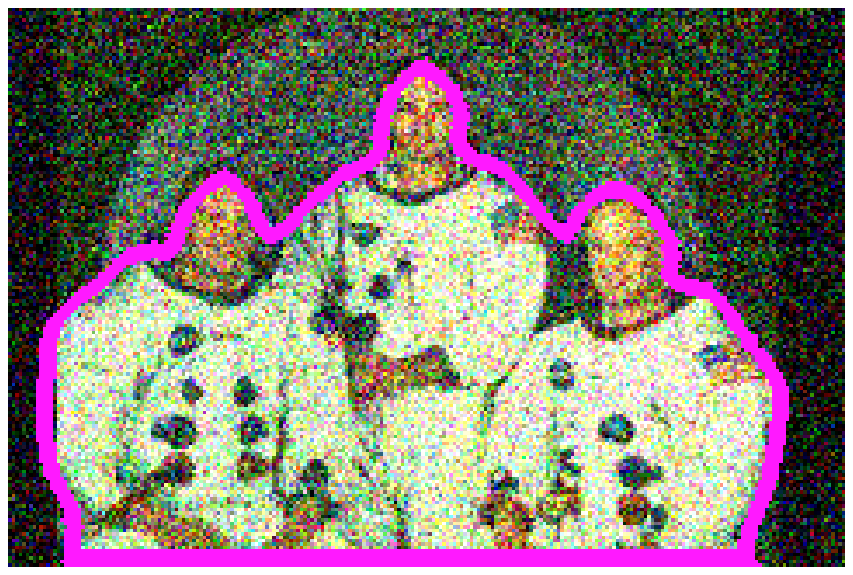,height=0.13\textwidth, width=0.2\textwidth}
\hspace{0.000001\textwidth}
\psfig{figure=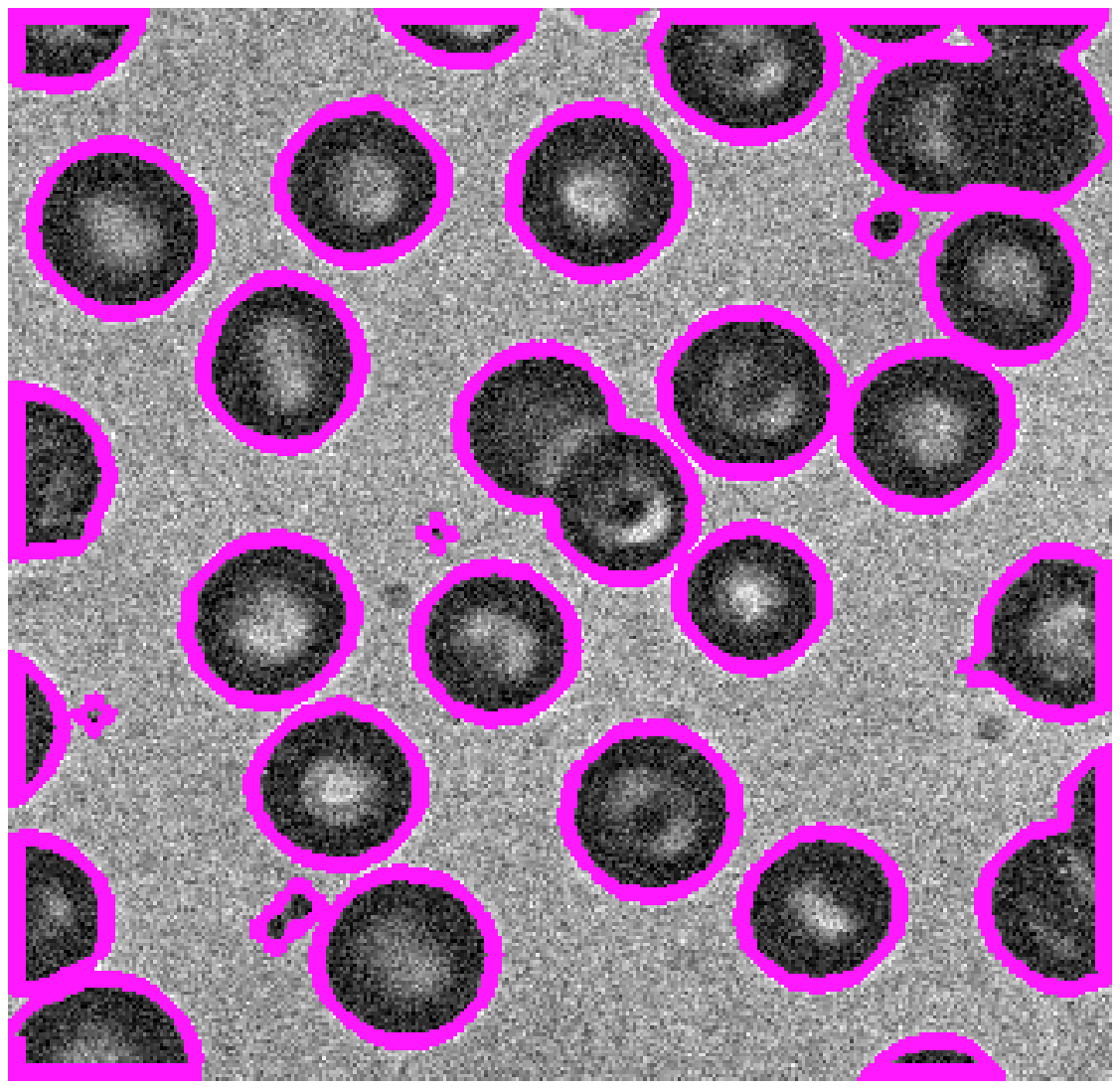,height=0.13\textwidth, width=0.2\textwidth}
\hspace{0.000001\textwidth}
\hspace{0.000001\textwidth}
\psfig{figure=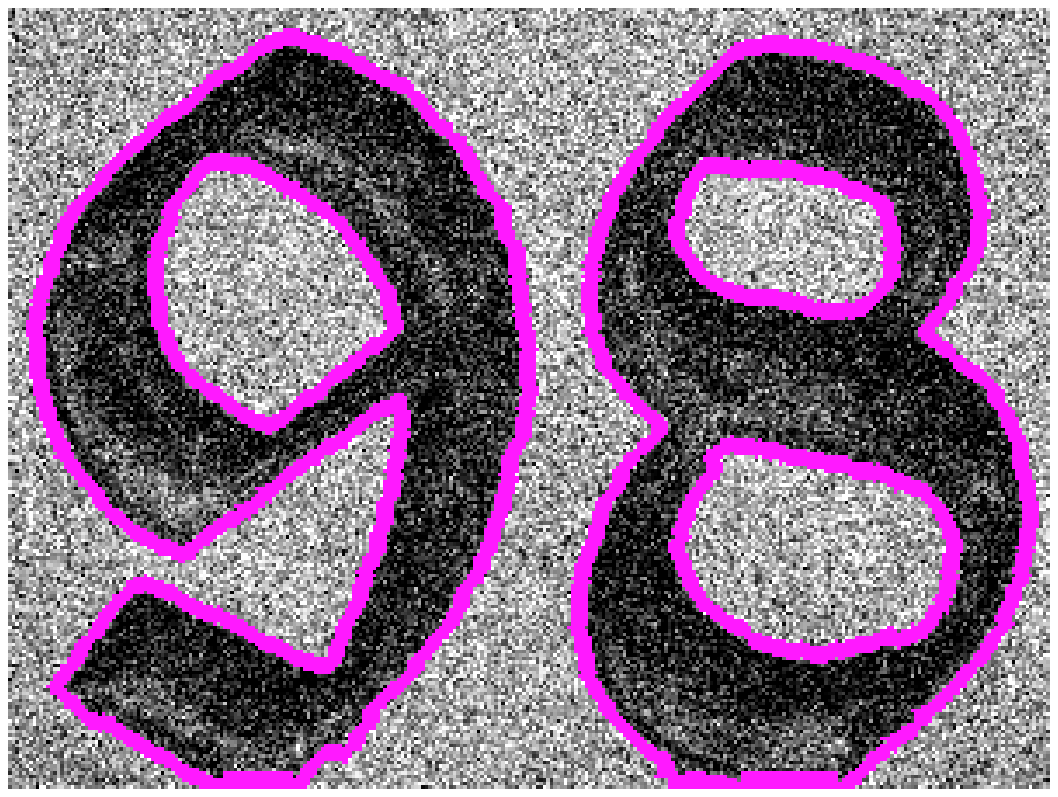,height=0.13\textwidth, width=0.2\textwidth}
\hspace{0.000001\textwidth}
\psfig{figure=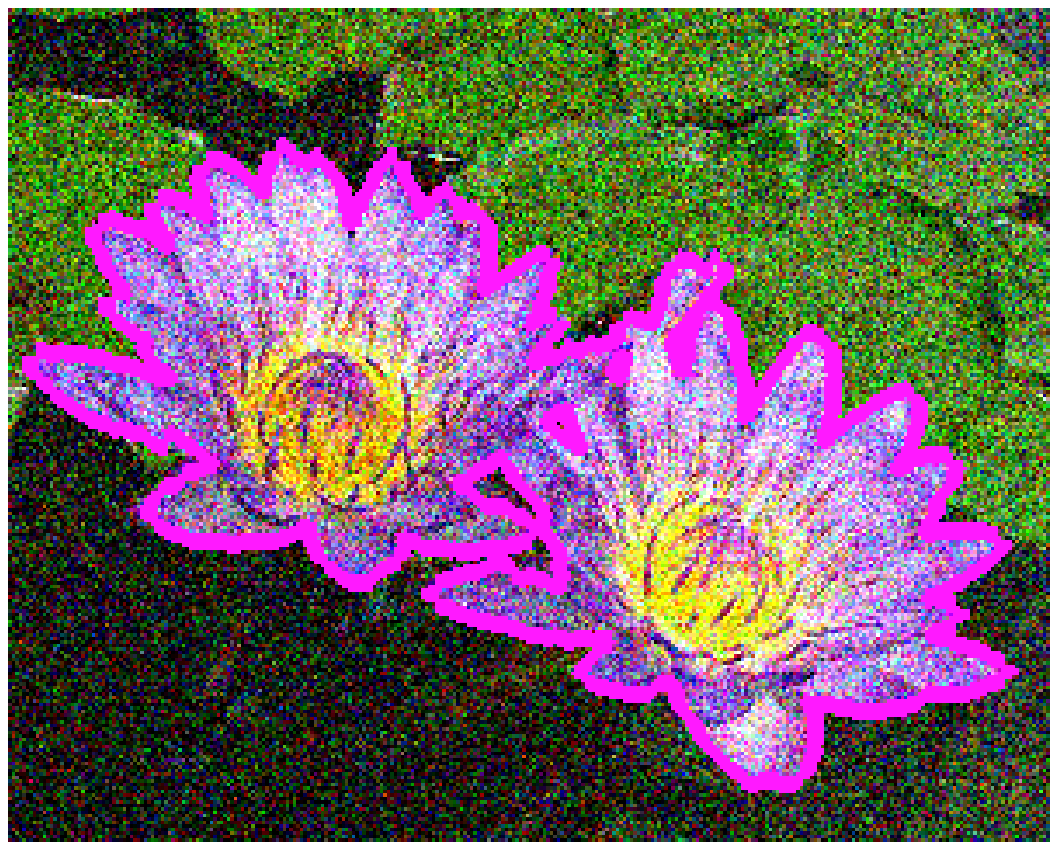,height=0.13\textwidth, width=0.2\textwidth}
\hspace{0.000001\textwidth}
\psfig{figure=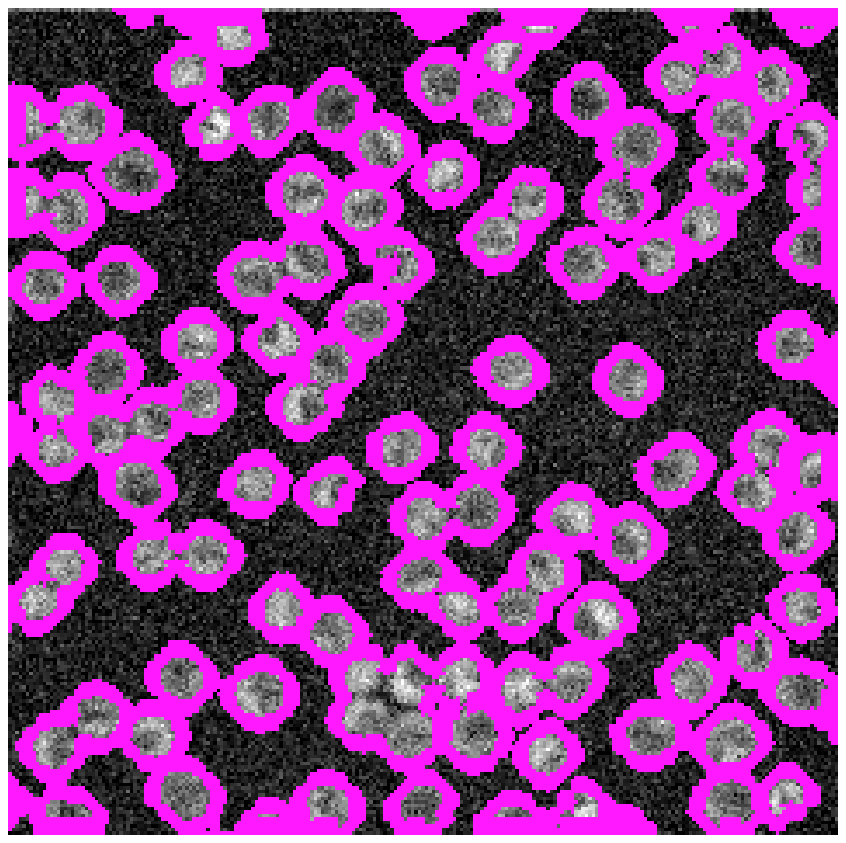,height=0.13\textwidth, width=0.2\textwidth}}
\centerline{(a)\hspace{0.17\textwidth} (b)\hspace{0.17\textwidth} (c)\hspace{0.17\textwidth} (d)\hspace{0.17\textwidth} (e)}
\caption{Visual comparison of segmentation results of corrupted (a) 323016 image by Gaussian noise ($\mu = 0.0$, $\sigma = 0.05$), (b) Blood Cell1 image with Gaussian noise ($\mu = 0.0$, $\sigma = 0.01$), (c) 98 image with Gaussian noise ($\mu = 0.0$, $\sigma = 0.05$), (d) Lily flower image with Gaussian noise ($\mu = 0.0$, $\sigma = 0.05$), (e) Blood Cell4 image by Gaussian noise ($\mu = 0.0$, $\sigma = 0.01$) using various techniques. \textbf{First row:} images with initial contours, \textbf{Second row:} segmentation results obtained by {\sc FEAC}, \textbf{Third row:} segmentation results obtained by {\sc NFACMKM}, \textbf{Fourth row:} segmentation results obtained by {\sc FACGK}, \textbf{Fifth row:} segmentation results obtained by {\sc LPFAC}, \textbf{Sixth row:} segmentation results obtained by {\sc FDFEAC}, \textbf{Seventh row:} segmentation results obtained by {\sc GLFEAC} and \textbf{Eighth row:} segmentation results obtained by {\sc RGLFEAC}.\label{figure_gn_seg}}
\end{figure}

\subsection{Segmentation of Corrupted Images by Gaussian Noise}

We conducted another experiment to analyze performance of these considered methods on segmentation of corrupted images by different level of Gaussian noise. Different levels of Gaussian noise are generated to corrupt the considered images. 
Region in-homogeneity is increased due to addition of Gaussian noise to the original images. Figure~\ref{figure_gn_seg} presents segmentation of the images corrupted by Gaussian noise. From the figure, it is observed that {\sc FEAC} is totally failed to produce proper segmentation of images under noisy environment. Due to incorporation of Gaussian kernel in the methods: {\sc NFACMKM} \& {\sc FACGK}, they are able to obtain good segmentation results for most of corrupted images. Local fuzzy energy based approaches: {\sc LPFAC} \& {\sc FDFEAC} obtained good segmentation for those images having lesser density noise. {\sc RGLFEAC} using local energy based on both spatial distance and gray level/color information provides the good segmentation results for most of the images. Table~\ref{table_gaussian} displays statistics on performance of all the methods.      

\begin{table}[htp]
\begin{center}
\begin{tabular}{|l|l|l|l|l|l|l|l|}\hline
Measures & \multicolumn{7}{|l|}{Techniques } \\\cline{2-8}
& {\sc m1} & {\sc m2} & {\sc m3} & {\sc m4} & {\sc m5} & {\sc m6} & {\sc m7}\\\cline{1-8}
Ave. Jacard error & 0.284 & 0.161 & 0.174 & 0.329 & 0.298 & 0.242 & \textcolor[rgb]{0.00,1.00,0.00}{0.068} \\\hline
Ave. F-measure    & 0.834& 0.910 & 0.902 & 0.745 &0.799 & 0.853& \textcolor[rgb]{1.00,0.00,0.00}{0.964}  \\\hline
\end{tabular}
\end{center}
\caption{Quantitative comparison among various techniques with respect to average Jacard error and average F-measure over $100$ corrupted images with Gaussian noise. {\sc m1: feac}, {\sc m2: nfacmkm}, {\sc m3: facgk}, {\sc m4: lpfac}, {\sc m5: fdfeac}, {\sc m6: glfeac} and {\sc m7: rglfeac}. Green colored numeric value indicates least average Jacard error corresponds to the best segmentation result. Whereas, red colored numeric value indicates highest F-measure corresponds to the best segmentation result. \label{table_gaussian}}
\end{table}

\begin{figure}
\centerline{
\psfig{figure=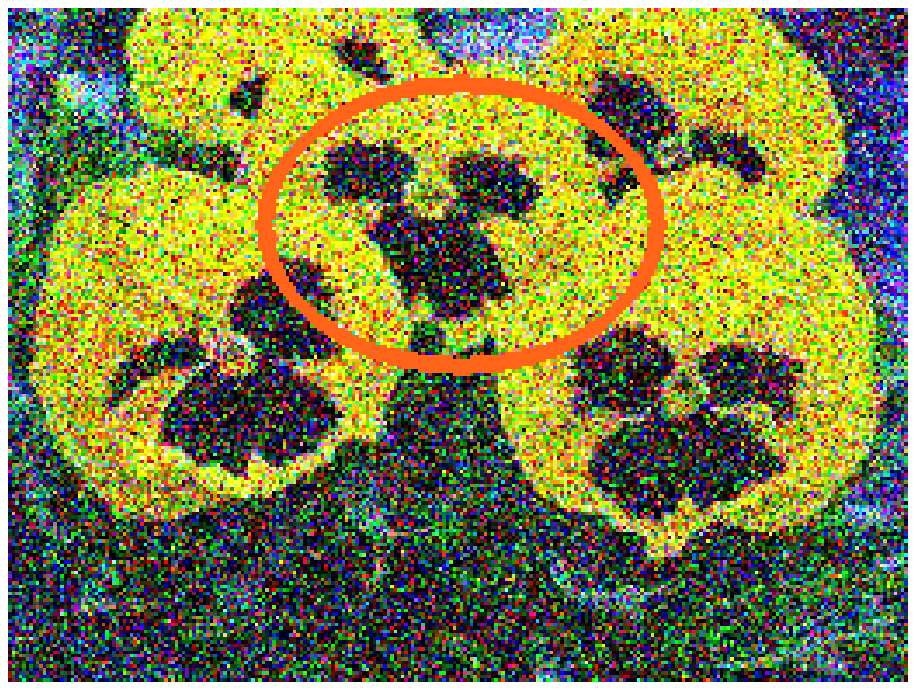,height=0.13\textwidth, width=0.2\textwidth}
\hspace{0.000001\textwidth}
\psfig{figure=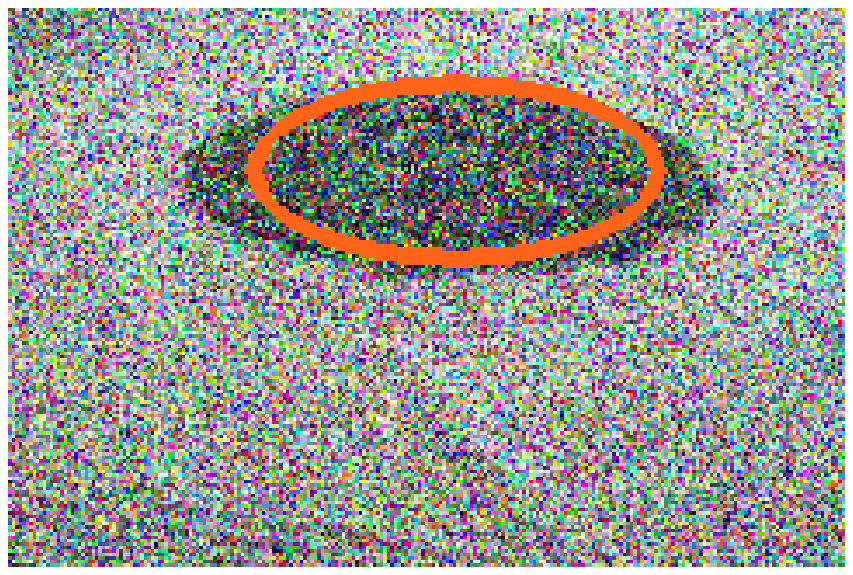,height=0.13\textwidth, width=0.2\textwidth}
\hspace{0.000001\textwidth}
\psfig{figure=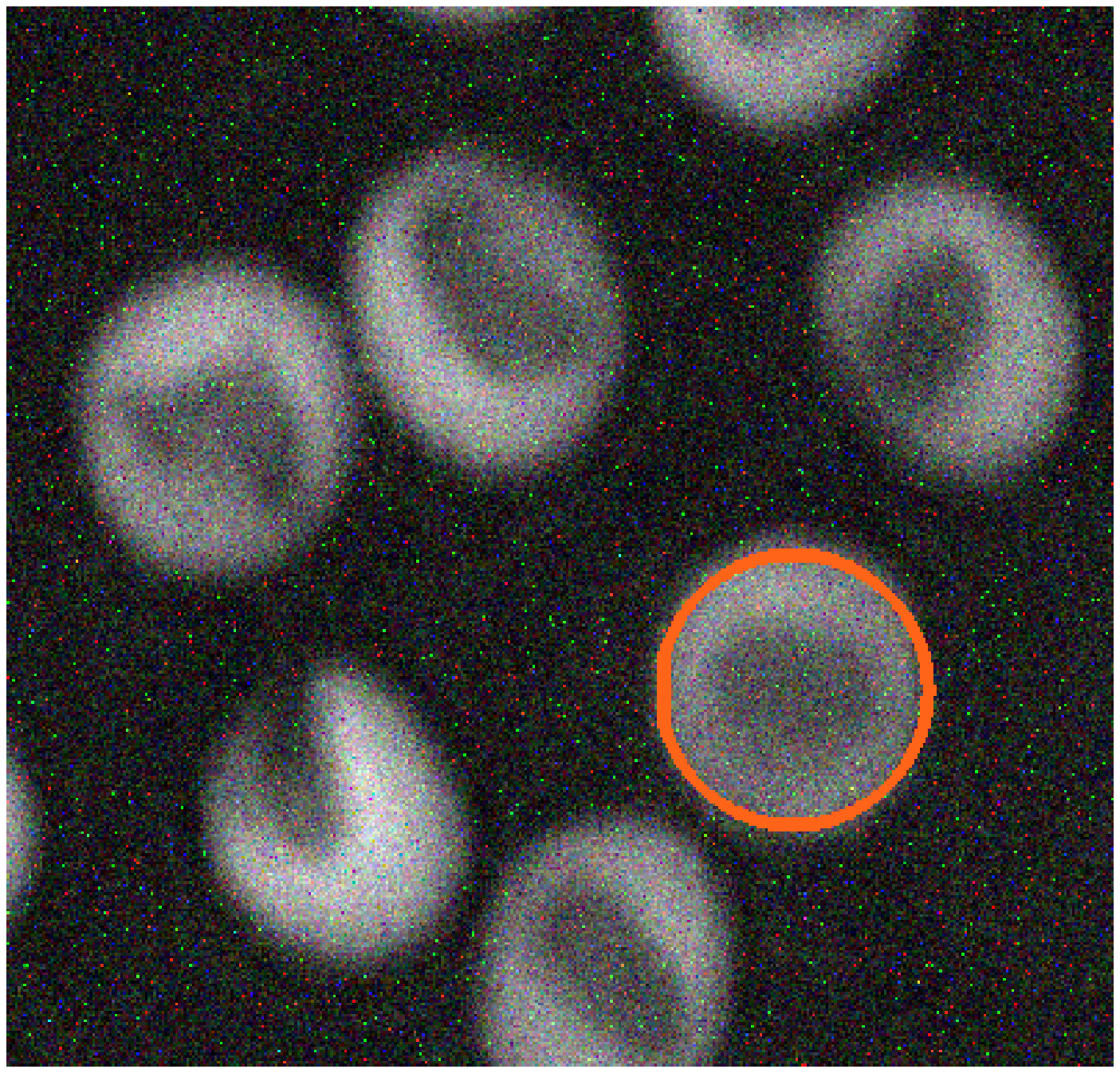,height=0.13\textwidth, width=0.2\textwidth}
\hspace{0.000001\textwidth}
\psfig{figure=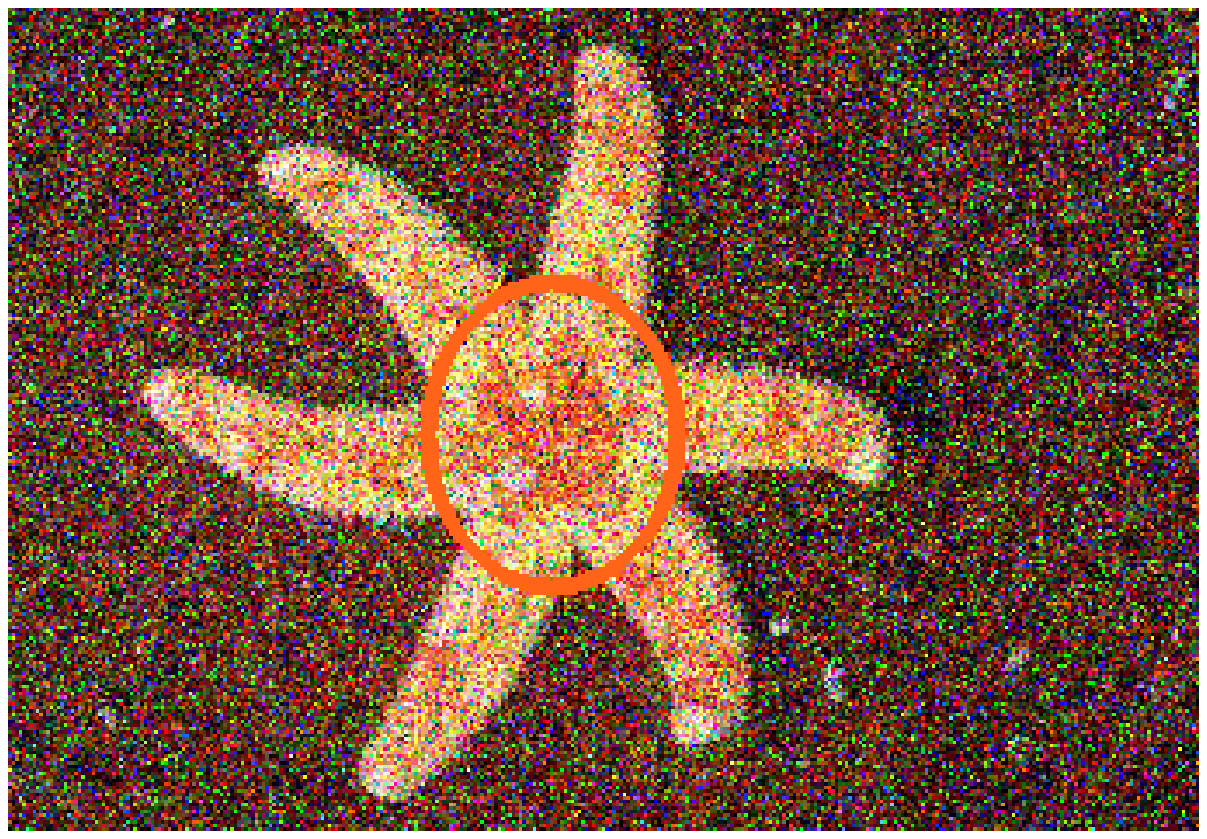,height=0.13\textwidth, width=0.2\textwidth}
\hspace{0.000001\textwidth}
\psfig{figure=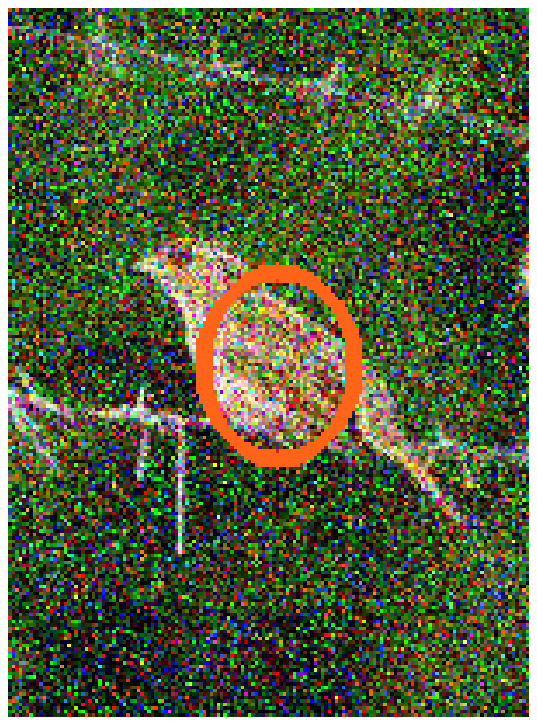,height=0.13\textwidth, width=0.2\textwidth}}
\vspace{0.000001\textwidth}
\centerline{
\psfig{figure=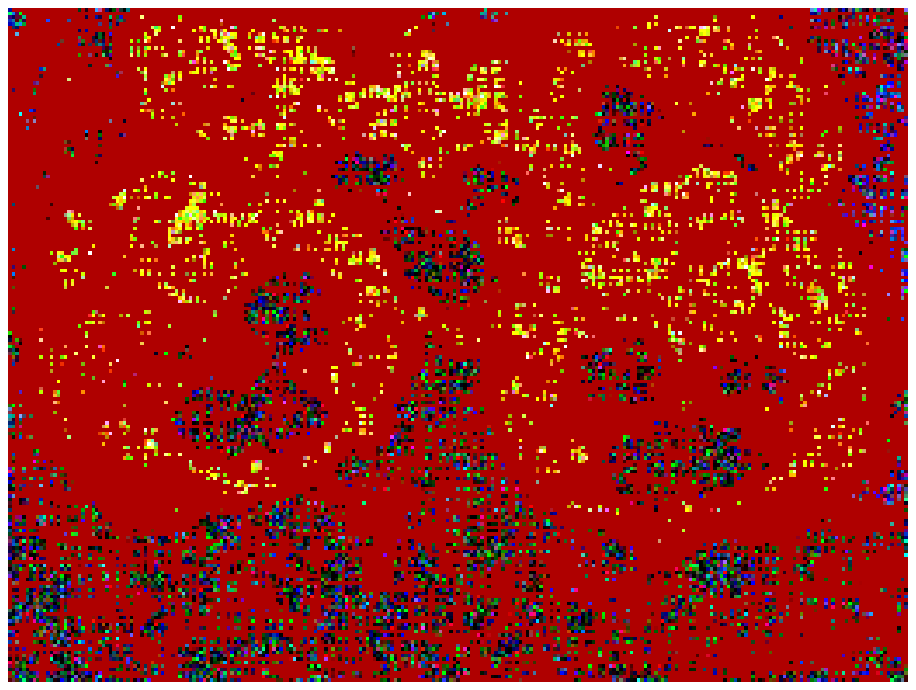,height=0.13\textwidth, width=0.2\textwidth}
\hspace{0.000001\textwidth}
\psfig{figure=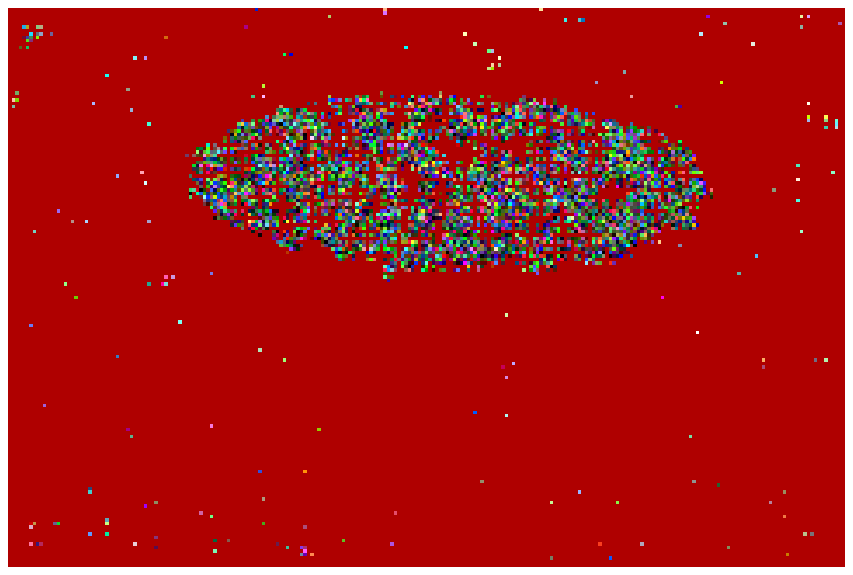,height=0.13\textwidth, width=0.2\textwidth}
\hspace{0.000001\textwidth}
\psfig{figure=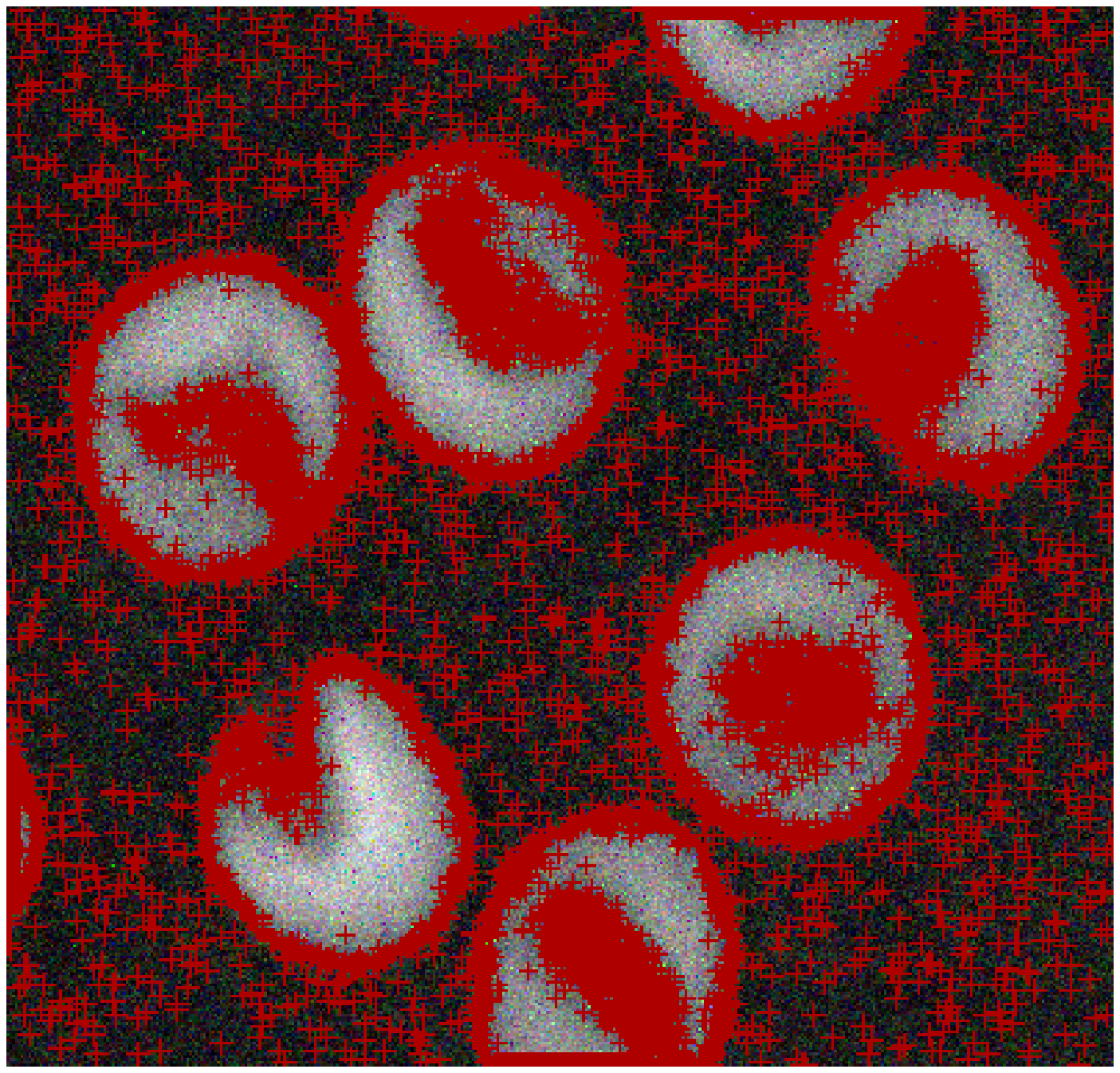,height=0.13\textwidth, width=0.2\textwidth}
\hspace{0.000001\textwidth}
\psfig{figure=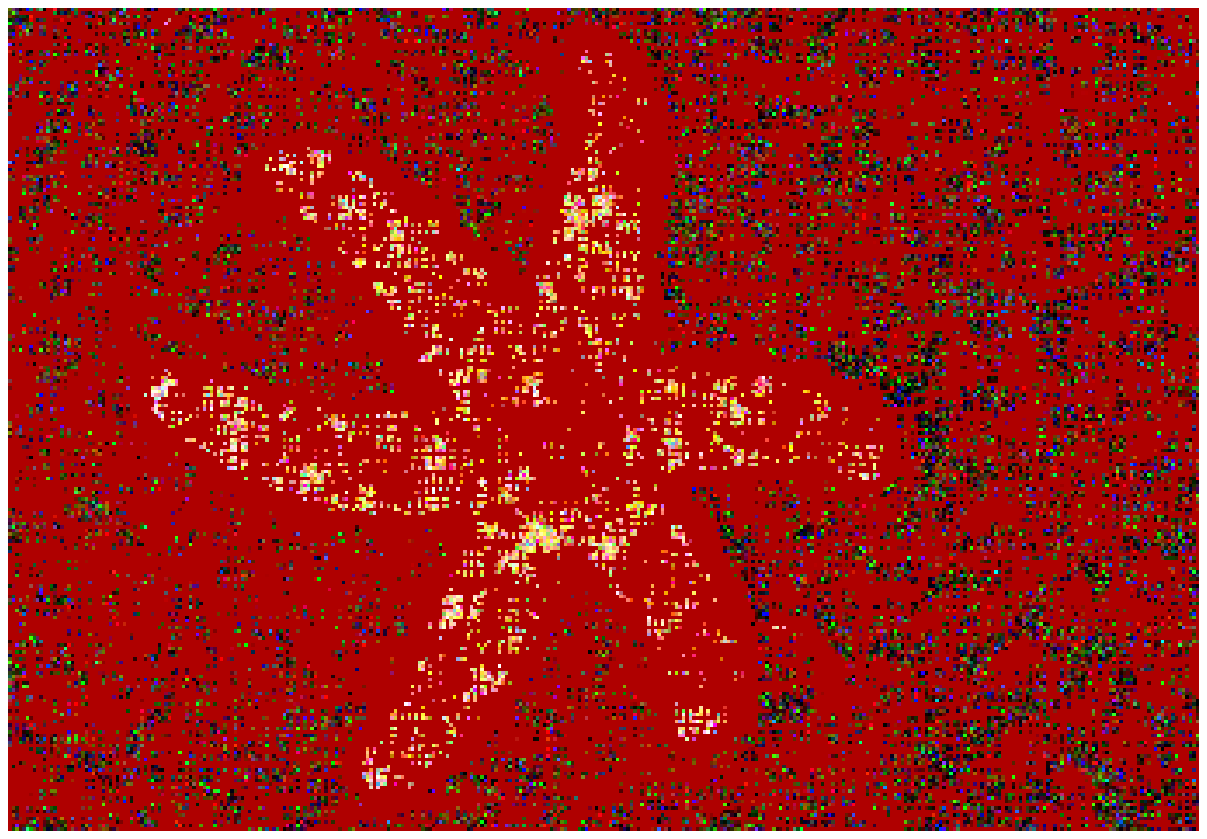,height=0.13\textwidth, width=0.2\textwidth}
\hspace{0.000001\textwidth}
\psfig{figure=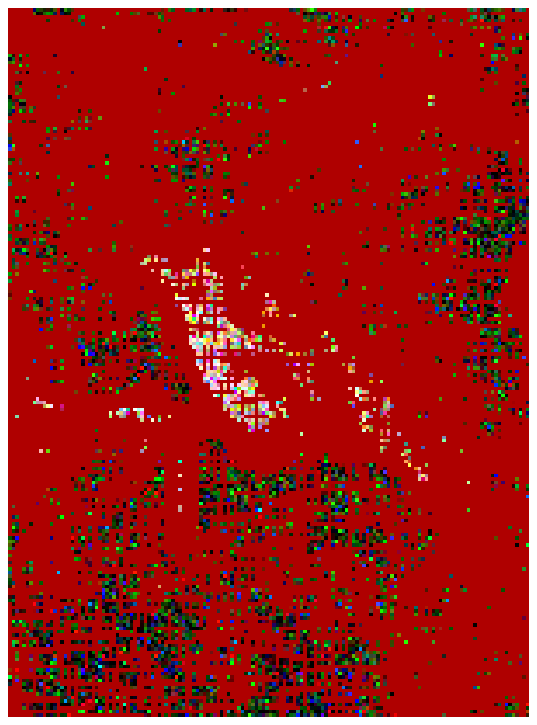,height=0.13\textwidth, width=0.2\textwidth}}
\vspace{0.000001\textwidth}
\centerline{
\psfig{figure=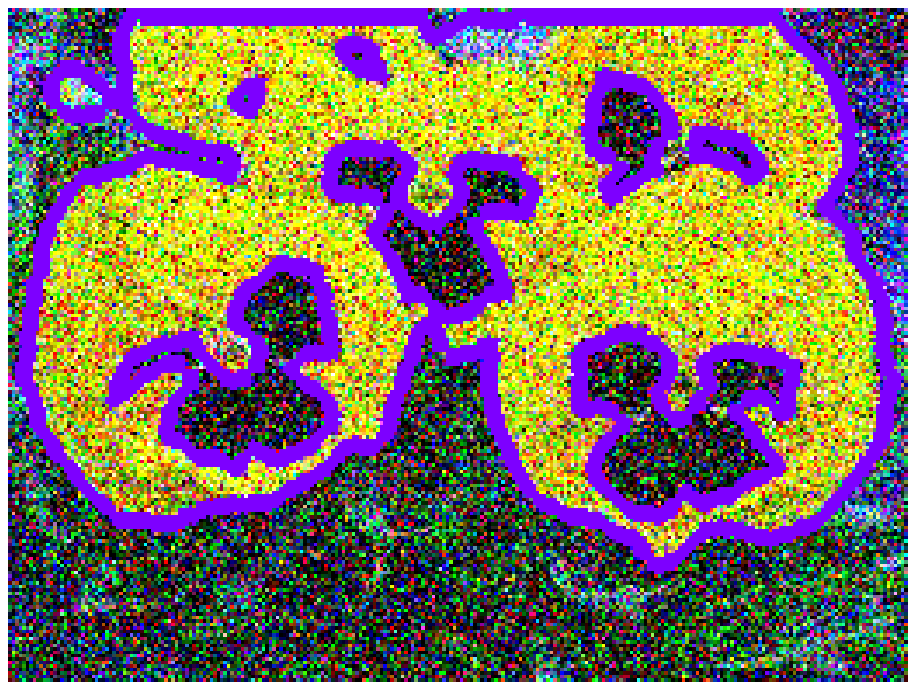,height=0.13\textwidth, width=0.2\textwidth}
\hspace{0.000001\textwidth}
\psfig{figure=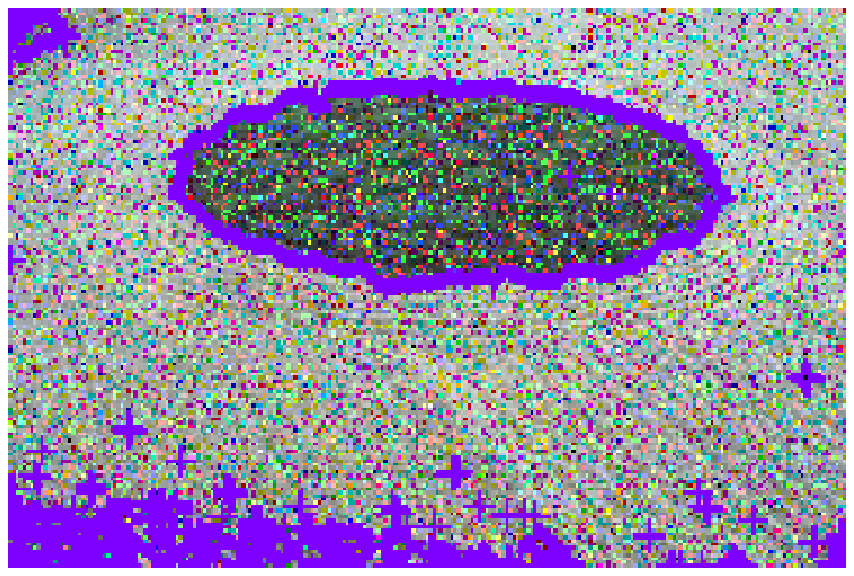,height=0.13\textwidth, width=0.2\textwidth}
\hspace{0.000001\textwidth}
\psfig{figure=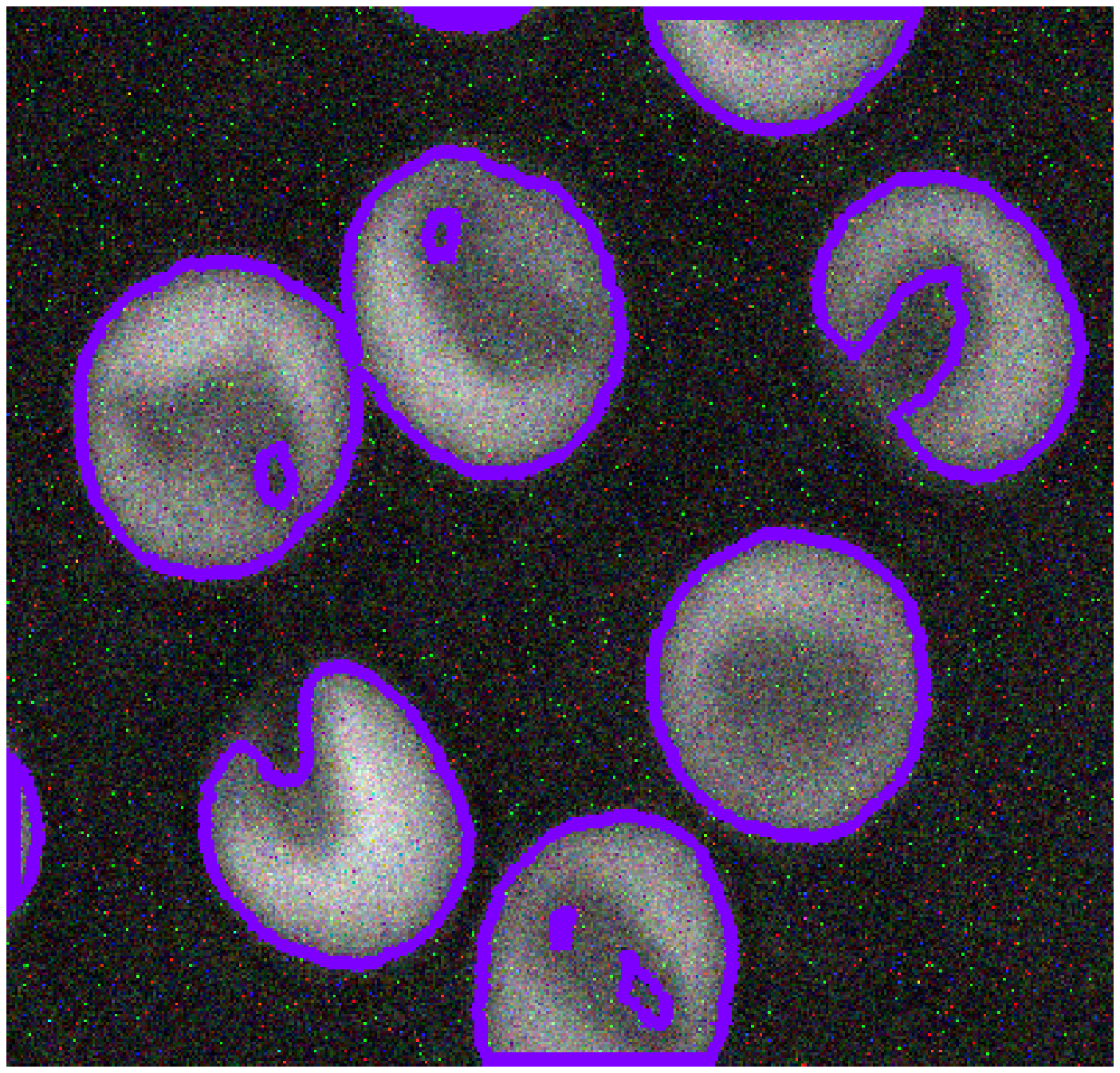,height=0.13\textwidth, width=0.2\textwidth}
\hspace{0.000001\textwidth}
\psfig{figure=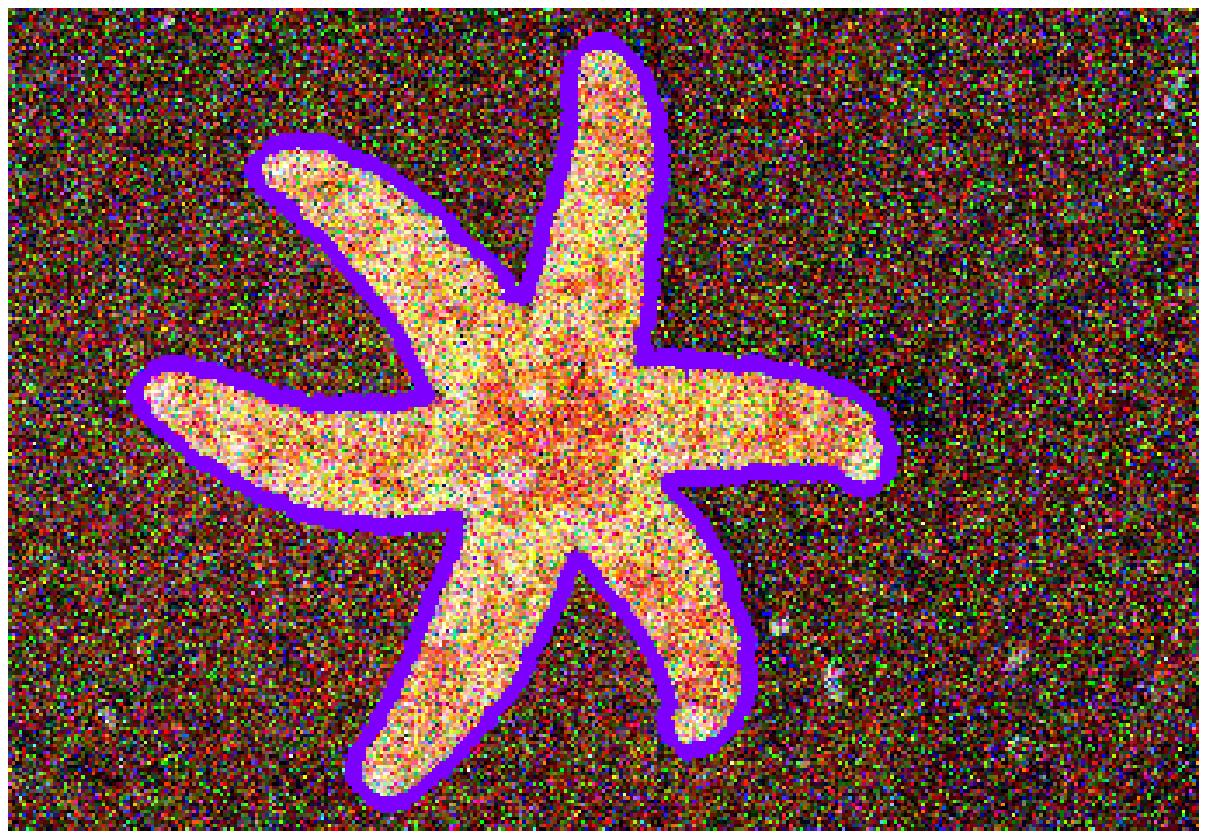,height=0.13\textwidth, width=0.2\textwidth}
\hspace{0.000001\textwidth}
\psfig{figure=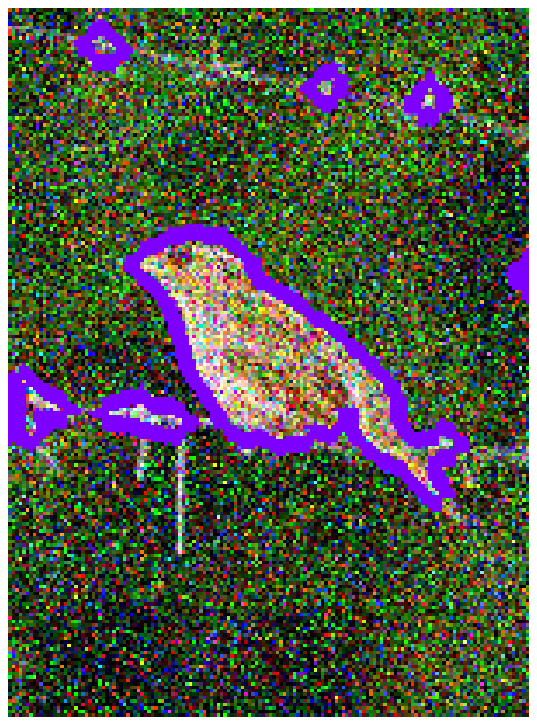,height=0.13\textwidth, width=0.2\textwidth}}
\vspace{0.000001\textwidth}
\centerline{
\psfig{figure=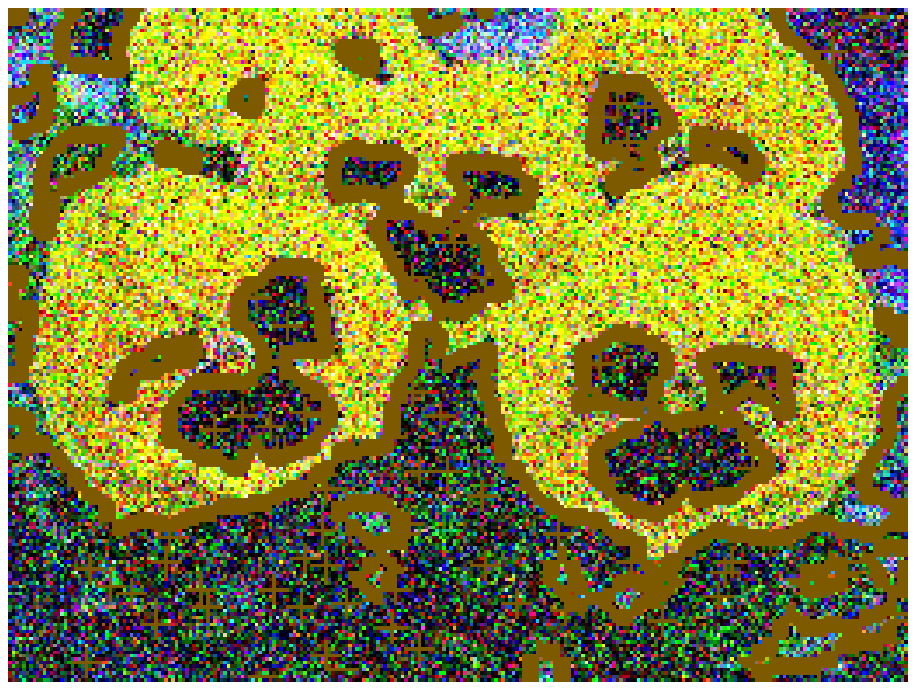,height=0.13\textwidth, width=0.2\textwidth}
\hspace{0.000001\textwidth}
\psfig{figure=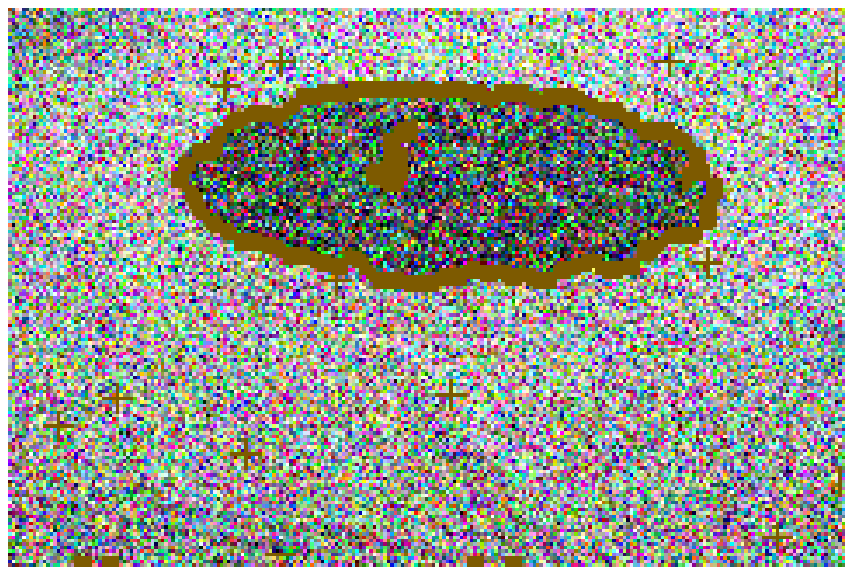,height=0.13\textwidth, width=0.2\textwidth}
\hspace{0.000001\textwidth}
\psfig{figure=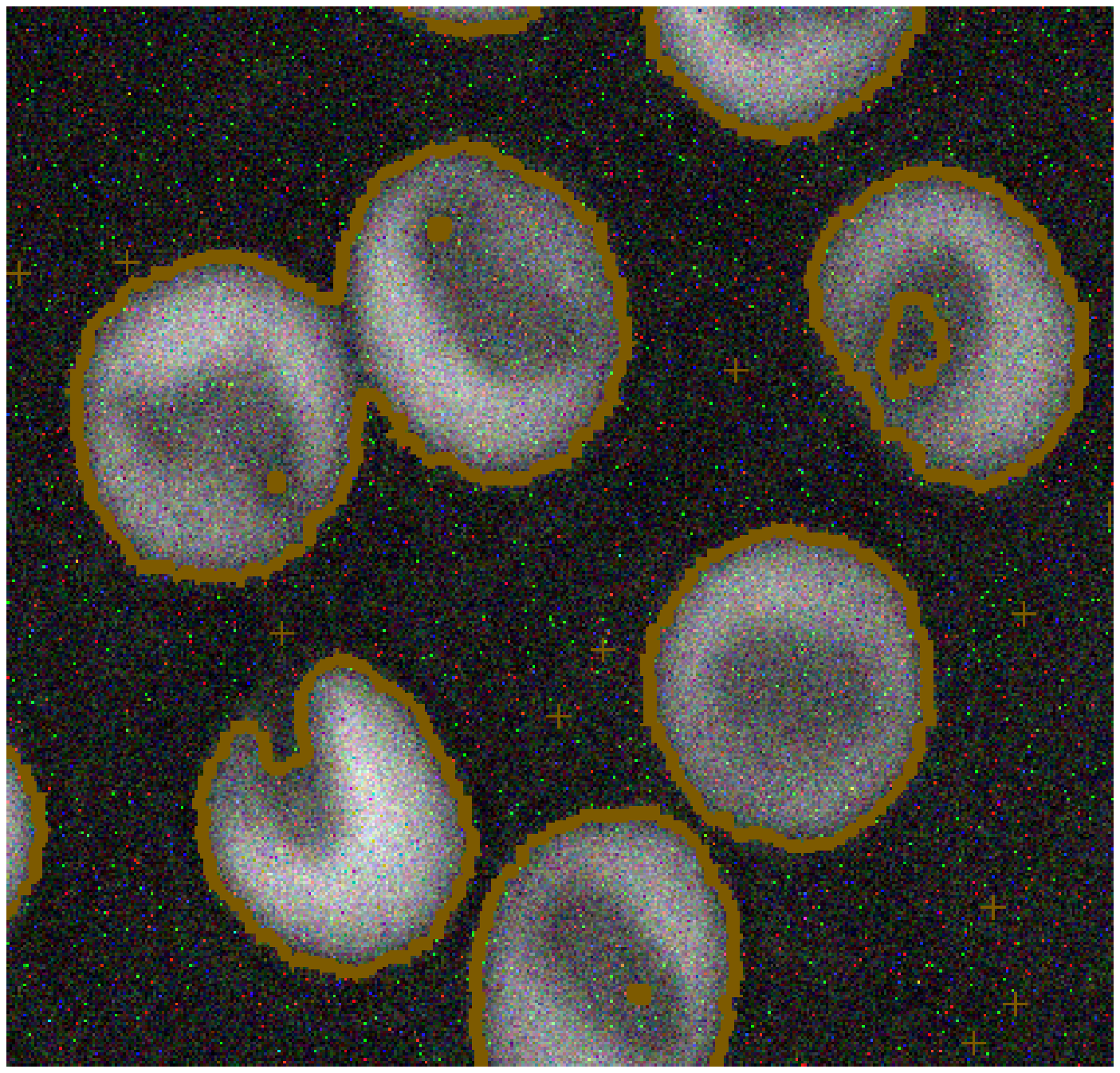,height=0.13\textwidth, width=0.2\textwidth}
\hspace{0.000001\textwidth}
\psfig{figure=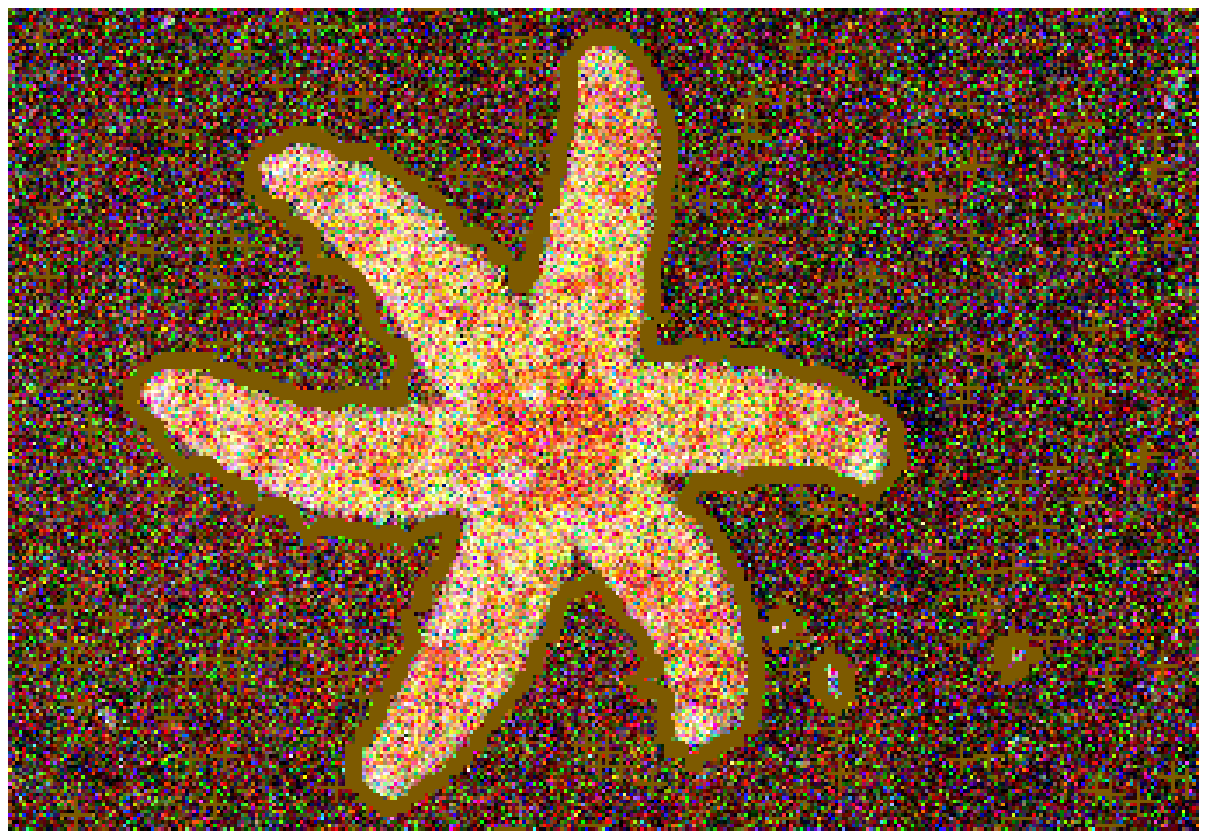,height=0.13\textwidth, width=0.2\textwidth}
\hspace{0.000001\textwidth}
\psfig{figure=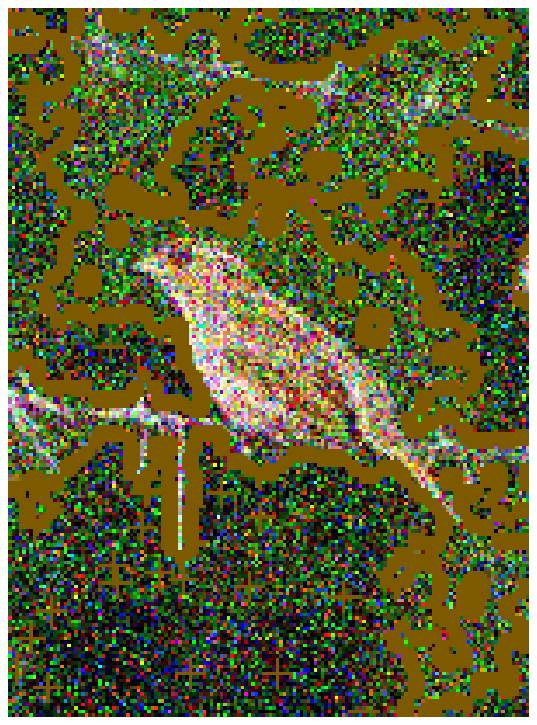,height=0.13\textwidth, width=0.2\textwidth}}
\vspace{0.000001\textwidth}
\centerline{
\psfig{figure=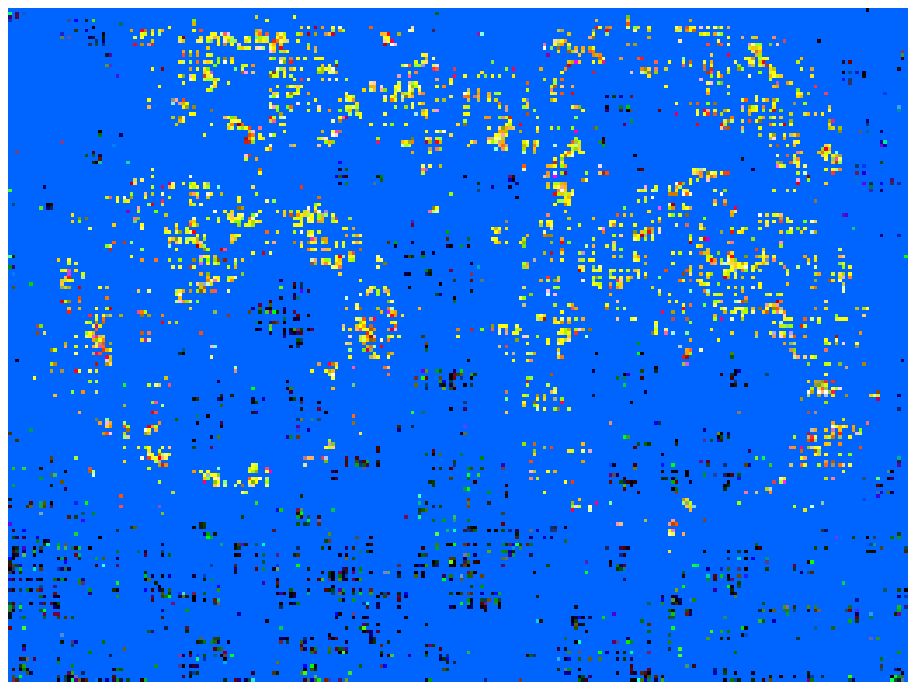,height=0.13\textwidth, width=0.2\textwidth}
\hspace{0.000001\textwidth}
\psfig{figure=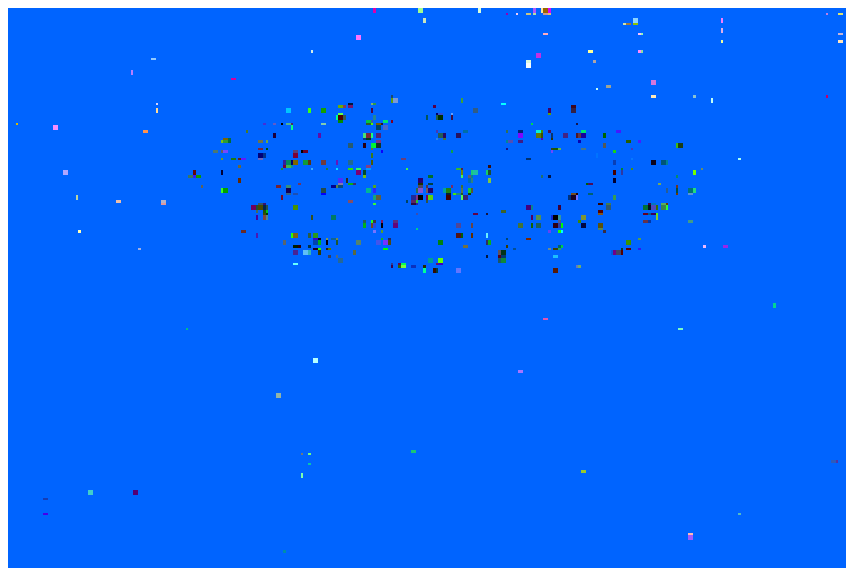,height=0.13\textwidth, width=0.2\textwidth}
\hspace{0.000001\textwidth}
\psfig{figure=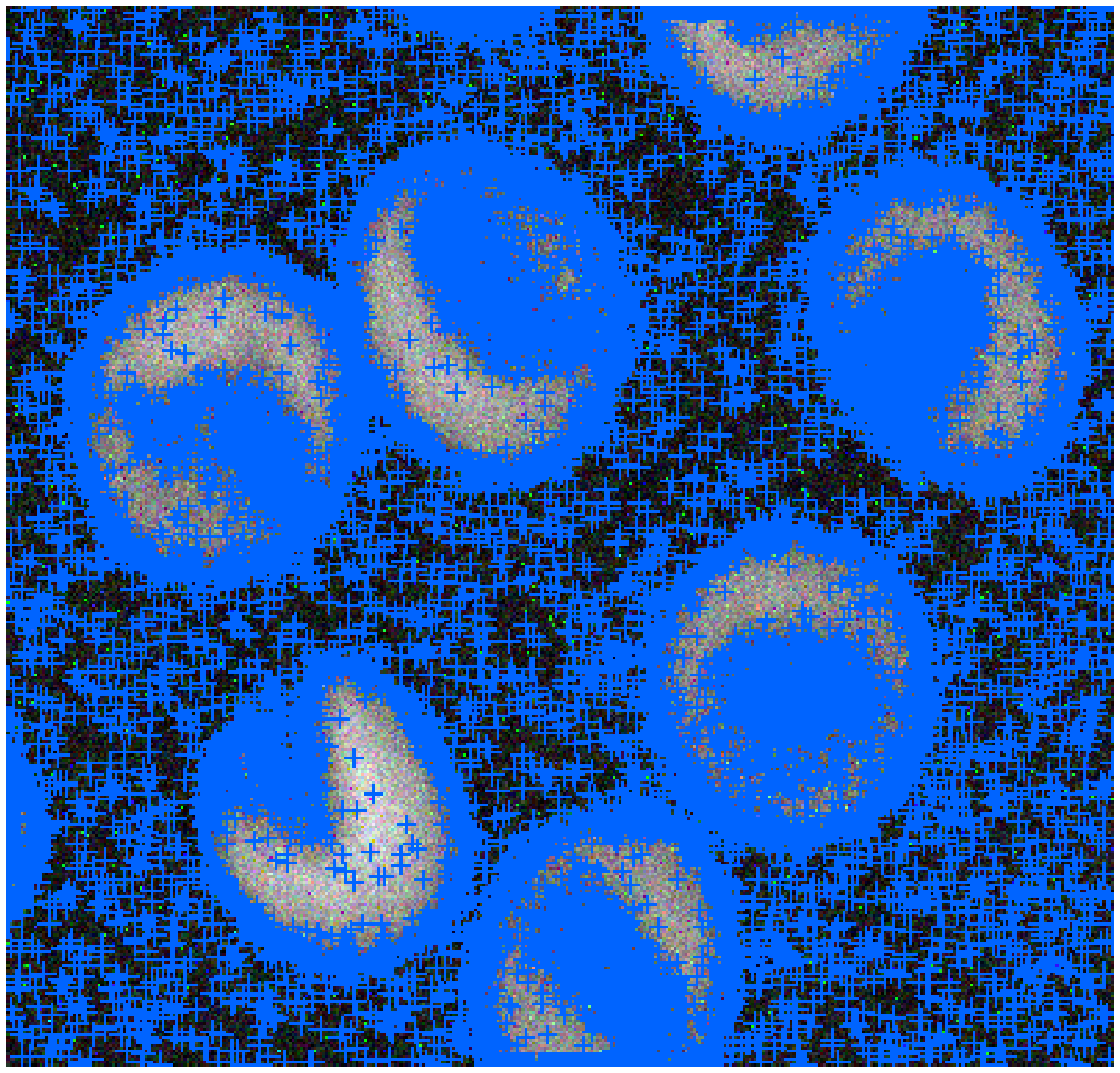,height=0.13\textwidth, width=0.2\textwidth}
\hspace{0.000001\textwidth}
\psfig{figure=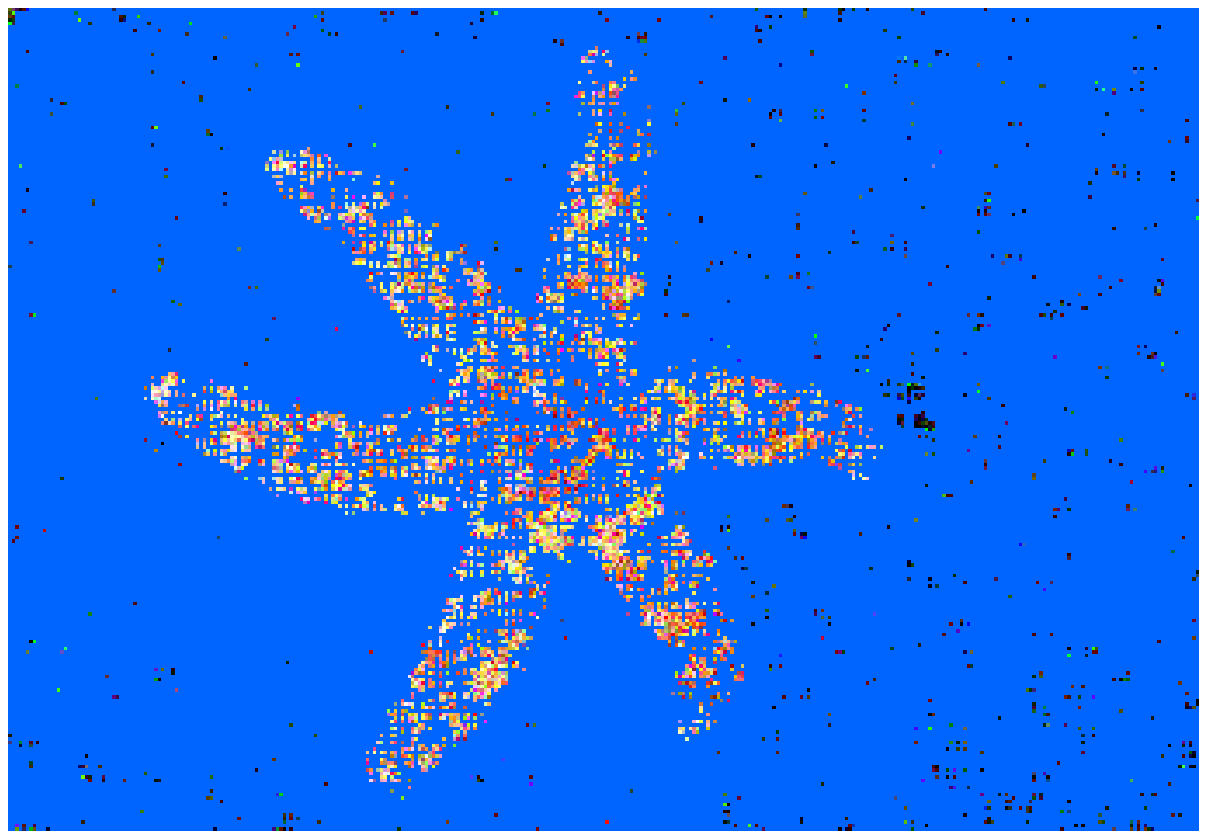,height=0.13\textwidth, width=0.2\textwidth}
\psfig{figure=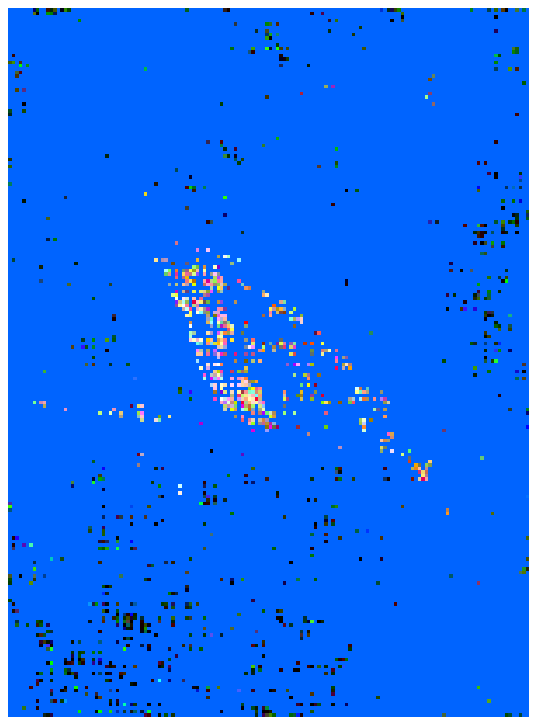,height=0.13\textwidth, width=0.2\textwidth}}
\vspace{0.000001\textwidth}
\centerline{
\psfig{figure=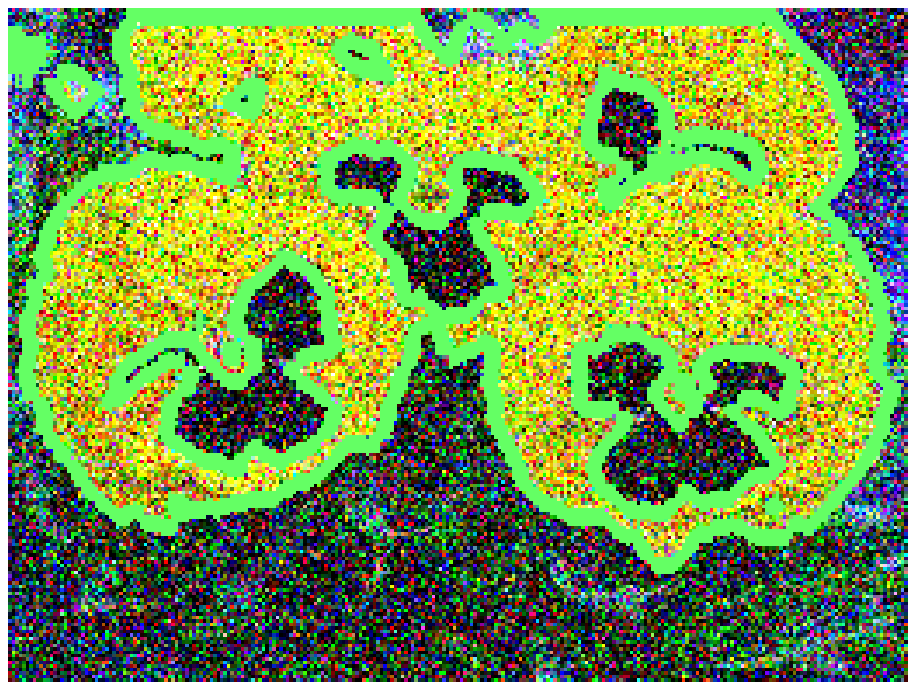,height=0.13\textwidth, width=0.2\textwidth}
\hspace{0.000001\textwidth}
\psfig{figure=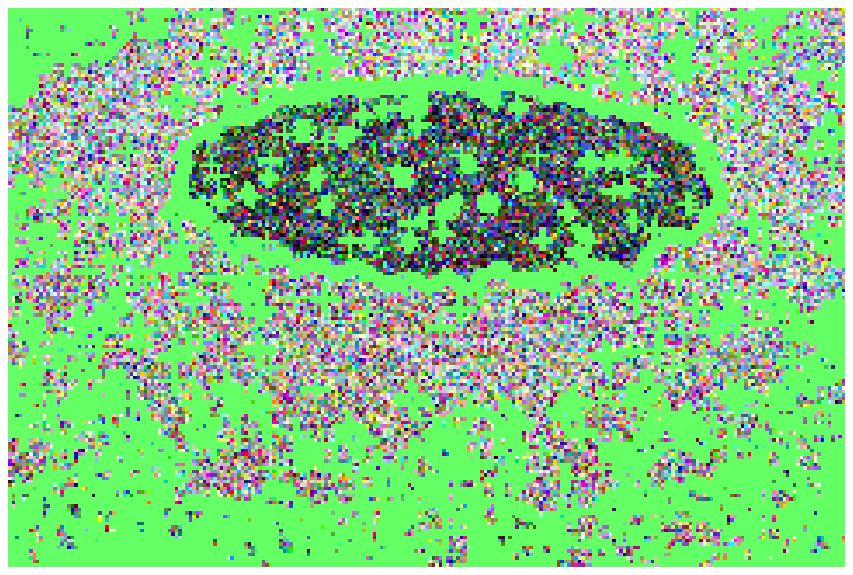,height=0.13\textwidth, width=0.2\textwidth}
\hspace{0.000001\textwidth}
\psfig{figure=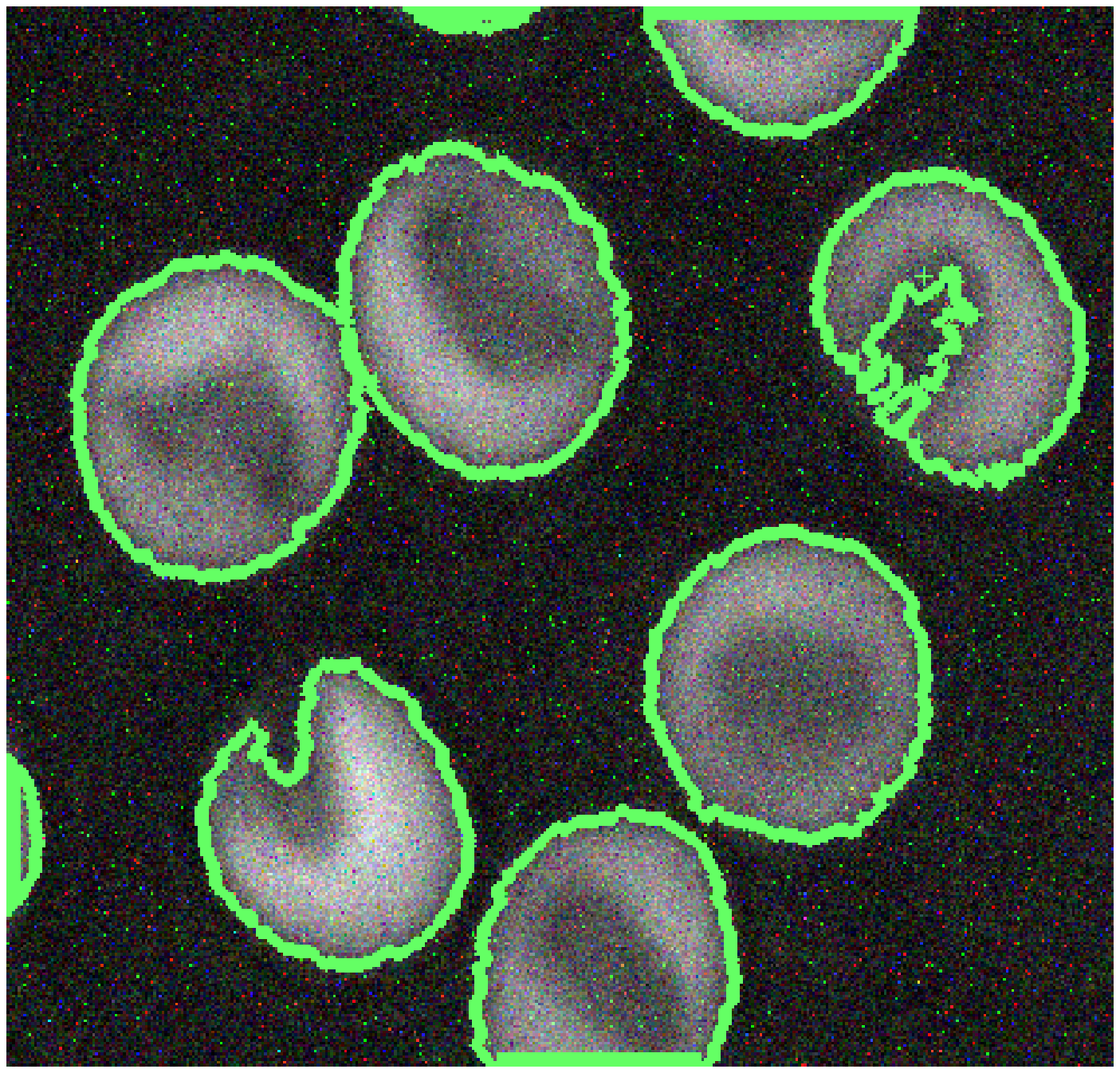,height=0.13\textwidth, width=0.2\textwidth}
\hspace{0.000001\textwidth}
\psfig{figure=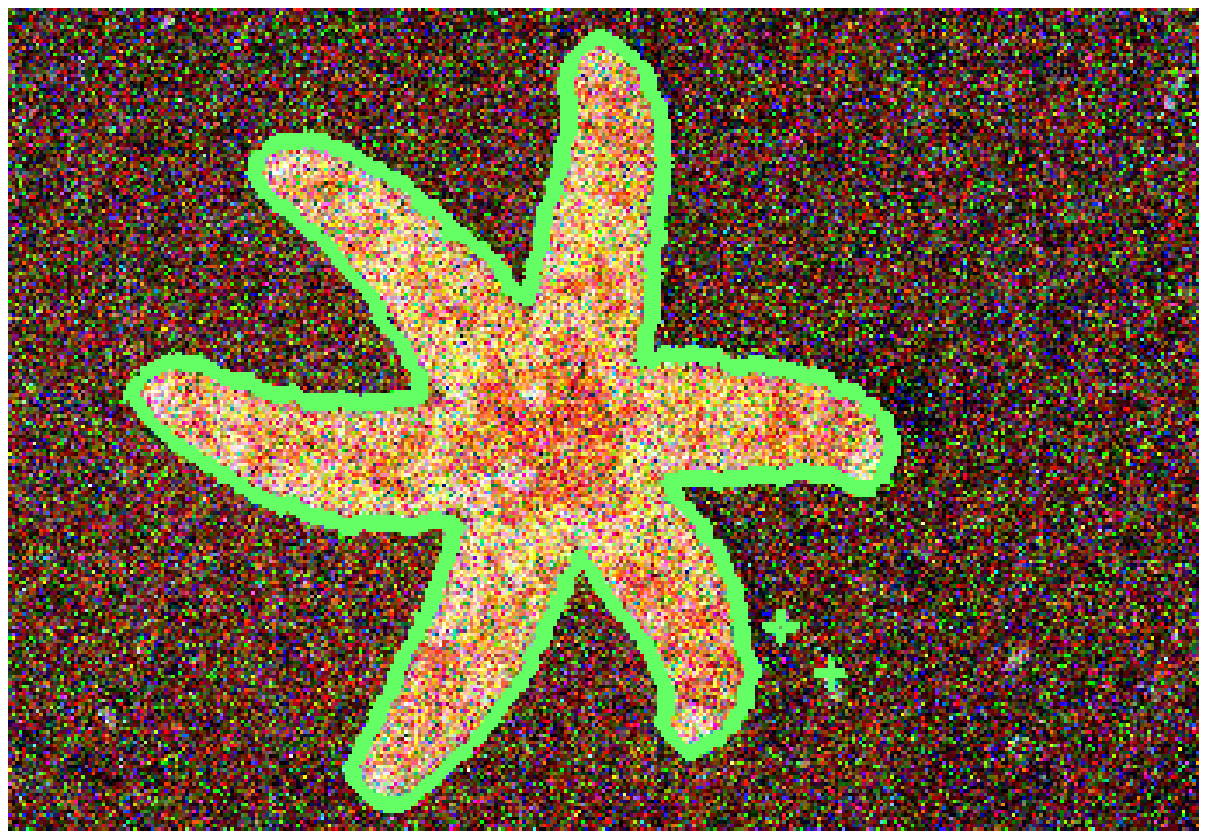,height=0.13\textwidth, width=0.2\textwidth}
\hspace{0.000001\textwidth}
\psfig{figure=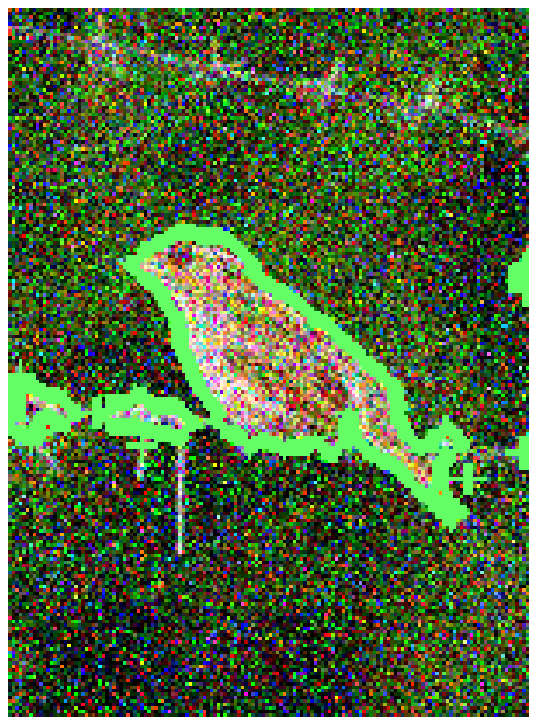,height=0.13\textwidth, width=0.2\textwidth}}
\vspace{0.000001\textwidth}
\centerline{
\psfig{figure=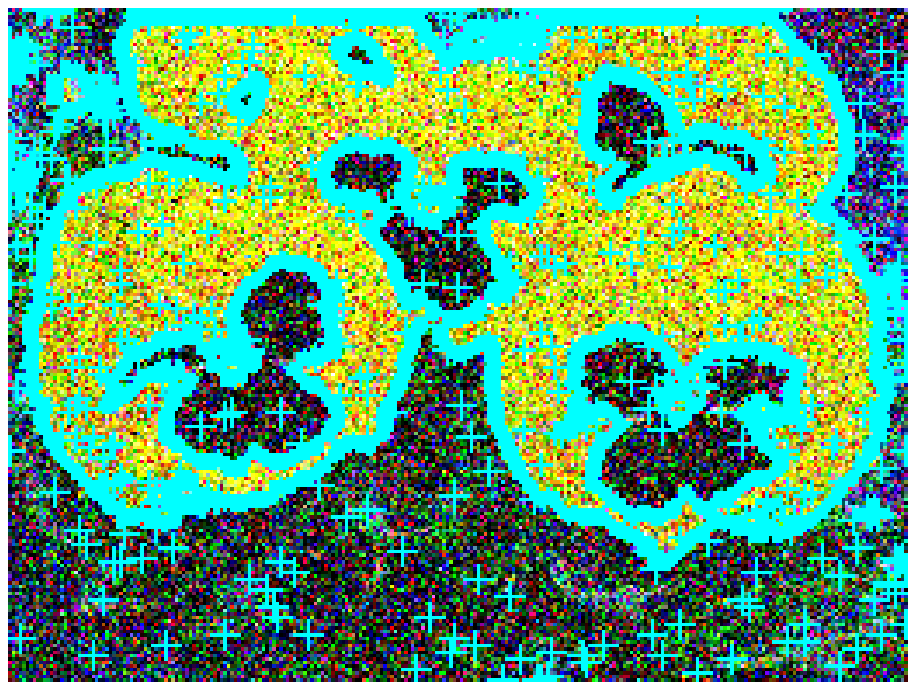,height=0.13\textwidth, width=0.2\textwidth}
\hspace{0.000001\textwidth}
\psfig{figure=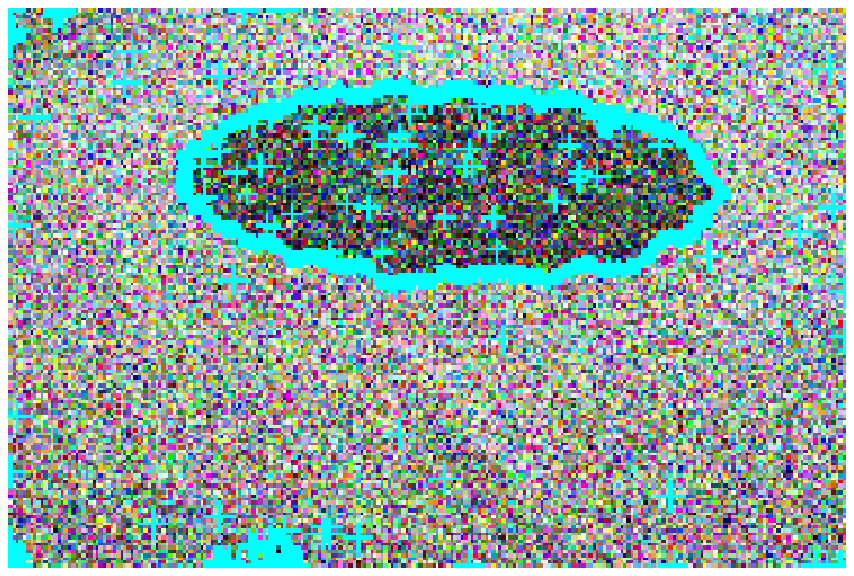,height=0.13\textwidth, width=0.2\textwidth}
\hspace{0.000001\textwidth}
\psfig{figure=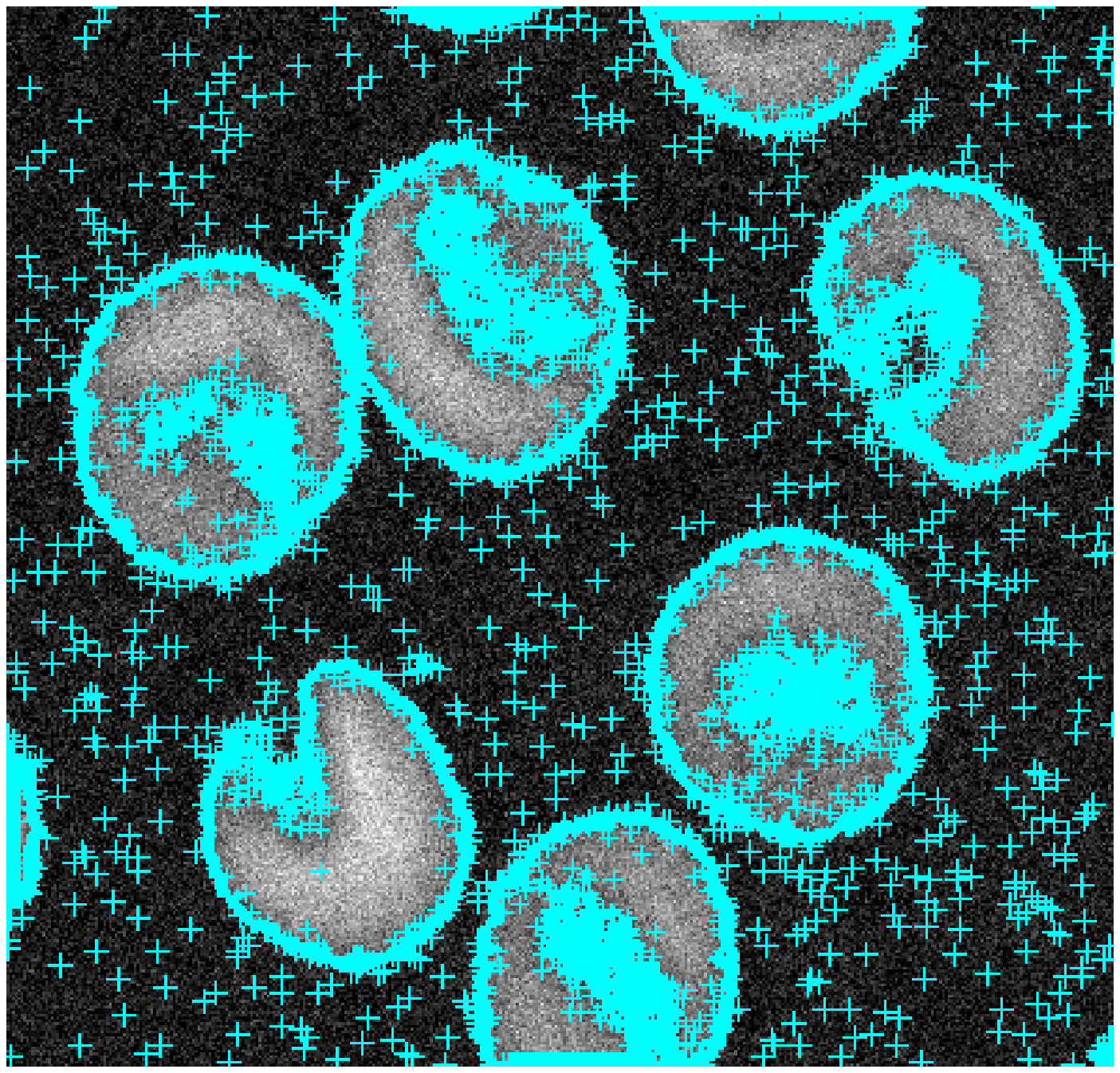,height=0.13\textwidth, width=0.2\textwidth}
\hspace{0.000001\textwidth}
\psfig{figure=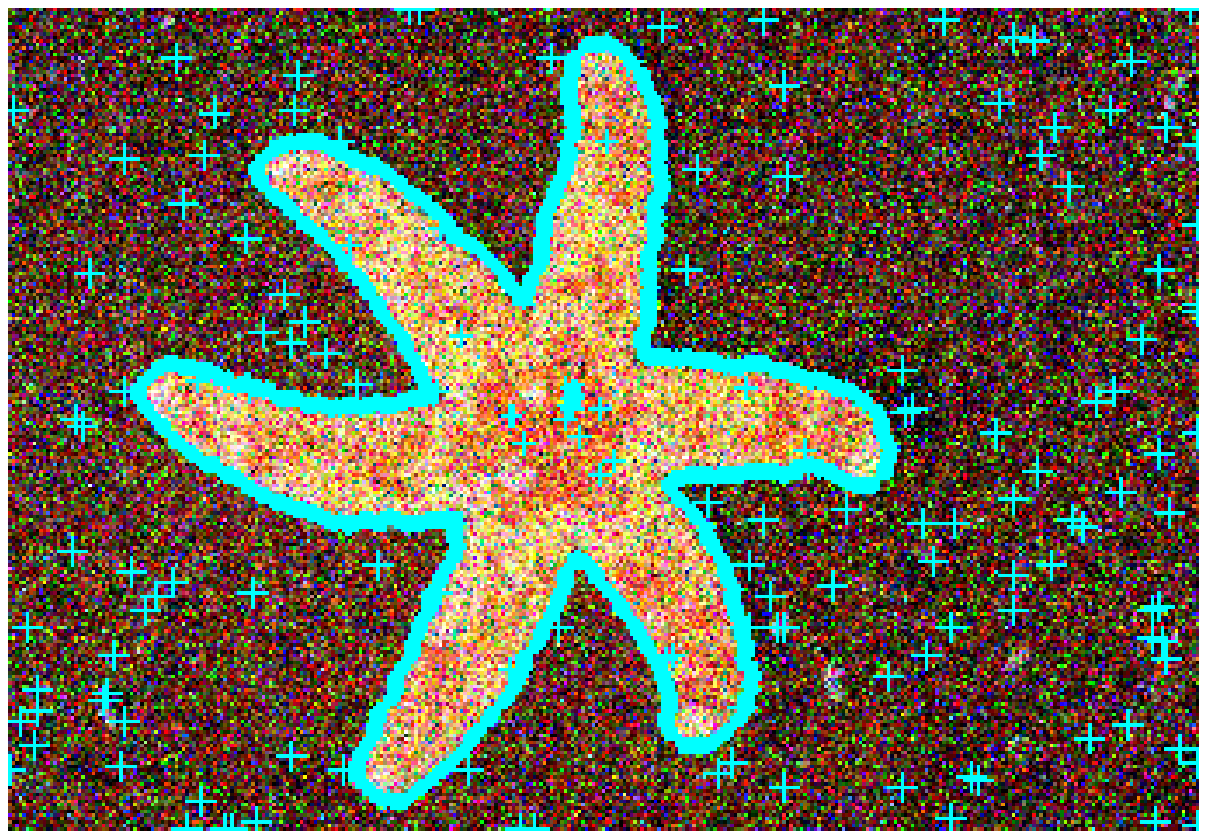,height=0.13\textwidth, width=0.2\textwidth}
\hspace{0.000001\textwidth}
\psfig{figure=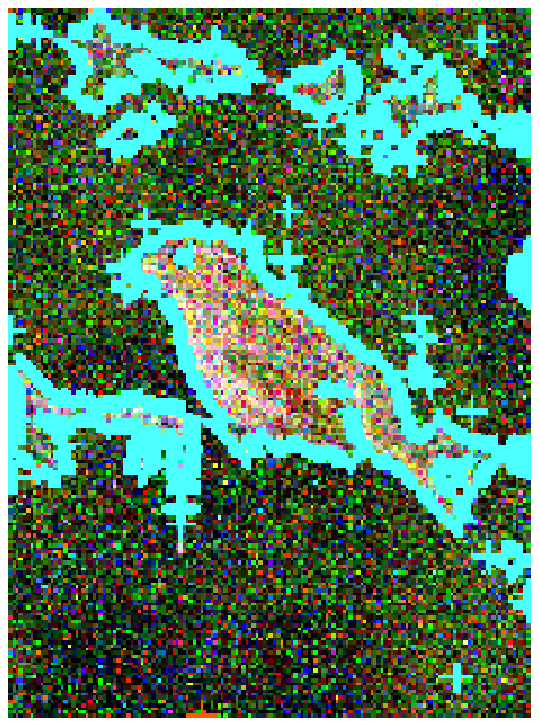,height=0.13\textwidth, width=0.2\textwidth}}
\vspace{0.000001\textwidth}
\centerline{
\psfig{figure=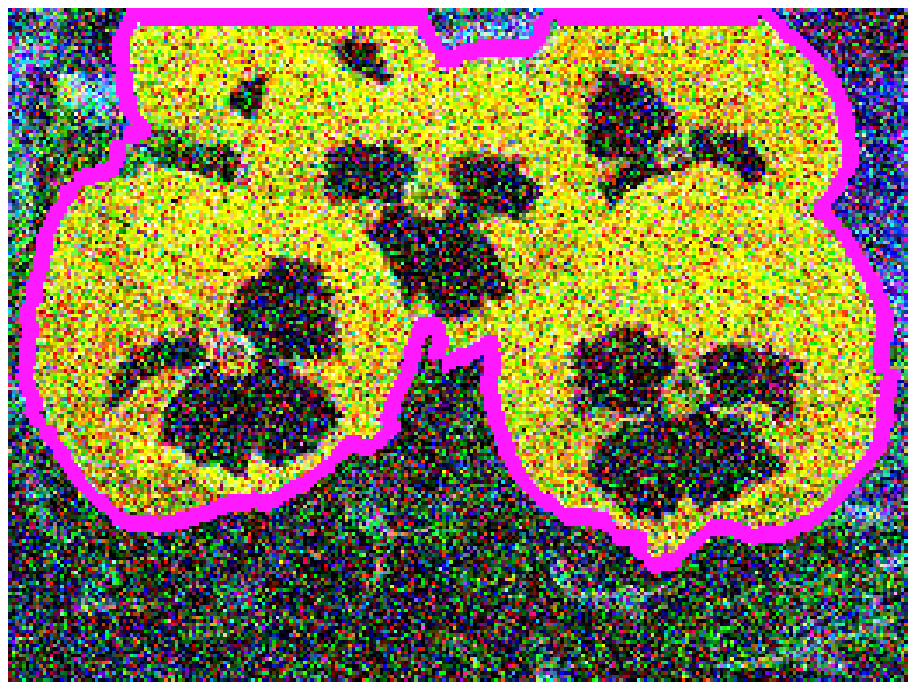,height=0.13\textwidth, width=0.2\textwidth}
\hspace{0.000001\textwidth}
\psfig{figure=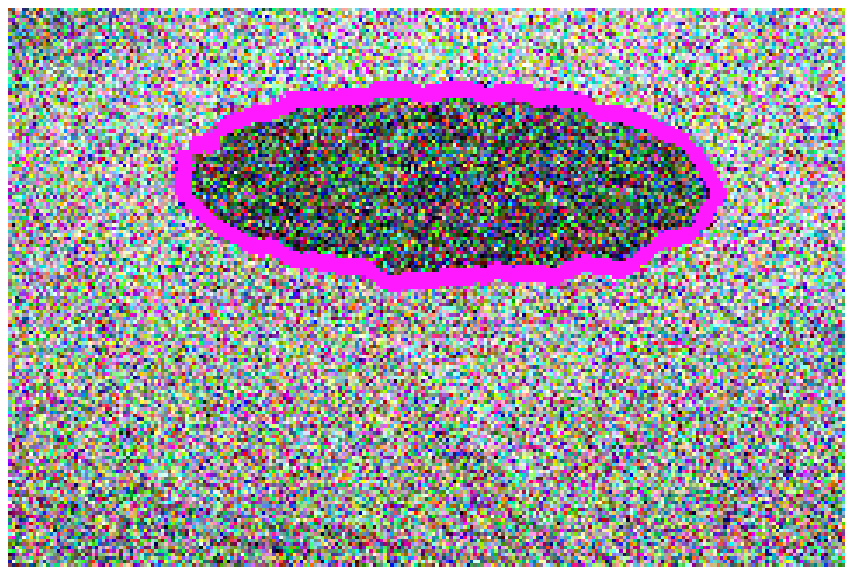,height=0.13\textwidth, width=0.2\textwidth}
\hspace{0.000001\textwidth}
\psfig{figure=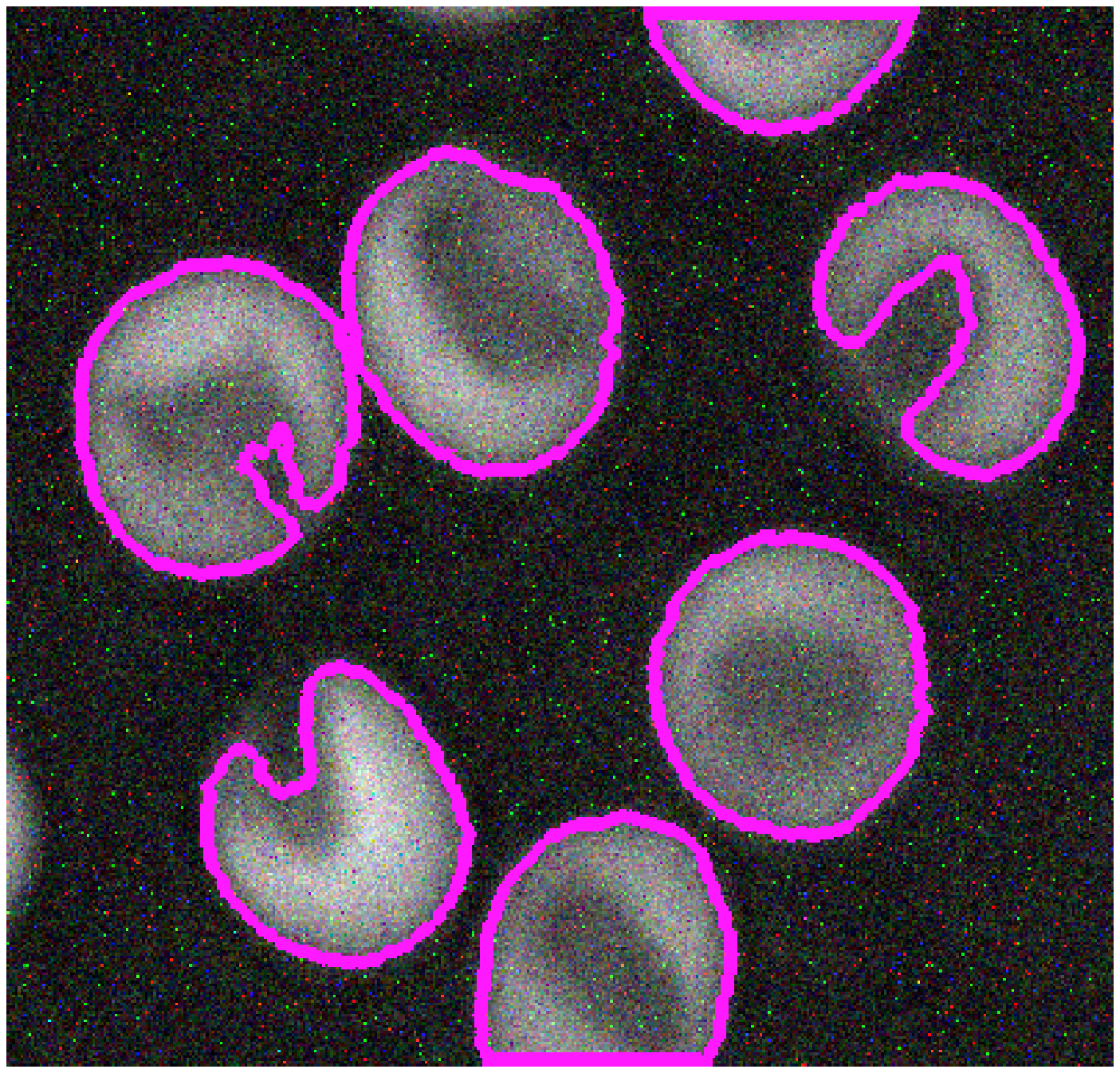,height=0.13\textwidth, width=0.2\textwidth}
\hspace{0.000001\textwidth}
\psfig{figure=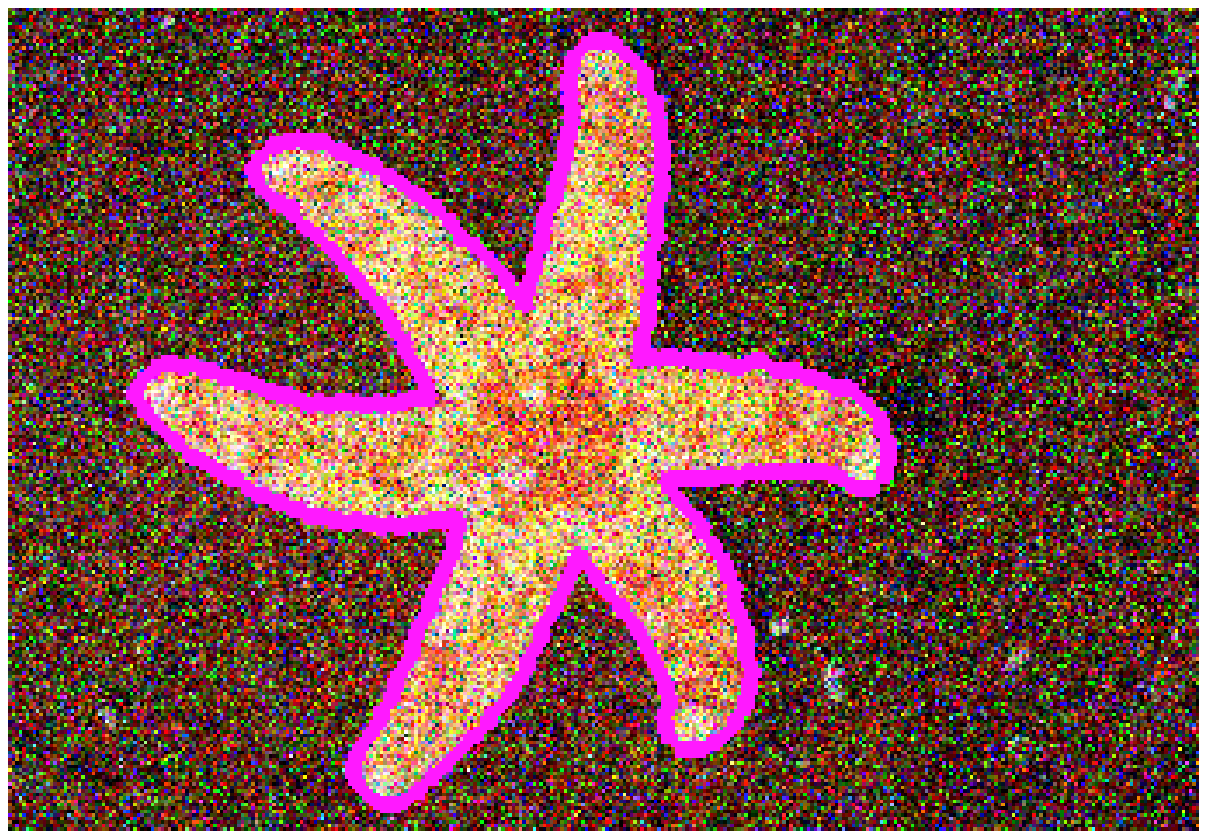,height=0.13\textwidth, width=0.2\textwidth}
\hspace{0.000001\textwidth}
\psfig{figure=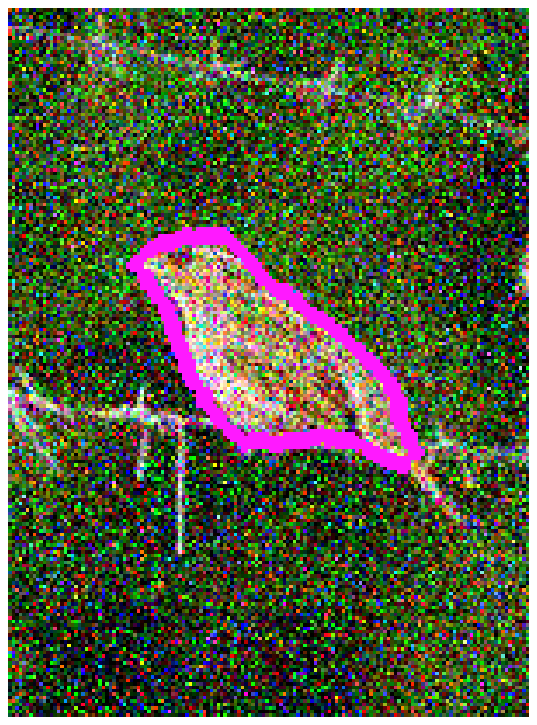,height=0.13\textwidth, width=0.2\textwidth}}
\centerline{(a)\hspace{0.17\textwidth} (b)\hspace{0.17\textwidth} (c)\hspace{0.17\textwidth} (d)\hspace{0.17\textwidth} (e)}
\caption{Visual comparison of segmentation results of corrupted (a) Yellow Pansy flower image with hybrid Gaussian ($\mu = 0.0$, $\sigma = 0.1$) and salt \& pepper noise (with probability $0.1$), (b) 86016 image with hybrid Gaussian ($\mu = 0.0$, $\sigma = 0.05$) and salt \& pepper noise (with probability $0.2$), (c) Cell4 image by hybrid Gaussian ($\mu = 0.0$, $\sigma = 0.01$) and salt \& pepper noise (with probability $0.01$), (d) Starfish image by hybrid Gaussian ($\mu = 0.0$, $\sigma = 0.05$) and salt \& pepper noise (with probability $0.1$), (e) Bird image with hybrid Gaussian ($\mu = 0.0$, $\sigma = 0.05$) and salt \& pepper noise (with probability $0.1$) using various techniques. \textbf{First row:} images with initial contours, \textbf{Second row:} segmentation results obtained by {\sc FEAC}, \textbf{Third row:} segmentation results obtained by {\sc NFACMKM}, \textbf{Fourth row:} segmentation results by {\sc FACGK}, \textbf{Fifth row:} segmentation results obtained by {\sc LPFAC}, \textbf{Sixth row:} segmentation results obtained by {\sc FDFEAC}, \textbf{Seventh row:} segmentation results obtained by {\sc GLFEAC} and \textbf{Eighth row:} segmentation results obtained by {\sc RGLFEAC}.\label{figure_hn_seg}}
\end{figure}

\subsection{Segmentation of Corrupted Images by Hybrid Noise}

To analyze the performance of the methods for segmentation of images under more complex environment, we considered the corrupted images with hybrid noise. In this experiment, mixture of different amount of Gaussian noise and salt \& pepper noise are considered to corrupt the original images. We considered all these methods to segment these corrupted images. Addition of different density level of noise creates different amount of region in-homogeneity in the images and it makes segmentation task is more difficult. Figure~\ref{figure_hn_seg} displays segmentation results of those corrupted images using various techniques. From this figure, it is observed that the global fuzzy energy based approach: {\sc FEAC} fails to segment images due to region in-homogeneity introduced by incorporation of hybrid noise.  Kernel based approaches: {\sc NFACMKM} and {\sc FACGK} are robust under noisy environment due to incorporation of Gaussian kernel into the energy function. Local energy based techniques: {\sc LPFAC} and {\sc FDFEAC} are not robust under hybrid noisy environment. For the corrupted images where density of noise level is less, only those images they produce good results. Combination of global and local fuzzy energy based techniques: {\sc GLFEAC} and {\sc RGLFEAC} are robust under hybrid noisy environment. Although, {\sc GLFEAC} fails to segment images with increasing density level of noise. Table~\ref{table_hybride} displays the similar finding.                

\begin{table}[htp]
\begin{center}
\begin{tabular}{|l|l|l|l|l|l|l|l|}\hline
Measures & \multicolumn{7}{|l|}{Techniques } \\\cline{2-8}
  & {\sc m1} & {\sc m2} & {\sc m3} & {\sc m4} & {\sc m5} & {\sc m6} & {\sc m7}\\\cline{1-8}
Ave. Jacard error & 0.311 & 0.165 & 0.200 & 0.376 & 0.304 & 0.312 & \textcolor[rgb]{0.00,1.00,0.00}{0.079} \\\hline
Ave. F-measure    & 0.814& 0.904 & 0.878 & 0.767 &0.796 & 0.815& \textcolor[rgb]{1.00,0.00,0.00}{0.952}  \\\hline
\end{tabular}
\end{center}
\caption{Quantitative comparison among various techniques with respect to average Jacard error and average F-measure over $100$ corrupted images with salt \& pepper and Gaussian noise. {\sc m1: feac}, {\sc m2: nfacmkm}, {\sc m3: facgk}, {\sc m4: lpfac}, {\sc m5: fdfeac}, {\sc m6: glfeac} and {\sc m7: rglfeac}. Green colored numeric value indicates least average Jacard error corresponds to the best segmentation result. Whereas, red colored numeric value indicates highest F-measure corresponds to the best segmentation result.\label{table_hybride}}
\end{table}

\section{Discussion and Future Work} \label{discussion}

Fuzzy energy based active contour model proposed by Krinidis and Chatzis~\cite{fuzzy_krinidis_2009} where energy terms are defined with global image information. This model produces good segmentation results for blurred, noisy and discontinuous edged images due to the incorpation of fuzzy logic into the function.     
However, it fails for images with gradual tonality variations, region in-homogeneity, background clutter, etc. Kernelized fuzzy energy active contour models~\cite{wu2015novel,badshah2018segmentation} obtained stable segmentation for images with region in-homogeneity and noise due to kernel function. On the other hand, fuzzy energy based active contour models~\cite{fang2016localized,shyu2012fuzzy} where energy terms are defined on local image information is also robust to the images having intensity in-homogeneity and noise. However, kernelized version of the global fuzzy energy based active contour models are more robust than the local fuzzy energy based active contour models on the level of noise and intensity in-homogeneity present in the images. Performance of the kernalised fuzzy active contour models degraded when the images contain more noise and in-homogeneous regions. In such cases, both local and global energy based active contour models~\cite{shyu2012global,mondal2016robust} provide better segmentation results than other models. However, with increasing density level of noise, they also fail to properly segment images. From the experimental results and their analysis, we observed that the global and local fuzzy energy based active contour models are better than all other models. 

All these discussed models require the intial contour to evolve the final contour. The main drawback of these models is the requirement of manually annotated images with initial contoures. One of the future direction will on automatic initialization of inital contour. Here, each pixel of the image is presented by hand-crafted features like color, texture, shape, etc. Since, features are not learned from the images, they may not robust for all types of images under the various conditions. In future, instead of considering hand-crafted features, learned feature using deep convolutional network from images will be considered. Since, all these models are working on pixel level, they need more computional time for doing segmentation of higher resolution images. Instead of pixel level, it is better to perform coarse level segmentation on super-pixel level then perform fine level segmentation on pixel level. These two step process will reduce the computional time for segmentation of high resolution images. All these models are designed for two regions segmentation (e.g. object and background). They can be extended to multi regions segmentation.

\section{Conclusions} \label{conclusion}

In this article, we review existing fuzzy energy based active contour models for image segmentation task from theoritical point of view. We also analyze robustness of all these models to segment various categories of images under variour complex conditions. Findings from experiments conclude that the global image information based fuzzy active contour models fail to properly segment the images having background clutter, noise, intensity in-homogeneity. However, kernelized versions of these global models are able to produce better segmentation results than the global methods due to the effectiveness of the kernel functions. On the contrary, local fuzzy energy based models obtain good results when density of noise level is limited. When level of noise density is increased, kernelize fuzzy energy based active contour models as well as local fuzzy energy based models fail to properly segment images. In such cases, the local and global fuzzy energy based models perform well. However, with increasing density level of noise, they also fail to properly segment images. We hope that this article helps to the reader to understand 
fuzzy energy based active models and several issues for image segmentation. We hope this article will help researchers toward designing new models for image segmentation. 

\bibliographystyle{spmpsci}

\bibliography{reference.bib}

\end{document}